\documentclass[a4paper, 11pt, oneside]{Thesis}  
\graphicspath{{Figures/}}  

\usepackage[notocbib]{apacite}
\usepackage{verbatim}  
\usepackage{vector}  
\usepackage{changepage}

\usepackage{graphicx}

\usepackage{tabularx}
\usepackage{adjustbox}

\usepackage{hyperref}

\usepackage{multirow}

\usepackage{paralist}
\usepackage{rotating}

\usepackage{floatrow}
\usepackage[colorinlistoftodos]{todonotes}

\usepackage{footmisc}

\usepackage{longtable,tabu}
\usepackage{enumitem}
\usepackage{multicol}

\newcommand{\ra}[1]{\renewcommand{\arraystretch}{#1}}

\theoremstyle{definition}
\newtheorem{exmp}{Example}[section]

\hypersetup{urlcolor=black, colorlinks=false}  

\begin{document}
\frontmatter	  

\title  {Extracting actionable information \\ from microtexts}
\authors  {\texorpdfstring
            {\href{a.hurriyetoglu@let.ru.nl}{Ali H\"{u}rriyeto\v{g}lu}}
            {Ali H\"{u}rriyeto\v{g}lu}
            }

\maketitle

\setstretch{1.3}  

\fancyhead{}  
\rhead{\thepage}  
\lhead{}  

\pagestyle{fancy}  

\mbox{}
\vfill

\begin{table}[ht]
\centering
\begin{tabular}{*{2}{m{0.48\textwidth}}}
\includegraphics[scale=0.5]{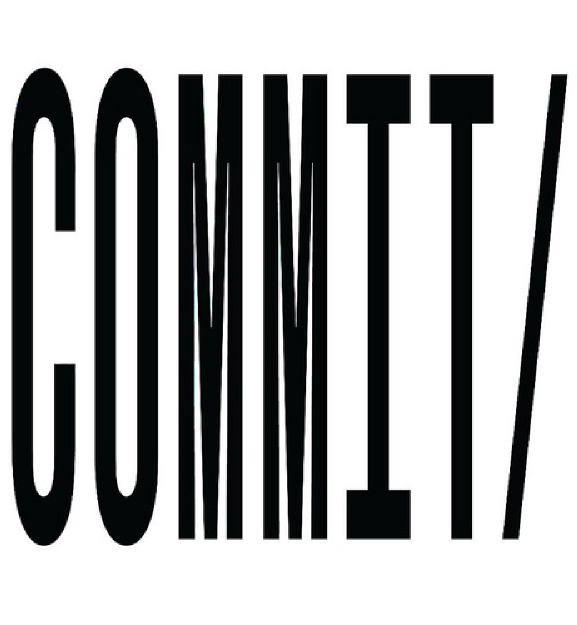} & The research was supported by the Dutch national research program COMMIT/.\\
\includegraphics[scale=0.5]{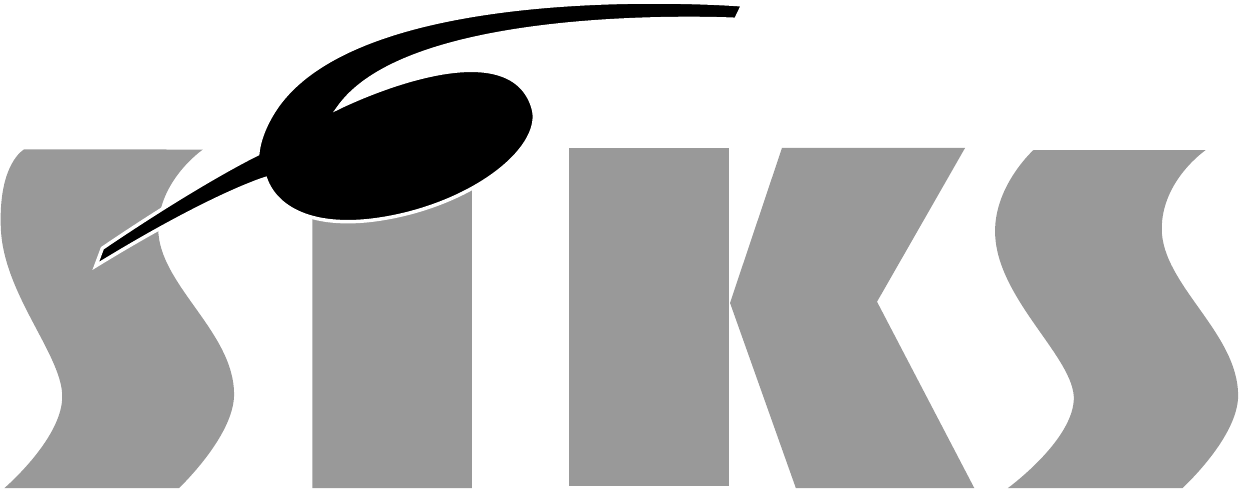} & SIKS Dissertation Series No. 2019-17. \newline The research reported in this thesis has been carried out under the auspices of SIKS, the Dutch Research School for Information and Knowledge Systems. \\

\end{tabular}
\label{tab:gt}
\end{table}

An electronic version of this dissertation is available at http://repository.ru.nl

ISBN: 978-94-028-1540-5

Copyright \textcopyright{} 2019 by Ali H\"{u}rriyeto\v{g}lu, Nijmegen, the Netherlands

Cover design by Burcu Hürriyetoğlu \textcopyright{}

Published under the terms of the Creative Commons Attribution License, CC BY 4.0 (http://creativecommons.org/licenses/by/4.0/), which permits unrestricted use, provided the original author and source are credited.


\chapter*{Acknowledgements}

The work presented in this book would be much harder without support from my family, friends, and colleagues.

First and foremost, I thank my supervisors prof. dr. Antal van den Bosch and Dr. Nelleke Oostdijk, for their invaluable support, guidance, enthusiasm, and optimism.

Being part of LAnguage MAchines (LAMA) group at Radboud University was a great experience. I met a lot of wonderful and interesting people. The round table meeting, hutje-op-de-hei, ATILA, the IRC channel, and lunches were fruitful sources of motivation and inspiration. I am grateful for this productive and relaxed work environment. Erkan, Florian, Iris, Kelly, Maarten, Martin, Wessel, Alessandro, Maria, and Ko were the key players of this pleasant atmosphere.

I thank all my co-authors and collaborators, who were in addition to my advisors and LAMA people, Piet Daas \& Marco Puts from Statistics Netherlands (CBS), Jurjen Wagemaker \& Ron Boortman from Floodtags, Christian Gudehus from Ruhr-University Bochum, for their invaluable contributions to the work in this thesis and to my development as a scientist. 

Florian and Erkan, we have done a lot together. Completing this together in terms of you supporting me again as being my paranymphs feels awesome. I appreciate it.

Serwan, Erkan, Ghiath, and Ali are the friends who were always there for me. I am very happy to have this invaluable company with you in this long and hard journey in life. 

Remy and Selman were the great people who made my time in Mountain View, CA, USA memorable during my internship at Netbase Solutions Inc. It would not be that much productive and cheerful without them. Their presence made me feel safe and at home.

I am grateful to prof. dr. Marteen de Rijke for welcoming me and Florian to his group on Fridays during the first year of my PhD studies. This experience provided me the context to understand information retrieval.

I would also like to thank the members of the doctoral committee, prof. dr. Martha Larson, prof. dr. ir. Wessel Kraaij, prof. dr. Lidwien van de Wijngaert, Dr. ir. Alessandro Bozzon, and Dr. Aswhin Ittoo. Moreover, I appreciate support of the anonymous reviewers who provided feedback to my submissions to scientific venues. Both the former and latter group of people significantly improved quality of my research and understanding scientific method.

There is not any dissertation that can be completed without institutional support. I appreciate support of Graduate School for the Humanities (GSH), Faculty of Arts, and Center for Language and Speech Technology (CLST) at Radboud University, COMMIT/ project, and the Netherlands Research School for Information and Knowledge Systems (SIKS). They provided the environment needed to complete this dissertation. 

I am always grateful to my family for their effort in standing by my side.

Last but not least, Sevara and Madina, you are meaning of happiness for me.

\begin{flushright}
Ali Hürriyetoğlu \\
Sarıyer, May 2019
\end{flushright}

\clearpage

 



\pagestyle{fancy}  

\lhead{\emph{Contents}}  
\tableofcontents  

\mainmatter	  
\pagestyle{fancy}  


\lhead{\emph{Introduction}}

\chapter{Introduction}




Microblogs such as Twitter represent a powerful source of information. Part of this information can be aggregated beyond the level of individual posts. Some of this aggregated information is referring to events that could or should be acted upon in the interest of e-governance, public safety, or other levels of public interest. Moreover, a significant amount of this information, if aggregated, could complement existing information networks in a non-trivial way. Here, we propose a semi-automatic method for extracting actionable information that serves this purpose. 

The term {\em event} denotes what happens to entities in a defined space and time~\cite{Casati+15}. Events that affect the behaviors and possibly the health, well-being, and other aspects of life of multiple people, varying from hundreds to millions, are the focus of our work. We are interested in both planned events (e.g. football matches and concerts) and unplanned events (e.g. natural disasters such as floods and earthquakes).

Aggregated information that can be acted upon is specified as actionable.\footnote{\url{http://www.oed.com/view/Entry/1941}, accessed June 10, 2018} Actionable information that can help to understand and handle or manage events may be detected at various levels: an estimated time to event, a graded  relevance estimate, an event's precise time and place, or an extraction of entities involved~\cite{Vieweg+10,Yin+12,Agichtein+08}.

Microtexts, which are posted on microblogs, are specified as short, user-dependent, 
minimally-edited texts in comparison to traditional writing products such as books, essays, and news articles~\cite{Ellen11,Khoury+14}.
\begin{sloppypar}
Users of a microblog platform may be professional authors but very often  they are non-professional authors whose writing skills vary widely and who are usually less concerned with readers' expectations as regards the well-formedness of a text. 
Generally, microtexts can be generated and published in short time spans with easy-to-use interfaces without being bound or dependent to a fixed place. Microbloggers may generate microtexts about anything they think, observe, or want to share through the microblogs.
\end{sloppypar}

In recent years, vast quantities of microtexts have been generated on microblogs that are known as social networking services, e.g., Twitter\footnote{\url{https://twitter.com} accessed June 10, 2018} and Instagram\footnote{\url{https://www.instagram.com}, accessed June 10, 2018}~\cite{Vallor16}. The network structure enables users of these platforms to connect with, influence, and interact with each other.
This additional social dimension affects the quality and quantity of the created microtexts on these platforms.

The form and function of the content on microblogs are determined by the microbloggers and can be anything that is within the scope of the used microblog's terms of service, which mostly only excludes excessive use of the platform and illegal behavior~\cite{Krumm+08}. 
Mostly, microbloggers can follow each other and create lists of microbloggers to keep track of the content flow on the microblog.

The form of microtexts is restricted only in terms of its length and its richness is supported by enabling the use of additional symbols to standard characters and punctuation for expressing additional meaning and structuring. The typical form of microblog content deviates from that found on the web and in traditional genres of written text in that it contains new types of entities such as user names, emoticons, highly flexible use and deviation of syntax, and spelling variation~\cite{Baldwin+13}.

The intended function of the conveyed information in microtexts is more liberal and more diverse than that of the information published through the standard media~\cite{Java+07,Zhao+09,Zhao+11b,Kavanaugh+14,Kwak+10}. The content is in principle as diverse as the different microbloggers on this platform~\cite{DeChoudhury+12b}. Distinct functions of the content on microblogs  include but are not limited to (dis)-approving microtexts by re-posting the same or commented version of a microtext, informing about observations, expressing opinions, reacting to discussions, sharing lyrics, quoting well known sayings, discussing history, or sharing artistic work. Moreover, microtexts may intend to mislead public by not reliably reflecting real-world events~\cite{Gupta+13}.

Often, microblogs allow the use of keywords or tags (referred to as hashtags on Twitter) to convey meta information about the context of a microtext. Tags serve the intended purpose to some extent, but they do not guarantee a high precision or a high recall when they are used for identifying relevant tweets about an event~\cite{Potts+2011}. Nevertheless, the use of tags in filtering microtexts remains a simple and straightforward strategy to select tweets. 

Given the aforementioned characteristics, the timely extraction of relevant and actionable subsets of microtexts from massive quantities of microtexts and in an arbitrary language has been observed to be a challenge~\cite{Imran+15,Sutton+08,Allaire16}. In order to handle events as efficiently as possible, it is important to understand at the earliest possible stage what information is available, specify what is relevant, apply the knowledge of what is relevant to new microtexts, and update the models we build as the event progresses. Meeting these requirements in a single coherent approach and at a high level of performance has not been tackled before.

We make use of the Twitter platform, which is a typical microblog, to develop and test our methodology. Twitter was established in 2006 and has around 313 million active users around the time we performed our studies.\footnote{\url{https://about.twitter.com/company}, accessed December 9, 2017} It allows its users to create posts that must be under a certain maximum length, so-called tweets.\footnote{Until November 2017 the length of a tweet was restricted to a maximum of 140 characters. Since then the limit was increased to 280.} Our research mainly utilizes the textual part, i.e. the microtext, of the tweets. On Twitter, a microtext consists of a sequence of printable characters, e.g. letters, digits, punctuation, and emoticons, which in addition to normal text, may contain references to user profiles on Twitter and links to external web content. The posting time of the tweets is used where the task has a temporal dimension such as time-to-event prediction.

\section{Research Questions}

This thesis is organized around a problem statement and three research questions. The overarching problem statement (PS) is:

\begin{adjustwidth}{1cm}{}
\textbf{PS: How can we develop an efficient automatic system that, with a high degree of precision and completeness, can identify actionable information about major events in a timely manner from microblogs while taking into account microtext and social media characteristics?}
\end{adjustwidth}

We focus on three types of actionable information, each of which is the topic of a research question. The prediction of the time to event, the detection of relevant information, and the extraction of target event information are addressed in research questions (RQ) 1, 2, and 3 respectively. 

\begin{adjustwidth}{1cm}{}
\textbf{RQ 1: To what extent can we detect patterns of content evolution in microtexts in order to generate time-to-event estimates from them?}
\end{adjustwidth} The studies under research question 1 explore linear and local regression and time series techniques to detect language use patterns that offer direct and indirect hints as to the start time of a social event, e.g. a football match or a music concert. 

\begin{adjustwidth}{1cm}{}
\textbf{RQ 2: How can we integrate domain expert knowledge and machine learning to identify relevant microtexts in a particular microtext collection?}
\end{adjustwidth} The task of discriminating between microtexts relevant to a certain event or a type of event and microtexts that are irrelevant despite containing certain clues such as event-related keywords is tackled in the scope of research question 2. The general applicability, speed, precision, and recall of the detection method are the primary concerns here. 

\begin{adjustwidth}{1cm}{}
\textbf{RQ 3: To what extent are rule-based and ML-based approaches complementary for classifying microtexts based on small datasets?}
\end{adjustwidth} Research question 3 focuses on microtext classification into certain topics from microtexts. We combine and evaluate the relevant information detection method with a linguistically oriented rule-based approach for extracting relevant information about various topics. 

The research questions are related to each other in a complementary and incremental manner. The results of each study in the scope of each research question feed into the following studies in the scope of the same or following research question.

\section{Data and Privacy}
We used the TwiNL framework\footnote{\url{http://www.ru.nl/lst/projects/twinl/}, accessed June 10, 2018}~\cite{TjongKimSang+13} and the Twitter API\footnote{\url{https://dev.twitter.com/rest/public}, accessed June 10, 2018} to collect tweets. For some use cases, we collected data using tweet IDs that were released by others in the scope of shared tasks. We provide details of the tweet collection(s) we used for each of the studies in the respective chapters.

Twitter data is a social media data type that has the potential to yield a biased sample or to be invalid~\cite{Tufekci14,Olteanu+16}. We take these restrictions into account while developing our methods and interpreting our results. When appropriate we discuss these potential biases and restrictions in more detail.

Microbloggers rightfully are concerned about their privacy when they post microtexts \cite{vanDenHoven+16}. In order to address this concern, we used only public tweets and never identified users in person in our research. For instance, we never use or report the unique user ID of a user. In all studies we normalize the screen names or user names to a single dummy value before we process the data. However, named entities remain as they occur in the text. Finally, the datasets are obtained and shared using only tweet IDs. Use of tweet IDs enable users who post these tweets to remove these tweets from our datasets. Consequently, we consider our data use in the research reported in this dissertation complies with the EU General Data Protection Regulation (GDPR).

\section{Contributions and Outline}

The results of our research suggests that if our goals are to be met or approximated, a balance needs to be, and in fact can be, struck between automation, available time, and human involvement when extracting actionable information from microblogs.

Aside from this global insight, we have three main contributions. First, we show that predicting time to event is possible for both in-domain and cross-domain scenarios. Second, we have developed a method which facilitates the definition of relevance for an analyst's context and the use of this definition to analyze new data. Finally, we integrate the machine learning based relevant information classification method with a rule-based information classification technique to classify microtexts. 

The outline of this thesis is as follows:

Chapter 2 is about time-to-event prediction. We report our experiments with linear and logistic regression and time series analysis. Our features vary from simple unigrams to detailed detection and handling of temporal expressions. We apply a form of distant supervision to collect our data by using hashtags. The use cases are mainly about football matches. We evaluated the models we created both on football matches and music concerts. This chapter is based on H\"{u}rriyeto\v{g}lu et al. \citeyear{Hurriyetoglu+13}, H\"{u}rriyeto\v{g}lu et al. \citeyear{Hurriyetoglu+14}, and H\"{u}rriyeto\v{g}lu et al. \citeyear{Hurriyetoglu+18}.

In Chapter 3 we introduce an interactive method that allows an expert to filter the relevant subset of a tweet collection. This approach provides experts with complete and precise information while relieving the experts from the task of crafting precise queries and risking the loss of precision or recall of the information they collect. Chapter 3 is based on H\"{u}rriyeto\v{g}lu et al. \citeyear{Hurriyetoglu+16a}, H\"{u}rriyeto\v{g}lu et al. \citeyear{Hurriyetoglu+16b}, H\"{u}rriyeto\v{g}lu et al. \citeyear{Hurriyetoglu+16c}, and H\"{u}rriyeto\v{g}lu et al. \citeyear{Hurriyetoglu+17b}.

In Chapter 4 we compare and integrate the relevant information detection approach we introduce in Chapter 3 with a rule-based information classification approach and compare the coverage and precision of the extracted information with manually annotated microtexts for topic-specific classification. This chapter is partly based on H\"{u}rriyeto\v{g}lu et al. \citeyear{Hurriyetoglu+17} and furthermore describes original work.

In Chapter 5 we summarize our contributions, formulate answers to our research questions, and provide an outlook towards future research.

\lhead{\emph{Time-to-Event Prediction}}
\chapter{Time-to-Event Detection}


\section{Introduction}

Social media produce data streams that are rich in content. Within the mass of microtexts posted on Twitter, for example, many sub-streams of messages (tweets) can be identified that refer to the same event in the real world. Some of these sub-streams refer to events that are yet to happen, reflecting the joint verbalized anticipation of a set of Twitter users towards these events. In addition to overt markers such as event-specific hashtags in messages on Twitter, much of the information on future events is present in the surface text stream. These come in the form of explicit as well as implicit cues: compare for example `the match starts in two hours' with `the players are on the field; can't wait for kickoff'. Identifying both types of linguistic cues may help disambiguate and pinpoint the starting time of an event, and therefore the remaining time to event (TTE).

Estimating the remaining time to event requires the identification of the future start time of an event in real clock time. This estimate is a core component of an alerting system that detects significant events (e.g. on the basis of mention frequency, a subtask not in focus in the present study) that places the event on the agenda and alerts users of the system. This alerting functionality is not only relevant for people interested in attending an event; it may also be relevant in situations requiring decision support to activate others to handle upcoming events, possibly with a commercial, safety, or security goal. A historical example of the latter category was {\em Project X Haren},\footnote{\url{http://en.wikipedia.org/wiki/Project_X_Haren}, accessed June 10, 2018} a violent riot on September 21, 2012, in Haren, the Netherlands, organized through social media. This event was abundantly announced on social media, with specific mentions of the date and place. Consequently, a national advisory committee, installed after the event, was asked to make recommendations to handle similar future events. The committee stressed that decision-support alerting systems on social media need to be developed, ``where the focus should be on the detection of collective patterns that are remarkable and may require action"~\cite[p. 31; our translation]{Cohen+13}.

Our research focuses on textual data published by humans via social media about particular events. If the starting point of the event in time is taken as the anchor $t=0$ point in time, texts can be viewed in relation to this point, and generalizations may be learned over texts at different distances in time to $t=0$. The goal of this study is to present new methods that are able to automatically estimate the time to event from a stream of microtext messages. These methods could serve as modules in news media mining systems\footnote{For instance, \url{https://www.anp.nl/product/anp-agenda}, accessed October 15, 2018} to fill upcoming event calendars. The methods should be able to work robustly in a stream of messages, and the dual goal would be to make (i) reliable predictions of the time to event (ii) as early as possible. Moreover, the system should be able, in the long run, to freely detect relevant future events that are not yet on any schedule we know in any language represented on social media. Predicting that an event is starting imminently is arguably less useful than being able to predict its start in a number of days. This implies that if a method requires a sample of tweets (e.g. with the same hashtag) to be gathered during some time frame, the frame should not be too long, otherwise predictions could come in too late to be relevant. 

The idea of publishing future calendars with potentially interesting events gathered (semi-)automatically for subscribers, possibly with personalization features and the option to harvest both social media and the general news, has been implemented already and is available through services such as Daybees\footnote{\url{http://daybees.com/}, accessed August 3, 2014} and Songkick\footnote{\url{https://www.songkick.com/}, accessed June 10, 2018}. To our knowledge, based on the public interfaces of these platforms, these services perform directed crawls of (structured) information sources, and identify exact date and time references in posts on these sources. Restricting the curation with explicit date mentions decreases the number of events that can be detected. These platforms also manually curate event information, or collect this through crowd-sourcing. However, non-automatic compilation of event data is costly, time consuming, and error prone, while it is also hard to keep the information up to date and ensure its correctness and completeness. 

\begin{sloppypar}
In this research we focus on developing a method for estimating the starting time of scheduled events, and use past and known events for a controlled experiment involving Dutch twitter messages. We study time-to-event prediction in a series of three connected experiments that are based on published work in scientific venues and build on each other's results. After we introduce (in Sections \ref{CH2_rel_research} and \ref{ch2DataCollection} respectively) related research and the data collections we used, we report on these studies. We start with a preliminary bag-of-words (BoW) based study to explore the potential of linear and logistic regression and time series models to identify the start time of football matches~\cite{Hurriyetoglu+13} in Section~\ref{DIRchapter}. Next, in Section~\ref{EMNLPchapter}, we focus on time series analysis of time expressions for the same task~\cite{Hurriyetoglu+14}. Finally, we add BoW and rule-based features to time expression features and include music concerts as an event type for the time-to-event (TTE) prediction task~\cite{Hurriyetoglu+18} in Section~\ref{NLPTchapter}. At the end of the chapter, conclusions derived from these studies and the overall evaluation of the developed TTE estimation method are provided.
\end{sloppypar}

\section{Related Research}
\label{CH2_rel_research}
The growing availability of digital texts with time stamps, such as e-mails, weblogs, and online news has spawned various types of studies on the analysis of patterns in texts over time. An early publication on the general applicability of time series analysis on time-stamped text is~\citeauthor{Kleinberg06}~\citeyear{Kleinberg06}. A more recent overview of future predictions using social media is~\citeauthor{Yu+12}~\citeyear{Yu+12}. A popular goal of time series analysis of texts is {\em event prediction}, where a correlation is sought between a point in the future and preliminary texts.

In recent years, there have been numerous studies in the fields of text mining and information retrieval directed at the development of approaches and systems that would make it possible to forecast events. The range of (types of) events targeted is quite broad and varies from predicting manifestations of societal unrest such as nation-wide strikes or uprisings~\cite{Ramakrishnan+14,Muthiah14,Kallus14}, to forecasting the events that may follow a natural disaster~\cite{Radinsky+12}. Studies that focus specifically on identifying future events are, for example, \citeauthor{Ricardo05} \citeyear{Ricardo05}; \citeauthor{Dias+11} \citeyear{Dias+11}; \citeauthor{Jatowt+11} \citeyear{Jatowt+11}; \citeauthor{Briscoe+15} \citeyear{Briscoe+15}. A review of the literature shows that while approaches are similar to the extent that they all attempt to learn automatically from available data, they are quite different as regards the information they employ. For example,~\citeauthor{Radinsky+12} ~\citeyear{Radinsky+12} attempt to learn from causality pairs (e.g. a flood causes people to flee) in long-ranging news articles to predict the event that is likely to follow the current event.~\citeauthor{Lee+14} ~\citeyear{Lee+14} exploit the up/down/stay labels in financial reports when trying to predict the movement of the stock market the next day.~\citeauthor{Redd+13} ~\citeyear{Redd+13} attempt to calculate the risk of veterans becoming homeless, by analyzing the medical records supplied by the U.S. Department of Veterans Affairs.

Predicting the type of event is one aspect of event forecasting; giving an estimate as to the time when the event is (likely) to take place is another. Many studies, such as the ones referred to above, focus on the event type rather than the event time, that is, they are more generally concerned with future events, but not particularly with predicting the specific date or hour of an event. The same goes for~\citeauthor{Noro+06} \citeyear{Noro+06}, who describe a system for the identification of the period in which an event will occur, such as in the morning or at night. And again, studies such as those by~\citeauthor{Becker+12} \citeyear{Becker+12} and~\citeauthor{Kawai+10} \citeyear{Kawai+10} focus more specifically on the type of information that is relevant for predicting event times, while they do not aim to give an exact time. Furthermore,~\citeauthor{Nakajima+14} \citeyear{Nakajima+14} extract (candidate) semantic and syntactic patterns with future reference from news articles in an attempt to improve the retrieval of future events. Finally,~\citeauthor{Noce+14} \citeyear{Noce+14} present a method that automatically extracts forward-looking statements from earnings call transcripts in order to support business analysts in predicting those future events that have economic relevance. 

There are also studies that are relevant in this context due to their focus on social media, and Twitter in particular. Research by~\citeauthor{Ritter+12} \citeyear{Ritter+12} is directed at creating a calendar of automatically detected events. They use explicit date mentions and words typical of a given event. They train on annotated open domain event mentions and use the TempEx tagger~\cite{Mani+00} for the detection of temporal expressions. Temporal expressions that point to certain periods such as `tonight' and `this morning' are used by~\citeauthor{Weerkamp+12} \citeyear{Weerkamp+12} to detect personal activities at such times. In the same line, Kunneman and van den Bosch~\citeyear{Kunneman+12} show that machine learning methods can differentiate between tweets posted before, during, and after a football match. 

\citeauthor{Hurriyetoglu+13} \citeyear{Hurriyetoglu+13} also use tweet streams that are related to football matches and attempt to estimate the time remaining to an event, using local regression over word time series. In a related study,~\citeauthor{Tops+13} \citeyear{Tops+13} use support vector machines to classify the TTE in automatically discretized categories. The results obtained in the latter two studies are at best about a day off in their predictions. Both studies also investigate the use of temporal expressions but fail to leverage the utility of this information source, presumably because they use limited sets of regular expressions: In each case fewer than 20 expressions were used.

The obvious baseline that we aim to surpass with our method is the detection of explicit temporal expressions from which the TTE could be inferred directly. Finding explicit temporal expressions can be achieved with rule-based temporal taggers such as the HeidelTime tagger~\cite{Strotgen+13}, which generally search for a small, fixed set of temporal expressions~\cite{Kanhabua+12,Ritter+12}. As it is apparent from studies such as~\citeauthor{Strotgen+13} \citeyear{Strotgen+13}, \citeauthor{Mani+00} \citeyear{Mani+00}, and~\citeauthor{Chang+12} \citeyear{Chang+12}, temporal taggers are successful in identifying temporal expressions in written texts as encountered in more traditional genres, such as news articles or official reports. They, in principle, can also be adapted to cope with various languages and genres 
(cf.~\citeauthor{Strotgen+13}, 2013). However, the focus is typically on temporal expressions that have a standard form and a straightforward interpretation. 

Various studies have shown that, while temporal expressions provide a reliable basis for the identification of future events, resolving the reference of a given temporal expression remains a challenge (cf.~\citeauthor{Kanhabua+12}, 2012; \citeauthor{Jatowt+13}, 2013; \citeauthor{Strotgen+13}, 2013; \citeauthor{Morency06}, 2006). In certain cases temporal expressions may even be obfuscated intentionally~\cite{Nguyen-Son+14} by deleting temporal information that can be misused to commit a crime against or invade the privacy of a user. Also, temporal taggers for languages other than English are not as successful and widely available as they are for English. Thus, basing TTE estimation only on temporal taggers is not optimal.

Detecting temporal expressions in social media text requires a larger degree of flexibility in recognizing the form of a temporal expression and identifying its value than it would in news text. Part of this flexibility may be gained by learning temporal distances from data rather than fixing them at knowledge-based values. \citeauthor{Blamey+13} \citeyear{Blamey+13} suggest estimating the values of temporal expressions on the basis of their distributions in the context of estimating creation time of photos on online social networks. \citeauthor{Hurriyetoglu+14} \citeyear{Hurriyetoglu+14} develop a method that relaxes and extends both the temporal pattern recognition and the value identification for temporal expressions.

\section{Data Collection}
\label{ch2DataCollection}
For our research we collected tweets referring to scheduled Dutch premier league football matches (FM) and music concerts (MC) in the Netherlands. These events trigger many anticipatory references on social media before they happen, containing numerous temporal expressions and other non-temporal implicit lexical clues on when they will happen.

We harvested all tweets from Twiqs.nl, an online database of Dutch tweets collected from December 2010 onwards \cite{TjongKimSang+13}. Both for football matches and music concerts we used event-specific hashtags to identify the event, i.e. we used the hashtag that, to the best of our knowledge, was the most distinctive for the event. The hashtags used for FM follow a convention where the first two or three letters of the names of the two teams playing against each other are concatenated, starting with the host team. An example is \#ajatwe for a football match in which Ajax is the host, and Twente is the away team. The MC hashtags are mostly concatenations of the first and last name of the artist, or concatenations of the words forming the band name. Although in the latter case for many of the hashtags shorter variants exist, we did not delve into the task of identifying such variants~\cite{Ozdikis+12,Wang+13} and used the full variants.

The FM dataset was collected by selecting the six best performing teams of the Dutch premier league in 2011 and 2012. We queried all matches in which these teams played against each other in the calendar years 2011 and 2012.\footnote{Ajax Amsterdam (aja), Feyenoord Rotterdam (fey), PSV Eindhoven (psv), FC Twente (twe), AZ Alkmaar (az), and FC Utrecht (utr).} The MC dataset contains tweets from concerts that took place in the Netherlands between January 2011 and September 2014. We restricted the data to tweets sent within eight days before the event.\footnote{An analysis of the tweet distribution shows that the 8-day window captures about 98\% of all tweets by means of the hashtags that we used.} We decided to refrain from extending the time frame to the point in time when the first tweet that mentions the hashtag was sent, because hashtags may denote a periodic event or a different event that takes place at another time, which may lead to inconsistencies that we did not aim to solve in the research reported here. Most issues having to do with periodicity, ambiguity and inconsistency are absent within the 8-day window, i.e. tweets with a particular event hashtag largely refer to the event that is upcoming within the next eight days. 

As noted above, the use of hashtags neither provides complete sets of tweets about the events nor does it ensure that only tweets pertaining to the main event are included \cite{Tufekci14}. We observed that some event hashtags from both data sets were used to denote other similar events that were to take place several days before the event we were targeting, such as a cup match instead of a league match between the same teams, or another concert by the same artist. For example, the teams Ajax and Twente played a league and a national cup match within a period of eight days (two consecutive Sundays). In case there was such a conflict, we aimed to estimate the TTE for the relatively bigger event, i.e. in terms of the available Dutch tweet count about it. For \#ajatwe, this was the league match. In so far as we were aware of related events taking place within the same 8-day window with comparable tweet counts, we did not include these events in our datasets.

Social media users tend to include various additional confusing hashtags other than the target hashtag in a tweet. We consider a hashtag to be confusing when it denotes an event different from the event designated by the hashtag creation rule. For football matches, this is the case for example when a user uses \#tweaja instead of \#ajatwe when referring to a home game for Ajax; for music concerts, we may encounter tweets with a specific hashtag where these tweets do not refer to the targeted event using the hashtag \#beyonce for a topic other than a Beyonc\'{e} concert. In these cases, the unrelated tweets are not removed, and are used as if they were referring to the main event. We aim for our approach to be resistant to such noise which after all, we find, is present in most social media data. 


In Table \ref{table:datasetTwCnt} we present an overview of the datasets that were used in the research reported on in the remainder of this chapter. We created a version without retweets for each dataset in order to be able to measure the effect of the retweets in our experiments. We used the simple pattern ``rt @" to identify retweets and create the `FM without retweets' and `MC without retweets' datasets. 
Events that have fewer than 15 tweets were eliminated. Consequently, the number of events in `MC without retweets' dropped to 32, since the number of tweets about three events became lower than 15 after removal of retweets. A different subset of this data was used for each experiment in the following subsections. 

\begin{table}[htb]
\ttabbox{\caption{Number of events and tweets for the FM and MC data sets (FM=football matches; MC=music concerts). In the `All' sets all tweets are included, that is, original posts and retweets.}\label{table:datasetTwCnt}}
{\begin{tabular*}{0.95\textwidth}{lcr@{\hskip 0.3in}r@{\hskip 0.3in}r@{\hskip 0.3in}r}
\toprule
{} &  \# of events & \multicolumn{4}{c}{\# of tweets} \\
\toprule
{} &     &  Min. &  Median &  Max. &   Total \\
\midrule
FM All              &  60 &      305 &  2,632 &    34,868 &  262,542 \\
FM without retweets &  60 &      191 &  1,345 &    23,976 &  139,537 \\
MC All              &  35 &       15 &   54 &     1,074 &    4,363 \\
MC without retweets &  32 &       15 &   55 &      674 &    3,479 \\
\bottomrule
\end{tabular*}
}

\end{table}

Each tweet in our data set has a time stamp of the moment (hour-minute-seconds) it was posted. Moreover, for each football match and each music concert we know exactly when it took place: the event start times were gathered from the websites Eredivisie.nl for football matches and lastfm.com for music concerts. This information is used to calculate for each tweet the actual time that remains to the start of the event, as well as to compute the absolute error in estimating the remaining time to event.

We would like to emphasize that the final data set contains all kinds of discussions that do not contribute to predicting time of event directly. The challenge we undertake is to make sense of this mixed and unstructured content for identifying temporal proximity of an event.

The football matches (FM All and FM without retweets) data sets were used to develop the time-to-event estimation method we suggest in this chapter. The music concerts data set was used in Section~\ref{NLPTchapter} in testing performance of the cross-domain applicability of the proposed method.

\section{Estimating the Time between Twitter Messages and Future Events}
\label{DIRchapter}

{\bf Based on:} Hürriyetoğlu, A., Kunneman, F., \& van den Bosch, A. (2013). Estimating the Time Between Twitter Messages and Future Events. In {\em Proceedings of the 13th Dutch-Belgian Workshop on Information Retrieval (pp. 20–23)}. Available from \url{http://ceur-ws.org/Vol-986/paper\_23.pdf} 

In this section, we describe and test three methods to estimate the remaining time between a series of microtexts (tweets) and the future event they refer to via a hashtag. Our system generates hourly forecasts. For comparison, two straightforward approaches, linear regression and local regression are applied to map hourly clusters of tweets directly onto time to event. To take changes over time into account, we develop a novel time series analysis approach that first derives word frequency time series from sets of tweets and then performs local regression to predict time to event from nearest-neighbor time series. We train and test on a single type of event, Dutch premier league football matches. Our results indicate that about four days or more before the event, the time series analysis produces relatively accurate time-to-event predictions that are about one day off; closer to the event, local regression offers the most accurate predictions. Local regression also outperforms both mean and median-based baselines, but on average none of the tested systems has a consistently strong performance through time.




\subsection{Introduction}

We test the predictive capabilities of three different approaches. The first system is based on linear regression and maps sets of tweets with the same hashtag posted during a particular hour to a time-to-event estimate. The second system attempts to do the same based on local regression. The third system uses time series analysis. It takes into account more than a single set of tweets: during a certain time period it samples several sets of tweets in fixed time frames, and derives time series information from individual word frequencies in these samples. It compares these word frequency time series profiles against a labeled training set of profiles in order to find similar patterns of change in word frequencies. The method then adopts local regression: finding a nearest-neighbor word frequency time series, the time to event stored with that neighbor is copied to the tested time series. With this third system, and with the comparison against the second system, we can test the hypothesis that it is useful to gather time series information (more specifically, patterns in word frequency changes) over a period of time.

The three systems are described in Section~\ref{regressionDIR}. Section~\ref{experimentsDIR} describes the overall experimental setup, including the baseline and the evaluation method used. The results are presented and analyzed in Section~\ref{resultsDIR}. We conclude with a discussion of the results and the following steps in Section~\ref{conclusionDIR}.




\subsection{Methods}
\label{regressionDIR}
The methods adopted in our study operate on streams of tweets, and generate hourly forecasts for the events that tweets with the same hashtag refer to. The single tweet is the smallest unit available for this task; we can also consider more than one tweet and aggregate tweets over a certain time frame. If these single tweets or sets of tweets are represented as a bag-of-words vector, the task can be cast as a regression problem: mapping a feature vector onto a continuous numeric output representing the time to event. In this study the smallest time unit is one hour, and all three methods work with this time frame.

\subsubsection{Linear and Local Regression}

In linear regression, each feature in the bag-of-words feature vector representing the presence or frequency of occurrence of a specific word can be regarded as a predictive variable to which a weight can be assigned that, in a simple linear function, multiplies the value of the predictive variable to generate a value for the response variable, the time to event. A multiple linear regression function can be approximated by finding the weights for a set of features that generates the response variable with the smallest error.

Local regression, or local learning \cite{Atkeson+97}, is the numeric variant of the $k$-nearest neighbor classifier. Given a test instance, it finds the closest $k$ training instances based on a similarity metric, and bases a local estimation of the numeric output by taking some average of the outcomes of the closest $k$ training instances. 

Linear regression and local regression can be considered baseline approaches, but are complementary. While in linear regression an overall pattern is generated to fit the whole training set, local regression only looks at local information for classification (the characteristics of single instances). Linear regression has a strong bias that is not suited to map Gaussian or other non-linear distributions. In contrast, local regression is unbiased and will adapt to any local distribution.

\subsubsection{Time Series Analysis}

Time series are data structures that contain multiple measurements of data features over time. If values of a feature change meaningfully over time, then time series analysis can be used to capture this pattern of change. Comparing new time series with memorized time series can reveal similarities that may lead to a prediction of a subsequent value or, in our case, the time to event. Our time series approach extends the local regression approach by not only considering single sets of aggregated tweets in a fixed time frame (e.g. one hour in our study), but creating sequences of these sets representing several consecutive hours of gathered tweets. Using the same bag-of-words representation as the local regression approach, we find nearest neighbors of sequences of bag-of-word vectors rather than single hour frames. The similarity between a test time series and a training time series of the same length is calculated by computing their Euclidean distance. In this study we did not further optimize any hyperparameters; we set $k=1$. 

The time series approach generates predictions by following the same strategy as the simple local regression approach: upon finding the nearest-neighbor training time series, the time to event of this training time series is taken as the time-to-event estimate of the test time series. In case of equidistant nearest neighbors, the average of their associated time to event is given as the prediction.

\subsection{Experimental Set-up}
\label{experimentsDIR}

\subsubsection{Training and Test Data Generation}

To generate training and test events we cut the set of the football match events, FM all, in two, resulting in a calender year (instead of a season) of matches for both training and testing. The events that happened in 2011 and 2012 were used as training and test data respectively. As the aim of our experiments was to estimate the time to event in terms of hours, we selected matches played on the same weekday and with the same starting time: Sundays at 2:30 PM (the most frequent starting time). This resulted in 12 matches as training data (totaling 54,081 tweets) and 14 matches as test events (40,204 tweets).

The goal of the experiments was to compare systems that generate hourly forecasts of the event start time for each test event. This was done based on the information in aggregated sets of tweets within the time span of an hour. The linear and local regression methods only operate on vectors representing one hour blocks. The time series analysis approach makes use of longer sequences of six hour blocks --- this number was empirically set in preliminary experiments.

The aggregated tweets were used as training instances for the linear and local regression methods. To maximize the number of training instances, we generated a sequence of overlapping instances using the minute as a finer-grained shift unit. At every minute, all tweets posted within the hour before the tweets in that minute were added to the instance. This resulted in 41,987 training instances that were generated from 54,081 tweets.

In order to reduce the feature space for the linear and local regression instances, we pruned every bag-of-word feature that occurred less than 500 times in the training set. Linear regression was applied by means of {\em R}.\footnote{\url{http://www.r-project.org/}, accessed June 10, 2018} Absolute occurrence counts of features were taken into account. For local regression we made use of the $k$-NN implementation as part of TiMBL\footnote{\url{http://ilk.uvt.nl/timbl}, accessed June 10, 2018}, setting $k=5$, using information gain feature weighting, and an overlap-based metric as a similar metric that does not count matches on zero values (features marking words that are absent in both test and training vectors). For $k$-NN, the binary values were used.

The time series analysis vectors are not filled with absolute occurrence counts, but with relative and smoothed frequencies. After having counted all words in each time frame, two frequencies are computed for each word. One is the overall frequency of a word which is calculated as the sum of its counts in all time frames, divided by the total number of tweets in all time frames in our 8-day window. This frequency ranges between $0$ (the word does not occur) and $1$ (the word occurs in every tweet). The other frequency is computed per time frame for each word, where the word count in that frame is divided by the number of tweets in the frame. The latter frequency is the basic element in our time series calculations.

\begin{sloppypar}
As many time frames contain only a small number of tweets, especially the frames more than a few days before the event, word counts are sparse as well. Besides taking longer time frames of more than a single sample size, frequencies can also be smoothed through typical time series analysis smoothing techniques such as moving average smoothing. We apply a pseudo-exponential moving average filter by replacing each word count by a weighted average of the word count at time frames $t$, $t-1$, and $t-2$, where $w_t=4$ (the weight at $t$ is set to 4), $w_{t-1}=2$, and $w_{t-2}=1$.
 \end{sloppypar}

\subsubsection{Baseline}

To ground our results better, we computed two baselines derived from the training set: the median and the mean time to event over all training tweets. For the median baseline, all tweets in the training set were ordered in time and the median time was identified. As we use one-hour time frames throughout our study, we round the median by the one-hour time frame it is in, which turns out to be $-3$ hours. The mean was computed by averaging the time to event of all tweets, and again rounded at the hour. The mean is $-26$ hours. Since relatively many tweets are posted just before a social event, these baselines form a competitive challenge.

\subsubsection{Evaluation}

A common metric for evaluating numeric predictions is the Root Mean Squared Error (RMSE), cf. Equation~\ref{rmse}. For all hourly forecasts made in $N$ hour frames, a sum is made of the squared differences between the actual value $v_{i}$ and the estimated value $e_{i}$; the (square) root is then taken to produce the RMSE of the prediction series. As errors go, a lower RMSE indicates a better approximation of the actual values.


\begin{equation}
 RMSE = ({\frac{1} {N}{\sum\limits_{i = 1}^N {(v_{i} - e_{i} } })^{2} })^{1/2}
 \label{rmse}
\end{equation}

\subsection{Results}
\label{resultsDIR}

\begin{sidewaystable}
\setlength{\tabcolsep}{0.2mm}
\begin{tabular}{l||r|r|r|r|r|r|r|r|r|r|r|r|r|r|r}
 & \multicolumn{8}{c|}{Spring 2012} & \multicolumn{6}{c}{Fall 2012} &  \\ 
 & azaja & feyaz & feyutr & psvfey & tweaja & twefey & tweutr & utraz & azfey & psvaz & twefey & utraz & utrpsv & utrtwe & Av (sd) \\ \hline 
Baseline Median & 63 & 49 & 54 & 62 & 38 & 64 & 96 & 71 & 62 & 67 & 62 & 66 & 61 & 62 & 63 (12) \\ 
Baseline Mean & 51 & 40 & 44 & 51 & 31 & 52 & 77 & 58 & 50 & 55 & 51 & 53 & 49 & 51 & 51 (10) \\ 
Linear regression & 52 & 42 & 59 & 54 & 410 & 41 & 41 & 33 & 111 & 31 & 110 & 54 & 37 & 68 & 82 (94) \\ 
Local regression & 48 & 44 & 35 & 41 & 43 & 43 & 31 & 20 & 57 & 40 & 52 & 48 & 34 & 52 & {\bf 43} (9) \\ 
Time Series & 48 & 50 & 42 & 43 & 45 & 41 & 63 & 70 & 48 & 58 & 46 & 71 & 59 & 63 & 54 (10) \\

\end{tabular}
\caption{Overall Root Mean Squared Error scores for each method: difference in hours between the estimated time to event and the actual time-to-event}
\label{tab:main_resultsDIR}
\end{sidewaystable}



Table~\ref{tab:main_resultsDIR} displays the averaged RMSE results on the 14 test events. The performance of the linear regression method is worse than both baselines, while the time series analysis outperforms the median baseline but lags behind the mean baseline. As the best performing method, which is local regression, is still an unsatisfactory 43 hours off (almost two days) on average, indicating that there is still a lot of improvement needed.

\begin{figure}[htb]
\begin{center}
\includegraphics[scale=0.525,angle=90,origin=c]{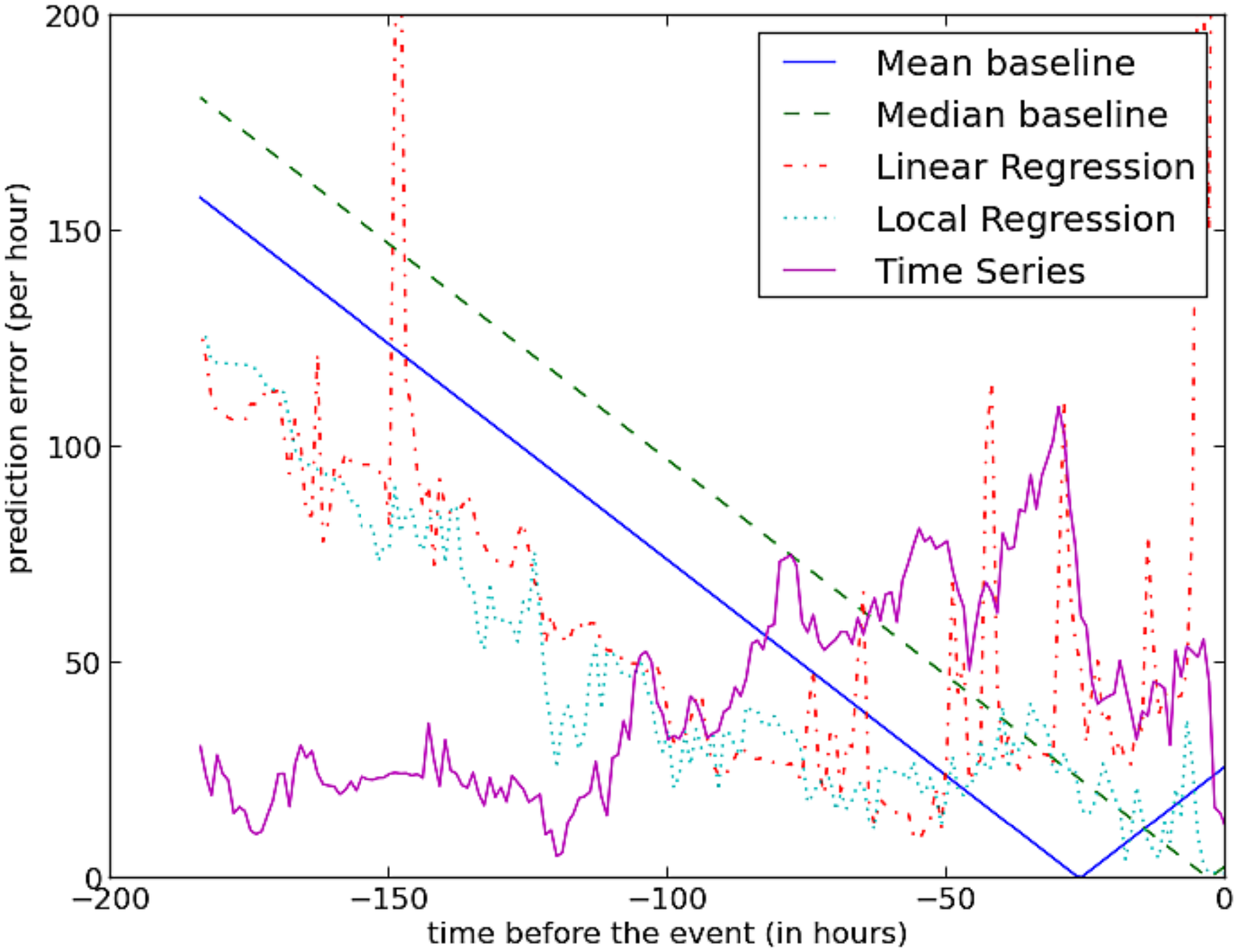}
\vspace{-3cm}
\caption{RMSE curves for the two baselines and the three methods for the last 192 hours before $t=0$.}
\label{fig:time}
\end{center}
\end{figure}

The performance of the different methods in terms of their RMSE according to hourly forecasts is plotted in Figure~\ref{fig:time}. In the left half of the graph the three systems outperform the baselines, except for an error peak of the linear regression method at around $t=-150$. Before $t = -100$ the time series prediction is performing rather well, with RMSE values averaging 23 hours, while the linear regression and local regression methods start at larger errors which decrease as time progresses. In the second half of the graph, however, only the local regression method appears to retain fairly low RMSE values at an average of 21 hours, while the linear regression method becomes increasingly erratic in its predictions. The time series analysis method also produces considerably higher RMSE values in the last days before the events.

\subsection{Conclusion}
\label{conclusionDIR}

In this study we explored and compared three approaches to time-to-event prediction on the basis of streams of tweets. We tested on the prediction of the time to event of fourteen football matches by generating hourly forecasts. Compared to two simplistic baselines based on the mean and median of the time to event of tweets sent before an event, only one of the three approaches, local regression, displays better overall RMSE values on the tested prediction range of $192 \ldots 0$ hours before the event. Linear regression generates erratic predictions and scores below both baselines. A novel time series approach that implements local regression based on sequences of samples of tweets performs better than the mean baseline, but worse than the median baseline.

Yet, the time series method generates quite accurate forecasts during the first half of the test period. Before $t < -100$ hours, i.e. earlier than four days before the event, predictions by the time series method are only about a day off (23 hours on average in this time range). When $ t \leq -100$, the local regression approach based on sets of tweets in hourly time frames is the better predictor, with quite low RMSE values close to $t=0$ (21 hours on average in this time range).

On the one hand, our results are not very strong: predictions that are on average more than two days off the actual time to event and that are at the same time only mildly better than the baselines cannot be considered precise. However, we observe that local regression and time series analysis methods have the strength of being in average precise and precise at an early phase of the event respectively. Since our ultimate aim is to generate precise estimates as early as possible, we will continue our exploration for an optimal solution in time series analysis in the following experiments. The first step in this endeavor will be analysis of temporal expression usage in the time series approach. 

Finally, we observed that RMSE is not suitable for evaluating performance of the methods that analyze microtexts. The measure highly penalizes outliers, which are abundant in microtexts. Consequently, we will be using MAE in order to measure the average performance of the approaches we suggest in the following studies.

\section{Estimating Time to Event from Tweets Using Temporal Expressions}
\label{EMNLPchapter}
{\bf Based on:} Hürriyetoğlu, A., Oostdijk, N., \& van den Bosch, A. (2014, April). Estimating time to event from tweets using temporal expressions. In {\em Proceedings of the 5th Workshop on Language Analysis for Social Media (LASM) (pp. 8–16)}. Gothenburg, Sweden: Association for Computational Linguistics. Available from \url{http://www.aclweb.org/anthology/W14-1302}


Given a stream of Twitter messages about an event, we investigate the predictive power of temporal expressions in the messages to estimate the time to event (TTE). From labeled training data we learn average TTE estimates of temporal expressions and combinations thereof, and define basic estimation rules to compute the time to event from temporal expressions, so that when they occur in a tweet that mentions an event we can generate a prediction. We show in a case study on football matches that our estimations are off by about eight hours on average in terms of mean absolute error (MAE).

\subsection{Introduction}

In this study we do not use a rule-based temporal tagger such as the HeidelTime tagger~\cite{Strotgen+13}, which searches for only a limited set of temporal expressions. Instead, we adopt an approach that uses a large set of temporal expressions, created by using
lexical items and generative rules, and a training method that automatically determines the TTE estimate to be associated with each temporal expression sequence in a data-driven way.\footnote{We distinguish between generative and estimation rules. The former are used to generate temporal expressions for recognition purposes. The latter enable encoding temporal logic for estimating time to event.}

Typically, rule-based systems are able to cover information provided by adverbs (`more' in `three more days') and relations between non-adjacent elements and encode temporal logic, while machine-learning-based systems do not make use of the temporal logic inherent to temporal expressions; they may identify `three more days' as a temporal expression but they lack the logical apparatus to compute that this implies a TTE of about $3 \times 24$ hours.\footnote{Although machine learning techniques have the potential to automatically learn approximating temporal logic, obtaining training data that can facilitate this learning process at an accuracy that can be encoded with estimation rules is challenging.} 
To make use of the best of both worlds we propose a hybrid system which uses information about the distribution of temporal expressions as they are used in forward-looking social media messages in a training set of known events, and combines this estimation method with an extensive set of linguistically-motivated patterns that capture a large space of possible Dutch temporal expressions.


For our experiment we used the `FM without retweets' set that was described in Section~\ref{ch2DataCollection}. This type of event generally triggers many anticipatory discussions on social media containing many temporal expressions. Given a held-out football match not used during training, our system predicts the time to the event based on individual tweets captured in a range from eight days before the event to the event time itself. Each estimation is based on the temporal expression(s) in a particular twitter message. The mean absolute error of the predictions for each of the 60 football matches in our data set is off by about eight hours. The results are generated in a leave-one-out cross-validation setup.\footnote{Tweet IDs, per tweet estimations, observed temporal expressions and estimation rules are available from \url{http://www.ru.nl/lst/resources/}.}

\begin{sloppypar}
This study starts with describing the overall experimental set up in Section~\ref{experimentsEACL}, including the temporal expressions that were used, our two baselines, and the evaluation method used. Next, in Section~\ref{resultsEACL} the results are presented. The results are analyzed and discussed in Section ~\ref{discussionEACL}. We conclude with a summary of our main findings and point to the following steps that are implemented based on the results of this study in Section~\ref{conclusionsEACL}.
\end{sloppypar}

\subsection{Experimental Set-Up}
\label{experimentsEACL}

We carried out a controlled case study in which we focused on Dutch premier league football matches as a type of scheduled event. As observed earlier, in Section~\ref{ch2DataCollection}, these types of matches have the advantage that they occur frequently, have a distinctive hashtag by convention, and often generate thousands to several tens of thousands of tweets per match. 

\begin{sloppypar}
Below we first describe the collection and composition of our data sets (Subsection~\ref{dataEACL}) and the temporal expressions which were used to base our predictions upon (Subsection~\ref{tempsEACL}). Then, in Subsection~\ref{baselineEACL}, we describe our baselines and evaluation method.
\end{sloppypar}

\subsubsection{Data Sets}
\label{dataEACL}


We use the `FM without retweets' data set in this study by making the assumption that the presence of a hashtag can be used as proxy for the topic addressed in a tweet. Observing that hashtags occur either as an element of the content inside or as a label at the end of a tweet text, we developed the hypothesis that the position of the hashtag may have an effect as regards the topicality of the tweet. Hashtags that occur in final position (i.e. they are tweet-final or are only followed by one or more other hashtags) are typically metatags and therefore possibly more reliable as topic identifiers than tweet non-final hashtags which behave more like common content words in context. In order to be able to investigate the possible effect that the position of the hashtag might have, we split our data in the following two subsets:

\begin{description}
\item[FIN] -- comprising tweets in which the hashtag occurs in final position (as defined above); 84,533 tweets.
\item[NFI] -- comprising tweets in which the hashtag occurs in non-final position; 53,608 tweets.
\end{description}

Each tweet in our data set has a time stamp of the moment (hour-minute-second) it was posted. Moreover, for each football match we know exactly when it took place. This information is used to calculate for each tweet the actual time that remains to the start of the event and the absolute error in estimating the time to event. 

\subsubsection{Temporal Expressions}
\label{tempsEACL}
In the context of this study temporal expressions are considered to be words or phrases which point to the point in time, the duration, or the frequency of an event. These may be exact, approximate, or even right out vague. Although in our current experiment we restrict ourselves to an eight-day period prior to an event, we chose to create a gross list of all possible temporal expressions we could think of, so that we would not run the risk of overlooking any items and the list can be used on future occasions even when the experimental setting is different. Thus the list also includes temporal expressions that refer to points in time outside the time span under investigation here, such as {\em gisteren} `yesterday' or {\em over een maand} `in a month from now', and items indicating duration or frequency such as {\em steeds} `continuously'/`time and again'. No attempt has been made to distinguish between items as regards time reference (future time, past time) as many items can be used in both fashions (compare for example {\em vanmiddag} in {\em vanmiddag ga ik naar de wedstrijd} `this afternoon I'm going to the match' vs {\em ik ben vanmiddag naar de wedstrijd geweest} `I went to the match this afternoon'. 

The list is quite comprehensive. Among the items included are single words, e.g. adverbs such as {\em nu} `now', {\em zometeen} `immediately', {\em straks} `later on', {\em vanavond} `this evening', nouns such as {\em zondagmiddag} `Sunday afternoon', and conjunctions such as {\em voordat} `before'), but also word combinations and phrases such as {\em komende woensdag} `next Wednesday’. Temporal expressions of the latter type were obtained by means of a set of 615 
lexical items and 70 rules, which generated a total of around 53,000 temporal expressions. Notwithstanding the impressive number of items included, the list is bound to be incomplete.\footnote{Not all temporal expressions generated by the rules will prove to be correct. Since incorrect items are unlikely to occur and therefore are considered to be harmless, we refrained from manually checking the resulting set.} In addition, there are patterns that match a couple of hundred thousand temporal expressions relating the number of minutes, hours, days, or time of day;\footnote{For examples see Table~\ref{tab:timexs} and Section~\ref{baselineEACL}.} they include items containing up to 9 words in a single temporal expression.

We included prepositional phrases rather than single prepositions so as to avoid generating too much noise. Many prepositions have several uses: they can be used to express time, but also for example location. Compare {\em voor} in {\em voor drie uur} `before three o'clock' and {\em voor het stadion} `in front of the stadium'. Moreover, prepositions are easily confused with parts of separable verbs which in Dutch are abundant. 

Various items on the list are inherently ambiguous and only in one of their senses can be considered temporal expressions. Examples are {\em week} `week' but also `weak' and {\em dag} `day' but also `goodbye'. For items like these, we found that the different senses could fairly easily be distinguished whenever the item was immediately preceded by an adjective such as {\em komende} and {\em volgende} (both meaning `next'). For a few highly frequent items this proved impossible. These are words like {\em zo} which can be either a temporal adverb (`in a minute'; cf. {\em zometeen}) or an intensifying adverb (`so'), {\em dan} `then' or `than', and {\em nog} `yet' or `another'. As we have presently no way of distinguishing between the different senses and these items have at best an extremely vague temporal sense so that they cannot be expected to contribute to estimating the time to event, we decided to discard these.\footnote{Note that {\em nog} does occur on the list as part of various multiword temporal expressions. Examples are {\em nog twee dagen} `another two days' and {\em nog 10 min} `10 more minutes'.}

In order to capture event targeted expressions, we treated domain terms such as {\em wedstrijd} `football match' as parts of temporal expressions in case they co-occur with a temporal expression, e.g. {\em morgen wedstrijd} `tomorrow football match'.

For the items on the list no provisions were made for handling any kind of spelling variation, with the single exception of a small group of words (including {\em 's morgens} `in the morning', {\em 's middags} `in the afternoon' and {\em 's avonds} `in the evening') which use in their standard spelling the archaic {\em 's}, and abbreviations. As many authors of tweets tend to spell these words as {\em smorgens}, {\em smiddags} and {\em savonds} we decided to include these forms as well.

The items on the list that were obtained through rule-based generation include temporal expressions such as {\em over 3 dagen} `in 3 days', {\em nog 5 minuten} `another 5 minutes', but also fixed temporal expressions such as clock times.\footnote{Dates are presently not covered by our patterns but will be added in the following experiments.} The patterns handle frequently observed variations in their notation, for example {\em drie uur} `three o'clock' may be written in full or as {\em 3:00}, {\em 3:00 uur}, {\em 3 u}, {\em 15.00}, etc.

Table~\ref{tab:timexs} shows example temporal expression estimates and applicable estimation rules. The median estimations are mostly lower than the mean estimations. The distribution of the time to event (TTE) for a single temporal expression often appears to be skewed towards lower values. The final column of the table displays the applicable estimation rules. The first six estimation rules subtract the time the tweet was posted (TT) from an average marker point such as `today 20.00' (i.e. 8 pm) for {\em vanavond} `tonight'. The second and third estimation rules from below state a TTE directly, {\em over 2 uur} `in 2 hours' is directly translated to a TTE of 2.

\begin{sidewaystable}
\begin{center}
\begin{tabular}{ll|rrr}
Temporal Expression & Gloss & Mean TTE & Median TTE & Rule \\
 \hline 
 vandaag & {\em today} & 5.63  &  3.09 & today 15:00 - TT \\
 vanavond & {\em tonight} & 8.40  & 4.78  & today 20:00 - TT \\
 morgen & {\em tomorrow} & 20.35  & 18.54  & tomorrow 15:00 - TT \\
 zondag & {\em Sunday} & 72.99 & 67.85  & Sunday 15:00 - TT \\
 vandaag 12.30 & {\em today 12.20} & 2.90  & 2.75 & today 12:30 - TT \\
 om 16.30 & {\em at 16.30} & 1.28 & 1.36 & today 16:30 - TT \\
 over 2 uur & {\em in 2 hours} & 6.78 & 1.97 & 2 h \\
 nog minder dan 1 u & {\em within 1 h} & 21.43 & 0.88 & 1 h \\
 in het weekend & {\em during the weekend} & 90.58 & 91.70 & {\em No Rule} \\
\end{tabular}
\caption{Examples of temporal expressions and their mean and median TTE estimation from training data. The final column lists the applicable estimation rule, if any. Estimation rules make use of the time of posting (Tweet Time, TT).}
\label{tab:timexs}
\end{center}
\end{sidewaystable}

\subsubsection{Evaluation and Baselines}
\label{baselineEACL}

Our approach to TTE estimation makes use of all temporal expressions in our temporal expression list that are found to occur in the tweets. A match may be for a single item in the list (e.g. {\em zondag} `Sunday') or any combination of items (e.g. {\em zondagmiddag}, {\em om 14.30 uur}, `Sunday afternoon', `at 2.30 pm'). There can be other words in between these expressions. We consider the longest match, from left to right, in case we encounter any overlap. 

The experiment adopts a leave-one-out cross-validation setup. Each iteration uses all tweets from 59 events as training data. All tweets from the single held-out event are used as test set. 

In the FIN data set there are 42,396 tweets with at least one temporal expression, in the NFI data set this is the case for 27,610 tweets. The number of tweets per event ranges from 66 to 7,152 (median: 402.5; mean 706.6) for the FIN data set and from 41 to 3,936 (median 258; mean 460.1) for the NFI data set. 

We calculate the TTE estimations for every tweet that contains at least one of the temporal expressions or a combination of these in the test set. The estimations for the test set are obtained as follows:

\begin{enumerate}
\setlength{\itemsep}{0mm}
\item For each match (a single temporal expression or a combination of temporal expressions) the mean or median value for TTE is used that was learned from the training set;

\item Temporal expressions that denote an exact amount of time are interpreted by means of estimation rules that we henceforth refer to as {\bf Exact rules}. This applies for example to temporal expressions answering to patterns such as {\em over N \{minuut $\mid$ minuten $\mid$ kwartier $\mid$ uur $\mid$ uren $\mid$ dag $\mid$ dagen $\mid$ week\}} `in N \{minute $\mid$ minutes $\mid$ quarter of an hour $\mid$ hour $\mid$ hours $\mid$ day $\mid$ days $\mid$ week\}'. Here the TTE is assumed to be the same as the N minutes, days or whatever is mentioned. The exact rules take precedence over the mean estimates learned from the training set;

\item A second set of estimation rules, referred to as the {\bf Dynamic rules}, is used to calculate the TTE dynamically, using the temporal expression and the tweet's time stamp. These estimation rules apply to instances such as {\em zondagmiddag om 3 uur} `Sunday afternoon at 3 p.m.'. Here we assume that this is a future time reference on the basis of the fact that the tweets were posted prior to the event. With temporal expressions that are underspecified in that they do not provide a specific point in time (hour), we postulate a particular time of day. For example, {\em vandaag} `today' is understood as `today at 3 p.m., {\em vanavond} `this evening' as 'this evening at 8 p.m. and {\em morgenochtend} `tomorrow morning' as `tomorrow morning at 10 a.m.'. Again, as was the case with the exact rules, these dynamic rules take precedence over the mean or median estimates learned from the training data.
\end{enumerate}

The results for the estimated TTE are evaluated in terms of the absolute error, i.e. the absolute difference in hours between the estimated TTE and the actual remaining time to the event.

We established two naive baselines: the mean and median TTE measured over all tweets of the FIN and NFI datasets. These baselines reflect a best guess when no information is available other than the tweet count and TTE of each tweet. The use of FIN and NFI for calculation of the baselines yields mean and median TTE as 22.82 and 3.63 hours before an event respectively. The low values of the baselines, especially the low median, reveal the skewedness of the data: most tweets referring to a football event are posted in the hours before the event.

\subsection{Results}
\label{resultsEACL}

Table~\ref{tab:main_resultsEACL} lists the overall mean absolute error (in number of hours) for the different variants. The results are reported separately for each of the two data sets (FIN and NFI) and for both sets aggregated (FIN+NFI).\footnote{Tweets that contain at least one temporal expression in FIN+NFI were used. The actual number of tweets that fall in this scope were provided in the {\em coverage} row. The coverage drops in relation to actual number of tweets that contain at least one temporal expression, since we did not assign a time-to-event value to basic temporal expressions that occur only once.} For each of these three variants, the table lists the mean absolute error when only the basic data-driven TTE estimations are used (`Basic'), when the Exact rules are added (`+Ex.'), when the Dynamic rules are added (`+Dyn'), and when both types of rules are added. The coverage of the combination (i.e. the number of tweets that match the expressions and the estimation rules) is listed in the bottom row of the table. 

A number of observations can be made. First, all training methods perform substantially better than the two baselines in all conditions. Second, the TTE training method using the median as estimation produces estimations that are about 1 hour more accurate than the mean-based estimations. Third, adding Dynamic rules has a larger positive effect on prediction error than adding Exact rules. 

\begin{sidewaystable}
\begin{center}
\begin{small}
\setlength{\tabcolsep}{1.2mm}
\renewcommand{\arraystretch}{1.3}
\begin{tabular}{l|rrrr|rrrr|rrrr}
System  & \multicolumn{4}{c|}{FIN} & \multicolumn{4}{c|}{NFI} & \multicolumn{4}{c}{FIN+NFI} \\ 
 & Basic & +Ex. & +Dyn. & +Both & Basic & +Ex. & +Dyn. & +Both & Basic & +Ex. & +Dyn. & +Both \\ 
 \hline 
Baseline Median & 21.09 & 21.07 & 21.16 & 21.14 & 18.67 & 18.72 & 18.79 & 18.84 & 20.20 & 20.20 & 20.27 & 20.27 \\ 
Baseline Mean   & 27.29 & 27.29 & 27.31 & 27.31 & 25.49 & 25.50 & 25.53 & 25.55 & 26.61 & 26.60 & 26.63 & 26.62 \\ 
\hline
Training Median & 10.38 & 10.28 & 7.68 & 7.62 & 11.09 & 11.04 & 8.65 & 8.50 & 10.61 & 10.54 & 8.03 & 7.99 \\ 
Training Mean   & 11.62 & 11.12 & 8.73 & 8.29 & 12.43 & 11.99 & 9.53 & 9.16 & 11.95 & 11.50 & 9.16 & 8.76 \\ 
\hline
Coverage     & 31,221 & 31,723 & 32.240 & 32,740 & 18,848 & 19,176 & 19,734 & 20,061 & 52,186 & 52,919 & 53,887 & 54,617 \\
\end{tabular}
\caption{Overall Mean Absolute Error for each method: difference in hours between the estimated time to event and the actual time to event, computed separately for the FIN and NFI subsets, and for the combination. For all variants a count of the number of matches is listed in the bottom row.}
\label{tab:main_resultsEACL}
\end{small}
\end{center}
\end{sidewaystable}

The bottom row in the table indicates that the estimation rules do not increase the coverage of the method substantially. When taken together and added to the basic TTE estimation, the Dynamic and Exact rules do improve over the Basic estimation by two to three hours.

Finally, although the differences are small, Table~\ref{tab:main_resultsEACL} reveals that training on hashtag-final tweets (FIN) produces slightly better overall results (7.62 hours off at best) than training on hashtag-non-final tweets (8.50 hours off) or the combination (7.99 hours off), despite the fact that the training set is smaller than that of the combination.

In the remainder of this section we report on systems that use all expressions and Exact and Dynamic rules.

\begin{figure}[htb]
\begin{center}
\includegraphics[scale=0.75]{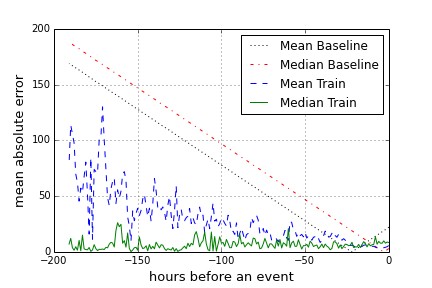}
\caption{Curves showing the absolute error (in hours) in estimating the time to event over an 8-day period (-192 to 0 hours) prior to the event. The two baselines are compared to the TTE estimation methods using the mean and median variant.}
\label{fig:meanMedianComparison}
\end{center}
\end{figure}


Whereas Table~\ref{tab:main_resultsEACL} displays the overall mean absolute errors of the different variants, Figure~\ref{fig:meanMedianComparison} displays the results in terms of mean absolute error at different points in time before the event, averaged over periods of one hour, for the two baselines and the FIN+NFI variant with the two training methods (i.e. taking the mean versus the median of the observed TTEs for a particular temporal expression). In contrast to Table~\ref{tab:main_resultsEACL}, in which only a mild difference could be observed between the median and mean variants of training, the figure shows a substantial difference. The estimations of the median training variant are considerably more accurate than the mean variant up to 24 hours before the event, after which the mean variant scores better. By virtue of the fact that the data is skewed (most tweets are posted within a few hours before the event) the two methods attain a similar overall mean absolute error, but it is clear that the median variant produces considerably more accurate predictions when the event is still more than a day away.

While Figure~\ref{fig:meanMedianComparison} provides insight into the effect of median versus mean-based training with the combined FIN+NFI dataset, we do not know whether training on either of the two subsets is advantageous at different points in time. Table~\ref{tab:ranges} shows the mean absolute error of systems trained with the median variant on the two subsets of tweets, FIN and NFI, as well as the combination FIN+NFI, split into nine time ranges. Interestingly, the combination does not produce the lowest errors close to the event. However, when the event is 24 hours away or more, both the FIN and NFI systems generate increasingly larger errors, while the FIN+NFI system continues to make quite accurate predictions, remaining under 10 hours off even for the longest TTEs, confirming what we already observed in Figure~\ref{fig:meanMedianComparison}.

\begin{table}[htb]
\begin{center}
\begin{tabular}{l|rrr}
TTE range (h) & FIN & NFI & FIN+NFI \\
 \hline 
      0 &  {\bf 2.58} &  3.07 & 8.51 \\
   1--4 &  {\bf 2.38} &  2.64 & 8.71 \\
   5--8 &  {\bf 3.02} &  3.08 & 8.94 \\
  9--12 &  {\bf 5.20} &  5.47 & 6.57 \\
 13--24 &  5.63 &  {\bf 5.54} & 6.09 \\
 25--48 & 13.14 & 15.59 & {\bf 5.81} \\
 49--96 & 17.20 & 20.72 & {\bf 6.93} \\
97--144 & 30.38 & 41.18 & {\bf 6.97} \\
$> 144$ & 55.45 & 70.08 & {\bf 9.41} \\
\end{tabular}
\caption{Mean Absolute Error for the FIN, NFI, and FIN+NFI systems in different TTE ranges.}
\label{tab:ranges}
\end{center}
\end{table}

\subsection{Analysis}
\label{discussionEACL}

One of the results observed in Table~\ref{tab:main_resultsEACL} was the relatively limited role of Exact rules, which were intended to deal with exact temporal expressions such as {\em nog 5 minuten} `5 more minutes' and {\em over een uur} `in one hour'. This can be explained by the fact that as long as the temporal expression is related to the event we are targeting, the point in time is denoted exactly by the temporal expression and the estimation obtained from the training data (the `Basic' performance) will already be accurate, leaving no room for the estimation rules to improve on this. The estimation rules that deal with dynamic temporal expressions, on the other hand, have quite some impact. 

\begin{sloppypar}
As explained in Section~\ref{tempsEACL}, our list of temporal expressions was a gross list, including items that were unlikely to occur in our present data. In all we observed 770 of the 53,000 items listed, 955 clock time pattern matches, and 764 patterns which contain number of days, hours, minutes etc. The temporal expressions observed most frequently in our data are:\footnote{The observed frequencies can be found between brackets.} {\em vandaag} `today' (10,037), {\em zondag} `Sunday' (6,840), {\em vanavond} `tonight' (5167), {\em straks} `later on' (5,108), {\em vanmiddag} `this afternoon' (4,331), {\em matchday} `match day' (2,803), {\em volgende week} `next week' (1,480) and {\em zometeen} `in a minute' (1,405).
\end{sloppypar}

Given the skewed distribution of tweets over the eight days prior to the event, it is not surprising to find that nearly all of the most frequent items refer to points in time within close range of the event. Apart from {\em nu} `now', all of these are somewhat vague about the exact point in time. There are, however, numerous items such as {\em om 12:30 uur} `at half past one' and {\em over ongeveer 45 minuten} `in about 45 minutes', which are very specific and therefore tend to appear with middle to low frequencies.\footnote{While an expression such as {\em om 12:30 uur} has a frequency of 116, {\em nog maar 8 uur en 35 minuten} `only 8 hours and 35 minutes from now' has a frequency of 1.} And while it is possible to state an exact point in time even when the event is in the more distant future, we find that there is a clear tendency to use underspecified temporal expressions as the event is still some time away. Thus, rather than {\em volgende week zondag om 14.30 uur} `next week Sunday at 2.30 p.m.' just {\em volgende week} is used, which makes it harder to estimate the time to event. 

Closer inspection of some of the temporal expressions which yielded large absolute errors suggests that these may be items that refer to subevents rather than the main event (i.e. the match) we are targeting. Examples are {\em eerst} `first', {\em daarna} `then', {\em vervolgens} `next', and {\em voordat} `before'.

\subsection{Conclusion}
\label{conclusionsEACL}

We have presented a method for the estimation of the TTE from single tweets referring to a future event. In a case study with Dutch football matches, we showed that estimations can be as accurate as about eight hours off, averaged over a time window of eight days. There is some variance in the 60 events on which we tested in a leave-one-out validation setup: errors ranged between 4 and 13 hours, plus one exceptionally badly predicted event with a 34-hour error.~\footnote{This is the case where two matches between the same two teams were played a week apart, i.e. premier league and cup match.}

The best system is able to stay within 10 hours of prediction error in the full eight-day window. This best system uses a large hand-designed set of temporal expressions that in a training phase have each been linked to a median TTE with which they occur in a training set.\footnote{The large set of hand-designed set of temporal expressions was required to ensure the approach we have developed is not significantly affected by any missing temporal expressions. Having showed that our approach is effective on this task, implementations of the method in the future can be based on the most frequent temporal expressions or be based on an available temporal tagger~\cite{Chang+12}} Together with these data-driven TTE estimates, the system uses a set of estimation rules that match on exact and indirect time references. In a comparative experiment we showed that this combination worked better than only having the data-driven estimations.

We then tested whether it was more profitable to train on tweets that had the event hashtag at the end, as this is presumed to be more likely a meta-tag, and thus a more reliable clue that the tweet is about the event than when the hashtag is not in final position. Indeed we find that the overall predictions are more accurate, but only in the final hours before the event (when most tweets are posted). 24 hours and earlier before the event it turns out to be better to train both on hashtag-final and hashtag-non-final tweets.

Finally, we observed that the two variants of our method of estimating TTEs for single temporal expressions, taking the mean or the median, leads to dramatically different results, especially when the event is still a few days away---when an accurate time to event is actually desirable. The median-based estimations, which are generally smaller than the mean-based estimations, lead to a system that largely stays under 10 hours of error.





Our study has a number of logical extensions that are implemented in the following section. First, our method is not bound to a single type of event, although we tested it in a controlled setting. With experiments on tweet streams related to different types of events the general applicability of the method could be tested: can we use the trained TTE estimations from our current study, or would we need to retrain per event type? 

Moreover, our method is limited to temporal expressions. For estimating the time to event on the basis of tweets that do not contain temporal expressions, we could benefit from term-based approaches that consider any word or word $n$-gram as potentially predictive \cite{Hurriyetoglu+13}. 

Finally, each tweet is evaluated individually in generating an estimate. However, estimates can be combined iteratively in order to have a final estimate that integrates estimations from previously posted tweets. 

The following section reports details of a study that extends the study reported in this section to include cross-domain evaluation, add word-based features, and integrate historical information, which is a window of previously posted tweets, in the estimate generation process.

\section{Estimating Time to Event based on Linguistic Cues on Twitter}
\label{NLPTchapter}


\begin{sloppypar}
{\bf Based on:} Hürriyetoğlu, A., Oostdijk, N., \& van den Bosch, A (2018). Estimating Time to Event based on Linguistic Cues on Twitter. In {\em K. Shaalan, A. E. Hassanien, \& F. Tolba (Eds.), Intelligent Natural Language Processing: Trends and Applications (Vol. 740)}. Springer International Publishing. Available from http://www.springer.com/cn/book/9783319670553
\end{sloppypar}


Given a stream of Twitter messages about an event, we investigate the predictive power of features generated from words and temporal expressions in the messages to estimate the time to event (TTE). From labeled training data average TTE values of the predictive features are learned, so that when they occur in an event-related tweet the TTE estimate can be provided for that tweet. We utilize temporal logic rules for estimation and a historical context integration function to improve the TTE estimation precision. In experiments on football matches and music concerts we show that the estimates of the method are off by four and ten hours in terms of mean absolute error on average, respectively. We find that the type and size of the event affect the estimation quality. An out-of-domain test on music concerts shows that models and hyperparameters trained and optimized on football matches can be used to estimate the remaining time to concerts. Moreover, mixing in concert events in training improves the precision of the average football event estimate.

\subsection{Introduction}

We extend the time-to-event estimation method that was reported in the previous section and implement it as an expert system that can process a stream of tweets in order to provide an estimate about the starting time of an event. Our ultimate goal is to provide an estimate for any type of event: football matches, music concerts, labour strikes, floods, etcetera. The reader should consider our study as a first implementation of a general time-to-event estimation framework. We focus on just two types of events, football matches and music concerts, as it is relatively straightforward to collect gold-standard event dates and times from databases for these types of events. We would like to stress, however, that the method is not restricted in any way to these two types; ultimately it should be applicable to any type of event, also event types for which no generic database of event dates and times is available.

In this study we explore a hybrid rule-based and data-driven method that exploits the explicit mentioning of temporal expressions but also other lexical phrases that implicitly encode time-to-event information, to arrive at accurate and early TTE estimates based on the Twitter stream. At this stage, our focus is only on estimating the time remaining to a given event. Deferring the fully automatic detection of events to future work, in our current study we identify events in Twitter microposts by event-specific hashtags. 


We aim to develop a method that can estimate the time to any event and that will assist users in discovering upcoming events in a period they are interested in. We automate the TTE estimation part of this task in a flexible manner so that we can generate continuous estimates based on realistic quantities of streaming data over time. Our estimation method offers a solution for the representation of vague terms, by inducing a continuous value for each possible predictive term from the training data.

The service offered by our method will be useful only if it generates accurate estimates of the time to event. Preferably, these accurate predictions should come as early as possible. We use a combination of rule-based, temporal, and lexical features to create a flexible setting that can be applied to events that may not contain all these types of information.

In the present study we use the same estimation method as ~\citeauthor{Hurriyetoglu+14}\citeyear{Hurriyetoglu+14}. We scale up the number of temporal expressions drastically compared to what the Heideltime tagger offers and other time-to-event estimation methods have used \cite{Ritter+12,Hurriyetoglu+13}. We also compare this approach to using any other lexical words or word skipgrams in order to give the best possible estimate of the remaining time to an event. Moreover, we implement a flexible feature generation and selection method, and a history function which uses the previous estimates as a context. In our evaluation we also look into the effects of cross-domain parameter and model transfer.

The remainder of this section is structured as follows. 
We start by introducing our time-to-event estimation method in Section~\ref{ttemethodNLPT}. Next, in Section~\ref{experimentsetupNLPT}, we explain 
the feature sets we designed, the training and test principles, the hyper-parameter optimization, and the evaluation method. We then, in Section~\ref{testresultsNLPT}, describe the results for football matches and music concerts by measuring the effect of cross-domain parameter and model transfer on the estimation quality. In Section~\ref{analysisNLPT} we analyze the results and summarize the insights obtained as regards various aspects of the TTE estimation method. Finally, Section~\ref{conclusionsNLPT} concludes this study with a summary of the main findings for each data set.

\subsection{Time-to-Event Estimation Method}
\label{ttemethodNLPT}

Our time-to-event (TTE) estimation method consists of training and estimation steps. First, training data is used to identify the predictive features and their values. Then, each tweet is assigned a TTE value based on the predictive features it contains and on the estimates from recent tweets for the same event stored in a fixed time buffer.

The training phase starts with feature extraction, generation, and selection. During feature extraction we first extract tokens as unigrams and bigrams and identify which of these are temporal expressions or occur in our lexicons. Then, we select the most informative features based on frequency and standard deviation of their occurrence using their temporal distribution from the final set of features. 

The estimation phase consists of two steps. First, an estimation function assigns a TTE estimate for a tweet based on the predictive features that occur in this tweet. Afterwards the estimate is adjusted by a historical function based on a buffer of previous estimations. Historical adjustment restricts consecutive estimates in the extent to which they deviate from each other.

As we aim to develop a method that can be applied to any type of event and any language, our approach is guided by the following considerations:

\begin{sloppypar}
\begin{enumerate}
\item We refrain from using any domain-specific prior knowledge such as `football matches occur in the afternoon or in the evening' which would hinder the prediction of matches held at unusual times, or more generally the application of trained models to different domains;


\item We use language analysis tools sparingly and provide the option to use only words so that the method can be easily adapted to languages for which detailed language analysis tools are not available, less well developed, or less suited for social media text;

\item We use basic statistics (e.g. computing the median) and straightforward time series analysis methods to keep the method efficient and scalable;

\item We use the full resolution of time to provide precise estimates. Our approach treats TTE as a continuous value: calculations are made and the results are reported using decimal fractions of an hour.
\end{enumerate}
\end{sloppypar}

\subsubsection{Features}
\label{features}

It is obvious to the human beholder that tweets referring to an event exhibit different patterns as the event draws closer. Not only do the temporal expressions that are used change (e.g. from `next Sunday' to `tomorrow afternoon'), the level of excitement rises as well, as the following examples show:

\begin{description}
\item[{122 hours before}:] {\em bj\"{o}rn kuipers is door de knvb aangesteld als scheidsrechter voor de wedstrijd fc utrecht psv van komende zondag 14.30 uur} (En: Bj\"{o}rn Kuipers has been assigned to be the referee for Sunday's fixture between FC Utrecht and PSV, played at 2.30 PM)
\item[{69 hours before}:] {\em zondag thuiswedstrijd nummer drie fc utrecht psv kijk ernaar uit voorbereidingen in volle gang} (En: Sunday the 3rd match of the season in our stadium FC Utrecht vs. PSV, excited, preparations in full swing)
\item[{27 hours before}:] {\em dick advocaat kan morgenmiddag tegen fc utrecht beschikken over een volledig fitte selectie} (En: Dick Advocaat has a fully healthy selection at his disposal tomorrow afternoon against FC Utrecht)
\item[{8 hours before}:] {\em werken hopen dat ik om 2 uur klaar ben want t is weer matchday} (En: working, hope I'm done by 2 PM, because it's matchday again)

\item[{3 hours before}:] {\em onderweg naar de galgenwaard voor de wedstrijd fc utrecht feyenoord \#utrfey} (En: on my way to Galgenwaard stadium for the football match fc utrecht feyenoord \#utrfey)

\item[{1 hour before}:] {\em wij zitten er klaar voor \#ajafey op de beamer at \#loods} (En: we are ready for \#ajafey by the data projector at \#loods) 

\item[{0 hours before}:] {\em rt @username zenuwen beginnen toch enorm toe te nemen} (En: RT @username starting to get really nervous)
\end{description}

The temporal order of different types of preparations for the event can be seen clearly in the tweets above. The stream starts with a referee assignment and continues with people expressing their excitement and planning to go to the event, until finally the event starts. Our goal is to learn to estimate the TTE from texts like these, that is, from tweets, along with their time stamps, by using linguistic clues that, explicitly or implicitly, refer to the time when an event is to take place. 

Below we introduce the three types of features that we use to estimate the TTE: temporal expressions, estimation rules, and word skipgrams. We extended the temporal expressions and estimation rules created in Section~\ref{EMNLPchapter} for this study. We provide
their characteristics and formal description in detail in the following subsection.

\subsubsection*{Temporal Expressions}



A sample of patterns that are used to generate the temporal expressions is listed below. The first and second examples provide temporal expressions without numerals and the third one illustrates how the numbers are included. The square and round brackets denote obligatory and optional items, respectively. The vertical bar is used to separate alternative items. Examples of derived temporal expressions are presented between curly brackets.

\begin{sloppypar}
\begin{enumerate}

\item $[$ in $\mid$ m.i.v. $\mid$ miv $\mid$ met ingang van $\mid$ na $\mid$ t/m $\mid$ tegen $\mid$ tot en met $\mid$ van $\mid$ vanaf $\mid$ voor$]$ + (de maand) + $[$ jan $\mid$ feb $\mid$ mrt $\mid$ apr $\mid$ mei $\mid$ jun $\mid$ jul $\mid$ aug $\mid$ sep $\mid$ okt $\mid$ nov $\mid$ dec $]$ $\rightarrow$ \{{\em in de maand Jan} `in the month of January', {\em met ingang van jul} `as of July' \}













\item een + $[$ minuut $\mid$ week $\mid$ maand $\mid$ jaar $]$ + (of wat) + $[$eerder $\mid$ later$]$ $\rightarrow$ \{{\em een minuut eerder} `one minute earlier', {\em een week later} `one week later' \}

\item N$>$1 + $[$minuten $\mid$ seconden $\mid$ uren $\mid$ weken $\mid$ maanden $\mid$ jaren $\mid$ eeuwen $]$ + [eerder $\mid$ eraan voorafgaand $\mid$ ervoor $\mid$ erna $\mid$ geleden $\mid$ voorafgaand $\mid$ later] $\rightarrow$ \{{\em 7 minuten erna} `seven minutes later', {\em 2 weken later} `after two weeks' \}

\end{enumerate}

\end{sloppypar}

The third example introduces N, a numeral that in this case varies between 2 and the maximal value of numerals in all expressions, 120.\footnote{The value of N was selected to cover the temporal expressions that fall into the time-to-event period we focus on.} The reason that $N>1$ in this example is that it generates expressions with plural forms such as `minutes', `seconds', etc.

Including numerals in the generation of patterns yields a total of 460,248 expressions expressing a specific number of minutes, hours, days, or time of day. 

Finally, combinations of a time, weekday or day of the month, e.g., `Monday 21:00', `Tuesday 18 September', and `1 apr 12:00' are included as well. Notwithstanding the substantial number of items included, the list is bound to be incomplete.\footnote{Dates, which denote the complete year, month and day of the month, are presently not covered by our patterns but will be added in future.}

Despite the large number of pre-generated temporal expressions, the number of expressions actually encountered in the FM data set is only 2,476; in the smaller MC set we find even fewer temporal expressions, viz. 398. 

This set also contains skipgrams, i.e. feature sequences that are created via concatenation of neighboring temporal expressions, while ignoring in-between tokens. Consider, for instance, the tweet {\em Volgende wedstrijd: Tegenstander: Palermo Datum: zondag 2 november Tijdstip: 20:45 Stadion: San Siro} `Next match: Opponent: Palermo Date: Sunday 2 November Time: 20:45 Stadium: San Siro'. The basic temporal features in this tweet are, in their original order, $<$zondag 2 november$>$ and $<$20:45$>$. The skipgram generation step will ignore the in-between tokens $<$Time$>$ and $<$:$>$, preserve the feature occurrence order, and result in $<$zondag 2 november, 20:45$>$ as a new skipgram feature. From this point onward we will refer to the entire set of basic and skipgram temporal features as \textbf{TFeats}.

We compare the performance of our temporal expression detection based on the TFeats list with that of the Heideltime Tagger\footnote{We used the Heideltime tagger (version 1.7) by enabling the interval tagger and configured NEWS type as genre.}, which is the only available temporal tagger for Dutch, on a small set of tweets, i.e. 18,607 tweets from the FM data set.\footnote{This subset is used to optimize the hyper-parameters as well.} There are 10,183 tweets that contain at least one temporal expression detected by the Heideltime tagger or matched by our list. 5,008 temporal expressions are identified by both. In addition, the Heideltime tagger detects 429 expressions that our list does not; vice versa, our list detects 2,131 expressions that Heideltime does not detect. In the latter category are {\em straks} `soon', {\em vanmiddag} `today', {\em dalijk} `immediately' (colloquial form of {\em dadelijk}), {\em nog even} `a bit', {\em over een uurtje} `in 1 hour', {\em over een paar uurtjes} `in a few hours', {\em nog maar 1 nachtje} `only one more night'. On the other hand, the Heideltime tagger detects expressions such as {\em de afgelopen 22 jaar} `the last 22 years', {\em 2 keer per jaar} `2 times a year', {\em het jaar 2012} `the year 2012'. This can easily be explained as our list currently by design focuses on temporal expressions that refer to the short term future, and not to the past or the long term. Also, the Heideltime tagger recognizes some expressions that we rule out intentionally due to their ambiguous interpretation. For instance, this is the case for {\em jan} `Jan' (name of person or the month of January) and {\em volgend} `next'. In sum, as we are focusing on upcoming events and our list has a higher coverage than Heideltime, we continue working with our TFeats list.

\subsubsection*{Estimation Rules}
\label{featsRules}

When we only want to use absolute forward-pointing temporal expressions as features we do not need temporal logic to understand their time-to-event value. These expressions provide the time to event directly, e.g. `in 30 minutes' indicates that the event will take place in 0.5 hours. We therefore introduce estimation rules that make use of temporal logic to define the temporal meaning of TFeats features that have a context dependent time-to-event value. We refer to these features as non-absolute temporal expressions.

Non-absolute, dynamic TTE temporal expressions such as days of the week, and date-bearing temporal expressions such as {\em 18 September} can on the one hand be detected relatively easily, but on the other hand require further computation based on temporal logic using the time the tweet was posted. 
For example, the TTE value of a future weekday should be calculated according to the referred day and time of the occurrence. Therefore we use the estimation rules list from \citeauthor{Hurriyetoglu+14} \citeyear{Hurriyetoglu+14} and extend it. We define estimation rules against the background of: 

\begin{enumerate}
\setlength{\itemsep}{0mm}
\item Adjacency: We specify only contiguous relations between words, i.e. without allowing any other word to occur in between; 
\item Limited scope: An estimation rule can indicate a period up to 8 days before an event; thus we do not cover temporal expressions such as {\em nog een maandje} `another month' and {\em over 2 jaar} `in 2 years';
\item Precision: We refrain from including estimation rules for highly ambiguous and frequent terms such as {\em nu} `now' and {\em morgen} `tomorrow';
\item Formality: We restrict the estimation rules to canonical (normal) forms. Thus we do not include estimation rules for expressions like {\em over 10 min} `in 10 min' and {\em zondag 18 9} `Sunday 18 9';
\item Approximation: We round the estimations to fractions of an hour with maximally two decimals; the estimation rule states that {\em minder dan een halfuur} `less than half an hour' corresponds to 0.5 hour;
\item Partial rules: We do not aim to parse all possible temporal expressions. Although using complex estimation rules and language normalization can increase the coverage and performance, this approach has its limits and will decrease practicality of the method. Therefore, we define estimation rules up to a certain length, which may cause a long temporal expression to be detected partially. A complex estimation rule would in principle recognize the temporal expression ``next week Sunday 20:00'' as one unit. But we only implement basic estimation rules that will recognize ``next week'' and ``Sunday 20:00'' as two different temporal expressions. Our method will combine their standalone values and yield one value for the string as a whole.

\end{enumerate}

As a result we have two sets of estimation rules, which we henceforth refer as \textbf{RFeats}. RFeats consists of {\bf Exact} and {\bf Dynamic} rules that were defined in the Section~\ref{baselineEACL} and extended in the scope of this study.



\subsubsection*{Word Skipgrams}
\label{WFeats}

In contrast to temporal expressions we also generated an open-ended feature set that draws on any word skipgram occurring in our training set. This feature set is crucial in discovering predictive features not covered by the time-related TFeats or RFeats. A generic method may have the potential of discovering expressions already present in TFeats, but with this feature type we expressly aim to capture any lexical expressions that do not contain any explicit start time, yet are predictive of start times. Since this feature type requires only a word list, it can be smoothly adapted to any language.

We first compiled a list of regular, correctly spelled Dutch words by combining the {\em OpenTaal flexievormen and basis-gekeurd} word lists.\footnote{\label{opentaalurl}We used the {\em OpenTaal flexievormen, basis-gekeurd, and basis-ongekeurd word lists from the URL}: \url{https://www.opentaal.org/bestanden/file/2-woordenlijst-v-2-10g-bronbestanden}, accessed June 10, 2018} From this initial list we then removed all stop words, foreign words, and entity names. The stop word list contains 839 entries. These are numerals, prepositions, articles, discourse connectives, interjections, exclamations, single letters, auxiliary verbs and any abbreviations of these. Foreign words were removed in order to avoid spam and unrelated tweets. Thus, we removed 
English words which are in the {\em OpenTaal flexievormen} or {\em OpenTaal basis-gekeurd} word lists as we come across them.\footnote{These are: {\em different, indeed, am, ever, field, indeed, more, none, or, wants}.} We also used two lists to identify the named entities: {\em Geonames}\footnote{\url{http://www.geonames.org/}, accessed June 10, 2018} for place names in the Netherlands and {\em OpenTaal basis-ongekeurd} for other named entities. The final set of words comprises 317,831 entries. The `FM without retweets' and `MC without retweets' data sets contain 17,646 and 2,617 of these entries, respectively.

These lexical resources were used to control the number and complexity of the features. 
In principle, the word lists could also be extracted from the set of the tweets that are used as training set.

Next, the words were combined to generate longer skipgram features based on the words that were found to occur in a tweet. For example, given the tweet {\em goed weekendje voor psv hopen dat het volgend weekend hét weekend wordt \#ajapsv bye tukkers} `a good weekend for psv hoping that next weekend will be the weekend \#ajapsv bye tukkers' we obtained the following words in their original order: $<$goed$>$, $<$weekendje$>$, $<$hopen$>$, $<$volgend$>$, $<$weekend$>$, $<$weekend$>$, $<$tukkers$>$. From this list of words, we then generated skipgrams up to $n=7$.\footnote{The range 7 was selected in order to benefit from any long-distance relations between words. The limited word count in a tweet hardly allows to implement higher ranges.} Retaining the order, we generated all possible combinations of the selected words. The feature set arrived at by this feature generation approach is henceforth referred as \textbf{WFeats}. 

\subsubsection{Feature Selection}

Each tweet in our data set has a time stamp for the exact time it was posted. Moreover, for each event we know precisely when it took place. This information is used to calculate for each feature the series of all occurrence times relative to the event start time, hereafter referred as the \textbf{time series} of a feature. The training starts with the selection of features that carry some information regarding the remaining time to an event, based on their frequency and standard deviation of occurrence times relative to an event start time. A feature time series should be longer than one to be taken into account, and should have a standard deviation below a certain threshold for the feature to be considered for the feature value assignment phase. The standard deviation threshold is based on a fixed number of highest quantile regions, a number which is optimized on a development set. Features that are in the highest standard deviation quantile regions are eliminated. 

\subsubsection{Feature Value Assignment}

The value of a feature time series is estimated by a training function. In the current study, the training function is either the mean or the median of the actual TTE values of the selected features encountered in the training data. The proper training function is selected on the basis of its performance on a development set. This method does not need any kind of frequency normalization. We consider this to be an advantage, as now there is no need to take into account daily periodicity or tweet distribution.


\begin{figure*}[htb]
\begin{center}
\includegraphics[scale=0.90]{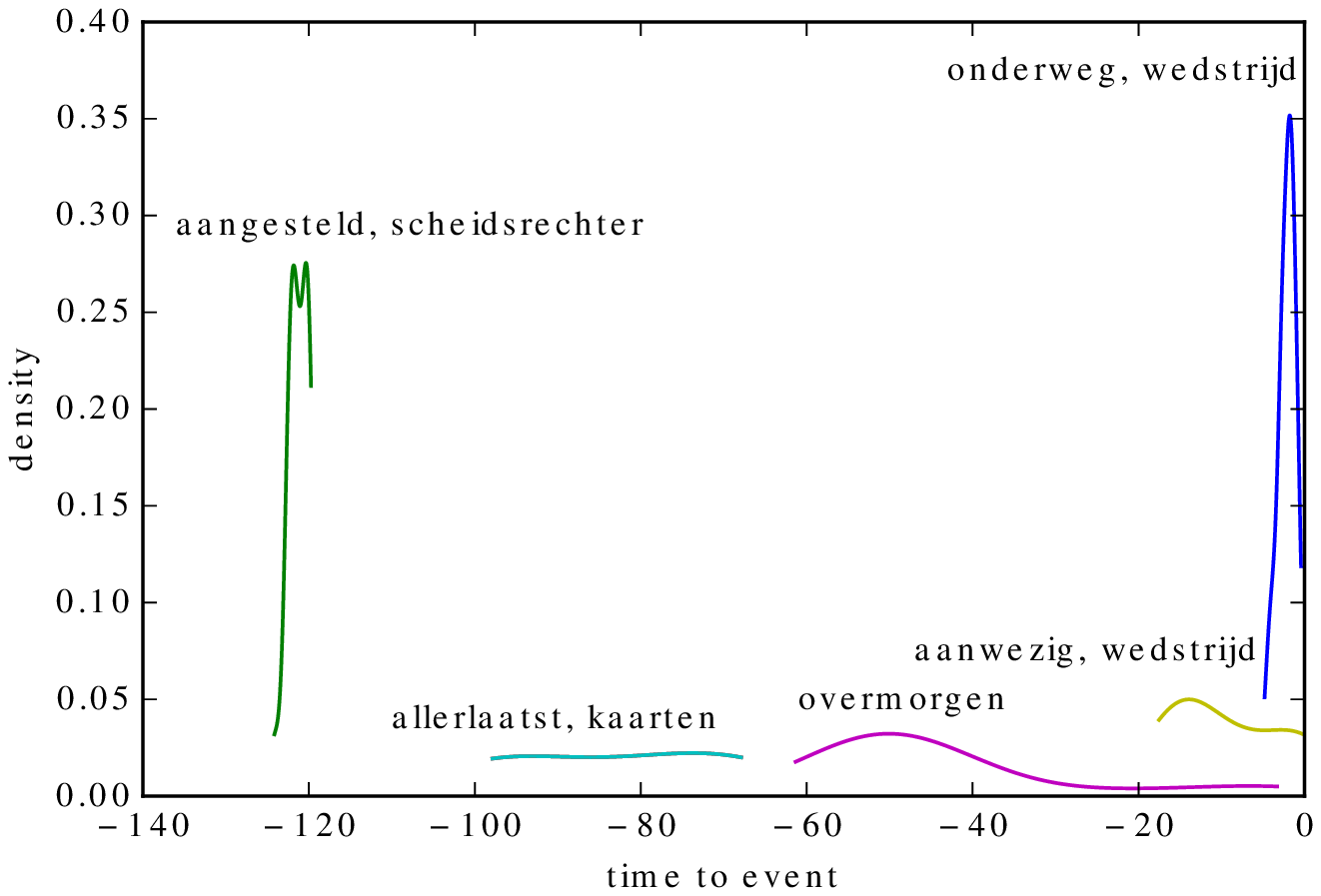}
\caption{Kernel density estimation using Gaussian kernels of some selected word skipgrams that show a predictive pattern about time of an event. They mostly indicate a phase of the event or preparations related to it.}
\label{fig:WFeatsTS}
\end{center}
\end{figure*}

\begin{figure*}[htb]
\begin{center}
\includegraphics[scale=0.90]{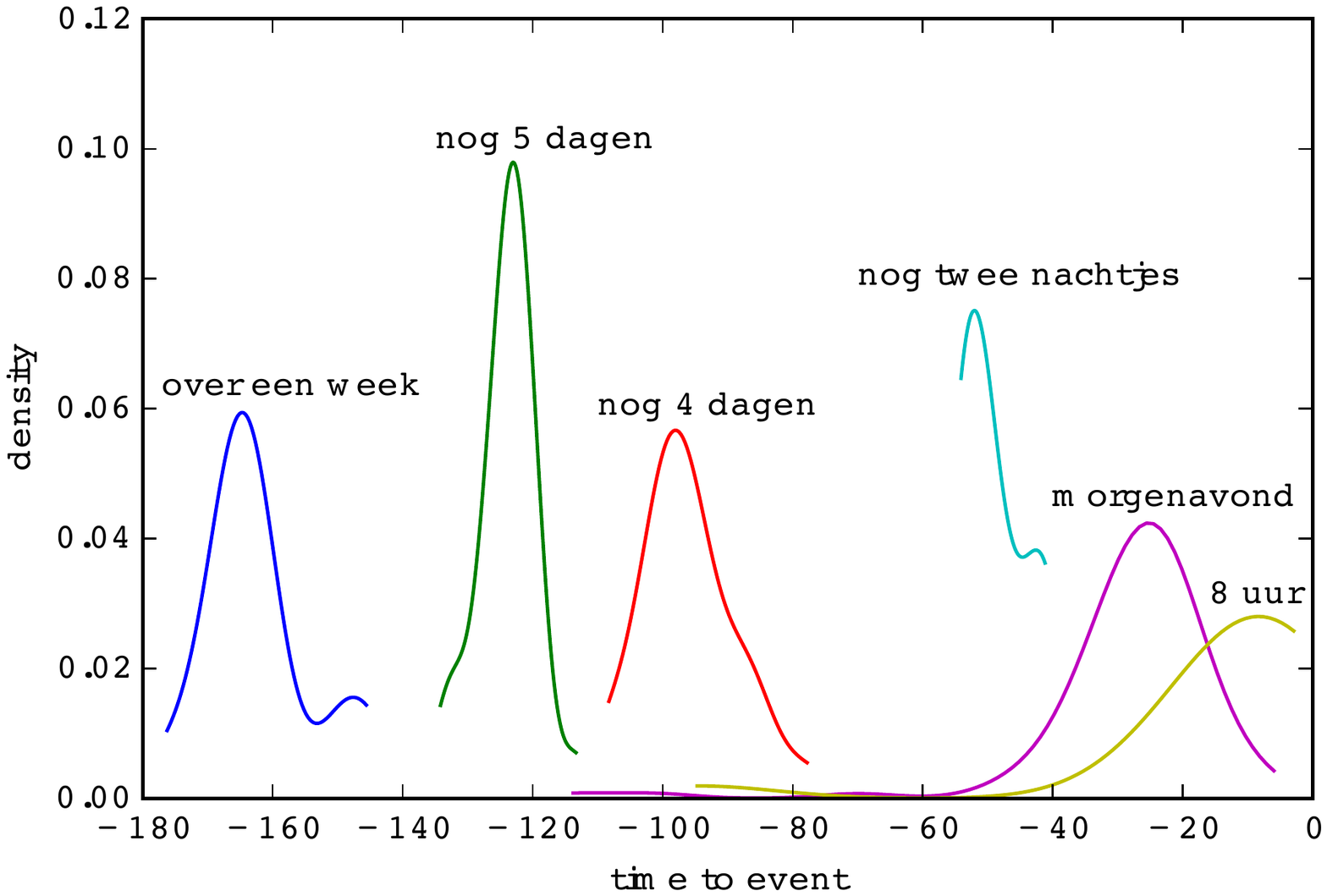}
\caption{Kernel density estimation using Gaussian kernels of some selected temporal features are illustrated to show the information these features carry. We can observe that the meaning of the temporal expressions comply with their temporal distribution before an event.}
\label{fig:TFeatsTS}
\end{center}
\end{figure*}

Figures~\ref{fig:WFeatsTS} and~\ref{fig:TFeatsTS} visualize the distribution of the TTE values of selected features from both feature sets. The distributions are fitted through kernel density estimation using Gaussian kernels.\footnote{We used the gaussian\_kde method from SciPy v0.14.0 URL:~\url{http://docs.scipy.org/doc/scipy/reference/generated/scipy.stats.gaussian_kde.html}, accessed June 10, 2018} Kernel density curves visualize the suitedness of a feature: the sharper the curve and the higher its peak (i.e. the higher its kurtosis), the more accurate the feature is for TTE estimation. Figure~\ref{fig:WFeatsTS} illustrates how peaks in WFeats features may inform about different phases of an event. The features {\em aangesteld, scheidsrechter} `appointed, referee', {\em aanwezig, wedstrijd} `present, game', and {\em onderweg, wedstrijd} `on the road, match' relate to different preparation phases of a football match. The feature {\em allerlaatst, kaarten} `latest, tickets' refers to the act of buying tickets; {\em overmorgen} `the day after tomorrow', a temporal expression (which is also included in the WFeats set) indicates the temporal distance to the event in terms of days. The curves are either sharply concentrated on particular points in time (e.g. `on the road, match' refers to travel within hours before the match), or fit a broad array of low-density data points.

In contrast, Figure~\ref{fig:TFeatsTS} displays kernel density curves of selected features from the TFeats set. The features are {\em over een week} `over a week', {\em nog 5 dagen} `another 5 days', {\em nog 4 dagen} `another 4 days', {\em nog twee nachtjes} `another two nights', {\em morgenavond} `tomorrow evening', and {\em 8 uur} `8 hours', and show relatively similar curves. This suggests that these temporal expressions tend to have a similar standard deviation of about a full day. Indeed, the expression {\em morgenavond} `tomorrow evening' may be used in the early morning of the day before up to the night of the day before.

\subsubsection{Time-to-Event Estimation for a Tweet}

The TTE estimate for a tweet will be calculated by the estimation function using the TTE estimates of all features observed in a particular tweet. 

Since TFeats and RFeats can be sparse as compared to WFeats, and since we want to keep the method modular, we provide estimates for each feature set separately. The TFeats and WFeats that occur in a tweet are evaluated by the estimation function to provide the estimation for each feature set. We use the mean or the median as an estimation function. 

To improve the estimate of a current tweet, we use a combination method for the available features in a tweet and a historical context integration function in which we take into account all estimates generated for the same event so far, and take the median of these earlier estimates as a fourth estimate besides those generated on the basis of TFeats, RFeats and WFeats. The combination was generated by using the estimates for a tweet in the following order: \begin{inparaenum}[(i)]
\item if the RFeats generate an estimate, which is the mean of all estimation rule values in a tweet, the TTE estimate for a tweet is that estimate;
\item else, use the TFeats-based model to generate an estimate; and
\item finally, if there is not yet an estimate for a tweet, use the WFeats-based model to generate an estimate.
\end{inparaenum} 
The priority order that places rule estimates before TFeats estimates, and TFeats estimates before WFeats estimates, is used in combination with the history function. Our method integrates the history as follows: \begin{inparaenum}[(i)]
\item rule estimates only use the history of rule estimates;
\item TFeats estimates use the history of rule estimates and TFeats estimates; and
\item WFeats estimates do not make any distinction between the source of previous estimates that enter the history window function calculation.
\end{inparaenum} In this manner the precise estimates generated by RFeats are not overwritten by the history of lower-precision estimates, which does happen in the performance-based combination. 

\subsection{Experimental Set-up}
\label{experimentsetupNLPT}

In this section we describe the experiments carried out using our TTE estimation method, the data sets gathered for the experiments, and the training and test regimes to which we subject our method. The section concludes with our evaluation method and the description of two baseline systems.

Retweets repeat the information of the source tweet and occur possibly much later in time than the original post. While including retweets could improve the performance under certain conditions~\cite{Batista+04}, results from a preliminary experiment we carried out using development data show that eliminating retweets yields better or comparable results for the TTE estimation task most of the time. Therefore, in the experiments reported here we used the datasets (FM and MC) without retweets. 

\subsubsection{Training and Test Regimes}
\label{traintestprincip}

After extracting the relevant features, our TTE estimation method proceeds with the feature selection, feature value assignment, and tweet TTE estimation steps. Features are selected according to their frequency and distribution over time. We then calculate the TTE value of selected features by means of a training function. Finally, the values of the features that co-occur in a single tweet are given to the estimation function, which on the basis of these single estimates generates an aggregated TTE estimation for the tweet.

The training and estimation functions are applied to TFeats and WFeats. The values of RFeats are not trained, but already set to certain values in the estimation rules themselves, as stated earlier.

\subsubsection{Evaluation and Baselines}
\label{baseline}

The method is evaluated on 51 of the 60 football matches and all 32 music concert events. Nine football matches (15\%), selected randomly, are held out as a development set to optimize the hyperparameters of the method.

Experiments on the test data are run in a leave-one-event-out cross-validation setup. In each fold the model is trained on all events except one, and tested on the held-out event; this is repeated for all events. 

We report two performance metrics in order to gain different quantified perspectives on the errors of the estimates generated by our systems. The error is represented as the difference between the actual TTE, $v_{i}$, and the estimated TTE, $e_{i}$. Mean Absolute Error (MAE), given in Equation~\ref{maeeq}, represents the average of the absolute value of the estimation errors over test examples $i = {1 \ldots N}$. Root Mean Squared Error (RMSE), given in Equation~\ref{rmse}, sums the squared errors; the sum divided by the number of predictions $N$; the (square) root is then taken to produce the RMSE of the estimations. RMSE penalizes outlier errors more than MAE does. 



\begin{equation}
 MAE ={\frac{1} {N}{\sum\limits_{i = 1}^N |{v_{i} - e_{i}| } } }
 \label{maeeq}
\end{equation}




We computed two straightforward baselines derived from the test set: the mean and median TTE of all tweets. They are computed by averaging and calculating the mean or the median of the TTE of all training tweets. The mean baseline estimate is approximately 21 hours, while the median baseline is approximately 4 hours. Baseline estimations are calculated by assigning every tweet the corresponding baseline as an estimate. For instance, all estimates for the median baseline will be the median of the tweets, which is 4 hours.

Although the baselines are simplistic, the distribution of the data make them quite strong. For example, 66\% of the tweets in the FM data set occur within 8 hours before the start of the football match; the median baseline generates an error of under 4 hours for 66\% of the tweets.

In addition to error measures, we take the coverage into account as well. The coverage reflects the percentage of the tweets for which an estimate is generated for an event. Evaluating coverage is important, as it reveals the recall of the method. Coverage is not recall (i.e. a match on a tweet does not mean that the estimate is correct) but it sets an upper bound on the percentage of relevant tweets a particular method is able to use. Having a high coverage is crucial in order to be able to handle events that have few tweets, to start generating estimations as early as possible, to apply a trained model to a different domain, and to increase the benefit of using the history of the tweet stream as a context. Thus, we seek a balance between a small error rate and high coverage. Twitter data typically contains many out-of-vocabulary words~\cite{Baldwin+13}, so it remains a challenge to attain a high coverage.

\subsubsection{Hyperparameter Optimization}
\label{hyperopt}

The performance of our method depends on three hyperparameters and two types of functions for which we tried to find an optimal combination of settings, by testing on the development set representing nine football matches:

\begin{description}
\item[Standard Deviation Threshold] -- the quantile cut-off point for the highly deviating terms, ranging from not eliminating any feature (0.0), to eliminating the highest standard deviating 40-quantile;
\item[Feature length] -- maximum number of words or temporal expressions captured in a feature, from 1 to 7;
\item[Window Size] -- the number of previous estimations used as a context to adjust the estimate of the current tweet, from 0 to the complete history of estimations for an event. 

\end{description}

The frequency threshold of features is not a hyperparameter in our experiments. Instead, we eliminate hapax features for features, which are WFeats and TFeats, that are assigned a time-to-event value from the training data.

We want to identify which functions should be used to learn the feature values and perform estimations. Therefore we test the following functions:

\begin{description}
\item[Training Function] -- calculates mean or median TTE of the features from all training tweets;
\item[Estimation Function] -- calculates mean or median value on the basis of all features occurring in a tweet.
\end{description}

\begin{figure*}[htb]
\begin{center}
\includegraphics[scale=0.90]{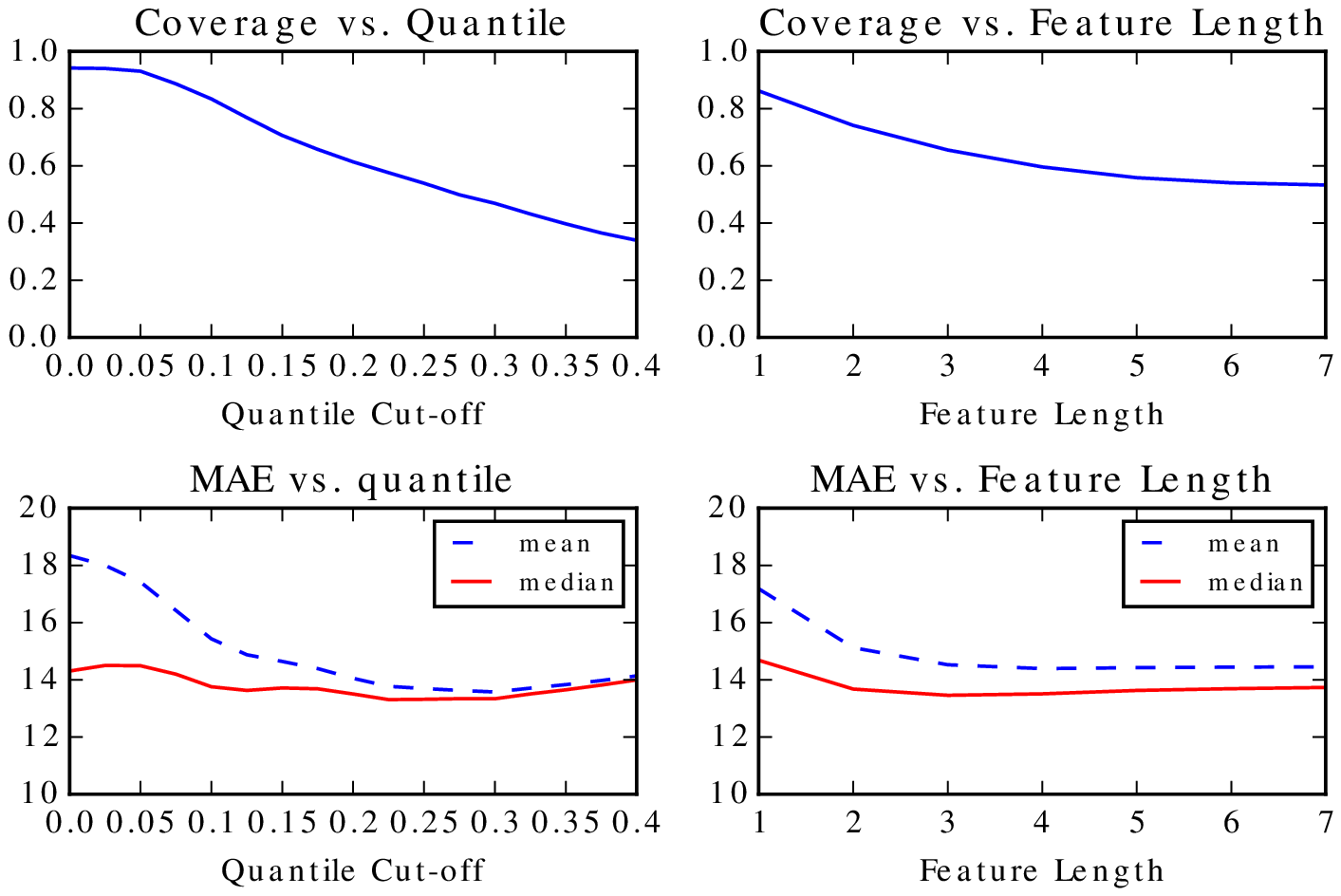}
\caption{MAE and coverage curves for different quantile cut-off and feature lengths for word skipgrams (WFeats). Longer features, higher quantile cut-off are mostly correlated with smaller MAE.}
\label{fig:subWFeats}
\end{center}
\end{figure*}

\begin{figure*}[htb]
\begin{center}
\includegraphics[scale=0.90]{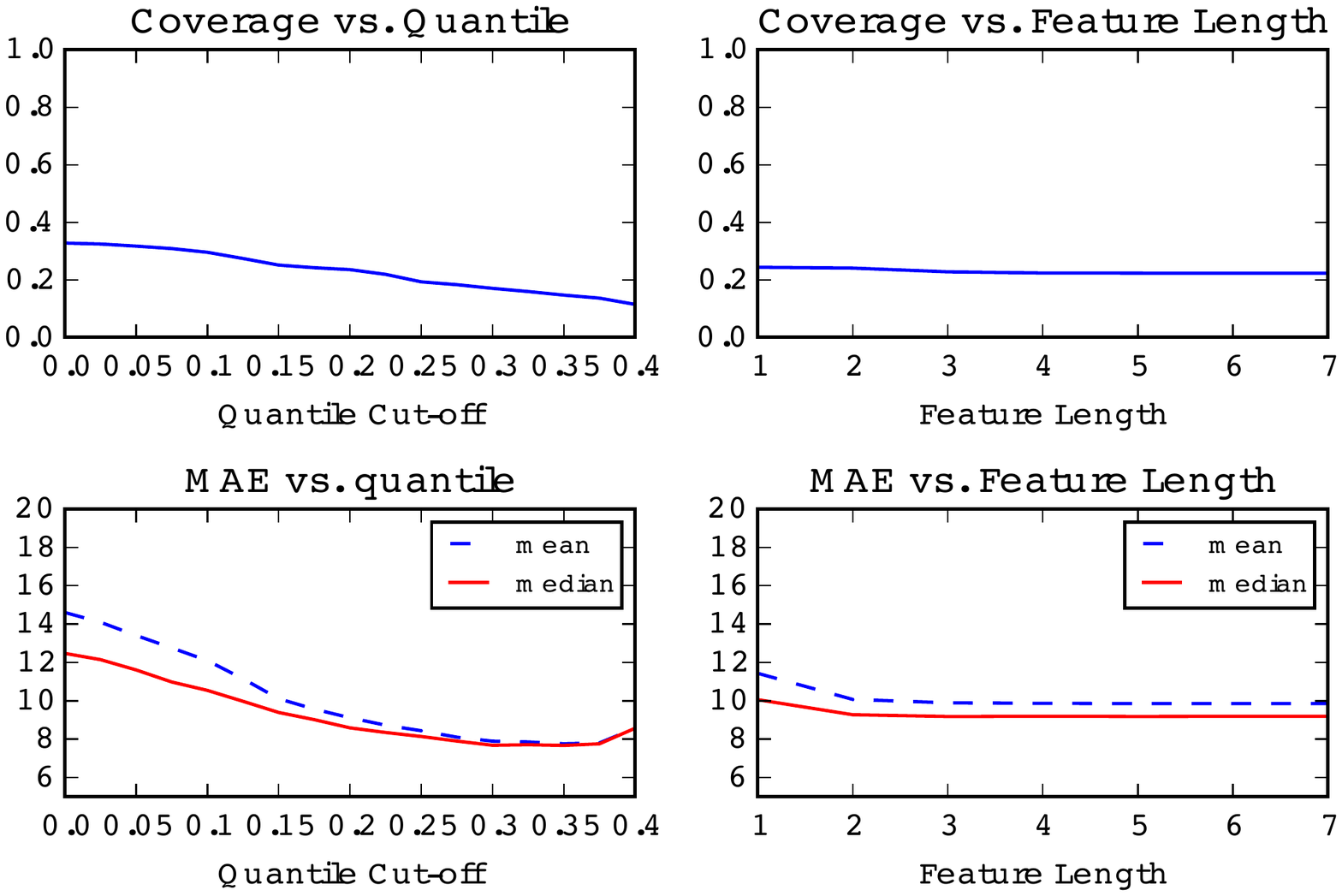}
\caption{MAE and coverage curves for different quantile cut-off and feature lengths for temporal expressions (TFeats). Longer features, higher quantile cut-off are mostly correlated with smaller MAE.}
\label{fig:subTemp}
\end{center}
\end{figure*}

Figures~\ref{fig:subWFeats} and~\ref{fig:subTemp} show how the feature length and quantile cut-off point affect the MAE and the coverage on the development data set. The larger the feature length and the higher quantile cut-off point, the lower both the coverage and MAE. We aim for a low MAE and a high coverage.


Long features, which consist of word skipgrams with n$\geq$2, are not used on their own but are added to the available feature set, which consists of continuous n-grams. It is perhaps surprising to observe that adding longer features causes coverage to drop. The reason for this is that longer features are less frequent and have smaller standard deviations than shorter features. Shorter features are spread over a long period which makes their standard deviation higher: this causes these short features to be eliminated by the quantile-threshold-based feature selection in favor of the longer features that occur less frequently. Additionally, selecting higher quantile cut-off points eliminates many features, which causes the coverage to decrease. 

Since our hyperparameter optimization shows that the MAE does not get better for features that are longer than 2, a feature length of $n=2$ is used for both WFeats and TFeats. The quantile cut-off point is 0.20 for WFeats as higher quantile cut-off points decrease the coverage; at the same time they do not improve the MAE. For TFeats, the quantile cut-off point is set at 0.25. These parameter values will be used for both optimizing the history window length on the development set and running the actual experiment. Using these parameters yields a reasonable coverage of 80\% on the development set.

Using the median function for both training and estimate generation provides the best result of 13.13 hours with WFeats. In this case the median of median TTE values of the selected features in a tweet is taken as the TTE estimation of a tweet. The median training and mean estimation, mean training and median estimation, and mean training and mean estimation yielded 13.15, 14.11, and 14.04 respectively.




              
        

\begin{figure*}[htb]
\begin{center}
\includegraphics[scale=0.90]{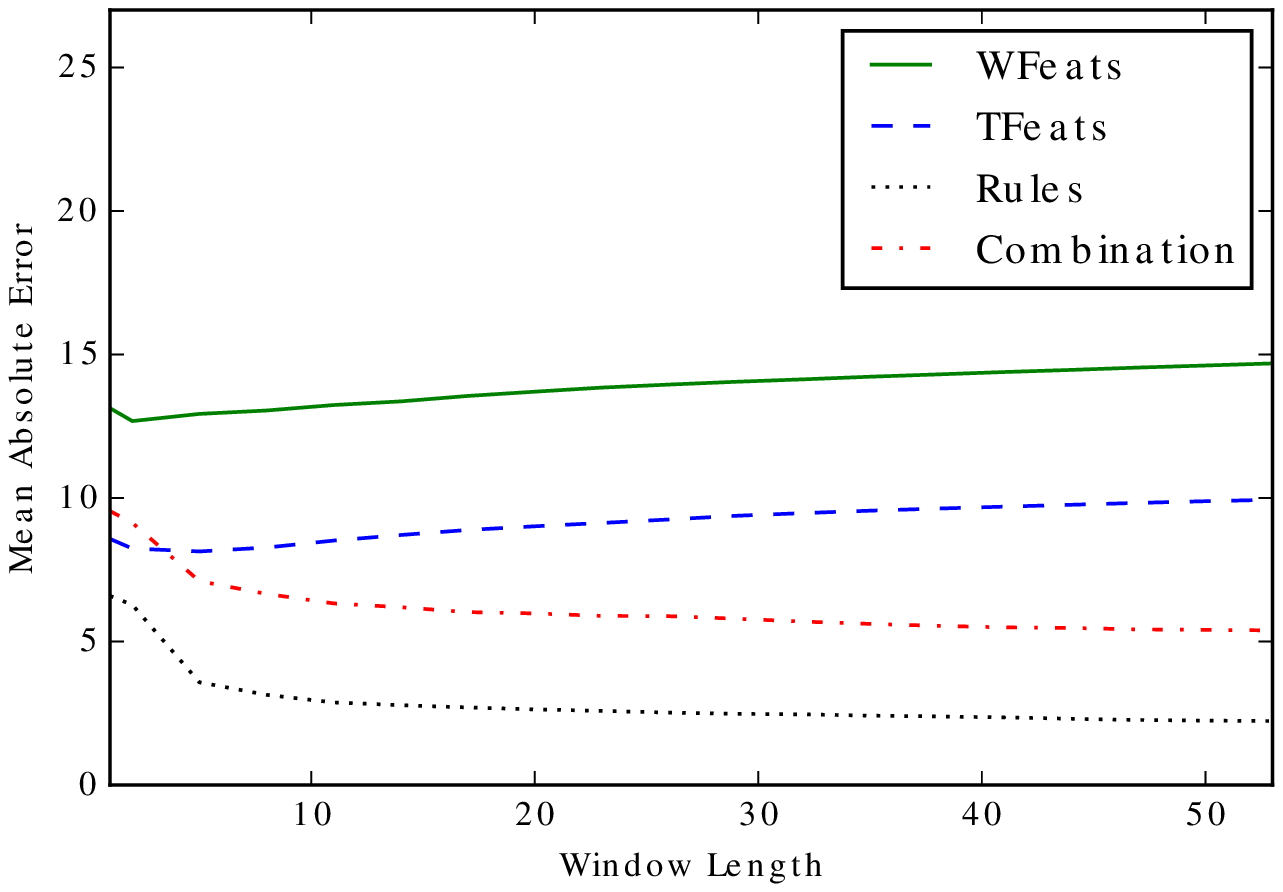}
\caption{MAE of estimates and window size of historical context function per feature set and their combination. The Rules and Combinations perform significantly better as they utilize the previous estimates.}
\label{fig:histWFeats}
\end{center}
\end{figure*}

\begin{figure*}[htb]
\begin{center}
\includegraphics[scale=0.90]{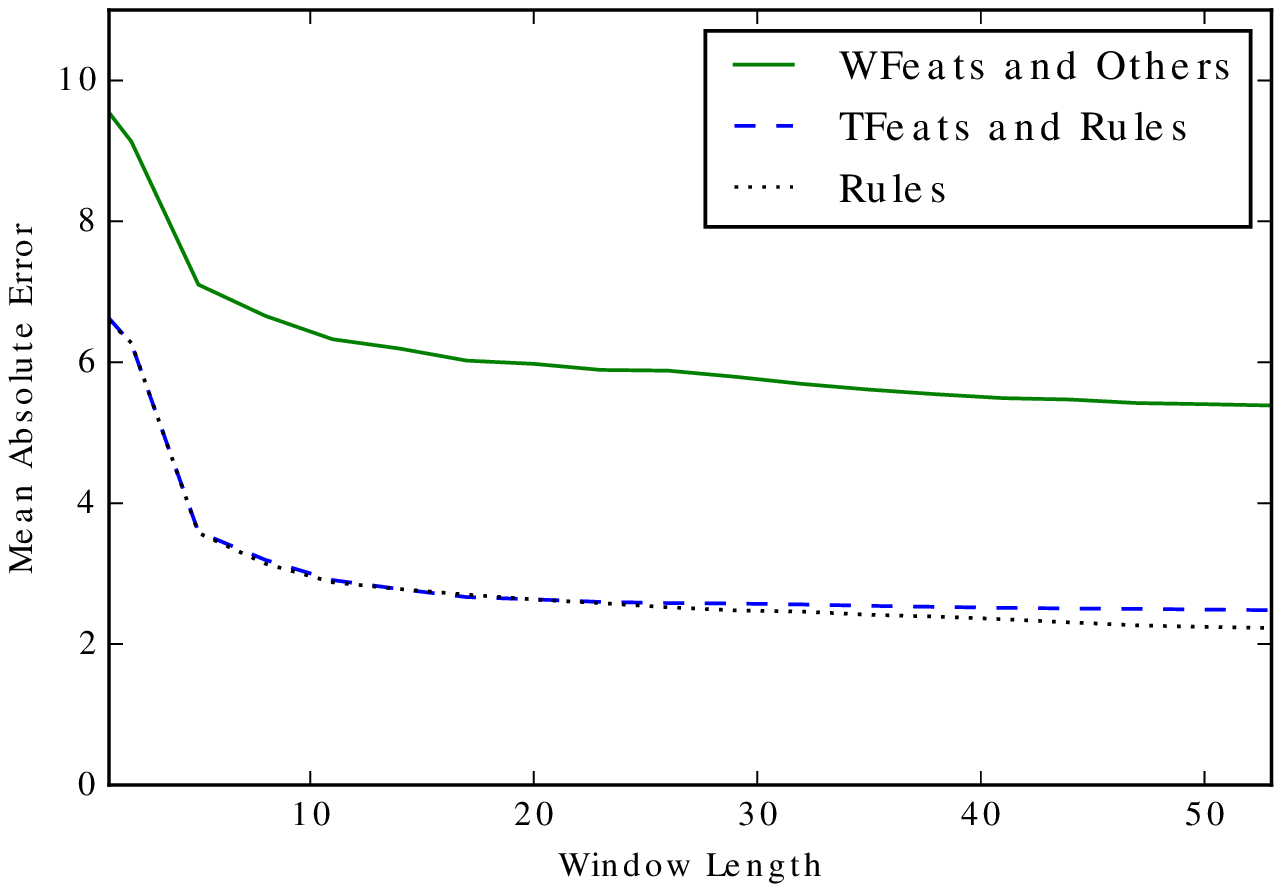}
\caption{Various feature set based estimates and the priority-based historical context integration. Applying priority to the estimates has a clear benefit.}
\label{fig:histAll}
\end{center}
\end{figure*}

The window size of the historical context integration function is considered to be a hyperparameter, and is therefore optimized as well. The historical context is included as follows: given a window size $|W|$, for every new estimate the context function will be applied to $W-1$ preceding estimates and the current tweet's estimate to generate the final estimate for the current tweet. In case the number of available estimates is smaller than $W-1$, the historical context function uses only the available ones. The history window function estimate is re-computed with each new tweet.

Figures~\ref{fig:histWFeats} and~\ref{fig:histAll} demonstrate how the window size affects the overall performance for each module in ~\ref{fig:histWFeats} and the performance-based combination in Figure~\ref{fig:histAll}. The former figure shows that the history function reinforces the estimation quality. In other words, using the historical context improves overall performance. The latter figure illustrates that the priority-based historical context integration improve the estimation performance further.

We aim to pick a window size that improves the estimation accuracy as much as possible while not increasing the MAE for any feature, combination, and integration type. Window size 15 answers to this specification. Thus, we will be using 15 as window size. Furthermore, we will use the priority-based historical context integration, since it provides more precise estimates than the feature type independent historical context integration. 

\subsection{Test Results}
\label{testresultsNLPT}

We test the method by running leave-one-event-out experiments in various domain, feature set and hyperparameter settings. First we perform in-domain experiments, where the system is trained and tested in the scope of a single domain, on the 51 FM events that were not used during the hyperparameter optimization phase, and all events in `MC without retweets' data. The in-domain experiment for MC data set is performed using the hyperparameters that were optimized on the development FM data set. We then measure the estimation performance of a model trained on FM data set on the MC data set. Finally we combine the data sets in order to test the contribution of each data set in an open-domain training and testing setting.

Table~\ref{table:resultstable} lists the performance of the method in terms of MAE, RMSE, and coverage for each experiment. The `Football Matches' and `Music Concerts' parts contain the results of the in-domain experiments; the domain union part provides the results for the experiment that uses both the FM and MC data sets to train and test the method. The model transfer part shows results for the experiment in which we use the FM data set for training and the MC data set as a test. Columns denote results obtained with the different feature sets: `RFeats', `TFeats', and `WFeats', priority based historical context integration in `All', and the two baselines: `Median Baseline' and `Mean Baseline'.

For the football matches, RFeats provide the best estimations with 3.42 hours MAE and 14.78 hours RMSE. RMSE values are higher than MAE values in proportion to the relatively large mistakes the method makes. Although RFeats need to be written manually and their coverage is limited to 27\%, their performance shows that they offer rather precise estimates. Without the RFeats, TFeats achieve 7.53 hours MAE and 24.39 hours RMSE. This indicates that having a temporal expressions list will provide a reasonable performance as well. WFeats, which do not need any resource other than a lexicon, yield TTE estimates with a MAE of 13.98 and a RMSE of 35.55. 

The integration approach, of which the results are listed in the column `All', succeeds in improving the overall estimation quality while increasing the coverage up to 85\% of the test tweets. Comparing these results to the mean and median baseline columns shows that our method outperforms both baselines by a clear margin.

\begin{table*}\centering
\ra{1.1}
\begin{tabular}{@{}lrrrrrr@{}}
\toprule
 & & & & & Mean & Median \\
 &  RFeats &  TFeats &  WFeats &  All &  Baseline &  Baseline \\
\midrule

Football Matches \\
 \phantom{a}     RMSE       &  14.78 &   24.39 &   35.55 &   21.62 &       38.34 &      41.67 \\
 \phantom{a}     MAE        &   3.42 &    7.53 &   13.98 &    5.95 &       24.44 &      18.28 \\
 \phantom{a}     Coverage   &   0.27 &    0.32 &    0.82 &    0.85 &        1.00 &       1.00 \\
\\
Music Concerts   \\  
 \phantom{a}     RMSE       &  25.49 &   31.37 &   50.68 &   38.44 &       54.07 &      61.75 \\
 \phantom{a}     MAE        &   9.59 &   13.83 &   26.13 &   15.80 &       38.93 &      34.98 \\
 \phantom{a}     Coverage   &   0.26 &    0.27 &    0.76 &    0.79 &        1.00 &       1.00 \\
\\
Domain Union     \\
 \phantom{a}     RMSE       &  15.19 &   24.16 &   35.76 &   22.38 &       38.97 &      42.43 \\
 \phantom{a}     MAE        &   3.59 &    7.63 &   14.25 &    6.24 &       24.95 &      18.79 \\
 \phantom{a}     Coverage   &   0.27 &    0.31 &    0.82 &    0.84 &        1.00 &       1.00 \\
\\
Model Transfer   \\
 \phantom{a}     RMSE       &  25.43 &   36.20 &   57.65 &   44.40 &       57.28 &      64.81 \\
 \phantom{a}     MAE        &   9.66 &   19.62 &   31.28 &   19.04 &       35.51 &      37.06 \\
 \phantom{a}      Coverage  &   0.26 &    0.22 &    0.74 &    0.77 &        1.00 &       1.00 \\
\bottomrule
\end{tabular}
\caption{In domain experiments for FM, MC, domain union and model transfer. The column `All' illustrates results of the integration approach. These results clearly show lower MAE and RMSE scores than the baselines.}
\label{table:resultstable}
\end{table*}

The errors and coverages obtained on music concerts listed in Table~\ref{table:resultstable} show that all features and their combinations lead to higher errors as compared to the FM dataset, roughly doubling their MAE. Notably, WFeats yield a MAE of more than one day. The different distribution of the tweets, also reflected in the higher baseline errors, appears to be causing a higher error range, but still our method performs well under the baseline.

The domain union part of Table~\ref{table:resultstable} shows results of an experiment in which we train and test in an open-domain setting. Both the 51 test events from the FM and 32 events from the MC data sets are used in a leave-one-out experiment. These results are comparable to FM in-domain experiment results.

The results of the cross-domain experiment in which the model is trained on FM data set and tested on the MC data set is represented in the `Model Transfer' part of Table~\ref{table:resultstable}. The performance is very close to the performance obtained in the in-domain experiment on the MC data set.

\subsection{Discussion}
\label{analysisNLPT}

In this section we take a closer look at some of the results in order to find explanations for some of the differences we find in the results.

       

\begin{figure*}[htb]
\begin{center}
\includegraphics[scale=0.90]{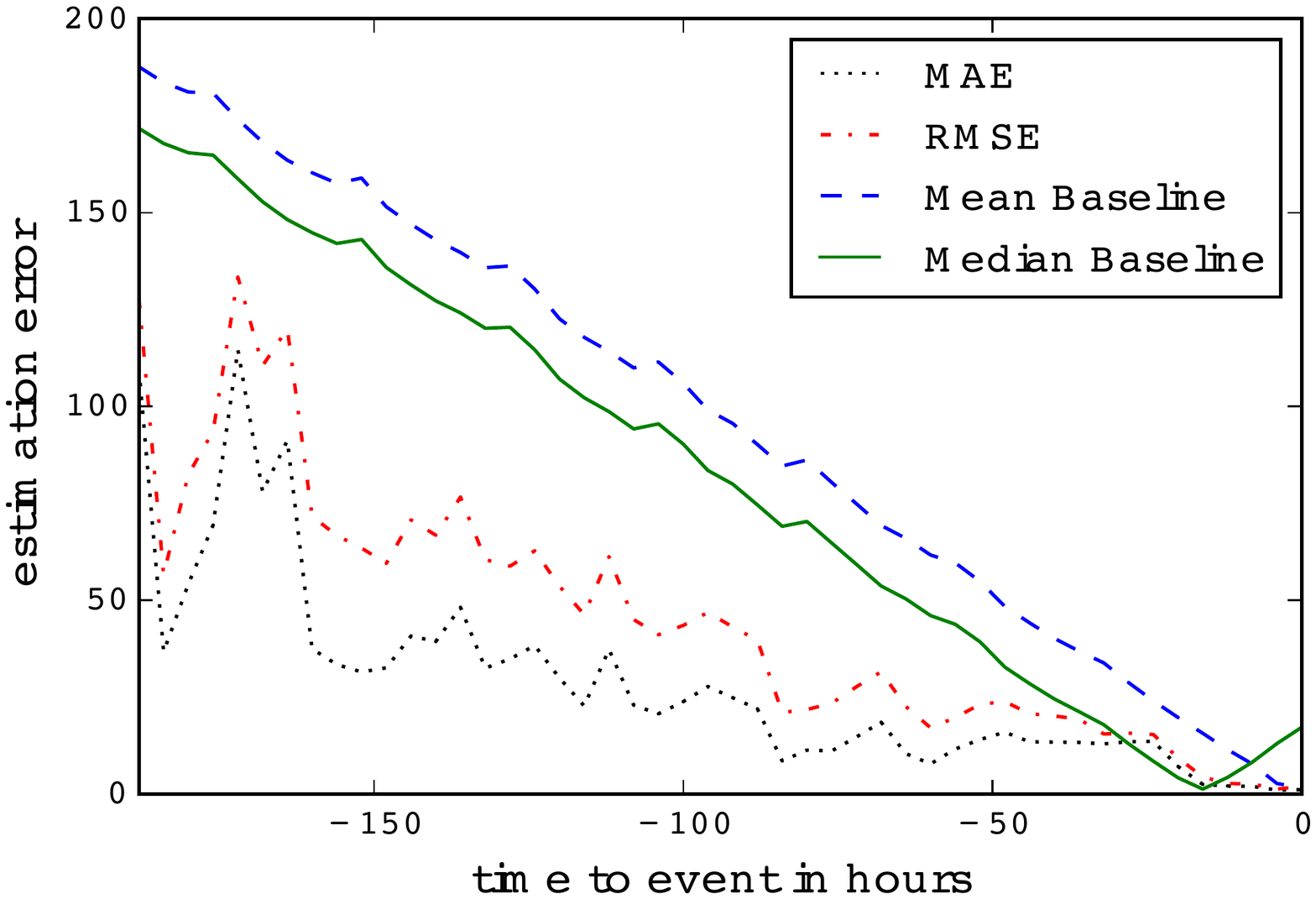}
\caption{Averaged mean estimation error, averaged in 4-hour frames, relative to event start time for the in-domain football matches experiment for the Integration and the baselines. Both MAE and RMSE errors are smaller than the baselines almost all the time.}
\label{fig:maeRmseRelStartTime}
\end{center}
\end{figure*}

\begin{figure*}[htb]
\begin{center}
\includegraphics[scale=0.90]{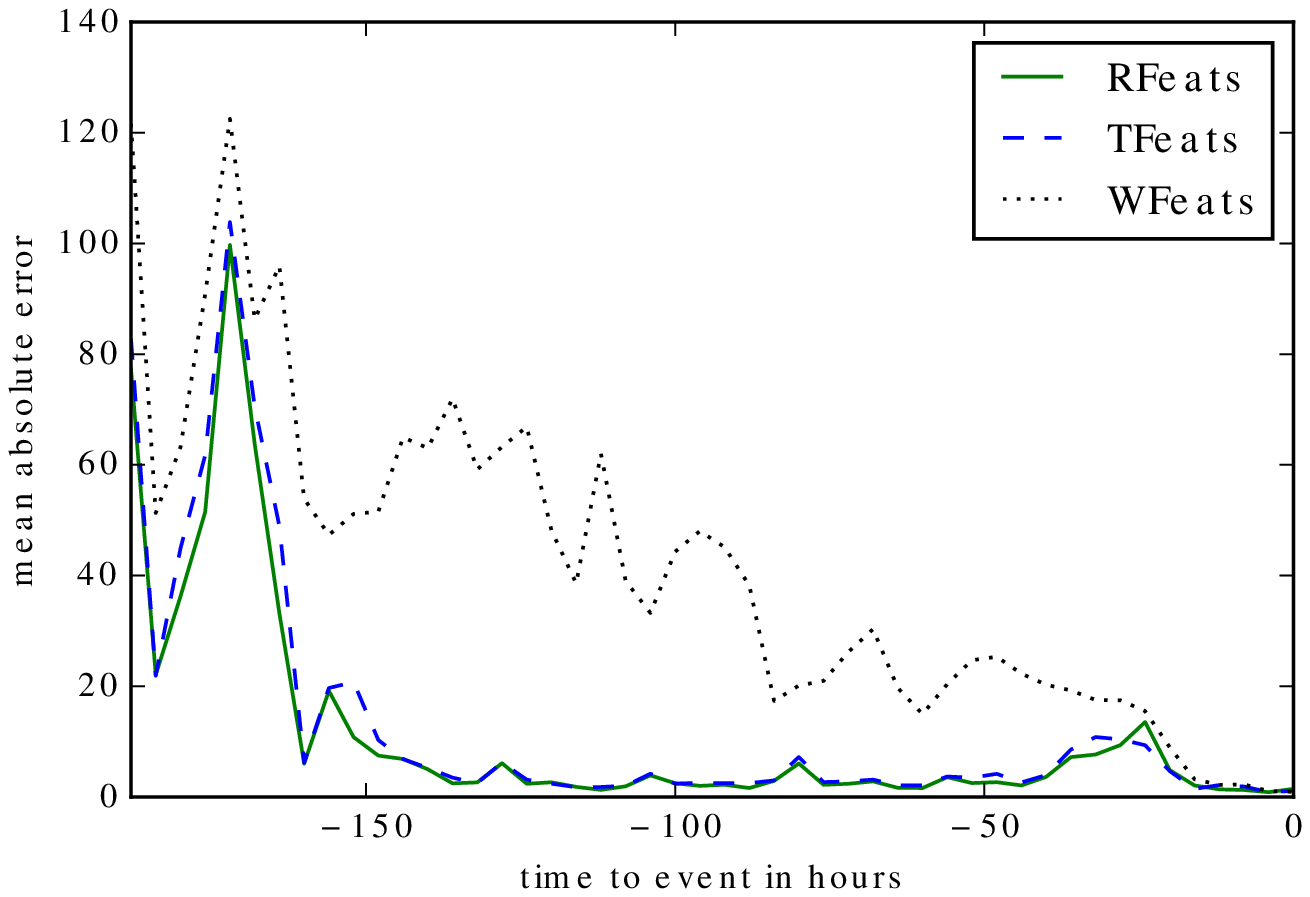}
\caption{Per feature set averaged mean estimation error, averaged in 4-hour frames, relative to event start time for the in-domain football matches experiment. The error decreases as the event time approaches. RFeats and TFeats based results are mostly quite precise.}
\label{fig:featSetsRelStartTime}
\end{center}
\end{figure*}

For the in-domain football matches experiment with our combination approach, the estimation quality in terms of MAE and RMSE relative to event start time of the integration-based method is represented in the Figure~\ref{fig:maeRmseRelStartTime}. The MAE of each feature type relative to historical context integration based on priority is displayed in the Figure~\ref{fig:featSetsRelStartTime}. 

Figure \ref{fig:maeRmseRelStartTime} shows that as the event draws closer, the accuracy of estimations increases. Within approximately 20 hours to the event, which is the period most of the tweets occur, our method estimates TTE nearly flawlessly.\footnote{97\% of the tweets occur before 150 hours of the event start time.} Figure~\ref{fig:featSetsRelStartTime} illustrates that all feature sets are relatively more successful in providing TTE estimates as the event start time approaches. RFeats and TFeats provide accurate estimates starting as early as around 150 hours before the event. In contrast, Wfeats produce relatively higher errors of 20 or more hours off up to a day before the event.


\begin{figure*}[htb]
\begin{center}
\includegraphics[scale=0.80]{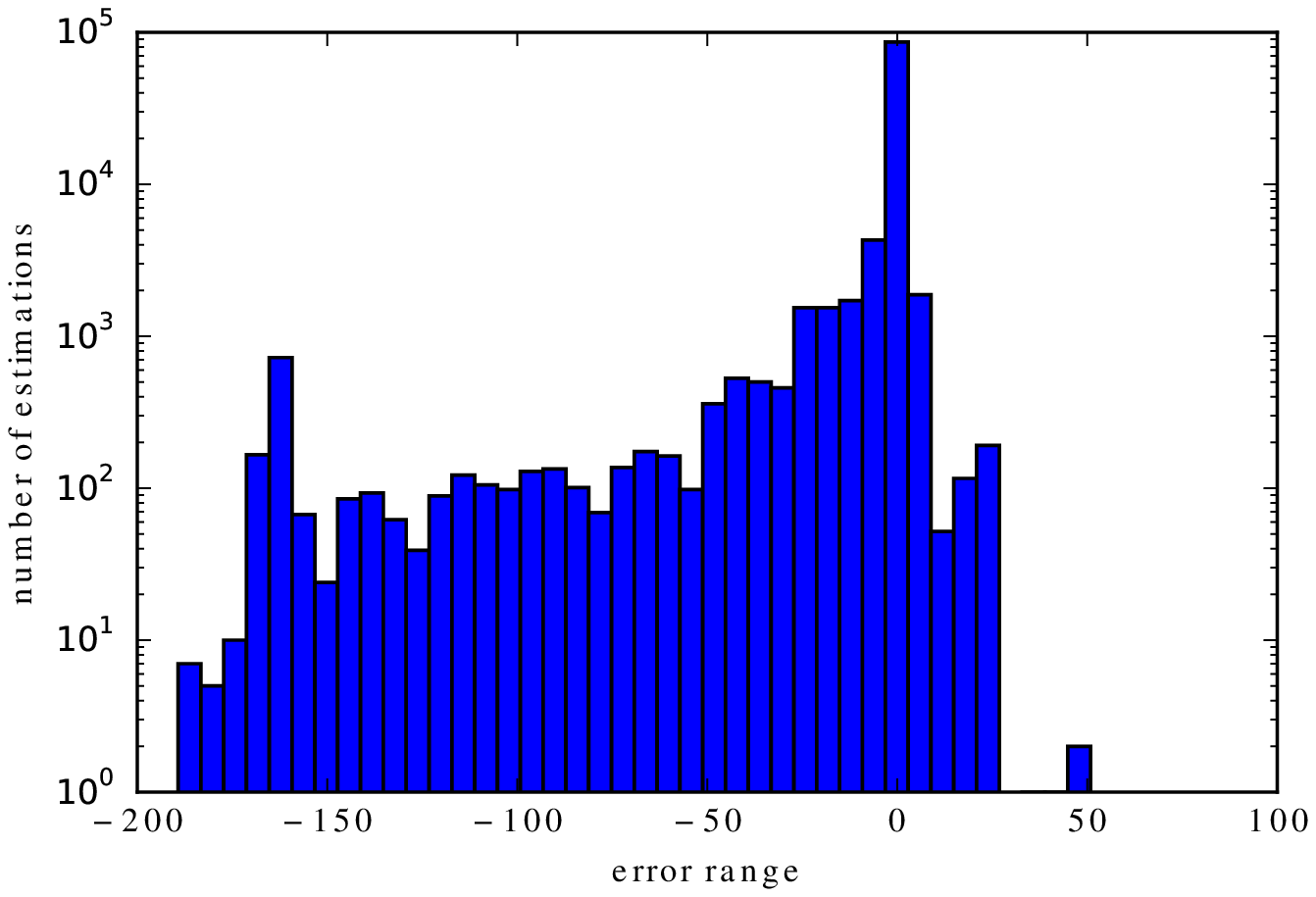}
\caption{Error distribution per estimate for the in-domain football matches experiment. Most of the estimates have no (0) error and only one estimate is drastically off on the positive side, 50 hours. The y-axis is at logarithmic scale.}
\label{fig:errorDistPerTweetHist}
\end{center}
\end{figure*}

\begin{figure*}[htb]
\begin{center}
\includegraphics[scale=0.80]{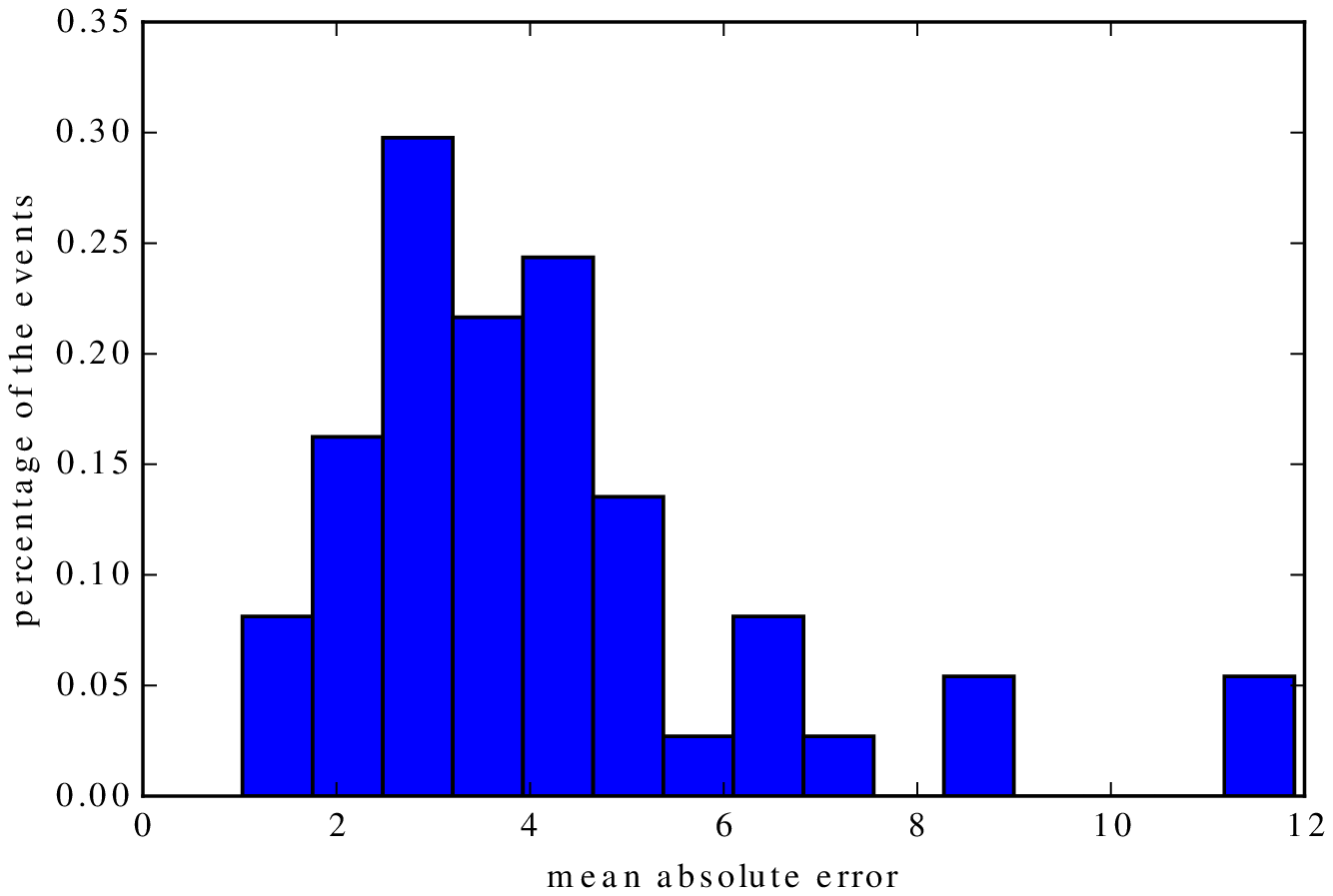}
\caption{MAE distribution per event for the in-domain football matches experiment. Most of the events have an estimation error of less than 4 hours on average.}
\label{fig:eventBasedErrDistHist}
\end{center}
\end{figure*}

Error distributions for tweets and events for the in-domain football matches experiment are displayed in Figure~\ref{fig:errorDistPerTweetHist}. This figure, which has a logarithmic y-axis, illustrates that the majority of the estimations have an error around zero, and that few estimations point to a time after the event as a starting time. Aggregated at the level of events, the per-event estimation performance can be observed in Figure~\ref{fig:eventBasedErrDistHist}. The mean absolute error of estimations for an event is mainly under 5 hours. There is just one outlier event for which it was not possible to estimate remaining time to it reliably. We observed that the football match \#ajatwe, which has an overlapping cup final match in 5 days of the targeted league match, affected the evaluation measures significantly by having the highest number of tweets by 23,976, and having 11.68 and 37.36 hours MAE and RMSE, respectively. 

\begin{sidewaystable}
\centering
\begin{tabular}{|m{0.02cm}|m{3cm}|m{2cm}|m{3cm}|m{3cm}|m{1cm}|m{1cm}|m{1cm}|m{1cm}|m{1cm}|}
\toprule
{} & Tweet &  RFeats &  WFeats &    TFeats &     TTE &  REst &  TEst &  WEst &  IEst \\

\midrule
1 &    nog 4 uurtje tot de klasieker \#feyaja volgen via radio want ik ben @ work &                &   radio, volgen radio, uurtje radio, uurtje & nog uurtje, uurtje &                          4.05 &  & 0.86 &         1.09 & 4.33 \\

\midrule
2 &         we leven er naar toe de 31ste titel wij zitten er klaar voor \#tweaja &                &                                        klaar, zitten klaar, zitten &                          &    1.07 &      &           &         0.51 &         0.90 \\

\midrule
3 &                  aanstaande zondag naar fc utrecht-afc ajax \#utraja \#afcajax &            aanstaande zondag &                                                                    &              &  112.87 &      115.37 &           &           &       115.13 \\

\midrule
4 &  onderweg naar de galgenwaard voor de wedstrijd fc utrecht feyenoord \#utrfey &               
 &                              onderweg, onderweg wedstrijd &                        &        2.88 &        &          &         2.07 &         3.37 \\

\midrule
5 &                                nu \#ajaaz kijken dadelijk \#psvutr kijken &                      &  \mbox{kijken dadelijk}, \mbox{dadelijk kijken}, \mbox{kijken kijken}, dadelijk, kijken &  \mbox{nu dadelijk}, dadelijk &    1.85 &       &         1.07 &         0.91 &         1.88 \\

\midrule
6 &                                     nu \#ajaaz kijken dadelijk \#psvutr kijken &                &  \mbox{kijken dadelijk}, \mbox{dadelijk kijken}, \mbox{kijken kijken}, dadelijk, kijken &  \mbox{nu dadelijk}, dadelijk &    1.85 &      &         1.07 &         0.91 &         1.95 \\

\bottomrule
\end{tabular}
\caption{Sample tweets, extracted features where applicable for RFeats, TFeats and WFeats. The estimates are in REst, TEst, WEst, and IEst for each feature set and their priority based historical context integration respectively.}
\label{table:sampleTweetsAndFeatsEsts}
\end{sidewaystable}

Table~\ref{table:sampleTweetsAndFeatsEsts} illustrates some extracted features for each type of feature, and the estimates based on these sets for six tweets. The second and fourth examples draw from WFeats only: {\em onderweg} `on the road', {\em onderweg, wedstrijd} `on the road, match' and {\em zitten klaar} `seated ready', {\em klaar} `ready', and {\em zitten} `seated' indicate that there is not much time left until the event starts; participants of the event are on their way to the event or ready to watch the game. These features provide a 0.51 and 2.07 hours estimate respectively, which are 0.56 and 0.81 hours off. The history function subsequently adjusts them to be just 0.17 and 0.49 hours, i.e. 6 and 29 minutes, off. 

The third example illustrates how RFeats perform. Although the estimate based on {\em aanstaande zondag} `next Sunday' is already fairly precise, the history function improves it by 0.24 hours. The remaining examples (1, 5, and 6) use TFeats. The first example's estimate is off by 3.09 hours, which the historical context improves it to be just 0.28 hours off. The fifth and sixth examples represent the same tweet for different events, i.e \#ajaaz and \#psvutr. Since every event provides a different history of tweets, they are adjusted differently. 


\begin{figure*}[htb]
\begin{center}
\includegraphics[scale=0.90]{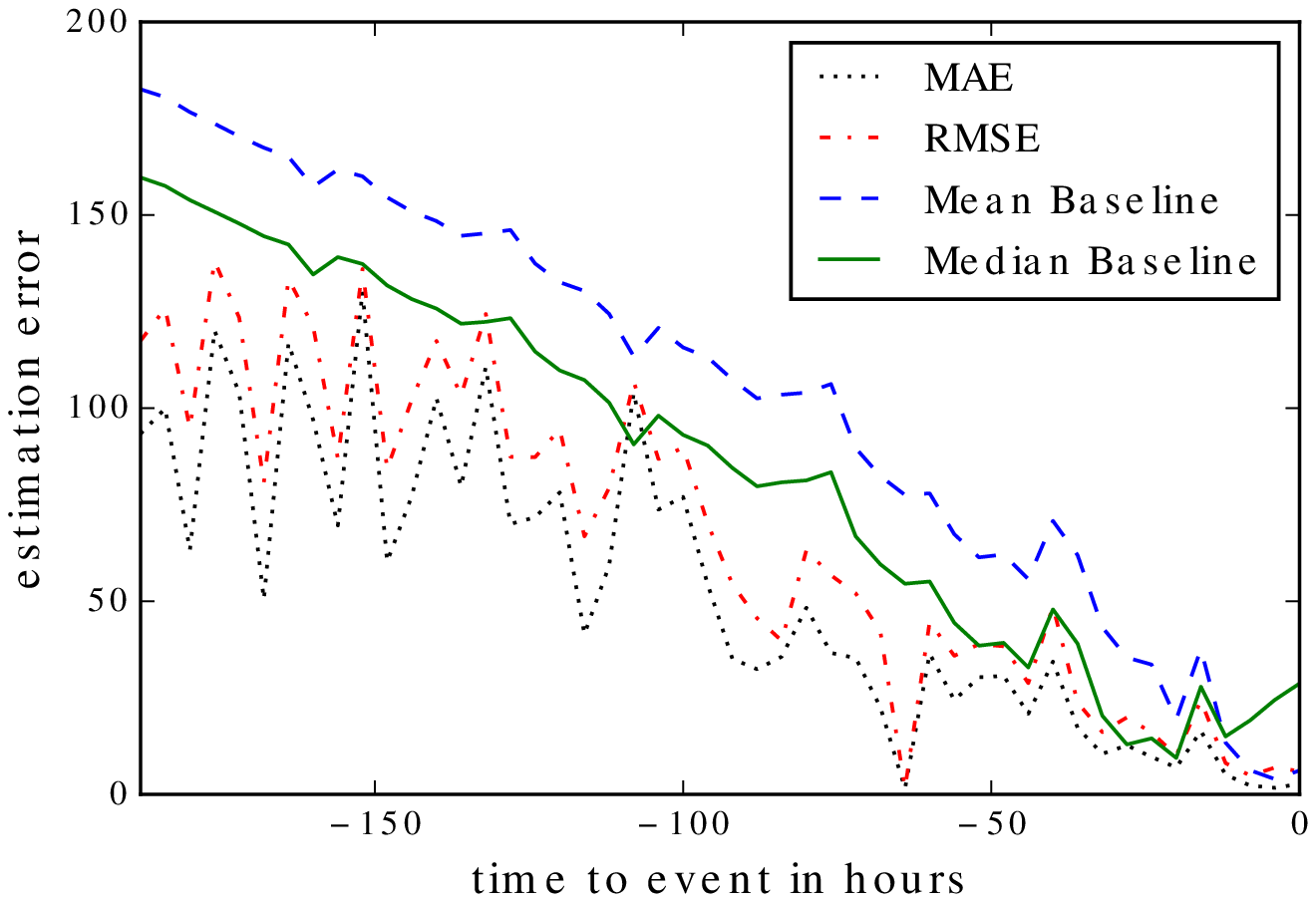}
\caption{MAE and RMSE for the integration and the baselines for the music concerts. The average error of our method is slightly lower than the baselines. The method has good estimates occasionally.}
\label{fig:maeRmseRelStartTime_MC}
\end{center}
\end{figure*}

\begin{figure*}[htb]
\begin{center}
\includegraphics[scale=0.90]{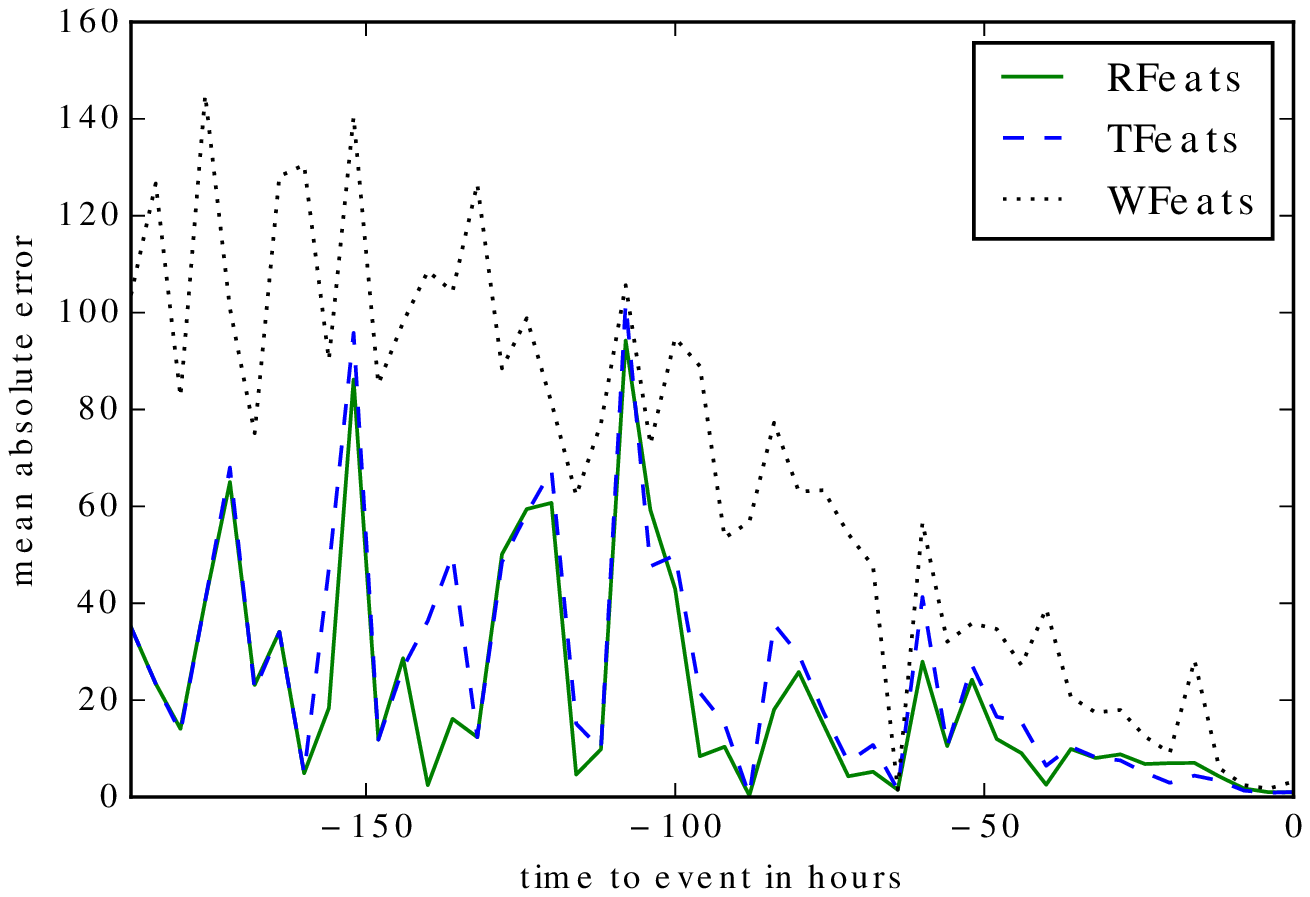}
\caption{MAE for each feature set for the music concerts. Our method starts to generate reliable and consistent results starting 50 hours before an event.}
\label{fig:featSetsRelStartTime_MC}
\end{center}
\end{figure*}

Repeating the analysis for the second domain, music concerts, Figures~\ref{fig:maeRmseRelStartTime_MC} and~\ref{fig:featSetsRelStartTime_MC} illustrate the estimation quality relative to event start time for the in-domain experiment on music concerts. Figure~\ref{fig:maeRmseRelStartTime_MC} shows that the mean error of the estimates remains under the baseline error most of the time, though the estimates are not as accurate as with the football matches. Errors do decrease as the event draws closer.

\begin{figure*}[htb]
\begin{center}
\includegraphics[scale=0.90]{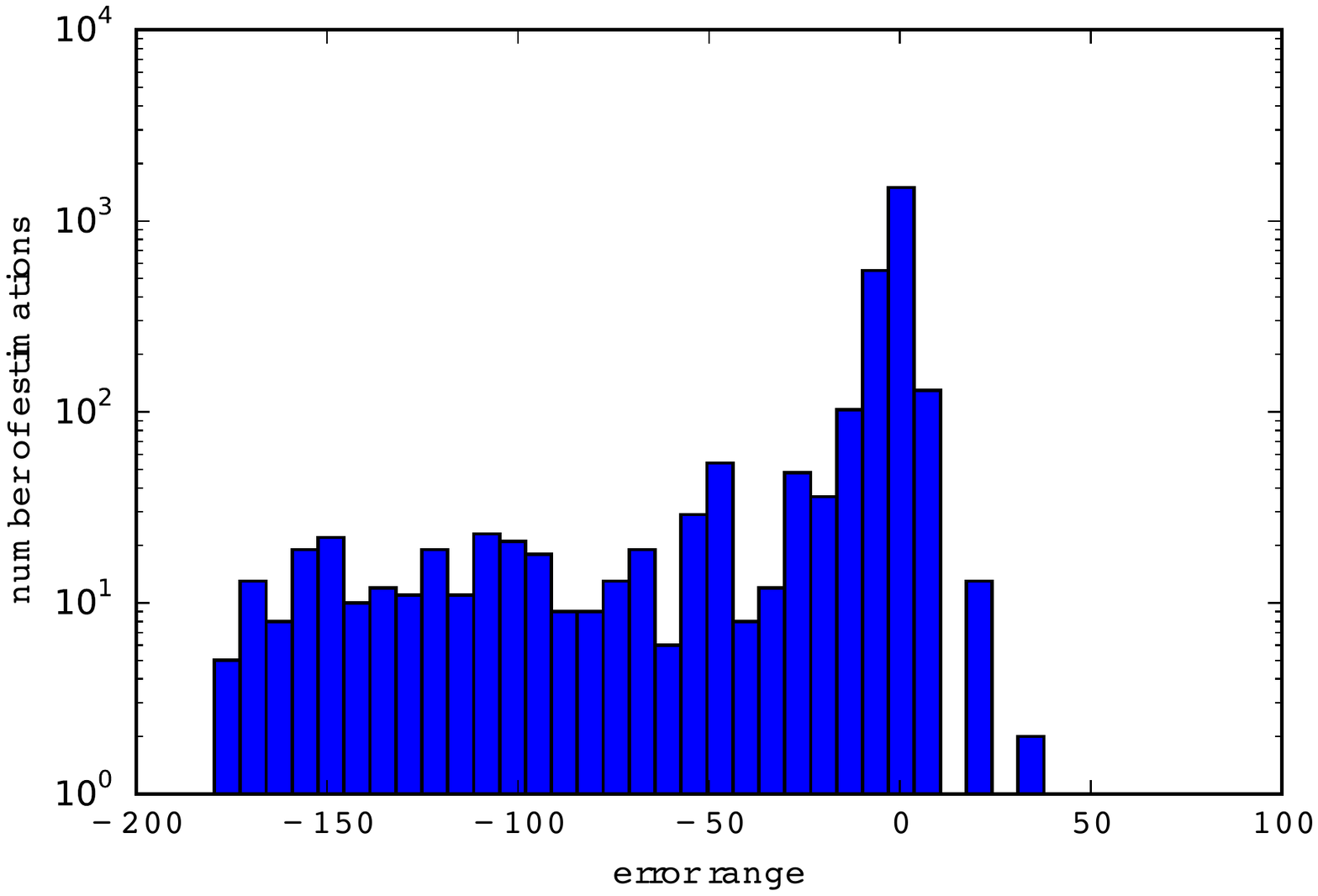}
\caption{Error Distribution per estimate in the music concerts data. Most of the estimates are around zero. The y-axis is at the logarithmic scale.}
\label{fig:errorDistPerTweetHist_MC}
\end{center}
\end{figure*}


\begin{figure*}[htb]
\begin{center}
\includegraphics[scale=0.90]{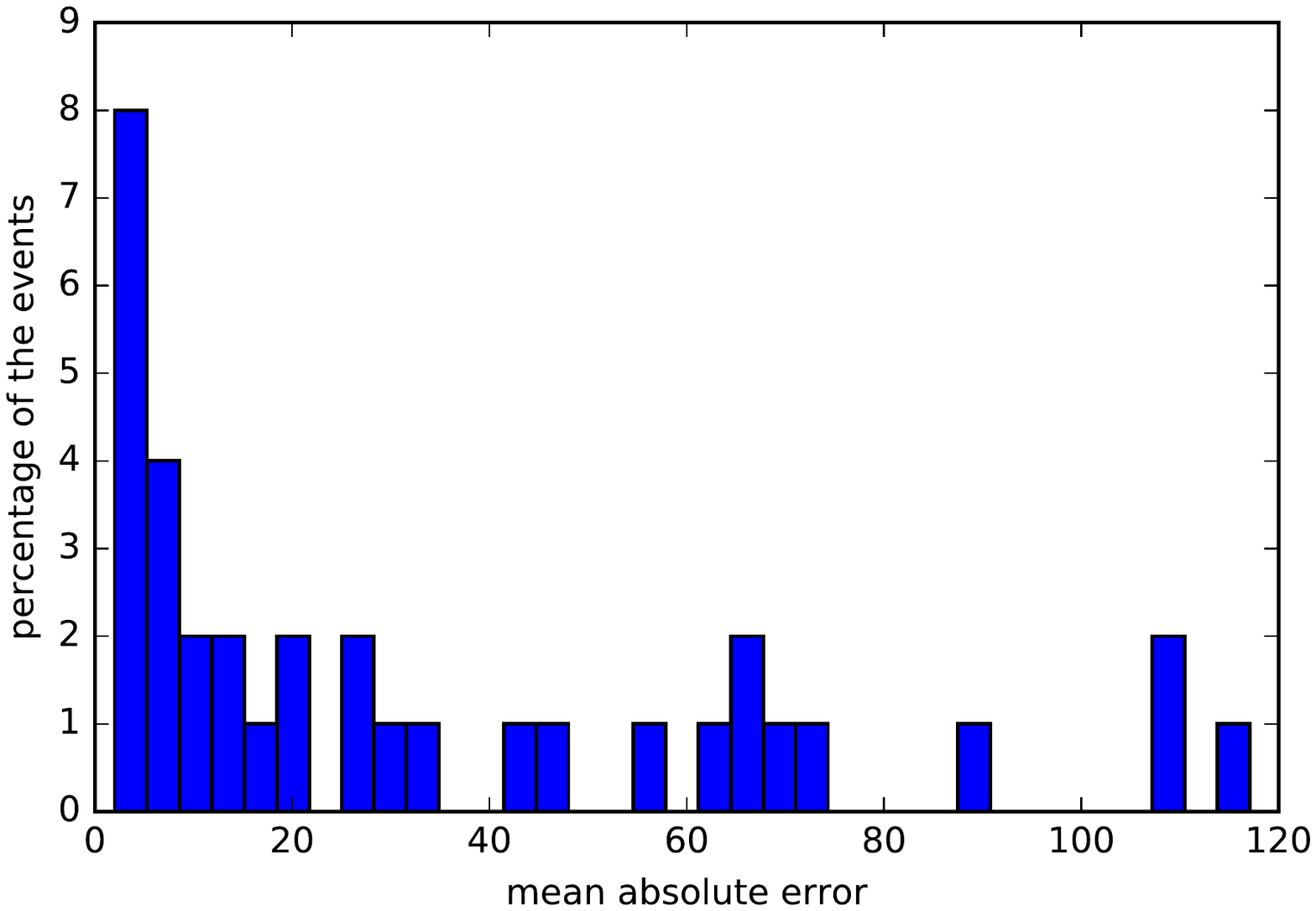}
\caption{MAE per event in the music concerts data. Except outlier events, the method yields results within a reasonable error margin.}
\label{fig:eventBasedErrDistHist_MC}
\end{center}
\end{figure*}

\begin{figure*}[htb]
\begin{center}
\includegraphics[scale=0.90]{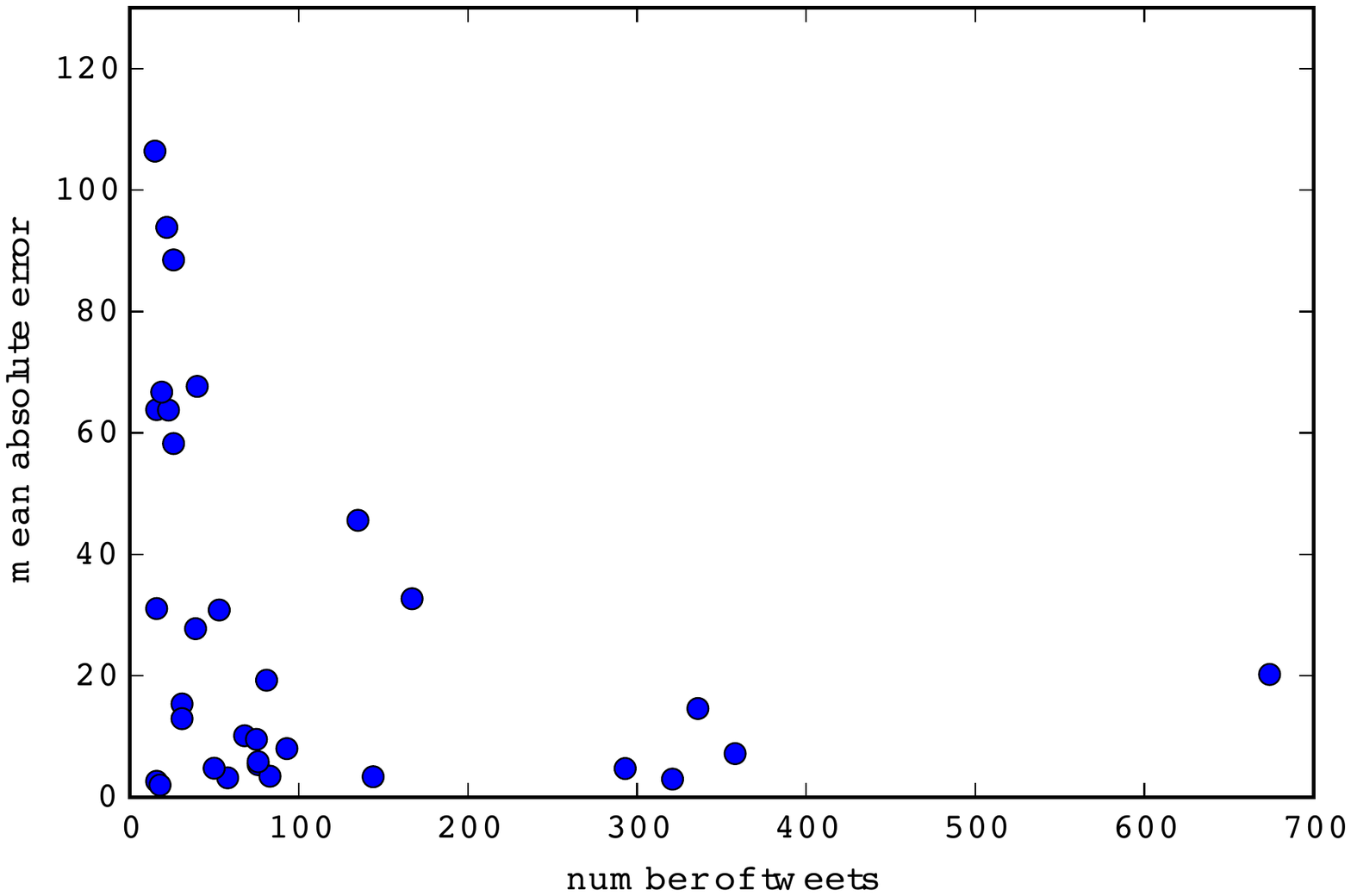}
\caption{MAE in relation to size of the event (MAE is expressed in number of hours, the size of the event in the number of tweets). Bigger events, which have more tweets, tend to have smaller errors.}
\label{fig:sizeMaeRelation}
\end{center}
\end{figure*}

As demonstrated in Figure~\ref{fig:eventBasedErrDistHist_MC} the accuracy of TTE estimation varies from one event to another. An analysis of the relation between the size of an event and the method performance (see also Figure~\ref{fig:sizeMaeRelation}) shows that there appears to be a correlation between the number of tweets referring to an event and the accuracy with which our method can estimate a TTE. For events for which we have less data, the MAE is generally higher. 

The `Domain union' experiment results show that mixing the two event types leads to errors that are comparable to the in-domain experiment for football matches. We take the size of each domain in terms of tweets and the individual domain performances into account to explain this. The proportion of the MC tweets to FM tweets is 0.025. This small amount of additional data is not expected to have a strong impact; results are expected to be, and are, close to the in-domain results. On the other hand, while the results of the domain union are slightly worse than the in-domain experiment on the FM data set, the per-domain results, which were filtered from all results for this experiment, are better than the in-domain experiments for the FM events. The WFeats features for FM event yield 13.85 hours MAE and 35.11 hours RMSE when music concerts are mixed in, compared to a slightly higher 13.98 hours MAE and 35.55 hours RMSE for the in-domain experiment on FM. Although the same improvement does not hold for the MC events, these results suggest that mixing different domains may contribute to a better overall performance.

Finally, we looked into features that enable the FM-based model to yield precise estimates on the MC data set as well. Successful features relate to sub-events related to large public events that require people to travel, park, and queue: {\em geparkeerd} `parked', {\em inpakken, richting} `take, direction', {\em afhalen} `pick up', {\em trein, vol} `train, full', {\em afwachting} `anticipation', {\em langzaam, vol} `slowly, full', {\em wachtend} `waiting', and {\em rij, wachten} `queue, waiting'. Moreover features such as {\em half, uurtje} `half, hour' prove that WFeats can learn temporal expressions as well. Some example TFeats features learned from FM that were useful with the MC data set are {\em over 35 minuten} `in 35 minutes', {\em rond 5 uur} `around 5 o'clock', {\em nog een paar minuutjes} `another few minutes', {\em nog een nachtje, dan} `one more night, then', and {\em nog een half uur, dan} `one more half an hour, then'. These results suggest that the models can be transferred across domains successfully to a fair extent.

\subsection{Conclusion}
\label{conclusionsNLPT}

We have presented a time-to-event estimation method that is able to infer the starting time of an event from a stream of tweets automatically by using linguistic cues. It is designed to be general in terms of domain and language, and to generate precise estimates even in cases in which there is not much information about an event.

We tested the method by estimating the TTE from single tweets referring to football matches and music concerts in Dutch. We showed that estimates can be as accurate as under 4 hours and 10 hours off, averaged over a time window of eight days, for football matches and music concerts respectively.

The method provided best results with an ordered procedure that prefers the prediction of temporal logic rules, which are estimation rules, and then backs off to estimates based on temporal expressions, followed by word skipgram features, of which the individual TTEs are estimated through median training. Comparing the precision of three types of features we found that temporal logic rules and an extensive temporal expression list outperformed word skipgrams in generating accurate estimates. On the other hand, word skipgrams demonstrated the potential of covering temporal expressions. We furthermore showed that integrating historical context based on previous estimates improves the overall estimation quality. Closer analysis of our results also reveals that estimation quality improves as more data are available and the event is getting closer. Finally, we presented results that hint at the possibility that optimized parameters and trained models can be transferred across domains.

\section{Conclusion}

We reported on our research that aims at characterizing the event information about start time of an event on social media and automatically using it to develop a method that can reliably predict time to event in this chapter. Various feature extraction and machine learning algorithms were explored in order to find the best combination for this task. 

As a result, we developed a method that produces accurate estimates using skipgrams and, in case available, is able to benefit from temporal information available in the text of the posts. Time series analysis techniques were used to combine these features for generating an estimate and integrate that estimate with the previous estimates for that event.

The studies in this chapter are performed on tweets collected using hashtags. However, hashtags enable collection of only a proportion of the tweets about an event and can be misleading. Therefore, we studied finding relevant posts in a collection of tweets collected using key terms in the following chapter. Detecting a big proportion of the tweets about an event, in addition to enhancing our understanding about an event, will increase chances of producing accurate time-to-event estimates for that event.

\lhead{\emph{Relevant Microtext Detection}}
\chapter{Relevant Document Detection}
\label{chapter3}

This chapter is based on the following studies:

Hürriyetoğlu, A., Gudehus, C., Oostdijk, N., \& van den Bosch, A. (2016). Relevancer: Finding and Labeling Relevant Information in Tweet Collections. In {\em E. Spiro \& Y.-Y. Ahn (Eds.), Social Informatics: 8th International Conference, SocInfo 2016, Bellevue, WA, USA, November 11-14, 2016, Proceedings, Part II (pp. 210–224)}. Cham: Springer International Publishing. Available from \url{http://dx.doi.org/10.1007/978-3-319-47874-6_15 doi: 10.1007/978-3-319-47874-6,_15}

Hürriyetoğlu, A., van den Bosch, J., \& Oostdijk, N. (2016a, September). Analysing the Role of Key Term Inflections in Knowledge Discovery on Twitter. In {\em Proceedings of the 1st international workshop on knowledge discovery on the web}. Cagliari, Italy. Available from \url{http://www.iascgroup.it/kdweb2016-program/accepted-papers.html}

Hürriyetoğlu, A., van den Bosch, A., \& Oostdijk, N. (2016b, December). Using Relevancer to Detect Relevant Tweets: The Nepal Earthquake Case. In {\em Working notes of FIRE 2016 - Forum for Information Retrieval Evaluation}. Kolkata, India. Available from \url{http://ceur-ws.org/Vol-1737/T2-6.pdf}

Hürriyetoǧlu, A., Oostdijk, N., Erkan Başar, M., \& van den Bosch, A. (2017). Supporting Experts to Handle Tweet Collections About Significant Events. In {\em F. Frasincar, A. Ittoo, L. M. Nguyen, \& E. Métais (Eds.), Natural Language Processing and Information Systems: 22nd International Conference on Applications of Natural Language to Information Systems, NLDB 2017, Liège, Belgium, June 21-23, 2017, Proceedings (pp. 138–141)}. Cham: Springer International Publishing. Available from \url{https://doi.org/10.1007/978-3-319-59569-6_14} 

\section{Introduction}
Microtexts on Twitter comprise a relatively new document type. Tweets are typically characterized by their shortness, impromptu nature, and the apparent lack of traditional written text structure. Unlike some news or blogging platforms, Twitter lacks facilities such as a widely known labeling system or a taxonomy for categorizing or tagging the posts. Consequently, collecting and analyzing tweets holds various challenges~\cite{Imran+15}. When collecting tweets, the use of one or more key terms in combination with a restriction to a particular geographical area or time period is prone to cause the final collection to be incomplete or unbalanced~\cite{Olteanu+14}, which may hamper our ability to leverage the available information, as we are unable to know what we have missed.\footnote{Hashtag use guarantees tweet set management to some extent and even then only for a limited set of users who are aware of what hashtag is associated with a particular topic or event. Related posts that do not carry a particular hashtag will be missed.}

Users who want to gather information from a keyword-based collection of tweets will often find that not all tweets are relevant for the task at hand. Tweets can be irrelevant for a particular task for a range of different reasons, for instance because they are posted by non-human accounts (bots), contain spam, refer to irrelevant events, or refer to an irrelevant sense of a keyword used in collecting the data. This variety is likely to be dynamic over time, and can be present in a static or continuously updated collection as well. 

In order to support the user in managing tweet collections, we developed a tool, Relevancer. Relevancer organizes content in terms of {\em information threads}. An information thread characterizes a specific, informationally related set of tweets. Relatedness is determined by the expert who uses the method. For example, the word `flood' has multiple senses, including `to cover or fill with water' but also `to come in great quantities'.\footnote{\url{http://www.oed.com/view/Entry/71808}, accessed June 10, 2018} A water manager will probably want to focus on only the water-related sense. At the same time, she will want to discriminate between different contextualizations of this sense: past, current, future events, effects, measures taken, etc. By incrementally clustering and labeling the tweets, the collection is analyzed into different information threads to enable this kind of task to organize the content presented in tweets efficiently.

Relevancer enables an expert to explore and label a tweet collection, for instance any set of tweets that has been collected by using keywords. The tool requires expert\footnote{An expert can be anybody who is able to make informed decisions about how to annotate tweet clusters in order to understand a tweet collection in a certain context in which she is knowledgeable.} feedback in terms of annotations of individual tweets, or sets (or clusters) of tweets in order to identify and label the information threads. Relevancer follows a strategy in which tweets are clustered by similarity, so that annotations can be applied efficiently to coherent clusters instead of to tweets individually. 
Our method advances the state of the art in terms of the efficient and relatively complete understanding and management of a non-standard, rich, and dynamic data source.

The strength of our approach is the ability to scale to a large collection while maintaining reasonable levels of precision and recall by understanding intrinsic characteristics of the used key terms on social media. Moreover, Relevancer shares the responsibility to strive for completeness and precision with the user.

This chapter reports four use cases that illustrate how Relevancer can be used to assist task holders in exploring a microtext collection, defining which microtexts are relevant to their needs, and using this definition to classify new microtexts in an automated setting using Relevancer. In each case Relevancer is used with a specific configuration and with incremental improvements to the applied method.

Our first use case describes the analysis of a tweet collection we gathered using the key terms `Rohingya' and `genocide' ~\cite{Hurriyetoglu+16b}. Next we investigate the effect of keyword inflections in a study involving a collection of tweets sampled through the English word `flood' ~\cite{Hurriyetoglu+16a}. The third use case reports on our participation in a shared task about analyzing earthquake-related tweets ~\cite{Hurriyetoglu+16c}. The fourth and final case study is about analyzing Dutch tweets gathered using the word `griep'. This use case demonstrates the application of our approach to the analysis of tweets in a language other than English and the performance of the generated classifier.

The chapter continues with related research in Section~\ref{relResCH3}. Then, in Section~\ref{relevancerCH3} Relevancer is described in more detail. The use cases are presented in Sections \ref{SectionSocInfo}, \ref{SectionKDWeb}, \ref{SectionFIRE}, and \ref{SectionNLDB} respectively. Finally, we draw overall conclusions in Section~\ref{conlusionCH3}.

\section{Related Research}
\label{relResCH3}
Identifying different uses of a word in different contexts has been subject of lexicographical studies~\cite{Eikmeyer+81}. Starting with Wordnet~\cite{Fellbaum98}, this field has benefited from computational tools to represent lexical information, which enabled the computational study of word meaning. On the basis of these resources and of corpora annotated with word sense information, word sense induction and disambiguation tasks were identified and grew into an important subfield of computational linguistics~\cite{Navigli09,Pedersen+06,Mccarthy+16}. This task is especially challenging for tweets, as they have a limited context~\cite{Gella+14}. Moreover, the diversity of the content found on Twitter~\cite{DeChoudhury+11} and the specific information needs of an expert require a more flexible definition than a sense or topic. This necessity led us to introduce the term `information thread', which can be seen as the contextualization of a sense.

Popular approaches of word sense induction on social media data are Latent Dirichlet allocation (LDA)~\cite{Gella+13,Lau+12}, and user graph analysis~\cite{Yang+15}. The success of the former method depends on the given number of topics, which is challenging to determine, and the latter assumes the availability of user communities. Both methods provide solutions that are not flexible enough to allow users to customize the output, i.e. the granularity of an information thread, based on a particular collection and the specific needs of an expert or a researcher. Therefore, Relevancer was designed in such a fashion that it is possible to discover and explore information threads without any a priori restrictions. 

\begin{sloppypar}
Since a microtext collection contains an initially unknown number of information threads, we propose an incremental-iterative search for information threads in which we include the human in the loop to determine the appropriate set of information threads. By means of this approach, we can manage the ambiguity of key terms as well as the uncertainty of the information threads that are present in a tweet collection.\footnote{The article in the following URL provides an excellent example of the ambiguity caused by lexical meaning and syntax: \url{http://speld.nl/2016/05/22/man-rijdt-met-180-kmu-over-a2-van-harkemase-boys/}, accessed June 10, 2018.} This solution enables experts both to spot the information they are looking for and the information that they are not aware exists in that collection.
\end{sloppypar}

Social science researchers have been seeking ways of utilizing social media data~\cite{Felt16} and have developed various tools~\cite{Borra+14} to this end.\footnote{Additional tools are {\url{https://wiki.ushahidi.com/display/WIKI/SwiftRiver}, accessed June 10, 2018}, {\url{https://github.com/qcri-social/AIDR/wiki/AIDR-Overview}, accessed June 10, 2018}, and {\url{https://github.com/JakobRogstadius/CrisisTracker}, accessed June 10, 2018}} Although these tools have many functions, they do not focus on identifying the uses of key terms. A researcher should continue to navigate the collection by specific key term combinations. Our study aims to enable researchers to discover expected as well as unforeseen uses of initial key terms. The practically enhanced understanding enables the analysis to be relatively precise, complete, and timely.


Enabling human intervention is crucial to ensuring high level performance in text analytics approaches~\cite{Chau12,Tanguy+16}. 
Our approach responds to this challenge and need by allowing the human expert to provide feedback at multiple phases of the analysis.


\section{Relevancer}
\label{relevancerCH3}
The Relevancer tool addresses a specific case for collecting and analyzing data using key terms from Twitter. Since the use and the interpretation of the key terms depends partly on the social context of a Twitter user and the point in time this term is used, often the senses and nuances or aspects that a word may have on social media cannot all be found in a standard dictionary. Therefore, Relevancer focuses on the automatic identification of sets of tweets that contain the same contextualization of a sense, namely tweets that convey the same meaning and nuance (or aspect) of a key term. We refer to such sets of tweets as information threads. 

The information thread concept allows a fine-grained management of all uses of a keyword. In the case of this study, this approach enables the user of the tool to focus on uses of a key term at any level of granularity. For instance, tweets about a certain event, which takes place at a certain time and place, and tweets about a type of event, without a particular place or time, can be processed either at the same level of abstraction or can be treated as separate threads, depending on the needs of the user.

While highly ambiguous words, such as `flood', require a proper analysis to identify and discriminate among its multiple uses in a collection, words that are relatively less ambiguous often need this analysis as well. For instance, a social scientist who collects data using the key term `genocide' may want to focus exclusively on the `extermination' sense of this word\footnote{\url{http://www.oed.com/view/Entry/77616}, accessed June 10, 2018} or on specific information threads that may be created and that related to past cases, current, possible future events, effects, measures taken, etc.

\begin{sloppypar}
Relevancer can help experts to come to grips with event-related microtext collections. The development of Relevancer has been driven and benefited from the following insights:
\begin{inparaenum}[(i)]
\item almost every event-related tweet collection 
contains also tweets about similar but irrelevant events~\cite{Hurriyetoglu+16b};
\item by taking into account the temporal distribution of the tweets about an event it is possible to achieve an increase in the quality of the information thread detection and a decrease in computation time~\cite{Hurriyetoglu+16c}; 
and
\item the use of inflection-aware key terms can decrease the degree of ambiguity~\cite{Hurriyetoglu+16a}.
\end{inparaenum}
\end{sloppypar}

This section describes our methodology and the way in which in Relevancer microtexts are selected, classified, and processed. The main steps are pre-processing, feature extraction, near-duplicate detection, information thread detection, cluster annotation, and classifier creation. Each of these steps is described in one of the following subsections below, from ~\ref{ch3DataPrep} to ~\ref{ch3CreateClassifier}. Finally, we discuss the scalability of the approach in \ref{ch3Scalability}.



\subsection{Data Preparation}
\label{ch3DataPrep}
A raw tweet collection that has been collected using one or more key terms consists of original posts and (near-)duplicates of these posts (retweets and posts with the nearly same content). The tweet content may contain URLs and user names. 
First we exclude the retweets, then normalize the URLs and user names, and finally remove the exact duplicates, for the following reasons.

Any tweet that has an indication of being a retweet is excluded because they do not contribute new information, except user information and how often a tweet is shared, but in all described studies we ignore user identity and the frequency of sharing a certain tweet. We use two types of retweet detection methods in a tweet. Retweets are detected on the basis of the retweet identifier of the tweet's JSON file or the existence of ``RT @'' at the beginning of a tweet text.

We proceed with normalizing the user names and URLs to ``usrusrusr'' and ``urlurlurl'' respectively. The normalization eliminates the token noise introduced by the huge number of different user names and URLs while preserving the abstract information that a user name or a URL is present in a tweet.

Finally, we detect and exclude exact duplicate tweets after normalizing the user names and URLs. We leave only one sample of these duplicate tweets. The importance of this step can be appreciated when we identify tweets such as exemplified in examples~\ref{examplegenocide1} and~\ref{examplegenocide2} below, which each were posted 5,111 and 2,819 times respectively.

\begin{exmp}
\label{examplegenocide1}
{\em usrusrusr The 2nd GENOCIDE against \#Biafrans as promised by \#Buhari has begun,3days of unreported aerial Bombardment in \#Biafraland}
\end{exmp}

\begin{exmp}
\label{examplegenocide2}
{\em .usrusrusr New must read by usrusrusr usrusrusr A genocide \& Human trafficking at a library in Sweden urlurlurl}
\end{exmp}


We have to note that by excluding duplicate tweets in terms of their text, we loose information about different events that are phrased in the same way. We consider this limitation as not affecting the goal of our study, since we focus on large scale events that are phrased in multiple ways by multiple people.

A final data preparation phase, which is near-duplicate elimination, is applied after the feature extraction step.

\subsection{Feature Extraction}

We represent tweets by features that each encode the occurrence of a string in the tweet. Features represent string types, or types for short, which may occur multiple times as tokens. Types are any space-delimited single or sequences of alphanumeric characters, combinations of special characters and alphanumeric characters such as words, numbers, user names (all normalized to a single type) and hashtags, emoticons, emojis, and sequences of punctuation marks.\footnote{Punctuation mark sequences are treated differently. Sequences of punctuation marks comprising two, three or four items are considered as one feature. If the punctuation marks sequence is longer than four, we split them from left to right in tokens of length 4 by ignoring the last part if it is a single punctuation mark. The limit of length 4 is used for punctuation mark combinations, since longer combinations can be rare, which may cause them not to be used at all.} 
Since none of the steps contains any language-specific optimizations, and the language and task-related information is provided by a human user, we consider our method to be language-independent.\footnote{For languages with scripts that do not have word segmentation (e.g. the space), we need to adapt the tokenizer.}

On the basis of tokenized text we generate unigrams and bigrams as features in the following steps of our method. We apply a dynamically calculated occurrence threshold to these features. The threshold is half of the log of the number of tweets, to base {\em e}, in a collection. If the frequency occurrence of a feature is below the threshold, it is excluded from the feature set. For instance, if the collection contains 100,000 tweets, the frequency cut-off will be 11, which is rounded down from 11.51.

\subsection{Near-duplicate Detection}

Most near-duplicate tweets occur because the same quotes are being used, the same message is tweeted, or the same news article is shared with other people. Since duplication does not add information and may harm efficiency and performance of the basic algorithms, we remove near-duplicate tweets by keeping only one of them in the collection. 

The near-duplicate tweet removal algorithm is based on the features described in the previous subsection and the cosine similarity of the tweets calculated based on these features. If the pairwise cosine similarity of two tweets is larger than 0.8, we assume that these tweets are near-duplicates of each other. In case the available memory does not allow for all tweets to be processed at once, we apply a recursive bucketing and removal procedure. The recursive removal starts with splitting the collection into buckets of tweets of a size that can be handled efficiently. After removing near-duplicates from each bucket, we merge the buckets and shuffle the remaining tweets, which remain unique after the recent iteration in their respective bucket. We repeat the removal step until we can no longer find duplicates in any bucket. 

For instance, the near-duplicate detection method recognizes the tweets in the examples~\ref{exampleActorMatt1} and~\ref{exampleActorMatt2} below as near-duplicates and leaves just one of them in the collection.

\begin{exmp}
\label{exampleActorMatt1}
{\em Actor Matt Dillon puts rare celebrity spotlight on Rohingya urlurlurl \#news}
\end{exmp}

\begin{exmp}
\label{exampleActorMatt2}
{\em urlurlurl Matt Dillon puts rare celebrity spotlight on Rohingya… urlurlurl} 
\end{exmp}


\subsection{Information Thread Detection}

The information thread detection step aims at finding sets of related tweets. In case a tweet collection was compiled using a particular keyword, different information threads may be detected by grouping tweets in sets where different contextualizations of the keyword are at play. In case the tweet collection came about without the use of a keyword, the thread detection step will still find related groups of tweets that are about certain uses of the words and word combinations in this collection. These groups will represent information threads.


The clustering process aims to identify coherent clusters of tweets. Each cluster ideally constitutes an information thread. We facilitate a classic and basic clustering algorithm, K-Means for collections smaller than 500 thousand tweets and MiniBatch K-Means for larger ones.\footnote{We used scikit-learn v0.17.1 for all machine learning tasks in this study (\url{http://scikit-learn.org}, accessed June 10, 2018).} We repeat the clustering and cluster selection steps in iterations until we reach the requested number of coherent clusters, which is provided by the expert.\footnote{We start the clustering with a relatively small k and increase k in the following iterations. Therefore, we consider this approach relaxes the requirement of determining best k before running the algorithm. The search for information threads may stop earlier in case a pre-determined number of good quality clusters is reached or the cluster coherence criteria reach unacceptable values.} Clusters that fit our coherence criteria are picked for annotation and the tweets they contain are excluded from the following iteration of clustering, which focuses on all tweets that have not been yet clustered and annotated. This procedure iterates until one of the stop criteria that are described in the following subsection is satisfied.

We observed that in tweet collections there are tweets that do not bear any clear relation to other tweets independent of being relevant or irrelevant to the use case. These tweets are either form incoherent clusters, which are erroneously identified as coherent by the clustering algorithm, or not placed in any cluster. The incoherent clusters are identified by the expert at the annotation step and excluded from the cluster set. The tweets that are not placed in any cluster remain untouched. We expect the available clusters that will be annotated by the experts to serve as training data and facilitate classification of the tweets in incoherent clusters and outlier tweets as relevant or not.


The clustering step of Relevancer facilitates three levels of parameters. First, the clustering parameters depend on the collection size in number of tweets (\textit{n}) and the time spent searching for clusters. Second, the cluster coherence parameters control the selection of clusters for annotation. Third, the parameter specifying the requested number of clusters sets the limit for the algorithm to stop in terms of cluster number. The clustering and coherence parameters are updated at each iteration (\textit{i}) based on the requested number of clusters (\textit{r}) and the number of already detected clusters in previous iterations. The function of these parameters is as follows.

\begin{description}

\item [Clustering parameters] There are two clustering parameters. \textit{k} is the number of expected clusters and \textit{t} is the number of initializations of the clustering algorithm before it delivers its result. These parameters are set at the beginning and updated at each iteration automatically. The value of the parameter \textit{k} of the K-Means algorithm is determined by Equation~\ref{keq} at each iteration. The parameter \textit{k} is equal to half of the square root of the tweet collection size at that iteration plus the number of previous iterations times the difference between the requested number of clusters and the detected number of clusters (\textit{a}). This adaptive behavior ensures that if we do not have sufficient clusters after several iterations, we will be searching for smaller clusters at each iteration.

\item [Coherence parameters] The three coherence parameters set thresholds for the distribution of the instances in a cluster relative to cluster center. The allowed Euclidean distance of the closest and farthest instance to the cluster center and the difference between these two parameters enables selecting coherent clusters algorithmically. Although these parameters have default values that are strict, they can be set by the expert as well. Adaptation, which is about relaxing the criteria, of the cluster coherence criteria steps is small if we are close to our target number of clusters.

\item [Requested number of clusters] The third layer of the parameters contains the requested number of clusters that should be generated for the expert (\textit{r}). This parameter is given as a stopping criterion for the exploration and as an indicator of the adaptation step for the value of the other parameters at each cycle. Since the available cluster number and value of this parameter are compared at the end of an iteration, the clustering step may yield more clusters than the requested number. In order to limit excessive number of clusters, the values of the coherence parameters are increased less as the available number of clusters gets closer to requested number.

\end{description}


The result of the clustering, which is evaluated by the coherence criteria, is the best score in terms of inertia after initializing the clustering process (\textit{t}) times in an iteration. As provided in Equation~\ref{teq}, (\textit{t}) is the log of the size of the tweet collection in number of tweets, to base 10, at the current iteration plus the number of iterations performed until that point times (\textit{t}), Equation~\ref{teq}. This formula ensures that the more time it takes to find coherent clusters, the more often the clustering will be initialized before it delivers any result.

\begin{equation}
k = {\frac{\sqrt n_{i}} 2} + (i\times({r - a_{i} }))
\label{keq}
\end{equation}

\begin{equation}
t = {\log_{10} n_{i}} + i
\label{teq}
\end{equation}

Some collections may not contain the requested number of coherent clusters as defined by the coherence parameters. In such a case, the adaptive relaxation (increase) of the coherence parameters stops at a level where the values of these parameters are too large to enable a sensible coherence-based cluster selection.\footnote{\label{coherencethresholds}This behavior is controlled by the coherence parameters \textit{min\_dist\_thres} and \textit{max\_dist\_thres}. The closest and farthest instances to the cluster center must not exceed \textit{min\_dist\_thres} and \textit{max\_dist\_thres} respectively. If the values of \textit{min\_dist\_thres} and \textit{max\_dist\_thres} exceed 0.85 and 0.99 at the same time due to adaptive increase in an iteration, we assume that the search for coherent clusters must stop.} This becomes the last iteration in search for coherent clusters. We think it is unrealistic to expect relatively good clusters in such a situation. In such a case, the available clusters are returned before they reach the number requested by the expert. 


\subsection{Cluster Annotation}

Automatically selected clusters are presented to an expert for identifying clusters that present certain information threads.\footnote{The annotation is designed to be done or coordinated by a single person in our setting.} Tweets in these clusters are presented based on their distance to the cluster center; the closer they are the higher their rank. After the annotation, each thread may consist of one or more clusters. In other words, similar clusters should be annotated with the same label. Clusters that are not clear or fall outside the focus of a Relevancer session should be labeled as incoherent and irrelevant respectively. This decision is taken by a human expert. The tweets that are in an incoherent labeled cluster are treated in a similar fashion as tweets that were not placed in any coherent cluster.

The example presented in Table~\ref{table:chVSin} shows tweets that are the closest to and farthest from the cluster center. The tweets closest to the cluster centre form a coherent set, those farthest removed clearly do not. Tweets in a coherent cluster (CH) have a clear relation that allows us to treat this group of tweets as pertaining to the same information thread. Incoherent clusters are summarized under IN1 and IN2. In the former group, the tweets are unrelated.\footnote{The expert may prefer to tolerate a few different tweets in the group in case the majority of the tweets are coherent and treat the cluster as coherent.} The latter group contains only a meta-relation that can be considered as an indication of being about liking some video, the rest of these posts are not related. The granularity level of the information thread definition determines the final decision for this particular cluster. In case the expert decides to define the labels as being about a video, the cluster will be annotated as coherent. Otherwise, this cluster should be annotated as incoherent.

\begin{table}[htb]
\ttabbox{\caption{Tweets that are the closest and the farthest to the cluster center for coherent (CH), incoherent type 1 (IN1) and type 2 (IN2) clusters }\label{table:chVSin}}
{\begin{tabular*}{0.87\textwidth}{|p{0.9cm}|p{11cm}|}
    
    \hline
    \multirow{2}{*}{CH}& $-$ myanmar rejects 'unbalanced' rohingya remarks in oslo (from usrusrusr urlurlurl\\
    & $-$ shining a spotlight on \#myanmar's \#rohingya crisis: usrusrusr remarks at oslo conf on persecution of rohingyas urlurlurl \\
    
    \hline
    \multirow{2}{*}{IN1}& $-$ un statement on \#burma shockingly tepid and misleading, and falls short in helping \#rohingya says usrusrusr usrusrusr urlurlurl\\
    & $-$ usrusrusr will they release statement on bengali genocide 10 months preceding '71 ?\\
    
    \hline
    \multirow{2}{*}{IN2}& $-$ i liked a usrusrusr video urlurlurl rwanda genocide 1994\\
    & $-$ i liked a usrusrusr video urlurlurl fukushima news genocide; all genocide paths lead to vienna and out of vienna \\
    \hline
\end{tabular*}}
\label{tab:multicol}
\end{table}

For instance, clusters that contain tweets like ``plain and simple: genocide urlurlurl'' and ``it's genocide out there right now'' can be gathered under the same label, e.g., actual ongoing events. If a cluster of tweets is about a particular event, a label can be created for that particular event as well. For instance, the tweet ``the plight of burma’s rohingya people urlurlurl'' is about a particular event related to the Rohingya people. If we want to focus on this event related to Rohingya, we should provide a specific label for this cluster and use it to specify this particular context. We can use `plight' as a label as well. In such a case, the context should specify cases relevant to this term.

In case the expert is not interested in or does not want to spend time on defining a separate label for a coherent cluster, the expert can attach the label irrelevant, which behaves as an umbrella label for all tweet groups (information threads) that are not the present focus of the expert. 

We developed a web-based user interface for presenting the clusters and assigning a label to the presented cluster.\footnote{\url{http://relevancer.science.ru.nl}, accessed June 10, 2018} We present all tweets in a cluster if the number of tweets is smaller than 20. Otherwise, we present the first and the last 10 tweets of a cluster. As explained before, the order of the tweets in a cluster is based on the relative distance to the cluster center; the first tweets are closest to the center, the last most removed from it. This setting provides an overview and enables spotting incoherent clusters effectively. A cluster can be skipped without providing a label for it. In such a case, that cluster is not included in the training set. If an expert finds that a cluster represents an information thread she is not currently interested in, the irrelevant label should be attached to that cluster explicitly.

At the end of this process, an expert is expected to have defined the information threads for this collection and have identified the clusters that are related to these threads. Tweets that are part of a certain information thread can be used to understand the related thread or to create an automatic classifier that can classify new tweets, e.g., ones that were not included in any selected cluster at the clustering step, in classes based on detected and labeled information threads.

\subsection{Creating a Classifier}
\label{ch3CreateClassifier}

Creating classifiers for tweet collections is a challenge that is mainly affected by the label selection of the expert annotator, the ambiguity of the keyword if the tweet collection was keyword-based, and the time period in which the tweets in the collection were posted. This time period may contain a different pattern of occurrences than seen in other periods. Consequently, the classes may be imbalanced or unrepresentative of the expected class occurrence patterns when applied to new data. 

The labeled tweet groups are used as training data for building automatic classifiers that can be applied `in the wild' to any earlier or later tweets, particularly if they are gathered while using the same query.

Relevancer facilitates the Naive Bayes and Support Vector Machine (SVM) algorithms for creating a classifier. These are classic, baseline supervised machine learning techniques. Naive Bayes and SVM have been noted to provide a comparable performance to sophisticated machine learning algorithms for short text classification~\cite{WangS+12}. We use Naive Bayes in case we need a short training time. We need time efficiency in cases where an expert prefers to update the current classifier frequently with new data or create another classifier after observing the results of a particular classifier. The other option, SVM, is used when either the training data is relatively small, which is mostly smaller than 100,000 microtexts, or the classifier does not need to be updated frequently. The experts determines which option suits their case.

The parameters of the machine learning algorithms are optimized by using a grid search on the training data. The performance of the classifier is evaluated based on 15\% of the training data, which is held out and not used at the optimization step. After optimization, the held-out data can be used as part of the training data.

\subsection{Scalability}
\label{ch3Scalability}

Relevancer applies various methods in order to remain scalable independently of the number of tweets and features used. The potentially large number of data points in tweet collections are the reason why this tool has scalability at the center of its design.

Processing is done by means of basic and fast algorithms. As observed above, depending on the size of the collection, K-means or MiniBatch K-Means algorithms are employed in order to rapidly identify candidate clusters. The main parameter k, the number of clusters parameter for K-Means, in these algorithms is calculated at each iteration in order to find more and smaller clusters than the previous iteration.\footnote{Equation~\ref{keq} enables this behavior.} Targeting more and smaller clusters increases the chance of identifying coherent clusters.

Tweets in coherent clusters are excluded from the part of the collection that enters the subsequent clustering iteration. This approach shrinks the collection size at each iteration. Moreover, the criteria for coherent cluster detection are relaxed at each step until certain thresholds are reached in order to prevent the clustering from being repeated too many times.

Finally the machine learning algorithm selection step allows the appropriate option to be chosen from the scalability point of view. For instance, the Naive Bayes classifier was chosen in order to create and evaluate a classifier within a reasonable period of time. The speed of this step enables users to decide whether they will use a particular classifier or need to generate and annotate additional coherent clusters immediately. This optimized cycle enables experts to provide feedback frequently without having to wait too long. As a result, the quality of the results increases with minimal input and time needed for a particular task. SVM should be used when fast learning is not required and sufficient amounts of training data are available.

The use of this approach is illustrated through Sections~\ref{SectionSocInfo} to \ref{SectionNLDB}. Each section focuses on a particular use case and the incremental improvements that were made to Relevancer.

\section{Finding and Labeling Relevant Information in Tweet Collections}
\label{SectionSocInfo}

The first use case we present is a study in which the Relevancer tool is evaluated on a tweet collection based on the key terms `genocide' and `Rohingya'. We retrieved a tweet collection from the public Twitter API~\footnote{\url{https://dev.twitter.com/rest/public}, accessed June 10, 2018} with these key terms between May 25 and July 7, 2015. The collection consists of 363,959 tweets. The number of tweets that contain only `genocide' or only `Rohingya' are 109,629 and 241,441 respectively; 12,889 tweets contain both terms.

\begin{figure}[htb]
\begin{center}
\includegraphics[scale=0.90]{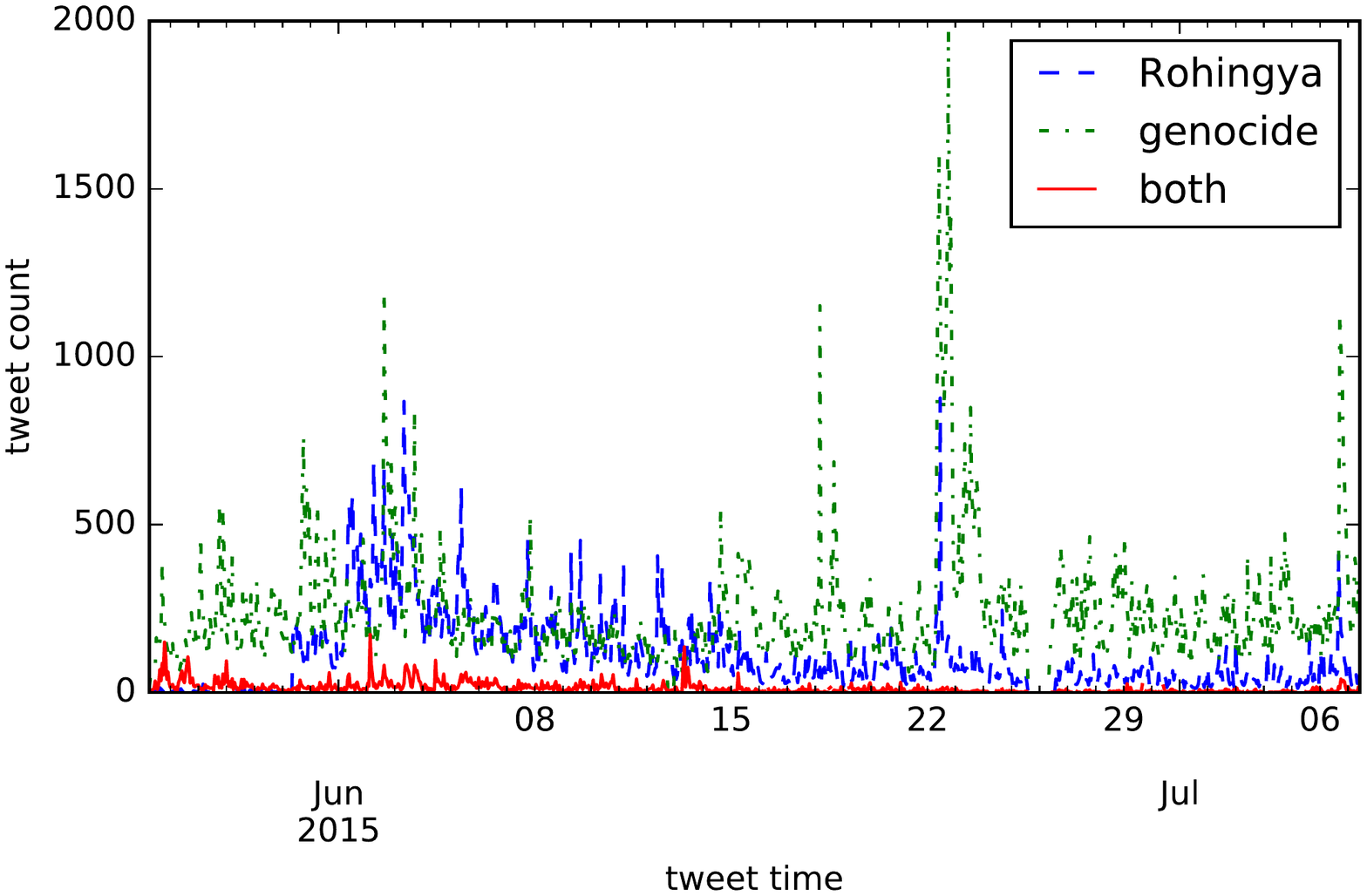}
\caption{Tweet distribution per key term}
\label{fig:tweetdist1}
\end{center}
\end{figure}

Figure~\ref{fig:tweetdist1} shows the distribution of the tweets, which contains a gap of a few days around July 26, 2015 due to a problem encountered during the data collection, for each subset. We can observe that there are many peaks and a relatively stable tweet ratio for each key term. Our analysis of the collection aims at understanding how the tweets in the peaks differ from the tweets that occur as part of a more constant flow of tweets.

\begin{figure}[htb]
\begin{center}
\includegraphics[scale=0.70]{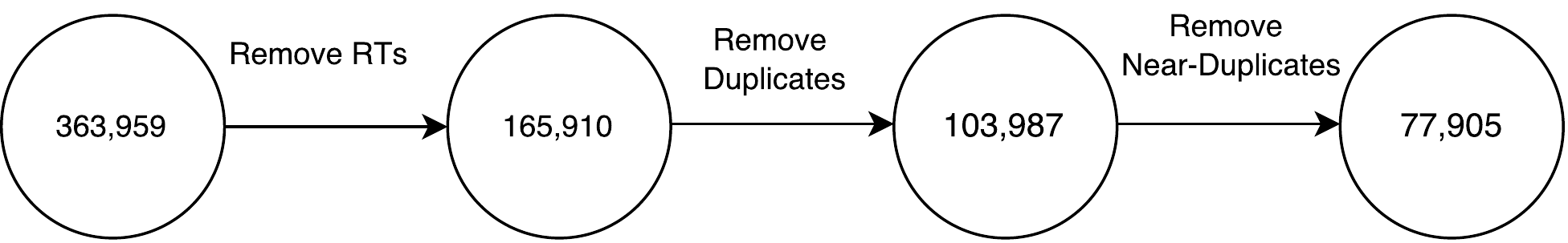}
\caption{Tweet volume change at each step of the preprocessing in the collection}
\label{fig:tweetsizeSCINF}
\end{center}
\end{figure}

As a first step we begin by cleaning the tweet collection. In all 198,049 retweets, 61,923 exact duplicates, and 26,082 near-duplicates are excluded from the original collection of 363,959 tweets, leaving 77,905 tweets in the collection. The summary of the evolution of the size of the tweet collection is presented in Figure~\ref{fig:tweetsizeSCINF}. At each step, a large portion of the data is detected as repetitive and excluded. This cleaning phase shows how the size of the collection depends on the preprocessing. 

\begin{figure}[htb]
\begin{center}
\includegraphics[scale=0.65]{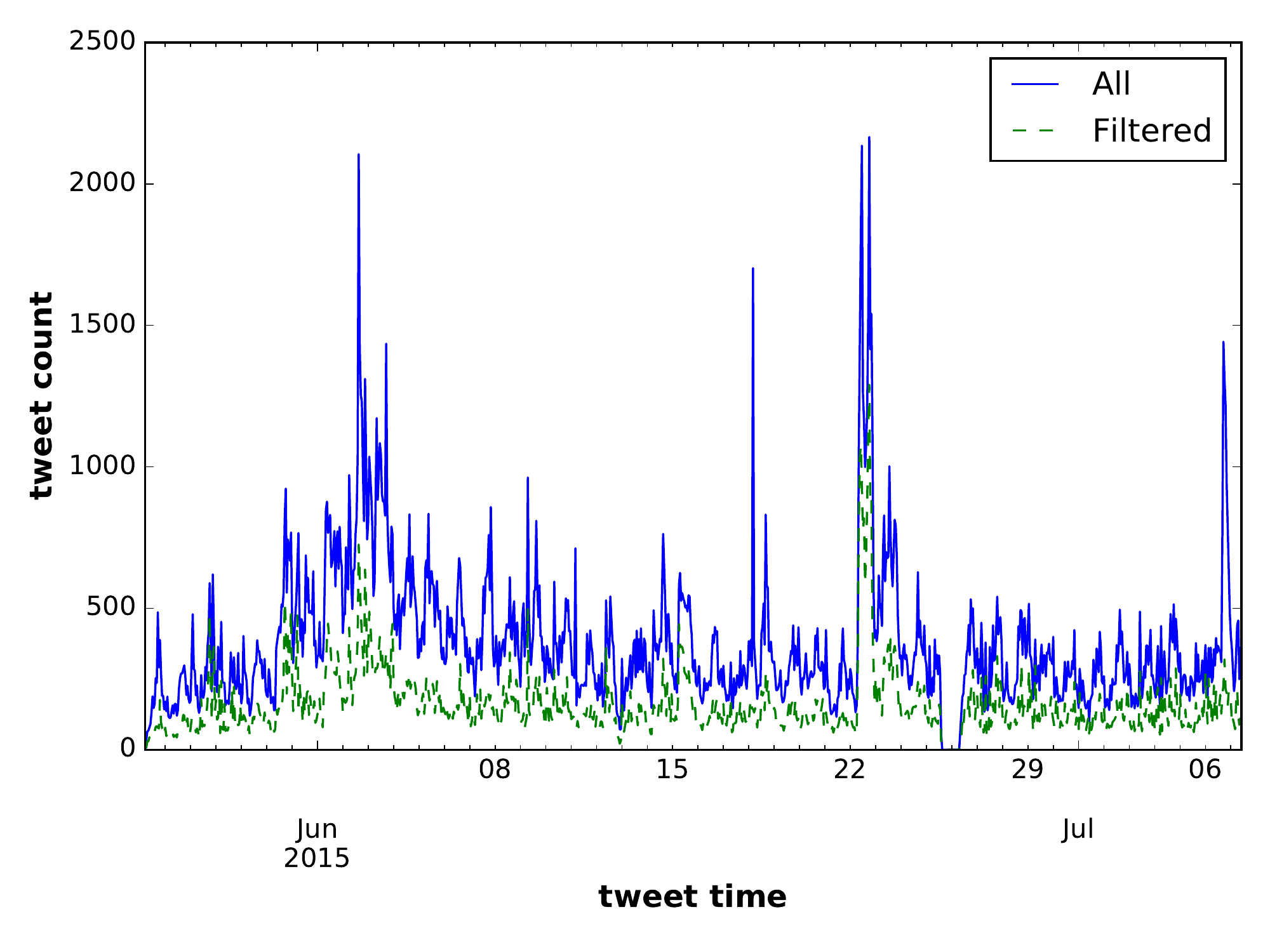}
\caption{Tweet volume change at each step of the preprocessing in the collection}
\label{fig:tweetsizeBeforeAfterCleaningSCINF}
\end{center}
\end{figure}

We provide the total tweet count distributions before and after cleaning steps, which are the lines `All' and `Filtered' respectively, in Figure~\ref{fig:tweetsizeBeforeAfterCleaningSCINF}. 
The Figure illustrates that peaks and trends in the tweet count are generally preserved. The reduced size of the data enables us to apply sophisticated analysis techniques to social media data without loosing the main characteristics of the collection.\footnote{We note that the repetition pattern analysis is valuable in its own right. However, this information is not within the scope of the present study.}

\begin{figure}[htb]
\begin{center}
\includegraphics[scale=0.65]{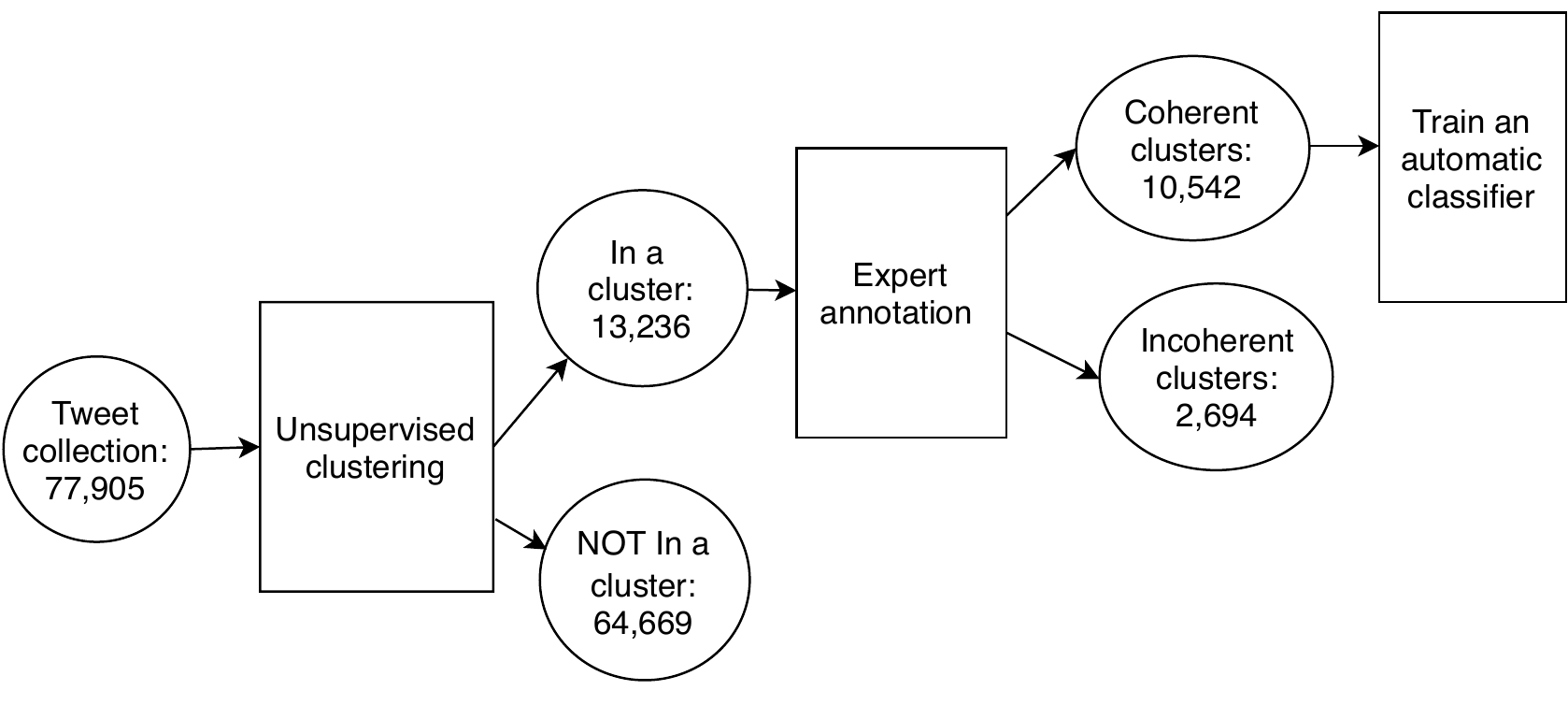}
\caption{Phases in the analysis process with the number of tweets at each step starting after duplicate and near-duplicate elimination}
\label{fig:sysdiagSCINF}
\end{center}
\end{figure}

Next, as described in the previous sections in more detail, detailed features are extracted. The data are then clustered yielding clusters of tweets to which an expert can attach labels. The labeled tweets are used to train an automatic classifier, which can be used to analyze the remaining tweets or a new collection. The analysis steps after duplicate and near-duplicate elimination of the tool are presented in Figure~\ref{fig:sysdiagSCINF}. The analysis steps and the results are explained below.


\begin{figure}[htb]
\begin{center}
\includegraphics[scale=0.90]{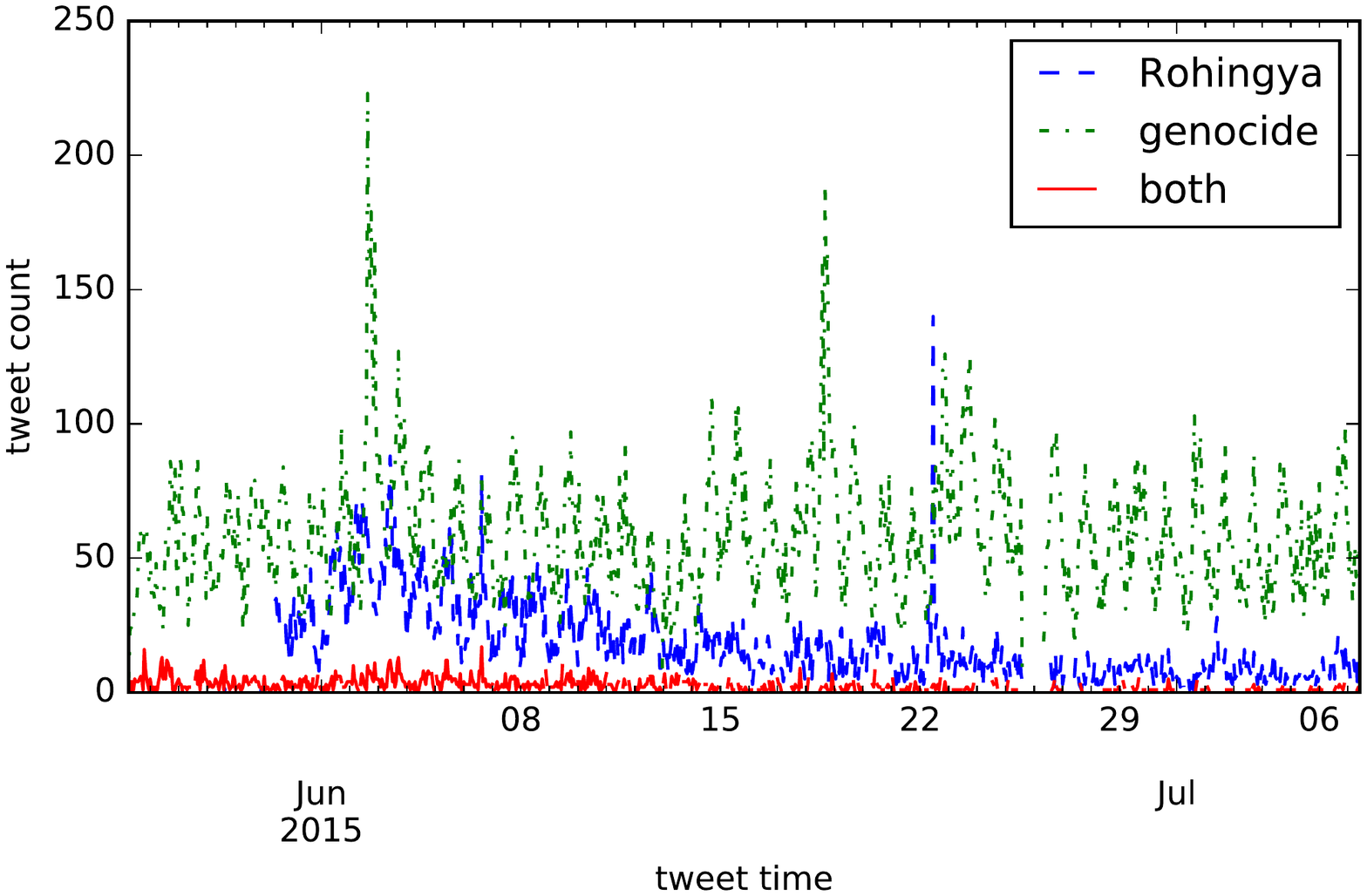}
\caption{Tweet distribution per key term after removing retweets, duplicates, and near-duplicates}
\label{fig:tweetdist2SCINF}
\end{center}
\end{figure}

Figure~\ref{fig:tweetdist2SCINF} illustrates the distribution of the tweets that remain in the collection after preprocessing each subset. We observe that the temporal distribution was changed after we eliminated the repetitive information. Large peaks in the `genocide' data were drastically reduced and some of the small peaks disappeared entirely. Thus, only peaks that consist of unique tweets in the `genocide' data remain and the peaks in tweets containing the key term `Rohingya' become apparent.

\subsection*{Clustering}
After removing the duplicates and near-duplicates, we clustered the remaining 77,905 tweets. Although the expert requested 100 clusters, the number of generated clusters was 145, which contain 13,236 tweets. The clustering parameters were set to begin with the following values: \begin{inparaenum}[(i)]
\item the Euclidean distances of the closest and farthest tweets to the cluster center have to be less than 0.725 and 0.875 respectively~\footnote{These values are strict values for our setting. The algorithm relaxes them based on the possibility of detecting clusters that can be annotated.}; and
\item the difference of the Euclidean distance to the cluster center between the closest and the farthest tweets in a cluster should not exceed 0.4.
\end{inparaenum} These values were designed to be strict in a way that at the first iteration almost no cluster satisfies these conditions. The automatic adaptation of these parameter's values after completion of each iteration sets them to suitable values for the respective collection.

\subsection*{Annotation and Results}

The annotation of the 145 clusters by a domain expert yielded the results in Table~\ref{table:clusterStats}. This process yielded eight labels: Actual cases (AC), Cultural genocide (CG), Historical cases (HC), Incoherent (IN), Indigenous people genocide (IPG), Irrelevant (IR), Jokes (JO), and Mass migration (MM). 

This step enabled the domain expert to understand the data by annotating only 17\% of the preprocessed tweets, which is 0.03\% of the complete collection, without the need of having to go over the whole set. Furthermore, annotating tweets in groups as suggested by the clustering algorithm improved the time efficiency of this process.

\begin{table}[htb]
\ttabbox{\caption{Number of labeled clusters and total number of tweets for each of the labels}\label{table:clusterStats}}
{\begin{tabular*}{0.77\textwidth}{lcr}
\toprule
{}  &  \# of clusters &  \# of tweets \\
\midrule
    Actual Cases (AC) & 48 & 4,937 \\
    Cultural Genocide (CG) & 7 & 375 \\
    Historical Cases (HC) & 22 & 1,530 \\
    Incoherent (IN) & 32 & 2,694 \\
    Indigenous People Genocide (IPG) & 1 & 109 \\ 
    Irrelevant (IR) & 30 & 3,365 \\ 
    Jokes (JO) & 1 & 30 \\ 
    Mass Migration (MM) & 4 & 226 \\
\midrule
    Total & 145 & 13,266\\
\bottomrule
\end{tabular*}
}\end{table}

Next, we used Relevancer to create an automatic classifier by using the annotated tweets. We merged the tweets that are under the JO (Jokes) label with the tweets under the Irrelevant label, since their number is relatively small compared to other tweet groups for generating a classifier for this class. Moreover, the incoherent clusters were not included in the training set. This leaves 10,572 tweets in the training set.

We used the same type characteristics explained in the feature extraction step to create the features used by the classifier. Since it yielded slightly better results than using only unigrams and bigrams, we extended the feature set by adding token trigrams in this study. The parameter optimization and the training was done on 85\% of the labeled data and the test was performed on the remaining 15\%. The only parameter of the Naive Bayes classifier, $\alpha$, was optimized on the training set to be 0.105 by testing with step size 0.105 between 0 and 2, 0 and 2 are included.

The performance of the classifier is summarized in Tables \ref{table:confusionMatrixGenocide} and \ref{table:classifierPerfGenocide} as a confusion matrix and evaluation summary. 
We observe that classes that have a clear topic, e.g., HC and CG, perform reasonably well. However, classes that are potentially a combination of many sub-topics, such as AC and IR, which contains JO labeled (each joke is a different sub-topic) tweets as well, perform relatively worse. Detailed analysis showed that the HC thread contains only a handful of past events that were referred to directly. On the other hand, there are many discussions in addition to clear event references in the AC thread. As a result, clusters that contain more or less similar language use work better than the clusters that contain widely diverse language use.

\begin{table}[htb]
\ttabbox{\caption{The rows and the columns represent the actual and the predicted labels of the test tweets. The diagonal provides the correct predictions.}\label{table:confusionMatrixGenocide}}
{\begin{tabular*}{0.86\textwidth}{lrrrrrr}
\toprule
{} &  AC &  CG &  HC &  IPG &  IR &  MM \\
\midrule
Actual Cases (AC)         &  \textbf{586} &  3 &  5 &  1 &  158 &  1 \\
Cultural Genocide (CG)         &  1 &  \textbf{42} &  0 &  0 &  3 &  0 \\
Historical Cases (HC)         &  26 &  0 &  \textbf{198} &  0 &  7 &  1 \\
Indigenous People Genocide (IPG)        &  2 &  1 &  0 &  \textbf{9} &  3 &  0 \\
Irrelevant (IR)          &  62 &  1 &  3 &  0 &  \textbf{441} &  1 \\
Mass Migration (MM)         &  8 &  0 &  0 &  0 &  0 &  \textbf{19} \\
\bottomrule
\end{tabular*}
}\end{table}

\begin{table}[htb]
\ttabbox{\caption{Precision, recall, and F1-score of the classifier on the test collection. The recall is based only on the test set, which is 15\% of the whole labelled dataset.}\label{table:classifierPerfGenocide}}
{\begin{tabular*}{0.85\textwidth}{lcccc}
\toprule
{} &  precision &  recall &  F1 &  support \\
\midrule
Actual Cases (AC)         &  .86  &    .78   &   .81   &    754 \\
Cultural Genocide (CG)         &  .89  &    .91   &   .90   &     46 \\
Historical Cases (HC)         &  .96  &    .85   &   .90   &    232 \\
Indigenous People Genocide (IPG)        &  .90  &    .60   &   .72   &     15 \\
Irrelevant (IR)          &  .72  &    .87   &   .79   &    508 \\
Mass Migration (MM)         &  .86  &    .70   &   .78   &     27 \\
\midrule
Avg/Total  &  .83  &   .82   &   .82   &   1,582 \\
\bottomrule
\end{tabular*}
}\end{table}

The result of this step is an automatic classifier that can be used to classify any tweet in the aforementioned six classes. Although the performance is relatively low for a class that is potentially a mix of various subtopics, the average scores (0.83, 0.82, and 0.82 for the precision, recall, and F1 respectively) were sufficient for using this classifier in understanding and organizing this collection.~\footnote{Thus, additional work was not performed in line of developing an SVM classifier for this task.} 

\subsection*{Conclusion}
\label{futureresSCINF}

In this case study, we demonstrated Relevancer's performance on a collection collected with the key terms `genocide' and `Rohingya'. Each step of the analysis process was explained in some detail in order to shed light both on the tool and the characteristics of this collection.

The results show that the requested number of clusters should not be too small. Otherwise, missing classes may cause the classifier not to perform well on tweets related to information threads not represented in the final set of annotated clusters. Second, the reported performance was measured on the held-out subset of the training set. This means that the test data has the same distribution as the subset of the training data that is used for the actual training. Therefore, generalization of those results to the actual social media context should be a concern of a further study. 

At the end of the analysis steps, the experts had successfully identified repetitive (79\% of the whole collection) and relevant information, analyzed coherent clusters, defined the main information threads, annotated the clusters, and created an automatic classifier that has an F1 score of .82.

The next case study is presented in the following section. It investigates the effect of key term inflections on the collection of microtexts from Twitter.




\section{Analysing the Role of Key Term Inflections in Knowledge Discovery on Twitter}
\label{SectionKDWeb}



Our methodology for collecting tweets and identifying event-related actionable information is based on the use of key terms and their inflections. Here, we investigate how the use of key term inflections has an impact on knowledge discovery in a tweet collection. 

The data consist of tweets we collected from the public Twitter API with the key term `flood' and its inflections `floods', `flooded', and `flooding' between December 17 and 31, 2015. In this experiment, the data were divided over different subsets (for flood, floods, flooded, and flooding respectively). We excluded tweets that contain the same pentagram (here: 5 consecutive words longer than 2 letters) before we apply the near-duplicate detection algorithm. After that the data were analyzed using Relevancer. The classifier generation step was not part of the analysis process. 

Detailed statistics of the collected tweets are represented in Table~\ref{table:tweetStats}. The number of collected tweets are provided in the column {\em \#All}. The columns {\em \#unique} and {\em unique\%} contain the counts after eliminating the retweets and (near-)duplicate tweets and the percentage of unique tweets in each subset of the collection. We observe that the volume of the tweets correlates with the volume of duplication. The higher the volume, the more duplicate information is posted. The ascending sorting of total and duplicate volume of the tweets per keyterm are `flood', `flooding', `floods', `flooded' and `flood', `floods', `flooding', `flooded'. The unique tweet ratio is highest for the term `flooded' and lowest for `flood'.



\begin{sidewaystable}
\ttabbox{\caption{Tweet statistics for the key term `flood' and its inflections}\label{table:tweetStats}}
{\begin{tabular*}{0.68\textwidth}{lccccccccccc}

\toprule
{}  &  \#All & \#unique & unique\% & \multicolumn{2}{c}{clustered} &  \multicolumn{2}{c}{relevant} &  \multicolumn{2}{c}{irrelevant} & \multicolumn{2}{c}{incoherent} \\
\midrule
{} & & & & \#raw & \% & \#raw & \% & \#raw & \% &  \#raw & \% \\
\midrule
    flood & 136,295 & 101,620 & 75 & 4,290 & 100 & 3,682 & 86 & 483 & 11 & 125 & 3\\
    floods & 55,384 & 47,312  & 86 & 2,429 & 100 & 818 & 34 & 1,291 & 53 & 320 & 13 \\
    flooded & 41,545 & 38,740 & 94 & 2,339 & 100 & 1,418 & 61 & 638 & 27 & 283 & 12 \\
    flooding & 77,280 & 66,920 & 87 & 3,003 & 100 & 1,420 & 47 & 1,573 & 52 & 10 & 1 \\
\bottomrule
\end{tabular*}
}
\end{sidewaystable}

After the subsets had been cleaned up, each subset of unique tweets was clustered in order to identify information threads. The clusters were annotated with the labels `relevant', `irrelevant, and `incoherent', which are the most general information threads that can be handled with our approach. We generated 50 clusters for each subset. The number of tweets that were placed in a cluster is presented in the column {\em clustered}. The clustering of the subset collected with the keyterm `floods' yielded both the most irrelevant (53\%) and incoherent (13\%) clusters. The keyterms `flood' and `flooded' contained the most relevant information, which are 86\% and 61\% of the related clusters respectively.\footnote{The {\em \%unique} column presents percentage in relation to the whole set before excluding the (near-)duplicates and the \% columns under {\em relevant}, {\em irrelevant}, and {\em incoherent} columns present the percentages in relation to the subset after excluding the (near-)duplicates.} The subset size and duplicate ratio difference between `flood' and `floods' was not affected by the bias of the clustering algorithm towards having more coherent clusters as the redundancy of the information increases. 


A cluster is {\em relevant}, if it is about a relevant information thread, e.g. an actionable insight, an observation, witness reaction, event information available from citizens or authorities that can help people avoid harm in the current use case. Otherwise, the label is {\em irrelevant}. Clusters that are not clearly about any information thread are labeled as {\em incoherent}. The respective columns in Table~\ref{table:tweetStats} provide information about the number of tweets in each thread. Having relatively many incoherent clusters from a subset points towards either the ambiguity of a term or by a large amount of uniquely different aspects, senses or threads covered by the respective subset.

Each inflection of the term `flood' has a different set of uses with a number of overlapping uses. A detailed cluster analysis reveals the characteristics of the information threads for the key term and its inflections. The key term `flood' (stem form) is mostly used by the authorities and automatic bots that provide updates about disasters in the form of a `flood alert' or `flood warning'. Moreover, mentioning multi-word terminological concepts, e.g, `flood advisory', enables tweets to fall in the same cluster. The irrelevant tweets using this term are mostly about ads or product names. The form `flooded' is mostly used for expressing observations and opinions toward a disaster event and news article related tweets. General comments and expressions of empathy toward the victims of disasters are found in tweets that contain the form `floods'. Finally, the form `flooding' mostly occurs in tweets that are about the consequences of a disaster. 

Another aspect that emerged from the cluster analysis is that common and specific multi-word expressions containing the key term or one of its inflections, e.g. `flooding back', `flooding timeline', form at least a cluster around them. Tweets that contain such expressions can be transferred from the remaining tweets, which are not put in any cluster, to a related cluster. For example, we identified 806 and 592 tweets that contain `flooding back' and `flooding timeline' respectively.

Finally, relevant named entities, such as the name of a storm, river, bridge, road, web platform, person, institution, or place, and emoticons enable the clustering algorithm to detect coherent clusters of tweets containing such named entities.

What the results of the study shows is that determining and handling separate uses of a key term and its inflections reveal different angles of the knowledge we can discover in tweet collections. 


\section{Using Relevancer to Detect Relevant Tweets: The Nepal Earthquake Case}
\label{SectionFIRE}


As a third use case, we describe our submission to the FIRE 2016 Microblog track {\em Information Extraction from Microblogs Posted during Disasters}~\cite{Ghosh+16}. The task in this track was to extract all relevant tweets pertaining to seven given topics from a set of tweets. These topics were about \begin{inparaenum}[(i)] \item available resources; \item required resources; \item available medical resources; \item required medical resources; \item locations where resources are available or needed; \item NGO or government activities; and \item infrastructure damage or restoration. \end{inparaenum} The tweet set was collected using key terms related to the Nepal 2015 Earthquake\footnote{\url{https://en.wikipedia.org/wiki/April_2015_Nepal_earthquake}, accessed June 10, 2018} by the task organizing team and distributed to the participating teams.

We used Relevancer and processed the data taking the following steps already detailed in the first sections of this chapter: (1) preprocessing the tweets, (2) clustering them, (3) manually labeling the coherent clusters, and (4) creating a classifier that can be used for classifying tweets that are not placed in any coherent cluster, and for classifying new (i.e. previously unseen) tweets using the labels defined in step (3). The data and the application of Relevancer are described below.

\subsection*{Data}

At the time of download (August 3, 2016), 49,660 tweet IDs were available out of the 50,068 tweet IDs provided for this task. The missing tweets had been deleted by the people who originally posted them. We used only the English tweets, 48,679 tweets in all, based on the language tag provided by the Twitter API. Tweets in this data set were already deduplicated by the task organisation team as much as possible.

\begin{figure}[htb]
\begin{center}
\includegraphics[scale=0.90]{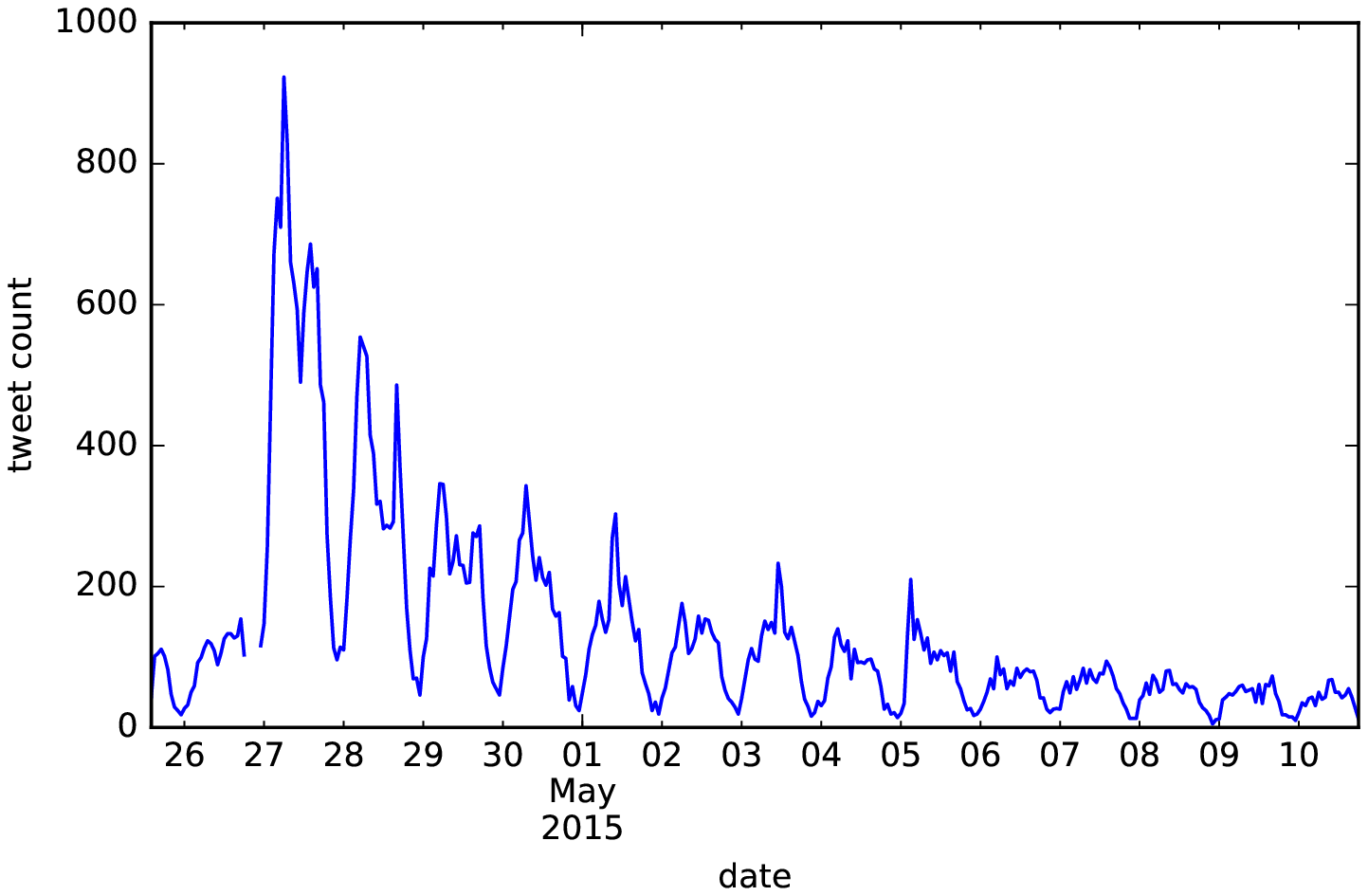}
\caption{Temporal distribution of the tweets}
\label{fig:sysdiag}
\end{center}
\end{figure}

The final tweet collection contains tweets that were posted between April 25, 2015 and May 5, 2015. The daily distribution of the tweets is visualized in Figure~\ref{fig:sysdiag}.

\subsection*{System Overview}

The typical analysis steps of Relevancer were applied to the data provided for this task. This study contains additional steps such as normalization, extending clusters, and bucketing tweets based on a time period in order to improve precision and recall of the tool. The bucketing, which splits the data into buckets of equally separated periods based on the posting time of the tweets, is introduced in this study. In this case, the focus of the Relevancer tool is the text and the date and time of posting of a tweet. The steps of Relevancer as applied in this use case are explained below. 


\subsubsection*{Normalization}

After inspection of the data, we decided to normalize a number of phenomena beyond the standard preprocessing steps described before. First, we removed certain automatically generated parts at the beginning and at the end of a tweet text. We determined these manually, e.g. `{\em live updates:}', `{\em I posted 10 photos on Facebook in the album}' and `{\em via usrusrusr}'. After that, words that end in `\ldots' were removed as well. These words are mostly incomplete due to the length restriction of a tweet text, and are usually at the end of tweets generated from within another application. Also, we eliminated any consecutive duplication of a token. Duplication of tokens mostly occurs with the dummy forms for user names and urls, and event-related key words and entities. For instance, two of three consecutive tokens at the beginning of the tweet {\em \#nepal: nepal: nepal earthquake: main language groups (10 may 2015) urlurlurl \#crisismanagement} were removed in this last step of normalization. This last step enables machine learning algorithms to be able to focus on the actual content of the tweets.

\subsubsection*{Clustering and Labeling}

The clustering step is aimed at finding topically coherent groups (information threads). These groups are labeled as {\bf relevant}, {\bf irrelevant}, or {\bf incoherent}. Coherent clusters were selected from the output of the clustering algorithm K-Means, with $k=200$, i.e. a preset number of 200 clusters. Coherency of a cluster is calculated based on the distance between the tweets in a particular cluster and the cluster center. Tweets that are in incoherent clusters (as determined by the algorithm) were clustered again by relaxing the coherence restrictions until the algorithm reaches the requested number of coherent clusters. The second stop criterion for the algorithm is the limit of the coherence parameter relaxation.

The coherent clusters were extended, specific for this use case, with the tweets that are not in any coherent cluster. This step was performed by iterating all coherent clusters in descending order of the total length of the tweets in a cluster and adding tweets that have a cosine similarity higher than 0.85 with respect to the center of a cluster to that respective cluster. The total number of tweets that were transferred to the clusters in this way was 847.

As bucketing is applied in this use case, the tool first searches for coherent clusters of tweets in each day separately. Then, in a second step it clusters all tweets from all days that previously were not placed in any coherent cluster. Applying the two steps sequentially enables Relevancer to detect local and global information threads as coherent clusters respectively.

For each cluster, a list of tweets is presented to an expert who then determines which are the relevant and irrelevant clusters.\footnote{The author had the role of being the expert for this task. A real scenario would require a domain expert.} Clusters that contain both relevant and irrelevant tweets are labeled as incoherent by the expert.\footnote{Although the algorithmic approach determines the clusters that were returned as coherent, the expert may not agree with it.} Relevant clusters are those which an expert considers to be relevant for the aim she wants to achieve. In the present context more specifically, clusters that are about a topic specified as relevant by the task organisation team should be labeled as relevant. Any other coherent cluster should be labeled as irrelevant. 

\subsubsection*{Creating the classifier}
The classifier was trained with the tweets labeled as relevant or irrelevant in the previous step. Tweets in the incoherent clusters were not included in the training set. The Naive Bayes method was used to train the classifier.

We used a small set of stop words, which were not included in the feature representation. These are a small set of key words (nouns), viz. {\em nepal, earthquake, quake, kathmandu} and their hashtag versions\footnote{This set was based on our observation as we did not have access to the key words that were used to collect this data set.}, the determiners {\em the, a, an}, the conjunctions {\em and, or}, the prepositions {\em to, of, from, with, in, on, for, at, by, about, under, above, after, before}, and the news-related words {\em breaking} and {\em news} and their hashtag versions. The normalized forms of the user names and URLs {\em usrusrusr} and {\em urlurlurl} are included in the stop word list as well.

We optimized the smoothing prior parameter $\alpha$ to be 0.31 by cross validation, comparing the classifier performance with equally separated 20 values of $\alpha$ between 0 and 2, including 0 and 2. Word unigrams and bigrams were used as features. The performance of the classifier on a set of 15\% held-out data is provided below in Tables \ref{table:confusionMatrix} and \ref{table:classifierPerf}. The rows and the columns represent the actual and the predicted labels of test tweets respectively. The diagonal provides the correct number of predictions.\footnote{Since we optimize the classifier for this collection, the performance of the classifier on unseen data is not analyzed here.}

\begin{table}[htb]
\ttabbox{\caption{Confusion matrix of the Naive Bayes classifier on test data.}\label{table:confusionMatrix}}
{\begin{tabular*}{0.45\textwidth}{lrrrrrr}
\toprule
{} &  Irrelevant &  Relevant  \\
\midrule
Irrelevant         &  \bf{720} &  34  \\
Relevant         &  33 &  \bf{257}  \\
\bottomrule
\end{tabular*}
}\end{table}

\begin{table}[htb]
\ttabbox{\caption{Precision, recall, and F1-score of the classifier on the test collection.}\label{table:classifierPerf}}
{\begin{tabular*}{0.6\textwidth}{lcccc}
\toprule
{} &  precision &  recall &  F1 &  support \\
\midrule
Irrelevant         &  .96  &    .95   &   .96   &    754 \\
Relevant        &  .88  &    .89   &   .88   &     290 \\
\midrule
Avg/Total  &  .94  &   .94   &   .94   &   1,044 \\
\bottomrule
\end{tabular*}
}\end{table}

The whole collection (48,679 tweets) was classified with the trained Naive Bayes classifier. 11,300 tweets were predicted as relevant. We continued the analysis with these relevant tweets.

\subsubsection*{Clustering and Labeling Relevant Tweets}

Relevant tweets, as predicted by the automatic classifier in the previous step, were clustered again without filtering them, based on the coherency criteria. In contrast to the first clustering step, the output of K-means was used as is, again with $k=200$. We annotated these clusters using the seven topics as predetermined by the task organizers. To the extent possible, incoherent clusters were labeled using the closest provided topic. Otherwise, the cluster was discarded. 

The clusters that have a topic label contain 8,654 tweets. 
The remaining clusters, containing 2,646 tweets, were discarded and not included in the submitted set.

\subsection*{Results}

The result of our submission was recorded under the ID relevancer\_ru\_nl. The performance of our results was evaluated by the organisation committee as precision at ranks 20, 1,000, and all, considering the tweets retrieved in the respective ranks. Our results, as announced by the organisation committee, are as follows: 0.3143 precision at rank 20, 0.1329 recall at rank 1,000,  0.0319 Mean Average Precision (MAP) at rank 1,000, and 0.0406 MAP considering all tweets in our submitted results.

We generated an additional calculation for our results based on the annotated tweets provided by task organizers. The overall precision and recall are 0.17 and 0.34 respectively. The performance for the topics FMT1 (available resources), FMT2 (required resources), FMT3 (available medical resources), FMT4 (required medical resources), FMT5 (resource availability at certain locations), FMT6 (NGO and governmental organization activities), and FMT7 (infrastructure damage and restoration reports) is provided in the Table~\ref{table:pertopicperf}.

\begin{table}[htb]
\ttabbox{\caption{Precision, recall, and F1-score of our submission and the percentage of the tweets in the annotated tweets per topic.}\label{table:pertopicperf}}
{\begin{tabular}{lcccc}
\toprule
{} &  precision &  recall &    F1 &  percentage \\
\midrule
FMT1 &       0.17 &    0.50 &  0.26 &        0.27 \\
FMT2 &       0.35 &    0.09 &  0.15 &        0.14 \\
FMT3 &       0.19 &    0.28 &  0.23 &        0.16 \\
FMT4 &       0.06 &    0.06 &  0.06 &        0.05 \\
FMT5 &       0.05 &    0.06 &  0.06 &        0.09 \\
FMT6 &       0.05 &    0.74 &  0.09 &        0.18 \\
FMT7 &       0.25 &    0.08 &  0.12 &        0.12 \\
\bottomrule
\end{tabular}
}\end{table}

On the basis of these results, we conclude that the success of our method differs widely across topics. In Table~\ref{table:pertopicperf}, we observe that there is a clear relation between the F1-score and the percentage of the tweets, which have 0.80 correlation coefficient, per topic in the manually annotated data. Consequently, we conclude that our method performs better in case the topic is represented well in the collection.

\subsection*{Conclusion}
In this study we applied the methodology supported by the Relevancer system in order to identify relevant information by enabling human input in terms of cluster labels. This method has yielded an average performance in comparison to other participating systems~\cite{Ghosh+16}, ranking seventh among fourteen submissions.

We observed that clustering tweets for each day separately enabled the unsupervised clustering algorithm to identify specific coherent clusters in a shorter time than the time spent on clustering the whole set. Moreover, this setting provided an overview that realistically changes each day, for each day following the day of the earthquake.

Our approach incorporates human input. In principle, an expert should be able to refine a tweet collection until she reaches a point where the time spent on a task is optimal and the performance is sufficient. However, with this particular task, an annotation manual was not available and the expert had to stop after one iteration without being sure to what extent certain information threads were actually relevant to the task at hand; for example, are (clusters of) tweets pertaining to providing or collecting funds for the disaster victims to be considered relevant or not?

It is important to note that the Relevancer system yields the results in random order, as it has no ranking mechanism that ranks posts for relative importance. We speculate that rank-based performance metrics as those used by the organizers of the challenge are not optimally suited for evaluating it.


\section{Identifying Flu Related Tweets in Dutch}
\label{SectionNLDB}


As a final use case we present a study in which we analyzed a collection of 229,494 Dutch tweets posted between December 16, 2010 and June 30th, 2013 and containing the key term `griep' (En: flu). After the preprocessing, retweets (24,019), tweets that were posted from outside the Netherlands (1,736) and irrelevant users (158), and exact duplicates (8,156) were eliminated.

Our use case aims at finding tweets reporting on personal flu experiences. Tweets in which users are reporting symptoms or declaring that they actually have or suspect having the flu are considered as relevant. Irrelevant tweets are mostly tweets containing general information and news about the flu, tweets about some celebrity suffering from the flu, or tweets in which users empathize or joke about the flu.

As a result of the preprocessing process, all retweets (24,019 tweets) have been removed, the tweet text has been transformed into a more standard form, and all duplicate tweets (8,156) have been eliminated. For this study we decided to keep the near-duplicate tweets in our collection, since we observed that Twitter users appear to use very similar phrases to describe their experiences and personal situation in the scope of the flu. The reason for this exception is that the flu topic related tweets are short and uttered similar to each other. Eliminating near-duplicates for this topic would cause a remarkable information loss.

The remaining 195,425 tweets were clustered in two steps. We started by working under the assumption that tweets that are posted around the same time have a higher probability of being related to each other, i.e.they are more likely to be related than the tweets that are posted some time apart from each other. Therefore, we split the tweet set in buckets of ten days each. We observed that this bucketing method increases the performance of and decreases the time spent by the clustering algorithm by decreasing the search space for determining similar tweets and the ambiguity across tweets~\cite{Hurriyetoglu+16c}. Moreover, bucketing allows the clustering to be performed in parallel for each bucket separately in this setting. Finally, the bucketing enables small information threads to be detected as they become more visible to the clustering algorithm. This was especially observed when there were news items or discussions about celebrities or significant events that last only a few days. After extracting these temporary tweet bursts as clusters in a clustering iteration, we could extract persistent topics in a successive clustering iteration.\footnote{Using a clustering algorithm that is more sophisticated than K-Means may remove the need of dividing the clustering task in this manner.}

We set the clustering algorithm to search for ten coherent clusters in each bucket. Then, we extract the tweets that are in a coherent cluster and search for ten other clusters in the remaining, un-clustered, tweets. In total, 10,473 tweets were put in 1,001 clusters. Since the temporal distribution of the tweets in the clusters has a correlation of 0.56 with the whole data, aggregated at a day level, we consider that the clustered part is weakly representative of the whole set. This characteristic enhances the understanding of the set and enables the classifier to have the right balance of the classes.

The time allowed for one annotator to label 306 of the clusters: 238 were found to be relevant (2,306 tweets), 196 irrelevant (2,189 tweets), and 101 incoherent (985 tweets). The tweets that are in the relevant or irrelevant clusters were used to train the SVM classifier. The performance of the classifier on the held-out 10\% (validation set) and unclustered part (a random selection from the tweets that were not placed in any cluster during the clustering phase and annotated by the same annotator) are listed in the Table~\ref{table:classifierHeldOutPerf}. The accuracy of the classifier is 0.95 in this setting.

\begin{table}[htb]
\ttabbox{\caption{Classifier performance in terms of precision (P), recall (R), and F1 on the validation (S1) and 255 (S2) unclustured tweets}\label{table:classifierHeldOutPerf}}
{\begin{tabular*}{0.95\textwidth}{lcccccccc}
\toprule
{} & & & validation set &&& &  unclustered \\
\toprule
{} &  P &  R &  F1 &  S1 &  P &  R &  F1 &  S2 \\
\midrule
relevant         &  .95  &    .96   &   .96   &     237 &  .66  &    .90   &   .76   &    144 \\
irrelevant         &  .96  &    .94   &   .95   &    213 &  .75  &    .39   &   .51   &     111 \\
\midrule
Avg/Total  &  .95  &   .95   &   .95   &   255 &  .70  &   .68   &   .65   &   255 \\
\bottomrule
\end{tabular*}
}\end{table}

The baseline of the classifier is the prediction of the majority class in the test set, in which the precision and recall of the minority class is undefined. Therefore, we compare the generated classifier and the baseline in terms of accuracy score, which are 0.67 and 0.56 respectively.

\section{Conclusion}
\label{conlusionCH3}
In this chapter we presented our research that aims at identifying relevant content on social media at a granularity level an expert determines. The evolution of the tool, which was illustrated at each subsequent use case, reflects how each step of the analysis was developed as a response to the needs that arise in the process of analyzing a tweet collection.

The novelty of our methodology is in dealing with all steps of the analysis from data collection to having a classifier that can be used to detect relevant information from microtext collections. It enables the experts to discover what is in a collection and make decisions about the granularity of their analysis. The result is a scalable tool that deals with all analysis, from data collection to having a classifier that can be used to detect relevant information at a reasonable performance.

The following chapter reports on our experiments to explore methods that can shed light on and improve the performance of Relevancer by integrating it with a linguistically motivated rule-based method. The motivation behind this exploration is to remedy the weakness of Relevancer in handling unbalanced class distributions and the limitations imposed by having a small training set. 

\lhead{\emph{Mixing Paradigms for Relevant Microtext Classification}}
\chapter{Mixing Paradigms for Relevant Microtext Classification}


\section{Introduction}

\begin{sloppypar}
Microtext classification can be performed using various methodologies. We test, compare, and integrate machine learning and rule-based approaches in this chapter. Each approach was first applied in the context of a shared task that did not provide any annotated data for training or evaluation before the submission. Next, they were applied in a case where annotated data is available from two different shared tasks.\footnote{It was possible to obtain the annotated data after completion of the task.}
\end{sloppypar}

These experiments shed light on the performance of machine learning and rule-based methods on microtext classification under various conditions in the context of an earthquake disaster. Different starting conditions and results were analyzed in order to identify and benefit from the strengths of each approach as regards different performance requirements, e.g. precision vs. recall. Our preliminary results show that integration of machine learning and rule-based methodologies can alleviate annotated data scarcity and class imbalance issues. 

The results of our experiments are reported in Section~\ref{CH4_extracting_humanitarian_inf} and ~\ref{CH4_comparing_integrating_rb_ml} for cases where annotated training and evaluation data is  both available and unavailable, respectively. 
Section~\ref{CH4_conclusion} concludes this chapter with the overall insights based on the experience gained through participating in relevant shared tasks and experiments performed in the scope of mixing rule and machine learning based paradigms.

\section{Classifying Humanitarian Information in Tweets}
\label{CH4_extracting_humanitarian_inf}

{\bf Based on:} Hürriyetoğlu, A., \& Oostdijk, N. (2017, April). Extracting Humanitarian Information from Tweets. In {\em Proceedings of the first international workshop on exploitation of social media for emergency relief and preparedness}. Aberdeen, United Kingdom. Available from \url{http://ceur-ws.org/Vol-1832/SMERP-2017-DC-RU-Retrieval.pdf}

In this section we describe the application of our methods to humanitarian information classification for microtexts and their performance in the scope of the SMERP 2017 Data Challenge task. Detecting and extracting the (scarce) relevant information from tweet collections as precisely, completely, and rapidly as possible is of the utmost importance during natural disasters and other emergency events. 
Following the experiments in the previous chapter, we remain focused on microtext classification.

Rather than using only a machine learning approach, we combined Relevancer with an expert-designed rule-based approach. Both are designed to satisfy the information needs of an expert by allowing experts to define and find the target information. The results of the current data challenge task demonstrate that it is realistic to expect a balanced performance across multiple metrics even under poor conditions, as the combination of the two paradigms can be leveraged; both approaches have weaknesses that the other approach can compensate for.

\subsection{Introduction}

This study describes our approach used in the text retrieval sub-track that was organized as part of the Exploitation of Social Media for Emergency Relief and Preparedness (SMERP 2017) Data Challenge Track, Task 1. In this task, participants were required to develop methodologies for extracting from a collection of microblogs (tweets) those tweets that are relevant to one or more of a given set of topics with high precision as well as high recall.\footnote{See also \url{http://computing.dcu.ie/~dganguly/smerp2017/index.html}, accessed June 10, 2018} The extracted tweets should be ranked based on their relevance. The topics were the following: resources available (T1), resources needed (T2), damage, restoration, and casualties (T3), and rescue activities of various NGOs and government organizations (T4). With each of the topics there was a short (one sentence) description and a more elaborate description in the form of a one-paragraph narrative.

The challenge was organized in two rounds.\footnote{The organizers consistently referred to these as `levels'.} The task in both rounds was essentially the same, but participants could benefit from the feedback they received after submitting their results for the first round.
The data were provided by the organizers of the challenge and consisted of tweets about the earthquake that occurred in Central Italy in August 2016.\footnote{\url{https://en.wikipedia.org/wiki/August_2016_Central_Italy_earthquake}, accessed June 10, 2018} The data for the first round of the challenge were tweets posted during the first day (24 hours) after the earthquake happened, while for the second round the data set was a collection of tweets posted during the next two days (day two and three) after the earthquake occurred. The data for the second round were released after round one had been completed. All data were made available in the form of tweet IDs (52,469 and 19,751 for rounds 1 and 2 respectively), along with a Python script for downloading them by means of the Twitter API. In our case the downloaded data sets comprised 52,422 (round 1) and 19,443 tweets (round 2) respectively.\footnote{At the time of download some tweets had been removed.}

For each round, we discarded tweets that \begin{inparaenum}[(i)]
\item were not marked as English by the Twitter API; 
\item did not contain the country name Italy or any region, city, municipality, or earthquake-related place in Italy;
\item had been posted by users that have a profile location other than Italy; this was determined by manually checking the most frequently occurring locations and listing the ones that are outside Italy; 
\item originated from an excluded user time zone; the time zones were identified manually and covered the ones that appeared to be the most common in the data sets; 
\item had a country meta-field other than Italy; and 
\item had fewer than 4 tokens after normalization and basic cleaning. 
\end{inparaenum} The filtering was applied in the order given above. After filtering the data sets consisted of 40,780 (round 1) and 17,019 (round 2) tweets. 

\begin{sloppypar}
We participated in this challenge with two completely different approaches which we developed and applied independently of each other. The first approach is a machine-learning approach implemented in the Relevancer tool, introduced in Chapter \ref{chapter3},~\cite{Hurriyetoglu+16b}, which offers a complete pipeline for analyzing tweet collections. The second approach is a linguistically motivated approach in which a lexicon and a set of hand-crafted rules are used in order to generate search queries. As the two approaches each have their particular strengths and weaknesses and we wanted to find out whether the combination would outperform each of the underlying approaches, we also submitted a run in which we combined them.
\end{sloppypar}

The structure of the remainder of this study is as follows. We first give a more elaborate description of our two separate approaches, starting with the machine learning approach in Section \ref{sec:CH4_SMERP_Approach1} and the rule-based approach in Section \ref{sec:CH4_SMERP_Approach2}. Then in Section \ref{sec:CH4_SMERP_Comb} we describe the combination of the results of these two approaches. Next, in Section \ref{CH4_SMERP_Results}, the results are presented, while in Section \ref{CH4_SMERP_Discussion} the most salient findings are discussed. Section \ref{CH4_SMERP_Conclusion} concludes this subsection with a summary of the main findings.

\subsection{Approach 1: Identifying Topics Using Relevancer}
\label{sec:CH4_SMERP_Approach1}

The main analysis steps supported by Relevancer are: preprocessing, clustering, manual labeling of coherent clusters, and creating a classifier for labeling previously unseen data. Below we provide details of the configuration that we used for the present task.

\begin{description}

\item [Pre-processing] 
RT elimination, the duplicate elimination, and near-duplicate elimination steps that are part of the standard Relevancer approach and in which retweets, exact- and near-duplicates are detected and eliminated were not applied in the scope of this task. The data set that was released was considered to have been preprocessed in this respect.
\item [Clustering] First we split the tweet set in buckets determined by periods, which are extended at each iteration of clustering. The bucket length starts with 1 hour and stops after the iteration in which the length of the period is equal to the whole period covered by the tweet set. Moreover, the feature extraction step was based on character n-grams (tri-, four- and five-grams) for this clustering. 

\item [Annotation] Coherent clusters that were identified by the algorithm automatically are presented to an expert who is asked to judge whether indeed a cluster is coherent and if so, to provide the appropriate label.\footnote{The experts in this setting were the task participants. The organizers of the task contributed to the expert knowledge in terms of topic definition and providing feedback on the topic assignment of the round 1 data.} For the present task, the topic labels are those determined by the task organization team (T1-T4). We introduced two additional labels: the irrelevant label for coherent clusters that are about any irrelevant topic and the incoherent label for clusters that contain tweets about multiple relevant topics or combinations of relevant and irrelevant topics.\footnote{The label definition affects the coherence judgment. Specificity of the labels determines the required level of the tweet similarity in a cluster.} The expert does not have to label all available clusters. For this task we annotated only one quarter of the clusters from each hour.

\item [Classifier Generation] The labeled clusters are used to create an automatic classifier. For the current task we trained a state-of-the-art Support Vector Machine (SVM) classifier by using standard default parameters. We trained the classifier with 90\% of the labeled tweets by cross-validation. The classifier was tested on the remaining labeled tweets, which were in the labeled clusters.\footnote{The performance scores are not considered to be representative due to the high degree of similarity of the training and test data.}

\item [Ranking] For our submission in round 1 we used the labels as they were obtained by the classifier. No ranking was involved. However, as the evaluation metrics used in this shared task expected the results to be ranked, for our submission in round 2 we classified the relevant tweets by means of a classifier based on these tweets and used the classifier confidence score for each tweet for ranking them. 

\end{description} 

\noindent We applied clustering with the value for the requested clusters parameter set to 50 (round 1) and 6 (round 2) per bucket, which yielded a total of 1,341 and 315 coherent clusters respectively. In the annotation step, 611 clusters from the round 1 and 315 clusters from round 2 were labeled with one of the topic labels T1-T4, irrelevant, or incoherent.\footnote{The annotation was performed by one person who has an expert role as suggested by the Relevancer methodology.} Since most of the annotated clusters were irrelevant or T3, we enriched this training set with the annotated tweets featuring in the FIRE - Forum for Information Retrieval Evaluation, Information Extraction from Microblogs posted during Disasters - 2016 task~\cite{Ghosh+16}. 

In preparing our submission for round 2 of the current task, we also included the positive feedback for both of our approaches from the round 1 submissions in the training set.

We used the following procedure for preparing the submission for round 2. First, we put the tweets from the clusters that were annotated with one of the task topics (T1-T4) directly in the submission with rank 1.~\footnote{The submission format allowed us to determine the rank of a document.} The number of these tweets per topic were as follows: 331 -- T1, 33 -- T2, 647 -- T3, and 134 -- T4. Then, the tweets from the incoherent clusters and the tweets that were not included in any cluster were classified by the SVM classifier created for this round. The tweets that were classified as one of the task topics were included in the submission. The second part is ranked lower than the annotated part and ranked based on the confidence of a classifier trained using these predicted tweets.

Table \ref{tab:CH4_SMERP_TopicAssignmentML} gives an overview of the submissions in rounds 1 and 2 that are based on the approach using the Relevancer tool. The submissions are identified as Relevancer and Relevancer with ranking (see also Section \ref{CH4_SMERP_Results}). 

\begin{table}
\centering
\begin{tabular}{|l|r|r|r|r|}
\hline
& \multicolumn{2}{|c|}{\textbf{Round 1}} & \multicolumn{2}{|c|}{\textbf{Round 2}} \\\hline
\textit{Topic(s)} & \textit{\# tweets} & \textit{\% tweets} & \textit{\# tweets} & \textit{\% tweets} \\\hline
T1&52&0.12&855&5.02 \\\hline
T2&22&0.05&173&1.02\\\hline
T3&5,622&13.79&3,422&20.11\\\hline
T4&50&0.12&507&2.98\\\hline
T0&35,034&85.91&12,062&70.87\\\hline
Total&40,780&100.00&17,019&100.00\\\hline\end{tabular}
\caption{Topic assignment using Relevancer}
\label{tab:CH4_SMERP_TopicAssignmentML}
\end{table}

\subsection{Approach 2: Topic Assignment by Rule-based Search Query Generation}
\label{sec:CH4_SMERP_Approach2}
In the second approach, a lexicon and a set of hand-crafted rules are used to generate search queries. As such it continues the line of research described in Oostdijk \& van Halteren~\cite{Oostdijk+13a,Oostdijk+13b} and Oostdijk et al.~\cite{Oostdijk+16} in which word n-grams are used for search. 

For the present task we compiled a dedicated (task-specific) lexicon and rule set from scratch. In the lexicon with each lexical item information is provided about the part of speech (e.g. noun, verb), semantic class (e.g. casualties, building structures, roads, equipment) and topic class (e.g. T1, T2). A typical example of a lexicon entry thus looks as follows:

\indent deaths\hspace{1cm}	N2B\hspace{.5cm}	T3
 
\noindent where \textit{deaths} is listed as a relevant term with N2B encoding the information that it is a plural noun of the semantic class [casualties] which is associated with topic 3. In addition to the four topic classes defined for the task, in the lexicon we introduced a fifth topic class, viz. T0, for items that rendered a tweet irrelevant. Thus T0 was used to mark a small set of words each word of which referred to a geographical location (country, city) outside Italy, for example \textit{Nepal}, \textit{Myanmar} and \textit{Thailand}.\footnote{More generally, T0 was assigned to all irrelevant tweets. See below.} 

The rule set consists of finite state rules that describe how lexical items can (combine to) form search queries made up of (multi-)word n-grams. Moreover, the rules also specify which of the constituent words determines the topic class for the search query. An example of a rule is

\indent NB1B\hspace{.5cm} *V10D

\noindent Here NB1B refers to items such as \textit{houses} and \textit{flats} while V10D refers to past participle verb forms expressing [damage] (\textit{damaged}, \textit{destroyed}, etc.). The asterisk indicates that in cases covered by this rule it is the verb that is deemed to determine the topic class for the multi-word n-gram that the rule describes. This means that if the lexicon lists \textit{destroyed} as V10D and T3, upon parsing the bigram \textit{houses destroyed} the rule will yield T3 as the result.

The ability to recognize multi-word n-grams is essential in the context of this challenge as most single key words on their own are not specific enough to identify relevant instances: with each topic the task is to identify tweets with specific mentions of resources, damage, etc. Thus the task/topic description for topic 3 explicitly states that tweets should be identified `which contain information related to infrastructure damage, restoration and casualties', where `a relevant message must mention the damage or restoration of some specific infrastructure resources such as structures (e.g., dams, houses, mobile towers), communication facilities, ...' and that `generalized statements without reference to infrastructure resources would not be relevant'. Accordingly, it is only when the words \textit{bridge} and \textit{collapsed} co-occur that a relevant instance is identified. 

As for each tweet we seek to match all possible search queries specified by the rules and the lexicon, it is possible that more than one match is found for a given tweet. If this is the case we apply the following heuristics: 
(a) multiple instances of the same topic class label are reduced to one (e.g. T3-T3-T3 becomes T3); 
(b) where more than one topic class label is assigned but one of these happens to be T0, then all labels except T0 are discarded (thus T0-T3 becomes T0); 
(c) where more than one topic label is assigned and these labels are different, we maintain the labels (e.g. T1-T3-T4 is a possible result).
Tweets for which no matches were found were assigned the T0 label.

The lexicon used for round 1 comprised around 950 items, while the rule set consisted of some 550 rules. For round 2 we extended both the lexicon and the rule set (to around 1,400 items and 1,750 rules respectively) with the aim to increase the coverage especially with respect to topics 1, 2 and 4.
Here we should note that, although upon declaration of the results for round 1 each participant received some feedback, we found that it contributed very little to improving our understanding of what exactly we were targeting with each of the topics. We only got confirmation -- and then only for a subset of tweets -- that tweets had been assigned the right topic. Thus we were left in the dark about whether tweets we deemed irrelevant were indeed irrelevant, while also for relevant tweets that might have been assigned the right topic but were not included in the evaluation set we were none the wiser.\footnote{The results from round 1 are discussed in more detail in Section \ref{CH4_SMERP_Discussion}. } 

The topic assignments we obtained for the two data sets are presented in Table \ref{tab:TopicAssignment}.

\begin{table}
\centering
\begin{tabular}{|l|r|r|r|r|}
\hline
& \multicolumn{2}{|c|}{\textbf{Round 1}} & \multicolumn{2}{|c|}{\textbf{Round 2}} \\\hline
\textit{Topic(s)} & \textit{\# tweets} & \textit{\% tweets} & \textit{\# tweets} & \textit{\% tweets} \\\hline
T1&91&0.22&206&1.21 \\\hline
T2&55&0.13&115&0.68\\\hline
T3&7,002&17.17&3,558&20.91\\\hline
T4&115&0.28&177&1.04\\\hline
Mult.&151&0.37&194&1.14\\\hline
T0&33,366&81.82&12,769&75.03\\\hline
Total&40,780&100.00&17,019&100.00\\\hline\end{tabular}
\caption{Topic assignment rule-based approach}
\label{tab:TopicAssignment}
\end{table}

In both data sets T3 (Damage, restoration and casualties reported) is by far the most frequent of the relevant topics. The number of cases where multiple topics were assigned to a tweet is relatively small (151/40,780 and 194/17,019 tweets resp.). Also in both datasets there is a large proportion of tweets that were labeled as irrelevant (T0, 81.22\% and 75.03\% resp.). We note that in the majority of cases it is the lack of positive evidence for one of the relevant topics that leads to the assignment of the irrelevant label.\footnote{In other words, it might be the case that these are not just truly irrelevant tweets, but also tweets that are falsely rejected because the lexicon and/or the rules are incomplete.} Thus for the data in round 1 only 2,514/33,366 tweets were assigned the T0 label on the basis of the lexicon (words referring to geographical locations outside Italy, see above). For the data in round 2 the same was true for 1,774/12,769 tweets.\footnote{Actually, in 209/2,514 tweets (round 1) and 281/1,774 tweets (round 2) one or more of the relevant topics were identified; yet these tweets were discarded on the basis that they presumably were not about Italy.}

For round 1 we submitted the output of this approach without any ranking (Rule-based in Table \ref{tab:CH4_SMERP_ResultsRound1Website}). For round 2 (cf. Table \ref{tab:CH4_SMERP_ResultsRound2}) there were two submissions based on this approach: one (Rule-based without ranking) similar to the one in round 1 and another one for which the results were ranked (Rule-based with ranking). In the latter case ranking was done by means of an SVM classifier trained on the results. The confidence score of the classifier was used as a rank.

\subsection{Combined Approach}
\label{sec:CH4_SMERP_Comb}

While analyzing the feedback on our submissions in round 1, we noted that, although the two approaches were partly in agreement as to what topic should be assigned to a given tweet, there was a tendency for the two approaches to obtain complementary sets of results, especially with the topic classes that had remained underrepresented in both submissions.\footnote{Thus for T2 there was no overlap at all in the confirmed results for the two submissions.} We speculated that this was due to the fact that each approach has its strengths and weaknesses. This then invited the question as to how we might benefit from combining the two approaches. 

Below we first provide a brief overview of how the approaches differ with regard to a number of aspects, before describing our first attempt at combining them. 

\begin{description}

\item [Role of the expert] Each approach requires and utilizes expert knowledge and effort at different stages. In the machine learning approach using Relevancer the expert is expected to (manually) verify the clusters and label them. In the rule-based approach the expert is needed for providing the lexicon and/or the rules.
\item [Granularity] The granularity of the information to be used as input and targeted as output is not the same across the approaches. The Relevancer approach can only control clusters. This can be inefficient in case the clusters contain information about multiple topics. By contrast, the linguistic approach has full control on the granularity of the details.
\item [Exploration] Unsupervised clustering helps the expert to understand what is in the data. The linguistic approach, on the other hand, relies on the interpretation of the expert. To the extent development data are available, they can be explored by the expert and contribute to insights as regards what linguistic rules are needed.
\item [Cost of start] The linguistic, rule-based approach does not require any training data. It can immediately start the analysis and yield results. The machine learning approach requires a substantial quantity of annotated data to be able to make reasonable predictions. These may be data that have already been annotated, or when no such data are available as yet, these may be obtained by annotating the clusters produced in the clustering step of Relevancer. The filtering and preprocessing of the data plays an important role in machine learning.
\item [Control over the output] In case of the rule-based approach it is always clear why a given tweet was assigned a particular topic: the output can straightforwardly be traced back to the rules and the lexicon. With the machine learning approach it is sometimes hard to understand why a particular tweet is picked as relevant or not.
\item [Reusability] Both approaches can re-use the knowledge they receive from experts in terms of annotations or linguistic definitions. The fine-grained definitions are more transferable than the basic topic label-based annotations.
\end{description}


\noindent One can imagine various ways in which to combine the two approaches. However, it is less obvious how to obtain the optimal combination. As a first attempt in round 2 we created a submission based on the intersection of the results of the two approaches (Rule-based without ranking and and Relevancer with ranking). The intersection contains only those tweets that were identified as relevant by both approaches and for which both approaches agreed on the topic class. We respected the ranking created in Relevancer with ranking for the combined submission. The results obtained by the combined approach are given in Table \ref{tab:TopicAssignmentComb}.

\begin{table}
\centering
\begin{tabular}{|l|r|r|}
\hline
& \multicolumn{2}{|c|}{\textbf{Round 2}} \\\hline
\textit{Topic(s)} & \textit{\# tweets} & \textit{\% tweets} \\\hline
T1&305& 2\\\hline
T2&120& 1\\\hline
T3&2,844&17\\\hline
T4&149& 1\\\hline
T0&13,601&79\\\hline
Total&17,019&100\\\hline\end{tabular}
\caption{Topic assignment combined approach}
\label{tab:TopicAssignmentComb}
\end{table}

\subsection{Results}
\label{CH4_SMERP_Results}

The submissions were evaluated by the organizers.\footnote{For more detailed information on the task and organization of the challenge, its participants, and the results achieved see Ghosh et al. \citeyear{Ghosh+17}.} Apart from the mean average precision (MAP) and recall that had originally been announced as evaluation metrics, two further metrics were used viz. bpref and precision@20, while recall was evaluated as recall@1000. As the organizers arranged for `evaluation for some of the top-ranked results of each submission' but eventually did not communicate what data of the submissions was evaluated (especially in the case of the non-ranked submissions), it remains unclear how the performance scores were arrived at. In Tables \ref{tab:CH4_SMERP_ResultsRound1Website} and \ref{tab:CH4_SMERP_ResultsRound2} the results for our submissions are summarized.



\begin{table}
\centering
\begin{tabular}{|l|p{2cm}|p{2cm}|p{2cm}|p{2cm}|}
\hline
\textit{Run ID}& \textit{bpref}& \textit{precision@20}& \textit{recall@1000} & \textit{MAP} \\\hline
Relevancer&0.1973&0.2625&0.0855&0.0375\\\hline
Rule-based&0.3153&0.2125&0.1913&0.0678\\\hline
\end{tabular}
\caption{Results obtained in Round 1 as evaluated by the organizers}
\label{tab:CH4_SMERP_ResultsRound1Website} 
\end{table}

\begin{table}
\centering
\begin{tabular}{|l|p{1.7cm}|p{2cm}|p{2cm}|p{1.7cm}|}
\hline
\textit{Run ID}& \textit{bpref}& \textit{precision@20}& \textit{recall@1000} & \textit{MAP} \\\hline
Relevancer with ranking&0.4724&0.4125&0.3367&0.1295\\\hline
Rule-based without ranking&0.3846&0.4125&0.2210&0.0853\\\hline
Rule-based with ranking&0.3846&0.4625&0.2771&0.1323\\\hline
Combined&0.3097&0.4125&0.2143&0.1093\\\hline
\end{tabular}
\caption{Results obtained in Round 2 as evaluated by the organizers}
\label{tab:CH4_SMERP_ResultsRound2}
\end{table}

\subsection{Discussion}
\label{CH4_SMERP_Discussion}

The task in this challenge proved quite hard. This was due to a number of factors. One of these was the selection and definition of the topics: topics T1 and T2 specifically were quite close, as both were concerned with resources; T1 was to be assigned to tweets in which the availability of some resource was mentioned while in the case of T2 tweets should mention the need of some resource. The definitions of the different topics left some room for interpretation and the absence of annotation guidelines was experienced to be a problem.

Another factor was the training dataset in both rounds we perceived to be highly imbalanced as regards to the distribution of the targeted topics.\footnote{T3 cases were 75\% of the whole dataset. There was not any special treatment of the class imbalance issue for the ML method.} Although we appreciate that this realistically reflects the development of an event -- you would indeed expect the tweets posted within the first 24 hours after the earthquake occurred to be about casualties and damage and only later tweets to ask for or report the availability of resources -- the underrepresentation in the data of all topics except T3 made it quite difficult to achieve a decent performance.

As already mentioned in Section \ref{sec:CH4_SMERP_Approach2}, the feedback on the submissions for round 1 was only about the positively evaluated entries of our own submissions. There was no information about the negatively evaluated submission entries. Moreover, not having any insight about the total annotated subset of the tweets made it impossible to infer anything about the subset that was marked as positive. This put the teams in unpredictably different conditions for round 2. Since the feedback was in proportion to the submission, having only two submissions was to our disadvantage.

As can be seen from the results in Tables \ref{tab:CH4_SMERP_ResultsRound1Website} and \ref{tab:CH4_SMERP_ResultsRound2} the performance achieved in round 2 shows an increase on all metrics when compared to that achieved in round 1. Since our approaches are inherently designed to benefit from experts over multiple interactions with the data, we consider this increase in performance significantly positive. 

The overall results also show that from all our submissions the one in round 2 using the Relevancer approach achieves the highest scores in terms of the bpref and recall@1000 metrics, while the ranked results from the rule-based approach has the highest scores for precision@20 and MAP. The Relevancer approach clearly benefited from the increase in the training data (feedback for the round 1 results for both our approaches and additional data from the FIRE 2016 task). For the rule-based approach the extensions to the lexicon and the rules presumably largely explain the increased performance, while the different scores for the two submissions in round 2 (one in which the results were ranked, the other without ranking) show how ranking boosts the scores.

\subsection{Conclusion}
\label{CH4_SMERP_Conclusion}

In this section we have described the approaches we used to prepare our submissions for the SMERP Data Challenge Task. Over the two rounds of the challenge we succeeded in improving our results, based on the experience we gained in round 1. We were ranked fourth and seventh and first, second, third, and fifth out of eight and twelve semi-automatic submissions respectively in the first and second rounds.

This task along with the issues that we came across provided us with a realistic setting in which we could measure the performance of our approaches. In a real use case, we would not have had any control on the information need of an expert, her annotation quality, her feedback on the output, and her performance evaluation. Therefore, from this point of view, we consider our participation and the results we achieved a success. 

As observed before, we expect that eventually the best result can be obtained by combining the two approaches. The combination of the outputs we attempted in the context of the current challenge is but one option, which as it turns out may be too simplistic. Therefore, we designed the study reported in the following section to explore the possibilities of having the two approaches interact and produce truly joint output.

\section{Comparing and Integrating Machine Learning and Rule-Based Microtext Classification}
\label{CH4_comparing_integrating_rb_ml}

This section compares and explores ways to integrate machine learning and rule-based microtext classification techniques. The performance of each method is analyzed and reported, both when used in a standalone fashion and when used in combination with the other method. The goal of this effort is to assess the effectiveness of combining these two approaches.

Our machine learning approach (ML) is a supervised machine learning model and the rule-based approach (RB) is an information extraction based classification approach, which   was described in Section~\ref{sec:CH4_SMERP_Approach2} and is loosely related to the approach of Wilson et al.~\cite{Wilson+05}. We aim to find what is similar and what is distinct between prediction performance of these two paradigms in order to derive a hybrid methodology that will yield a higher performance, preferably both in terms of precision and recall. The level of performance of the standalone implementations in terms of numbers is not of primary importance.\footnote{Annotating a high number of tweets for evaluation of these systems is a challenge that includes inconsistencies in labeling~\cite{Ghosh+16}.}

We use gold standard data (tweets) released under the {\em First Workshop on the Exploitation of Social Media for Emergency Relief and Preparedness}\/ (SMERP 2017) and the {\em Microblog track: Information Extraction from Microblogs}\/ of the Forum for Information Retrieval Evaluation (FIRE 2016) for training and testing our approaches respectively. These data sets have a naturally unbalanced class distribution and tweets may be annotated with multiple labels. 

We compare ML and RB in terms of their standalone performance on the training and test data, the number of predictions where the other approach does not yield any prediction on test data, and their performance on the part where only one of the approaches yields a prediction.

We also explored integrating the two approaches at result level by intersecting and unifying their predictions and at training data level by using complete and filtered versions of the RB output as training data.


We provide details of these experiments below as follows. First, we discuss related research in Section~\ref{CH4_Integration_relstudies}. Then, the data that we used is described in detail in Section~\ref{CH4_Integration_datasets}. In Section~\ref{CH4_Integration_baseline}, we define the baseline we used to evaluate our results.
The creation and results of the standalone systems are presented in Sections~\ref{CH4_Integration_building_ml_classifier} and~\ref{CH4_Integration_rule_based_sys} for ML and RB respectively. Sections~\ref{CH4_Integration_comparing_ml_rb} and ~\ref{CH4_Integration_integrating_ml_rb} report on the details of comparing the performance of the two approaches with each other and integrating them. In the two final Sections (Sections~\ref{CH4_Integration_discussion} and ~\ref{CH4_Integration_conclusion}) we discuss the results and present our conclusions.

\subsection{Related Studies}
\label{CH4_Integration_relstudies}

\begin{sloppypar}
Different information classification approaches, including data-driven and knowledge-driven approaches, have been compared mainly in terms of effort required to develop them, their complexity, the task they attempt to solve, their performance, and interpretability of their predictions~\cite{Pang+2002}. 
The strengths and weaknesses of each approach direct us to apply a particular setting for almost each use case for the given conditions, e.g. availability of annotated data or linguistic expertise. Therefore, the integration of the data-driven and the knowlegde-based approaches has been subject of many studies. Aim of the integration is to benefit from the strengths and overcome the weaknesses of each of the approaches and thus contribute to the development of robust information classification and extraction systems~\cite{Borthwick+98,Srihari+00}.
\end{sloppypar}

As a sample task, the named entity recognition studies have proposed developing hybrid approaches by solving distinct parts of the task using different approaches \cite{Saha+08}, applying correction rules to machine learning results~\cite{Gali+08,Praveen+08}, and solving the task incrementally by applying dictionary-, rule-, and n-gram-based approaches in succession~\cite{Chaudhuri+08}. The tasks of information classification and extraction in the medical domain~\cite{Xu+12} and sentiment analysis~\cite{Melville+09} benefit from hybrid approaches in terms of solving the task incrementally and using a manually compiled lexicon respectively.

Another recent study focuses on using rule-based approaches for assisting in creating training data  as practiced by Ratner et al.~\cite{Ratner+18}. This approach uses manually compiled rules to generate training data for machine learning.

In FIRE 2016, a semi-supervised model that is in line of our efforts was developed by Lkhagvasuren, Gonçalves, and Saias~\citeyear{lkhagvasuren+16}. They used key terms for each topic to create a training set and then train a classifier on this data. This approach is similar to our RB approach in respect to having a topic-specific lexicon and similar to our ML approach in terms of using a classifier. The results of this submission was not comparable to the other submissions since their results were only for three classes, which were {\em available}, {\em required}, and {\em other}.

We consider our experiment as a continuation of comparing and integrating ML and RB classification efforts. The domain and the microtext characteristics of the text we are handling in our experiment are the most distinguishing properties of our current experiment.

\subsection{Data Sets}
\label{CH4_Integration_datasets}

We use annotated tweets that were released under scope of {\em First workshop on: Exploitation of Social Media for Emergency Relief and Preparedness} (SMERP) 2017  shared task as training data~\cite{Ghosh+17}.\footnote{\url{http://www.computing.dcu.ie/~dganguly/smerp2017/}, accessed June 10, 2018} The annotation labels represent the four target topics of the task. These topics are
\begin{inparaenum}[(i)]
\item {\bf T1} available resources;
\item {\bf T2} required resources;
\item {\bf T3} damage or casualties;
and
\item {\bf T4} rescue activities.
\end{inparaenum} This data set was collected during the earthquake that happened in Italy in August, 2016.\footnote{\url{https://en.wikipedia.org/wiki/August_2016_Central_Italy_earthquake}, accessed June 10, 2018}


The systems developed are tested on a similarly annotated data set that was released under the shared task {\em Microblog track: Information Extraction from Microblogs Posted during Disasters} organized in the scope of {\em Forum for Information Retrieval 2016} \cite{Ghosh+16} (FIRE).\footnote{\url{https://sites.google.com/site/fire2016microblogtrack/information-extraction-from-microblogs-posted-during-disasters}, accessed June 10, 2018} This data set is about the Nepal earthquake that happened in April, 2015.\footnote{\url{https://en.wikipedia.org/wiki/April_2015_Nepal_earthquake}, accessed June 10, 2018} Since FIRE contains seven topics, we use only tweets about topics, which are T1, T2, T3, and T4, that match between FIRE and SMERP data from FIRE. FMT1, FMT2, FMT3, FMT4 in FIRE dataset have been named as T1, T2, T3, and T4 respectively in this study. FMT5, FMT6, and FMT7 were excluded from the experiment reported in this chapter.  

The RB system was mostly developed on the basis of the data available in round 1, and the feedback we get for our submission to round 1. Extra time between round 1 and round 2 enabled us to add more lexical patterns and rules. Moreover, we used the output of the RB system on this larger set as training data for the ML system in our integration experiments. Our test on data from Nepal is prone to have missing data due to exclusion of the tweets that contain place references outside Italy. This phenomena may cause the performance on Nepal data to be lower than it could be when the RB system is adjusted to recognize tweets that contain reference to Nepal in-scope.

The SMERP and FIRE data sets contain 1,224 and 1,290 annotated tweets and 1,318 and 1,481 labels attached to them respectively. These tweets are labeled with one, two, or three of the aforementioned topic labels. Table~\ref{table:labelcooccurrences} presents the label co-occurrence patterns.\footnote{We left some cells empty to represent the number of co-occurring labels only once, e.g, T1 and T2 is not repeated in the T2, T1 column.} The diagonal shows the number of tweets that have only one label. In both data sets, T1 and T4 are found to be the two labels that co-occur most frequently. 
There are actually only three tweets that are annotated with three labels in these data sets, one tweet is labeled T1, T2, T4 in both data sets and one tweet is labeled with T1, T3, and T4 in SMERP.

\begin{table}[!htbp]
\ttabbox{\caption{Number of label co-occurring across the training (SMERP) and test set (FIRE)}\label{table:labelcooccurrences}}
{\begin{tabular*}{0.66\textwidth}{lcccccccc}
\toprule
{} & & SMERP && && FIRE & &\\
\midrule
{} &  T1 &  T2 &  T3 &  T4 {} &  T1 &  T2 &  T3 &  T4\\
\midrule
T1         &  {\bf 119}  &    2   &   1   &   51 & {\bf 409} & 3 & 8 & 151 \\
T2        &  -  &    {\bf 73}   &  2   &     30 & - & {\bf 261} & 23 & 3 \\
T3         &  -  &    -   &   {\bf 851}   &   4 & - & - & {\bf 215} & 1 \\
T4        &  -  &    -   &   -   &   {\bf 89} & - & - & - & {\bf 215} \\
\bottomrule
\end{tabular*}
}\end{table}

We observe class imbalance in both the SMERP and FIRE datasets, i.e. the majority of the labels are T3 and T1 in SMERP and FIRE respectively.

We apply multi-label machine learning and rule-based information classification to take into account the multiple topic labels for a tweet.

\subsection{Baseline}
\label{CH4_Integration_baseline}

We calculated a single and a double majority label and an all-labels-prediction based baselines in order to assess the performance of the methods we are using. The majority labels are identified in the training data (SMERP) and applied to test data (FIRE) to measure the baseline performance. The single-majority-label baseline predicts all as T3 and the double-majority-label baseline predicts all as T1 and T3. The all-labels-prediction baseline assigns all possible labels to all tweets. The scores for each baseline are provided in Table~\ref{table:baselinescores}.

\begin{table}[!htbp]
\ttabbox{\caption{Baseline scores for single and double majority and all-labels-prediction}\label{table:baselinescores}}
{\begin{tabular*}{0.55\textwidth}{lccc}
\toprule
{}  &  precision &  recall & F1 \\
\midrule
single-majority-label          &    .03   &   .17  & .05 \\
double-majority-label          &    .20   &  .55 & .29 \\
all-labels-prediction    &    .32   &   1.0 & .48 \\
\bottomrule
\end{tabular*}
}\end{table}

The different class distribution between the training and test sets causes single and double majority based baselines to perform poorly on the test set. The third baseline, all-labels-prediction, yields relatively good results. Therefore, we will be comparing our results to the third baseline.

\subsection{Building a Machine Learning Based Classifier}
\label{CH4_Integration_building_ml_classifier}
We trained an SVM classifier using TF-IDF weighted bag-of-words (BoW) features.\footnote{We used Scikit-learn version 0.19.1 for creating ML models~\cite{Pedregosa+11}.} We optimized the feature extraction and the classifier parameters jointly in a cross-validation setting on 80\% of the training data. The grid search for hyper-parameter optimization of the classifier yielded the values `char' for ngram\_type, between 3 and 4 for ngram\_range, and 0.9 for max\_df for the feature extraction and the value C=2 for the SVM classifier.\footnote{The tested values of the hyper-parameters are char, char\_wb, and word for ngram\_type, all valid combinations between 1 and 6 for ngram\_range, 0.8 and 0.9 for max\_df, and 0.01, 0.67, 1.33, and 2 for C.} We tested the created classifier using 80\% of the training data on the remaining 20\% held-out training data. The precision, recall, and F1 scores of this classifier are presented in Table~\ref{table:classifierPerfBasicML} under the SMERP-20\% column. These results show that the classifier has the capability to predict topic of out-of-sample tweets that are about the same disaster.

The right side of Table~\ref{table:classifierPerfBasicML} provides the performance of the classifier that was trained using the optimized parameters and all of the training data, both the 80\% that was used for hyper-parameter optimization and the 20\% that was held-out from hyper-parameter optimization. 


\begin{table}[!htbp]
\ttabbox{\caption{Precision, recall, and F1-score of the ML classifier on the held-out and test collection.}\label{table:classifierPerfBasicML}}
{
\setlength\tabcolsep{3.7pt}
\begin{tabular*}{0.99\textwidth}{lcccccccc}
\toprule
{} & & SMERP - 20\% && && FIRE & &\\
\midrule
{} &  precision &  recall &  F1 &  support {} &  precision &  recall &  F1 &  support\\
\midrule
T1         &  .83  &    .76   &   .79   &    33 & .65 & .49 & .56 & 572 \\
T2        &  .88  &    .75   &   .81   &     20 & .63 & .41 & .49 & 291 \\
T3         &  .99  &    .99   &   .99   &    173 & .81 & .81 & .81 & 247 \\
T4        &  .91  &    .78   &   .84   &     27 & .62 & .24 & .35 & 371 \\
\midrule
Avg/Total  &  .96  &   .92   &   .94   &   253 & .67 & .47 & .54 & 1,481 \\
\bottomrule
\end{tabular*}
}\end{table}

The results suggests that this approach yields high performance on tweets about the same disaster. However, the classifier performance decreases for all topics on tweets about a different disaster, represented by the FIRE data. The performance on topic T3 of the test data remains remarkably high.

\subsection{Rule-Based System}
\label{CH4_Integration_rule_based_sys}
We used the rule-based system we developed for the SMERP shared task~\cite{Hurriyetoglu+17} in this study. 
This system was developed using the raw data that was released by task organizers at the beginning of the task as not annotated. The topic descriptions were used to analyze the data and determine what should be covered for each topic. As a result, the lexicon and rule set were created manually based on that insight. Under these conditions the precision, recall, and F1 scores on all of the training data, which is annotated {\em SMERP} data, and the {\em FIRE} are reported in the Table~\ref{table:classifierPerfBasicRules}. 

 
\begin{table}[htb]
\ttabbox{\caption{Precision, recall, and F1-score of the RB on the test collection.}\label{table:classifierPerfBasicRules}}
{
\setlength\tabcolsep{3.7pt}
\begin{tabular*}{0.99\textwidth}{lcccccccc}
\toprule
{} & & SMERP - 100\% && && FIRE & &\\
\midrule
{} &  precision &  recall &  F1 &  support {} &  precision &  recall &  F1 &  support\\
\midrule
T1         &  .32  &    .13   &   .19   &    175 & .68 & .30 & .41 & 572 \\
T2        &  .38  &    .20   &   .27   &     108 & .65 & .24 & .35 & 291 \\
T3         &  .95  &    .78   &   .86   &    859 & .72 & .24 & .36 & 247 \\
T4        &  .87  &    .48   &   .62   &     176 & .62 & .08 & .15 & 371 \\
\midrule
Avg/Total  &  .81  &   .61   &   .69   &   1,318 & .67 & .22 & .33 & 1,481 \\
\bottomrule
\end{tabular*}
}\end{table}
 
On the one hand, we observe that the rule-based system is able to predict the majority topic (T3) and one of the minority topics (T4) relatively better than T1 and T2 in SMERP data. On the other hand, it yields consistent precision across different topics on the test data. The low recall scores cause the F1 score to be lower than the baseline on FIRE. The performance of the RB approach remains consistent and does not drop drastically on the test set.

The following subsections will provide the analysis of the performance difference and the integration experiment results of the described ML and RB approaches.


\subsection{Comparing Machine Learning and Rule-Based System Results}
\label{CH4_Integration_comparing_ml_rb}
In this subsection, we analyze the predictions of each system and compare them with each other in detail on the test data. The analysis is based on the number of predictions and their correctness. The difference between what each approach is capturing in comparison to the other approach and, in case both approaches yield a prediction, the exact and partial correctness of these predictions are inspected.

Tables~\ref{table:classifierPerfBasicML} and~\ref{table:classifierPerfBasicRules} illustrate that the precision of the ML approach decreases on the FIRE data set for all topics, while the precision of the RB approach decreases for T3 and T4 and increases for T1 and T2 respectively. The performance drop for the ML method is expected due to its limitedness to the training data. But having a precision increase for the RB method demonstrates the power of the domain and linguistic expertise that can be provided in terms of a dedicated lexicon and a set of rules. The RB approach yields comparable or better precision scores than the ML approach on all topics but the majority topic in the training set. ML ensures higher recall than RB. The consistent significantly high precision and recall of the ML approach on majority topic (T3) from SMERP, which is the training data for this model, illustrates the efficiency of ML approach on handling majority topics. On the other hand, although it is a known fact that the precision of RB methods is relatively high in comparison to ML approaches, observing this phenomenon on the minority topics that involve a significant increase of the score on new data set (FIRE) but not on majority topics is a remarkable result in the setting we report in this experiment.

The tweet and label prediction counts on FIRE are presented in Table~\ref{table:predictioncounts} for each approach. On the one hand the `Only ML' and Only RB' columns show the number of tweets that received a prediction only by ML or only by RB respectively. On the other hand, the `ML and RB' and `No Prediction' columns show the count of tweets that get a prediction from both ML and RB approaches and none of the approaches respectively.\footnote{`No Prediction' means that a system does not yield any of the topic labels as output.}

\begin{table}[!htbp]
\ttabbox{\caption{Prediction count analysis on FIRE}\label{table:predictioncounts}}
{\begin{tabular*}{0.85\textwidth}{lccccc}
\toprule
{} &  Only ML &  Only RB &  ML and RB & No Prediction & ML or RB \\
\midrule
T1         &  237  &    59   &   129  & 147 & 425 \\
T2        &  123  &    27   &  67 & 74  & 217 \\
T3        &  149  &    4   &   66 & 28  & 219 \\
T4        &  159  &    40   &   105 & 67  & 304 \\
\midrule
Total    &  668  &    114   &   309 & 281  & 1,009 \\
\bottomrule
\end{tabular*}
}\end{table}

An analysis of the columns in the Table~\ref{table:predictioncounts} reveals the strengths of each approach on a data set other than the training set accessible to these approaches at the system development phase in terms of number of predictions. We focus on the prediction quality for each column in this table in the remaining part of this section.

The {\em Only ML} column contains tweets that receive a prediction only from the ML system. The Table~\ref{table:onlyMLpredictScores} shows that the performance is comparable to the performance on all of the FIRE dataset (Table~\ref{table:classifierPerfBasicML}) except the higher precision for T3 and T4.\footnote{Since the recall was calculated on the restricted data set, we do not consider it as informative as the recall score calculated on the whole test set.}

\begin{table}[!htbp]
\ttabbox{\caption{Performance of the ML system where RB fails to predict any label for FIRE dataset.}\label{table:onlyMLpredictScores}}
{\begin{tabular*}{0.55\textwidth}{lcccc}
\toprule
{}  &     precision &   recall &    F1   & support \\
\midrule
T1          & .62   &    .78   &   .69   &    237 \\
T2          & .62   &    .59   &   .60   &    123 \\
T3          & .83   &    .93   &   .87   &    149 \\
T4          & .64   &    .33   &   .43   &    159 \\
\midrule
Avg/Total & .67  &     .67   &   .65   &    668  \\
\bottomrule
\end{tabular*}
}\end{table}

The {\em Only RB} column contains tweets that receive a prediction only from the RB system. The performance of the RB approach on these tweets is summarized in Table~\ref{table:onlyRBpredictScores}. We observe that the precision is remarkably high for T1 and T4.\footnote{We refer to minority classes in the training set.} The precision of T3 is significantly lower than the overall performance on FIRE (Table~\ref{table:classifierPerfBasicRules}), which drops from 0.72 to .30. Finally, T2 precision slightly drops from 0.65 to 0.56.

\begin{table}[!htbp]
\ttabbox{\caption{Performance of the RB system where ML fails to predict any label for FIRE dataset. }\label{table:onlyRBpredictScores}}
{\begin{tabular*}{0.55\textwidth}{lcccc}
\toprule
{}  &  precision &  recall & F1 & support \\
\midrule
T1          &    .73   &   .92  & .81 & 59 \\
T2          &    .56   &  .70 & .62 & 27 \\
T3    &    .30   &  .75  & .43 & 4 \\
T4    &    .69   &   .28 & .39 & 40 \\
\midrule
Avg/Total    &   .67   &   .67   &   .63    &   130  \\
\bottomrule
\end{tabular*}
}\end{table}

We calculated the cosine similarity of the tweets that have a prediction only from the ML or only from the RB approaches as an attempt to explain the difference in prediction performance.\footnote{We created a document that contains all tweets from the respective subset for each of the subsets.} The actual value of the cosine similarity between {\em Only ML}, {\em Only RB}, and {\em No Prediction} subsets to SMERP data is 0.102, 0.038, and 0.076 respectively. The cosine similarity of all of the FIRE set to SMERP set is 0.116. We interpret these scores as that the ML is successful where the training data and test data are similar, while RB is able to capture the least similar parts of the training and test data. 

{\em No Prediction} The number of tweets without any prediction is remarkably high. The RB approach succeeds in increasing the recall for T1 and T2 but the recall on T4 is low when applied to FIRE data. The recall of ML approach decreases for all topics. The recall of an approach may decrease due to information in the new data not being available or expressed differently in the data used to build the classification systems. An analysis toward the information related to blood donations in the train and test data showed that the SMERP data contains the phrases `donate blood' and `blood donation' whereas FIRE data contains `blood donors' and `blood requirements'. 


The columns {\em ML and RB} (intersection) and {\em ML or RB} (union) will be analyzed in the following section, where we provide the integration performance of the ML and RB approaches.

\subsection{Integrating ML and RB Approaches at the Result Level}
\label{CH4_Integration_integrating_ml_rb}
The performance difference directed us to integrate ML and RB approaches. We first integrate them at the result level. Next, we use raw predictions of the RB from Section~\ref{CH4_Integration_rule_based_sys} and a filtered version of them as training data for the ML approach. We use results on FIRE dataset for this comparison and experiment.  

The result level integration was performed by calculating the intersection and union of the predicted labels for each tweet by both of the approaches. The integration results in Table~\ref{table:classifierIntegration} demonstrate the results of this integration. These results show that overall, intersecting ML and RB yields better precision, while taking the union of ML and RB yields better recall.

\begin{table}[htb]
\ttabbox{\caption{Precision, recall, and F1-score of the classifier result integrations on the FIRE dataset.}\label{table:classifierIntegration}}
{
\begin{tabular*}{0.95\textwidth}{lccccccc}
\toprule
{} & & Intersection && & Union & &\\
\midrule
{} &        precision &  recall &   F1 {}  &  precision &  recall &    F1 &  support\\
\midrule
T1         &  .77     &   .16   &   .26    &    .64    &   .63    &   .63 &    572 \\
T2         &  .93     &   .14   &   .24    &    .59    &   .51    &   .54 &    291 \\
T3         &  .86     &   .23   &   .36    &    .77    &   .83    &   .80 &    247 \\
T4         &  .55     &   .03   &   .06    &    .63    &   .29    &   .40 &    371 \\
\midrule
Avg/Total  &  .76     &   .14   &   .23    &    .65    &   .56    &   .59  &   1,481 \\
\bottomrule
\end{tabular*}
}\end{table}

The precision of the topics T1, T2, and T3 increases in the intersection, but precision of T4 decreases. Moreover, in the union only T4's precision increases. T4's high precision originates from the high precision in the {\em Only RB} predictions. 

As another way of integrating the two approaches, we use the output of the RB for all of the data released in the scope of the SMERP shared task as training data for the ML approach.\footnote{Although we have used only the annotated part of the SMERP data in our ML experiments, RB enables access to the part of the released data that was not annotated as well.} The distinguishing aspect of this integration approach is that this training data creation step does not involve using gold standard annotated data. The output of the RB consists of 4,250 tweets, which contain 309, 211, 3,698, and 255 labels for T1, T2, T3, and T4 respectively.\footnote{There are more labels in total than the number of tweets due to multiple topic assignment to some tweets by RB.}

\begin{table}[htb]
\ttabbox{\caption{Precision, recall, and F1-score of the ML approach trained on RB output on SMERP dataset on FIRE dataset.}\label{table:RBasTrainingForML}}
{
\setlength\tabcolsep{5pt}
\begin{tabular*}{0.99\textwidth}{lcccccccc}
\toprule
{} & & SMERP && && FIRE & &\\
\midrule
{} &  precision &  recall &  F1 &  support {} &  precision &  recall &  F1 &  support\\
\midrule
T1         &  .45  &    .27   &   .34   &    175 & .61 & .45 & .52 & 572 \\
T2        &  .40  &    .26   &   .31   &     108 & .53 & .24 & .33 & 291 \\
T3         &  .91  &    .99   &   .95   &    859 & .48 & .89 & .63 & 247 \\
T4        &  .85  &    .50   &   .63   &     176 & .41 & .02 & .05 & 371 \\
\midrule
Avg/Total  &  .80  &   .77   &   .77   &   1,318 & .52 & .38 & .38 & 1,481 \\
\bottomrule
\end{tabular*}
}\end{table}

Table~\ref{table:RBasTrainingForML} presents results of the experiment that facilitates RB output as training data for the ML approach. When compared to Table~\ref{table:classifierPerfBasicRules}, this table show that RB output can be improved using the ML approach. Using the RB output as training data for the ML approach yielded an improvement over the RB approach of 8 F-score points for the SMERP data, from 69 to 77, and a 5 F-score point improvement for the FIRE data, from 33 to 38. Most of this gain was obtained from an increase in recall. Another benefit is the precision increase from 32 to 45 F-score and from 38 to 40 for T1 and T2 on SMERP data respectively.

As a final attempt at combining the two approaches, we aimed to improve the output of the RB approach by excluding the tweets that are not predicted at least one same label by the ML classifier created using the gold standard annotated data. As a result of this filtering, the final RB output data contained 113, 103, 3,519, and 144 labels for the topics T1, T2, T3, and T4 respectively. The performance of this hybrid method on FIRE is reported in Table~\ref{table:hybridByFiltering}. 

\begin{table}[!htbp]
\ttabbox{\caption{Performance of the hybrid system where RB output is filtered using ML predictions before being used as training data for ML.}\label{table:hybridByFiltering}}
{\begin{tabular*}{0.58\textwidth}{lcccc}
\toprule
{}  &     precision &   recall &    F1   & support \\
\midrule
T1          & .49   &   .16   &   .24   &    572 \\
T2          & .50   &   .21   &   .29   &    291 \\
T3          & .27   &   .98   &   .43   &    247 \\
T4          & .53   &   .03   &   .05   &    371 \\
\midrule
Avg/Total   & .46   &   .27   &   .23   &   1,481  \\
\bottomrule
\end{tabular*}
}\end{table}

The integration using filtered RB output as training data did not yield any significant improvement in comparison to using raw RB output. Therefore, we designed and performed the following experiment.

We observed that half of the instances of the minority topics were eliminated by the filtering applied in our aforementioned experiment. The observation that the classifier yields poor performance on minority topics in the training data, which are T1, T2, and T4, we filtered out only the T3 predicted tweets that are not predicted as T3 by the ML approach from the RB result. Then we used this selectively filtered RB output as training data for the ML approach. The final training set consists of 309, 211, 3,555, and 255 for T1, T2, T3, and T4. The results of this attempt are presented in Table~\ref{table:hybridByFilteringOnlyT3}.

\begin{table}[!htbp]
\ttabbox{\caption{Performance of the ML system where RB fails to predict any label on test data. Recall was calculated for this subset.}\label{table:hybridByFilteringOnlyT3}}
{\begin{tabular*}{0.58\textwidth}{lcccc}
\toprule
{}  &     precision &   recall &    F1   & support \\
\midrule
T1          & .61  &    .44  &    .51   &    572 \\
T2          & .52  &    .24  &    .33   &    291 \\
T3          & .54  &    .90  &    .68   &    247 \\
T4          & .43  &    .03  &    .05    &   371 \\
\midrule
Avg/Total   & .54  &    .37  &    .39   &   1,481 \\
\bottomrule
\end{tabular*}
}\end{table}

This last attempt yielded significantly higher scores than eliminating non-matching predictions for all topics on FIRE data. 

\subsection{Discussion}
\label{CH4_Integration_discussion}

The ML approach was proved to be relatively successful in predicting the training data in general and the majority topic in the test data. The RB approach was able to predict the minority topics better than ML. Since the RB approach had access only to the task instructions, without being able to observe the annotated data, we consider it remarkably successful. Each of the approaches was more successful on a distinct part of the test data in comparison to the other approach.

Our integration efforts yield critical clues about how to integrate these two distinct methods. In case annotated data is available, the intersection of the RB and ML predictions will yield higher precision and lower recall, while the union will yield lower precision and higher recall.

RB output can be used to train an ML classifier in case annotated data is not available. Such a classifier provides relatively better results than the raw RB output on the minority classes and slightly lower performance on the majority class on the training data. This classifier yields lower precision and higher recall than using raw RB output on the test data.

The availability of annotated data can facilitate the use of an ML classifier trained on it to filter RB output before creating a classifier from the RB output. This approach yields a slightly better performance if the filtering is applied only to the majority topic in the annotated data.

\subsection{Conclusion}
\label{CH4_Integration_conclusion}

We developed and tested machine learning and rule-based approaches for classifying tweets in four topics that are present at various ratios in the training and test sets. The majority and minority topics were drastically different from each other across the training sets as well. The systems developed started with different conditions, approached the problem from different angles, and yielded different results on different parts of the test set. These observations directed us to design hybrid systems that integrate these approaches.

The standalone system results showed that each approach performs well on a distinct part of the test set. This observation was confirmed by having a decrease in prediction precision of some topics when we use the intersection of the predictions as the final prediction from each approach. The union of the predictions yielded a balanced precision and recall and a higher F1 score. The result of the integration experiments yielded improved performance in case the output of RB is refined using the ML approach. 

In case annotated data is available, the best scenario is building an ML system and observing the data to develop an RB system. Combinations of the predictions from these two systems will yield the highest precision in case their predictions are intersected and the highest recall in case their predictions are unified. The output of ML is the best for the majority topics in the training set, whereas the RB system performs better on the minority topics for both the training and test sets.

In case annotated data is not available, using the RB output as training data for building an ML based classifier will yield the best results for the training data in terms of F1 score. However the performance of this classifier will be better only for the recall on the test data in comparison to raw RB output.


\section{Conclusion}
\label{CH4_conclusion}

Our work in the scope of this chapter showed that microtext classification can facilitate detecting actionable insights from tweet collections in disaster scenarios, but both machine-learning and rule-based approaches show weaknesses. We determined how the performance of these approaches depends on the starting conditions in terms of available resources. As a result, our conclusion is that availability of annotated data, domain and linguistic expertise together determine the path that should be followed and the performance that can be obtained in a use case.

Our participation in a shared task, in which our system was ranked best in the second round, shed light on the performance of our approaches in relation to other participating teams. Moreover, it provided us some insights about what can be potentially feasible and valuable to tackle the given task. The main conclusion is that extracting information about minority topics is the challenge of this kind of tasks and should be tackled through a combination of machine learning and rule-based approaches. 

\lhead{\emph{Conclusions}}
\chapter{Conclusions}
In this thesis we reported on a number of studies that we conducted with the aim of defining, detecting, and extracting actionable information from social media, more specifically from microtext collections. Each study provided insights into and proposed novel solutions for challenges in this field. As a result, we developed a comprehensive set of methodologies and software tools to tackle these challenges. 

In this last chapter, we iterate over the research questions and the problem statement that formed a basis for our research. The final subsections summarize the contributions of our studies and, extrapolating from our current findings, outline the direction that future research may take.

\section{Answers to Research Questions}

The first set of studies, which are reported in Chapter 2, centered around research question R1, which we repeat below:

\begin{adjustwidth}{1cm}{}
\textbf{RQ 1: To what extent can we detect patterns of content evolution in microtexts in order to generate time-to-event estimates from them?}
\end{adjustwidth}

The event time affects the nature and the quantity of the information posted about that event on social media. 
We have observed a particular occurrence pattern of microtext-content that informs about the starting time of social events, i.e events that are anticipated and attended by groups of (possibly many) people and are hence referred to a lot on social media. The details related to time of, preparations for it, and excitement toward the event occur in microtext-content extensively as the event time approaches. We reported the characteristics of this content evolution and the method we developed to utilize these characteristics for estimating the time of an upcoming football match or concert. 

Our experiments were designed to measure the effectiveness and usefulness of various microtext aggregation, feature extraction and selection, and estimate generation techniques. We first treated all microtexts that were posted within the period of one hour as one document and applied linear and local regression using bag-of-word (BoW) features. Then we focused on temporal expressions, word-skipgrams, and manually defined features, which are pertaining to the interpretation of temporal expressions, 
for feature extraction and mean and median as feature value assignment and estimate generation functions to measure their TTE estimation performance per tweet. Based on the results of these preliminary studies, we suggested a method that generates a time-to-event estimate for each tweet, combines these estimates with previous estimates by using rule-based and skip-gram based features in a certain order.

The final method is able to quite successfully estimate the time to event for in-domain and cross-domain training and test settings, which is under 4 and 10 hours off respectively.

The studies related to the second research question (R2),

\begin{adjustwidth}{1cm}{}
\textbf{RQ 2: How can we integrate domain expert knowledge and machine learning to identify relevant microtexts in a particular microtext collection?}
\end{adjustwidth}

had the overarching aim of extracting relevant information from microtext sets. Since relevance is a function of the data, an expert, and a target task, we focused on a semi-automatic solution that combines expert knowledge and automatic procedures. 

Our research confirmed the linguistic redundancy reported by Zanzotto et al.~\citeyear{Zanzotto+11}, which ascertain that 30\% of the tweets entail already posted information. Indeed, we observed that in many tweet collections, e.g. those collected on the basis of keywords or hashtags, a substantial amount of tweets in the collection contain repetitive information. Thus, we developed a method to collect data by taking inflections into account effectively, to eliminate exact- and near-duplicates, to extract a representative sample of the microtext collection in terms of clusters, to label them effectively, and to use this annotation for training a classifier that can be used to label new microtexts automatically. We tested and improved this method on uses cases about genocide, earthquake, and the flu domains. Consequently, the classifiers that could be created using this methodology are able to distinguish between relevant and irrelevant microtexts, and then to classify relevant microtexts into further fine-grained topics relevant to the target of the study.
We reported on estimated accuracies of between 0.67 and 0.95 F-score on new data.

Selected microtexts contain a lot of information that can be extracted for a detailed overview of an event. Therefore, we investigated the extent to which we can extract this detailed information in the scope of the third research question (R3) repeated below:

\begin{adjustwidth}{1cm}{}
\textbf{RQ 3: To what extent are rule-based and ML-based approaches complementary for classifying microtexts based on small datasets?}
\end{adjustwidth}

We compared a rule-based and a machine learning method for classifying microtexts in four topic classes. We reported on the performance of our approaches applied to detailed microtext classification, a relatively underexplored area of research.
The results suggested that rule-based and machine-learning-based approaches perform well on different parts of microtext collections. Therefore, combining their predictions tends to yield higher performance scores than applying them stand-alone. For instance, applying ML on the output of the rule-based system improves the recall of all topics and precision of the topic of interest, which was a minority topic in our case study. Moreover, filtering the rule-based system output for the majority topics using the stand-alone ML system enhances the performance that can be obtained from the rule-base system. The robustness of our results was tested by designing our experiments using noisy, imbalanced, and scarce data.

Our studies related to each of three research questions yielded approaches that estimate time to event, detect relevant documents, and classify relevant documents into topical classes. 
The performance of these approaches was evaluated in various case studies and improved in multiple iterations based on observations and results of previous iterations. Major improvements that were derived in this manner were performing priority-based historical context integration, using the temporal distribution in clustering the microtext collections, and integrating rule-based and machine learning based approaches for time-to-event estimation, relevant document detection, and relevant information classification approaches respectively. For instance, from our experiments we conclude that rule-based and machine-learning-based approaches are indeed complementary, and their complementarity can be exploited to improve on precision, recall, or both. Which combination method works best  (e.g. taking the intersection or union, or including predictions of one route into the other) remains an empirical question. The topics/domains of the use cases were football matches and music concerts for time-to-event estimation, flood, genocide, flu, and earthquake for the relevant microtext detection, and earthquake for the microtext classification methods.


\section{Answer to Problem Statement}

\begin{adjustwidth}{1cm}{}
\textbf{PS: How can we develop an efficient automatic system that, with a high degree of precision and completeness, can identify actionable information about major events in a timely manner from microblogs while taking into account microtext and social media characteristics?}
\end{adjustwidth}

Every experiment that was performed in the scope of this dissertation provided insights about what should be the next step in the line specified in our problem statement. Identifying characteristics of the microtext collections about events and utilizing this information to tackle this challenge was at the core of our efforts. Full automation of the process has been the ultimate aim of our work. Therefore, the first parts of Chapters 2 and 3 focused only on automatic approaches. The results from Chapter 3 showed us that microtext collections tend to contain substantial amounts of irrelevant posts, and that we require the input of experts to distinguish relevant from irrelevant data. Consequently, we directed our efforts to study incorporation of the users' insight in describing what is relevant to their use case and transferring this description to a machine learning model. This attempt yielded a machine learning based methodology, Relevancer, that performs well on detecting irrelevant information. 
However, detecting information about relevant topics that are relatively less frequent than the majority topics in a collection was only possible to a limited degree with this approach. Fine-grained topics may not be represented as clusters at a degree that can be labeled and be used by supervised machine learning techniques. Consequently, we integrated this approach with a linguistically motivated rule-based approach and obtained a robust system that alleviated this minority-class problem.

We tested parts of the system on collections from various domains, such as flood, genocide, flu, and earthquake and we applied the hybrid rule-based and ML-based system to earthquake disaster data. The cross-domain and cross-event success of the time-to-event estimation method, which is trained on football events and tested on music events, and of the information classification, which is trained on earthquake data from Italy and tested on earthquake data from Nepal, respectively, show that our system is robust enough to handle real-world variation.


\section{Thesis Contributions}

In sum, we have made the following contributions to the growing field of event information extraction from social media:
\begin{enumerate}
\item We developed a feature extraction and time-to-event estimation method for predicting the start time of social events. This methodology is tested on microtexts in in-domain and cross-domain scenarios, for which the estimations are under 4 and 10 hours off on average respectively. 

\item The results of our experiments show that to counteract negative effects of outlier microtexts when estimating time to event requires using the median as training function, integrating a history of estimates, and using the mean absolute error as a performance measure.

\item Our studies provided insight into the degree of redundancy of the content on social media. Machine learning techniques were used to facilitate the need of experts to filter irrelevant data and zoom in on what is relevant for the task at hand.

\item We showed that by exploiting the temporal sub-structure of microtext distributions over time decreases the time spent on clustering and improves the chance of obtaining a representative cluster set.

\item We suggested a method to integrate rule-based and machine learning oriented approaches to alleviate annotated data scarcity and class imbalance issues.
\end{enumerate}

\section{Outlook}
Although the approaches we developed deliver reasonable to sufficient performance, there still are areas for continued development. We have identified lines of research that have the potential to improve the approach we developed in the scope of this dissertation. We iterate over these ideas in this subsection.

Our time-to-event estimation method has a number of logical extensions in the line of used data, features, applied operations, and applicability to different domains. 

TTE estimation should remain reliable on event data that is noisier than the data collected by using a single hashtag -- this could be tested using the output of our relevant microtext detection or some automatic event detection methods that detect events. Moreover, the method should also be applied to data that come from different social media sources, e.g Facebook or internet forum posts, to investigate its robustness towards source type. The TTE method was evaluated on baselines we have determined. However, this method should be compared to other proposed time series methods as well~\cite{Hmamouche+19}.

Further analysis and improvement of the word skipgram based features have the potential to make the time-to-event estimation method more flexible by decreasing the need for temporal expression extraction and rule generation steps. In case use of temporal expressions, determining the relevance of temporal expressions in case there are several such expressions in a single message would have considerable importance as well.

Social media data contain relatively many outlier instances that affect the performance of our methods drastically. Our results show that using the median as a value assignment and estimation function and adjusting estimations with a window of preceding estimations remedy this issue to some extent. Further research toward understanding and handling these outliers has the potential to improve the robustness of our approach.
The number of posted tweets is a significant indicator of event time. We observed this phenomenon in our baselines. Based on this insight, we anticipate that analyzing changes in tweet frequencies relative to an event may support the feature selection and TTE estimation phases. 

Finally, the time-to-event estimation method could be extended by moving from football to other scheduled events, and from scheduled events to unscheduled events, the ultimate goal of a forecasting system like this. This extension can be achieved by applying event detection and classification systems as proposed by Kunneman and van den Bosch~\citeyear{Kunneman+14b} and Van Noord et al.~\citeyear{Noord+17} respectively.

Improving the feature extraction by including skip-grams, word embeddings~\cite{Mikolov+13}, considering other types of information present in the social media platform (e.g. features that characterize the user, such as numbers of followers and personal descriptions), and introducing more sophisticated clustering~\cite{Fahad+14} and cluster evaluation metrics~\cite{Lee+12} have the potential to improve clustering step of the Relevancer approach. Moreover, incorporating the feedback from the experts about which users' posts or hashtags should best be ignored or included will improve the labeling process. This information can be used to update and continuously evaluate the clusters and the classifiers. The posting user and included hashtags have the potential to provide information about certain information threads. This information can enable the annotation step to be expanded to the tweets that are not in any cluster but contain information thread specific hashtags or users.


We do not carry out any post-processing on the clusters in the Relevancer approach. However, identifying outlier samples in a cluster in terms of the distance to the cluster center in order to refine the clusters can enhance obtaining coherent clusters. This could enable Relevancer to detect cleaner microtext clusters in fewer iterations. Application of the method proposed by Henelius et al.~\citeyear{Henelius+16} could improve the quality of the clusters as well. 





In general, machine learning approaches miss relatively `small' topics. The clustering and classifying steps should be improved to yield coherent clusters for small topics and to utilize the information about small topics
in the automatic classification respectively. The results of the rule-based approach demonstrated their potential in remedying this issue. Further exploration of how to best make use of the rule-based approach should benefit the use of machine learning both for clustering and classification. 




Microtexts may intend to mislead public by spreading false information or fake news. Our methodology attempts to restrict effect of this kind of microtexts by excluding duplicate microtexts and facilitate information in an aggregated manner. The effectiveness of our approach should be analyzed in detail. Moreover, the credibility analysis of the microtexts and users who post them should be made a standard step of microtext collection analysis studies. 

Fully automatizing microtext analysis has been our goal since the first day of this research project. Our efforts in this direction informed us about the extent this automation can be realized. We mostly first developed an automated approach, then we extended and improved it by integrating human intervention at various steps of the automated approach. Our experience confirms previous work that states that a well designed human intervention or contribution in design, realization, or evaluation of an information system either improves its performance or enables its realization. As our studies and results directed us toward its necessity and value, we were inspired from previous studies in designing human involvement and customized our approaches to benefit from human input. Consequently, our contribution to existing body of research in this line has become the confirmation of the value of human intervention in extracting actionable information from microtexts.






\addtocontents{toc}{\vspace{2em}} 

\appendix 

\addtocontents{toc}{\vspace{2em}}  
\backmatter

\label{Bibliography}
\lhead{\emph{Bibliography}}  
\bibliographystyle{apacite}  
\bibliography{myadnext}

\begin{thebibliography}{}

\bibitem [\protect \citeauthoryear {%
Agichtein%
, Castillo%
, Donato%
, Gionis%
\BCBL {}\ \BBA {} Mishne%
}{%
Agichtein%
\ \protect \BOthers {.}}{%
{\protect \APACyear {2008}}%
}]{%
Agichtein+08}
\APACinsertmetastar {%
Agichtein+08}%
\begin{APACrefauthors}%
Agichtein, E.%
, Castillo, C.%
, Donato, D.%
, Gionis, A.%
\BCBL {}\ \BBA {} Mishne, G.%
\end{APACrefauthors}%
\unskip\
\newblock
\APACrefYearMonthDay{2008}{}{}.
\newblock
{\BBOQ}\APACrefatitle {{Finding High-quality Content in Social Media}}
  {{Finding High-quality Content in Social Media}}.{\BBCQ}
\newblock
\BIn{} \APACrefbtitle {{Proceedings of the 2008 International Conference on Web
  Search and Data Mining}} {{Proceedings of the 2008 International Conference
  on Web Search and Data Mining}}\ (\BPGS\ 183--194).
\newblock
\APACaddressPublisher{New York, NY, USA}{ACM}.
\newblock
\begin{APACrefURL} \url{http://doi.acm.org/10.1145/1341531.1341557}
  \end{APACrefURL}
\newblock
\begin{APACrefDOI} \doi{10.1145/1341531.1341557} \end{APACrefDOI}
\PrintBackRefs{\CurrentBib}

\bibitem [\protect \citeauthoryear {%
Allaire%
}{%
Allaire%
}{%
{\protect \APACyear {2016}}%
}]{%
Allaire16}
\APACinsertmetastar {%
Allaire16}%
\begin{APACrefauthors}%
Allaire, M\BPBI C.%
\end{APACrefauthors}%
\unskip\
\newblock
\APACrefYearMonthDay{2016}{}{}.
\newblock
{\BBOQ}\APACrefatitle {{Disaster loss and social media: Can online information
  increase flood resilience?}} {{Disaster loss and social media: Can online
  information increase flood resilience?}}{\BBCQ}
\newblock
\APACjournalVolNumPages{{Water Resources Research}}{52}{9}{7408--7423}.
\newblock
\begin{APACrefURL} \url{http://dx.doi.org/10.1002/2016WR019243}
  \end{APACrefURL}
\newblock
\begin{APACrefDOI} \doi{10.1002/2016WR019243} \end{APACrefDOI}
\PrintBackRefs{\CurrentBib}

\bibitem [\protect \citeauthoryear {%
Atkeson%
, Moore%
\BCBL {}\ \BBA {} Schaal%
}{%
Atkeson%
\ \protect \BOthers {.}}{%
{\protect \APACyear {1997}}%
}]{%
Atkeson+97}
\APACinsertmetastar {%
Atkeson+97}%
\begin{APACrefauthors}%
Atkeson, C.%
, Moore, A.%
\BCBL {}\ \BBA {} Schaal, S.%
\end{APACrefauthors}%
\unskip\
\newblock
\APACrefYearMonthDay{1997}{}{}.
\newblock
{\BBOQ}\APACrefatitle {Locally weighted learning} {Locally weighted
  learning}.{\BBCQ}
\newblock
\APACjournalVolNumPages{{Artificial Intelligence Review}}{11}{1--5}{11--73}.
\PrintBackRefs{\CurrentBib}

\bibitem [\protect \citeauthoryear {%
Baeza~Yates%
}{%
Baeza~Yates%
}{%
{\protect \APACyear {2005}}%
}]{%
Ricardo05}
\APACinsertmetastar {%
Ricardo05}%
\begin{APACrefauthors}%
Baeza~Yates, R.%
\end{APACrefauthors}%
\unskip\
\newblock
\APACrefYearMonthDay{2005}{}{}.
\newblock
{\BBOQ}\APACrefatitle {Searching the future} {Searching the future}.{\BBCQ}
\newblock
\BIn{} \APACrefbtitle {{ACM SIGIR Workshop on Mathematical/Formal Methods for
  Information Retrieval (MF/IR 2005)}.} {{ACM SIGIR Workshop on
  Mathematical/Formal Methods for Information Retrieval (MF/IR 2005)}.}
\PrintBackRefs{\CurrentBib}

\bibitem [\protect \citeauthoryear {%
Baldwin%
, Cook%
, Lui%
, MacKinlay%
\BCBL {}\ \BBA {} Wang%
}{%
Baldwin%
\ \protect \BOthers {.}}{%
{\protect \APACyear {2013}}%
}]{%
Baldwin+13}
\APACinsertmetastar {%
Baldwin+13}%
\begin{APACrefauthors}%
Baldwin, T.%
, Cook, P.%
, Lui, M.%
, MacKinlay, A.%
\BCBL {}\ \BBA {} Wang, L.%
\end{APACrefauthors}%
\unskip\
\newblock
\APACrefYearMonthDay{2013}{}{}.
\newblock
{\BBOQ}\APACrefatitle {{How noisy social media text, how diffrnt social media
  sources}} {{How noisy social media text, how diffrnt social media
  sources}}.{\BBCQ}
\newblock
\BIn{} \APACrefbtitle {{Proceedings of the 6th International Joint Conference
  on Natural Language Processing (IJCNLP 2013)}} {{Proceedings of the 6th
  International Joint Conference on Natural Language Processing (IJCNLP
  2013)}}\ (\BPGS\ 356--364).
\PrintBackRefs{\CurrentBib}

\bibitem [\protect \citeauthoryear {%
Batista%
, Prati%
\BCBL {}\ \BBA {} Monard%
}{%
Batista%
\ \protect \BOthers {.}}{%
{\protect \APACyear {2004}}%
}]{%
Batista+04}
\APACinsertmetastar {%
Batista+04}%
\begin{APACrefauthors}%
Batista, G\BPBI E\BPBI A\BPBI P\BPBI A.%
, Prati, R\BPBI C.%
\BCBL {}\ \BBA {} Monard, M\BPBI C.%
\end{APACrefauthors}%
\unskip\
\newblock
\APACrefYearMonthDay{2004}{{\APACmonth{06}}}{}.
\newblock
{\BBOQ}\APACrefatitle {{A Study of the Behavior of Several Methods for
  Balancing Machine Learning Training Data}} {{A Study of the Behavior of
  Several Methods for Balancing Machine Learning Training Data}}.{\BBCQ}
\newblock
\APACjournalVolNumPages{{SIGKDD Explor. Newsl.}}{6}{1}{20--29}.
\newblock
\begin{APACrefURL} \url{http://doi.acm.org/10.1145/1007730.1007735}
  \end{APACrefURL}
\newblock
\begin{APACrefDOI} \doi{10.1145/1007730.1007735} \end{APACrefDOI}
\PrintBackRefs{\CurrentBib}

\bibitem [\protect \citeauthoryear {%
Becker%
, Iter%
, Naaman%
\BCBL {}\ \BBA {} Gravano%
}{%
Becker%
\ \protect \BOthers {.}}{%
{\protect \APACyear {2012}}%
}]{%
Becker+12}
\APACinsertmetastar {%
Becker+12}%
\begin{APACrefauthors}%
Becker, H.%
, Iter, D.%
, Naaman, M.%
\BCBL {}\ \BBA {} Gravano, L.%
\end{APACrefauthors}%
\unskip\
\newblock
\APACrefYearMonthDay{2012}{}{}.
\newblock
{\BBOQ}\APACrefatitle {{Identifying content for planned events across social
  media sites}} {{Identifying content for planned events across social media
  sites}}.{\BBCQ}
\newblock
\BIn{} \APACrefbtitle {{Proceedings of the fifth ACM international conference
  on Web search and data mining}} {{Proceedings of the fifth ACM international
  conference on Web search and data mining}}\ (\BPGS\ 533--542).
\newblock
\APACaddressPublisher{New York, NY, USA}{ACM}.
\newblock
\begin{APACrefURL} \url{http://doi.acm.org/10.1145/2124295.2124360}
  \end{APACrefURL}
\newblock
\begin{APACrefDOI} \doi{10.1145/2124295.2124360} \end{APACrefDOI}
\PrintBackRefs{\CurrentBib}

\bibitem [\protect \citeauthoryear {%
Blamey%
, Crick%
\BCBL {}\ \BBA {} Oatley%
}{%
Blamey%
\ \protect \BOthers {.}}{%
{\protect \APACyear {2013}}%
}]{%
Blamey+13}
\APACinsertmetastar {%
Blamey+13}%
\begin{APACrefauthors}%
Blamey, B.%
, Crick, T.%
\BCBL {}\ \BBA {} Oatley, G.%
\end{APACrefauthors}%
\unskip\
\newblock
\APACrefYearMonthDay{2013}{}{}.
\newblock
{\BBOQ}\APACrefatitle {{`The First Day of Summer': Parsing Temporal Expressions
  with Distributed Semantics}} {{`The First Day of Summer': Parsing Temporal
  Expressions with Distributed Semantics}}.{\BBCQ}
\newblock
\BIn{} M.~Bramer\ \BBA {} M.~Petridis\ (\BEDS), \APACrefbtitle {{Research and
  Development in Intelligent Systems XXX}} {{Research and Development in
  Intelligent Systems XXX}}\ (\BPG~389-402).
\newblock
\APACaddressPublisher{}{{Springer International Publishing}}.
\newblock
\begin{APACrefURL} \url{http://dx.doi.org/10.1007/978-3-319-02621-3\_29}
  \end{APACrefURL}
\newblock
\begin{APACrefDOI} \doi{10.1007/978-3-319-02621-3\_29} \end{APACrefDOI}
\PrintBackRefs{\CurrentBib}

\bibitem [\protect \citeauthoryear {%
Borra%
\ \BBA {} Rieder%
}{%
Borra%
\ \BBA {} Rieder%
}{%
{\protect \APACyear {2014}}%
}]{%
Borra+14}
\APACinsertmetastar {%
Borra+14}%
\begin{APACrefauthors}%
Borra, E.%
\BCBT {}\ \BBA {} Rieder, B.%
\end{APACrefauthors}%
\unskip\
\newblock
\APACrefYearMonthDay{2014}{}{}.
\newblock
{\BBOQ}\APACrefatitle {{Programmed method: developing a toolset for capturing
  and analyzing tweets}} {{Programmed method: developing a toolset for
  capturing and analyzing tweets}}.{\BBCQ}
\newblock
\APACjournalVolNumPages{Aslib Journal of Information
  Management}{66}{3}{262--278}.
\newblock
\begin{APACrefURL} \url{http://dx.doi.org/10.1108/AJIM-09-2013-0094}
  \end{APACrefURL}
\newblock
\begin{APACrefDOI} \doi{10.1108/AJIM-09-2013-0094} \end{APACrefDOI}
\PrintBackRefs{\CurrentBib}

\bibitem [\protect \citeauthoryear {%
Borthwick%
, Sterling%
, Agichtein%
\BCBL {}\ \BBA {} Grishman%
}{%
Borthwick%
\ \protect \BOthers {.}}{%
{\protect \APACyear {1998}}%
}]{%
Borthwick+98}
\APACinsertmetastar {%
Borthwick+98}%
\begin{APACrefauthors}%
Borthwick, A.%
, Sterling, J.%
, Agichtein, E.%
\BCBL {}\ \BBA {} Grishman, R.%
\end{APACrefauthors}%
\unskip\
\newblock
\APACrefYearMonthDay{1998}{}{}.
\newblock
{\BBOQ}\APACrefatitle {{Exploiting Diverse Knowledge Sources via Maximum
  Entropy in Named Entity Recognition}} {{Exploiting Diverse Knowledge Sources
  via Maximum Entropy in Named Entity Recognition}}.{\BBCQ}
\newblock
\BIn{} \APACrefbtitle {{Proceedings of the Sixth Workshop on Very Large
  Corpora}} {{Proceedings of the Sixth Workshop on Very Large Corpora}}\
  (\BPGS\ 152--160).
\newblock
\begin{APACrefURL}
  \url{http://citeseerx.ist.psu.edu/viewdoc/summary?doi=10.1.1.14.8357}
  \end{APACrefURL}
\PrintBackRefs{\CurrentBib}

\bibitem [\protect \citeauthoryear {%
Briscoe%
, Appling%
\BCBL {}\ \BBA {} Schlosser%
}{%
Briscoe%
\ \protect \BOthers {.}}{%
{\protect \APACyear {2015}}%
}]{%
Briscoe+15}
\APACinsertmetastar {%
Briscoe+15}%
\begin{APACrefauthors}%
Briscoe, E.%
, Appling, S.%
\BCBL {}\ \BBA {} Schlosser, J.%
\end{APACrefauthors}%
\unskip\
\newblock
\APACrefYearMonthDay{2015}{}{}.
\newblock
{\BBOQ}\APACrefatitle {{Passive Crowd Sourcing for Technology Prediction}}
  {{Passive Crowd Sourcing for Technology Prediction}}.{\BBCQ}
\newblock
\BIn{} N.~Agarwal, K.~Xu\BCBL {}\ \BBA {} N.~Osgood\ (\BEDS), \APACrefbtitle
  {{Social Computing, Behavioral-Cultural Modeling, and Prediction}} {{Social
  Computing, Behavioral-Cultural Modeling, and Prediction}}\ (\BVOL\ 9021,
  \BPG~264-269).
\newblock
\APACaddressPublisher{}{Springer International Publishing}.
\newblock
\begin{APACrefURL} \url{http://dx.doi.org/10.1007/978-3-319-16268-3\_28}
  \end{APACrefURL}
\newblock
\begin{APACrefDOI} \doi{10.1007/978-3-319-16268-3\_28} \end{APACrefDOI}
\PrintBackRefs{\CurrentBib}

\bibitem [\protect \citeauthoryear {%
Casati%
\ \BBA {} Varzi%
}{%
Casati%
\ \BBA {} Varzi%
}{%
{\protect \APACyear {2015}}%
}]{%
Casati+15}
\APACinsertmetastar {%
Casati+15}%
\begin{APACrefauthors}%
Casati, R.%
\BCBT {}\ \BBA {} Varzi, A.%
\end{APACrefauthors}%
\unskip\
\newblock
\APACrefYearMonthDay{2015}{}{}.
\newblock
{\BBOQ}\APACrefatitle {Events} {Events}.{\BBCQ}
\newblock
\BIn{} E\BPBI N.~Zalta\ (\BED), \APACrefbtitle {The Stanford Encyclopedia of
  Philosophy} {The stanford encyclopedia of philosophy}\ (\PrintOrdinal{Winter
  2015}\ \BEd).
\newblock
\APACaddressPublisher{}{Metaphysics Research Lab, Stanford University}.
\newblock
\APAChowpublished
  {\url{https://plato.stanford.edu/archives/win2015/entries/events/}}.
\PrintBackRefs{\CurrentBib}

\bibitem [\protect \citeauthoryear {%
Chang%
\ \BBA {} Manning%
}{%
Chang%
\ \BBA {} Manning%
}{%
{\protect \APACyear {2012}}%
}]{%
Chang+12}
\APACinsertmetastar {%
Chang+12}%
\begin{APACrefauthors}%
Chang, A\BPBI X.%
\BCBT {}\ \BBA {} Manning, C.%
\end{APACrefauthors}%
\unskip\
\newblock
\APACrefYearMonthDay{2012}{may}{}.
\newblock
{\BBOQ}\APACrefatitle {{SUTime: A library for recognizing and normalizing time
  expressions}} {{SUTime: A library for recognizing and normalizing time
  expressions}}.{\BBCQ}
\newblock
\BIn{} N\BPBI C\BPBI C.~Chair)\ \BOthers {.}\ (\BEDS), \APACrefbtitle
  {{Proceedings of the Eight International Conference on Language Resources and
  Evaluation (LREC'12)}.} {{Proceedings of the Eight International Conference
  on Language Resources and Evaluation (LREC'12)}.}
\newblock
\APACaddressPublisher{Istanbul, Turkey}{European Language Resources Association
  (ELRA)}.
\newblock
\begin{APACrefURL}
  \url{http://www.lrec-conf.org/proceedings/lrec2012/summaries/284.html}
  \end{APACrefURL}
\PrintBackRefs{\CurrentBib}

\bibitem [\protect \citeauthoryear {%
Chau%
}{%
Chau%
}{%
{\protect \APACyear {2012}}%
}]{%
Chau12}
\APACinsertmetastar {%
Chau12}%
\begin{APACrefauthors}%
Chau, D\BPBI H.%
\end{APACrefauthors}%
\unskip\
\newblock
\APACrefYearMonthDay{2012}{}{}.
\newblock
\APACrefbtitle {Data Mining Meets HCI: Making Sense of Large Graphs} {Data
  mining meets hci: Making sense of large graphs}\ \APACbVolEdTR{}{\BTR{}}.
\newblock
\APACaddressInstitution{}{DTIC Document}.
\newblock
\begin{APACrefURL} \url{http://repository.cmu.edu/dissertations/94/}
  \end{APACrefURL}
\PrintBackRefs{\CurrentBib}

\bibitem [\protect \citeauthoryear {%
Chaudhuri%
\ \BBA {} Bhattacharya%
}{%
Chaudhuri%
\ \BBA {} Bhattacharya%
}{%
{\protect \APACyear {2008}}%
}]{%
Chaudhuri+08}
\APACinsertmetastar {%
Chaudhuri+08}%
\begin{APACrefauthors}%
Chaudhuri, B\BPBI B.%
\BCBT {}\ \BBA {} Bhattacharya, S.%
\end{APACrefauthors}%
\unskip\
\newblock
\APACrefYearMonthDay{2008}{}{}.
\newblock
{\BBOQ}\APACrefatitle {{An experiment on automatic detection of named entities
  in Bangla}} {{An experiment on automatic detection of named entities in
  Bangla}}.{\BBCQ}
\newblock
\BIn{} \APACrefbtitle {{Proceedings of the IJCNLP-08 Workshop on NER for South
  and South East Asian Languages,Hyderabad, India. Asian Federation of Natural
  Language Processing}} {{Proceedings of the IJCNLP-08 Workshop on NER for
  South and South East Asian Languages,Hyderabad, India. Asian Federation of
  Natural Language Processing}}\ (\BPGS\ 75--81).
\PrintBackRefs{\CurrentBib}

\bibitem [\protect \citeauthoryear {%
Cohen%
, Brink%
, Adang%
, Dijk%
\BCBL {}\ \BBA {} Boeschoten%
}{%
Cohen%
\ \protect \BOthers {.}}{%
{\protect \APACyear {2013}}%
}]{%
Cohen+13}
\APACinsertmetastar {%
Cohen+13}%
\begin{APACrefauthors}%
Cohen, M\BPBI J.%
, Brink, G\BPBI J\BPBI M.%
, Adang, O\BPBI M\BPBI J.%
, Dijk, J\BPBI A\BPBI G\BPBI M.%
\BCBL {}\ \BBA {} Boeschoten, T.%
\end{APACrefauthors}%
\unskip\
\newblock
\APACrefYearMonthDay{2013}{}{}.
\newblock
\APACrefbtitle {Twee werelden: You only live once} {Twee werelden: You only
  live once}\ \APACbVolEdTR{}{\BTR{}}.
\newblock
\APACaddressInstitution{The Hague, The Netherlands}{Ministerie van Veiligheid
  en Justitie}.
\PrintBackRefs{\CurrentBib}

\bibitem [\protect \citeauthoryear {%
De~Choudhury%
, Counts%
\BCBL {}\ \BBA {} Czerwinski%
}{%
De~Choudhury%
\ \protect \BOthers {.}}{%
{\protect \APACyear {2011}}%
}]{%
DeChoudhury+11}
\APACinsertmetastar {%
DeChoudhury+11}%
\begin{APACrefauthors}%
De~Choudhury, M.%
, Counts, S.%
\BCBL {}\ \BBA {} Czerwinski, M.%
\end{APACrefauthors}%
\unskip\
\newblock
\APACrefYearMonthDay{2011}{July}{}.
\newblock
\APACrefbtitle {{Find Me the Right Content! Diversity-based Sampling of Social
  Media Content for Topic-centric Search.}} {{Find Me the Right Content!
  Diversity-based Sampling of Social Media Content for Topic-centric Search.}}
\newblock
\APACaddressPublisher{}{Int'l AAAI Conference on Weblogs and Social Media}.
\newblock
\begin{APACrefURL}
  \url{https://www.microsoft.com/en-us/research/publication/find-me-the-right-content-diversity-based-sampling-of-social-media-content-for-topic-centric-search/}
  \end{APACrefURL}
\PrintBackRefs{\CurrentBib}

\bibitem [\protect \citeauthoryear {%
De~Choudhury%
, Diakopoulos%
\BCBL {}\ \BBA {} Naaman%
}{%
De~Choudhury%
\ \protect \BOthers {.}}{%
{\protect \APACyear {2012}}%
}]{%
DeChoudhury+12b}
\APACinsertmetastar {%
DeChoudhury+12b}%
\begin{APACrefauthors}%
De~Choudhury, M.%
, Diakopoulos, N.%
\BCBL {}\ \BBA {} Naaman, M.%
\end{APACrefauthors}%
\unskip\
\newblock
\APACrefYearMonthDay{2012}{}{}.
\newblock
{\BBOQ}\APACrefatitle {Unfolding the Event Landscape on Twitter: Classification
  and Exploration of User Categories} {Unfolding the event landscape on
  twitter: Classification and exploration of user categories}.{\BBCQ}
\newblock
\BIn{} \APACrefbtitle {{Proceedings of the ACM 2012 Conference on Computer
  Supported Cooperative Work}} {{Proceedings of the ACM 2012 Conference on
  Computer Supported Cooperative Work}}\ (\BPGS\ 241--244).
\newblock
\APACaddressPublisher{New York, NY, USA}{ACM}.
\newblock
\begin{APACrefURL} \url{http://doi.acm.org/10.1145/2145204.2145242}
  \end{APACrefURL}
\newblock
\begin{APACrefDOI} \doi{10.1145/2145204.2145242} \end{APACrefDOI}
\PrintBackRefs{\CurrentBib}

\bibitem [\protect \citeauthoryear {%
Dias%
, Campos%
\BCBL {}\ \BBA {} Jorge%
}{%
Dias%
\ \protect \BOthers {.}}{%
{\protect \APACyear {2011}}%
}]{%
Dias+11}
\APACinsertmetastar {%
Dias+11}%
\begin{APACrefauthors}%
Dias, G.%
, Campos, R.%
\BCBL {}\ \BBA {} Jorge, A.%
\end{APACrefauthors}%
\unskip\
\newblock
\APACrefYearMonthDay{2011}{}{}.
\newblock
{\BBOQ}\APACrefatitle {Future Retrieval: What Does the Future Talk About?}
  {Future retrieval: What does the future talk about?}{\BBCQ}
\newblock
\BIn{} \APACrefbtitle {{Proceedings SIGIR2011 Workshop on Enriching Information
  Retrieval (ENIR2011)}.} {{Proceedings SIGIR2011 Workshop on Enriching
  Information Retrieval (ENIR2011)}.}
\PrintBackRefs{\CurrentBib}

\bibitem [\protect \citeauthoryear {%
Eikmeyer%
\ \BBA {} Rieser%
}{%
Eikmeyer%
\ \BBA {} Rieser%
}{%
{\protect \APACyear {1981}}%
}]{%
Eikmeyer+81}
\APACinsertmetastar {%
Eikmeyer+81}%
\begin{APACrefauthors}%
Eikmeyer, H\BPBI J.%
\BCBT {}\ \BBA {} Rieser, H.%
\end{APACrefauthors}%
\unskip\
\newblock
\APACrefYear{1981}.
\newblock
\APACrefbtitle {Words, worlds, and contexts: New approaches in word semantics}
  {Words, worlds, and contexts: New approaches in word semantics}\ (\BVOL~6).
\newblock
\APACaddressPublisher{}{Walter de Gruyter GmbH \& Co KG}.
\PrintBackRefs{\CurrentBib}

\bibitem [\protect \citeauthoryear {%
Ellen%
}{%
Ellen%
}{%
{\protect \APACyear {2011}}%
}]{%
Ellen11}
\APACinsertmetastar {%
Ellen11}%
\begin{APACrefauthors}%
Ellen, J.%
\end{APACrefauthors}%
\unskip\
\newblock
\APACrefYearMonthDay{2011}{}{}.
\newblock
{\BBOQ}\APACrefatitle {ALL ABOUT MICROTEXT - A Working Definition and a Survey
  of Current Microtext Research within Artificial Intelligence and Natural
  Language Processing} {All about microtext - a working definition and a survey
  of current microtext research within artificial intelligence and natural
  language processing}.{\BBCQ}
\newblock
\BIn{} \APACrefbtitle {{Proceedings of the 3rd International Conference on
  Agents and Artificial Intelligence - Volume 1: ICAART,}} {{Proceedings of the
  3rd International Conference on Agents and Artificial Intelligence - Volume
  1: ICAART,}}\ (\BPG~329-336).
\newblock
\begin{APACrefDOI} \doi{10.5220/0003179903290336} \end{APACrefDOI}
\PrintBackRefs{\CurrentBib}

\bibitem [\protect \citeauthoryear {%
{Fahad}%
\ \protect \BOthers {.}}{%
{Fahad}%
\ \protect \BOthers {.}}{%
{\protect \APACyear {2014}}%
}]{%
Fahad+14}
\APACinsertmetastar {%
Fahad+14}%
\begin{APACrefauthors}%
{Fahad}, A.%
, {Alshatri}, N.%
, {Tari}, Z.%
, {Alamri}, A.%
, {Khalil}, I.%
, {Zomaya}, A\BPBI Y.%
\BDBL {}{Bouras}, A.%
\end{APACrefauthors}%
\unskip\
\newblock
\APACrefYearMonthDay{2014}{Sep.}{}.
\newblock
{\BBOQ}\APACrefatitle {A Survey of Clustering Algorithms for Big Data: Taxonomy
  and Empirical Analysis} {A survey of clustering algorithms for big data:
  Taxonomy and empirical analysis}.{\BBCQ}
\newblock
\APACjournalVolNumPages{IEEE Transactions on Emerging Topics in
  Computing}{2}{3}{267-279}.
\newblock
\begin{APACrefDOI} \doi{10.1109/TETC.2014.2330519} \end{APACrefDOI}
\PrintBackRefs{\CurrentBib}

\bibitem [\protect \citeauthoryear {%
Fellbaum%
}{%
Fellbaum%
}{%
{\protect \APACyear {1998}}%
}]{%
Fellbaum98}
\APACinsertmetastar {%
Fellbaum98}%
\begin{APACrefauthors}%
Fellbaum, C.%
\end{APACrefauthors}%
\unskip\
\newblock
\APACrefYear{1998}.
\newblock
\APACrefbtitle {WordNet} {Wordnet}.
\newblock
\APACaddressPublisher{}{Wiley Online Library}.
\PrintBackRefs{\CurrentBib}

\bibitem [\protect \citeauthoryear {%
Felt%
}{%
Felt%
}{%
{\protect \APACyear {2016}}%
}]{%
Felt16}
\APACinsertmetastar {%
Felt16}%
\begin{APACrefauthors}%
Felt, M.%
\end{APACrefauthors}%
\unskip\
\newblock
\APACrefYearMonthDay{2016}{}{}.
\newblock
{\BBOQ}\APACrefatitle {{Social media and the social sciences: How researchers
  employ Big Data analytics}} {{Social media and the social sciences: How
  researchers employ Big Data analytics}}.{\BBCQ}
\newblock
\APACjournalVolNumPages{Big Data {\&} Society}{3}{1}{1--15}.
\newblock
\begin{APACrefURL}
  \url{http://bds.sagepub.com/lookup/doi/10.1177/2053951716645828}
  \end{APACrefURL}
\newblock
\begin{APACrefDOI} \doi{10.1177/2053951716645828} \end{APACrefDOI}
\PrintBackRefs{\CurrentBib}

\bibitem [\protect \citeauthoryear {%
Gali%
, Surana%
, Vaidya%
, Shishtla%
\BCBL {}\ \BBA {} Sharma%
}{%
Gali%
\ \protect \BOthers {.}}{%
{\protect \APACyear {2008}}%
}]{%
Gali+08}
\APACinsertmetastar {%
Gali+08}%
\begin{APACrefauthors}%
Gali, K.%
, Surana, H.%
, Vaidya, A.%
, Shishtla, P.%
\BCBL {}\ \BBA {} Sharma, D\BPBI M.%
\end{APACrefauthors}%
\unskip\
\newblock
\APACrefYearMonthDay{2008}{}{}.
\newblock
{\BBOQ}\APACrefatitle {Aggregating Machine Learning and Rule Based Heuristics
  for Named Entity Recognition.} {Aggregating machine learning and rule based
  heuristics for named entity recognition.}{\BBCQ}
\newblock
\BIn{} \APACrefbtitle {{Proceedings of the IJCNLP-08 Workshop on NER for South
  and South East Asian Languages,Hyderabad, India. Asian Federation of Natural
  Language Processing}} {{Proceedings of the IJCNLP-08 Workshop on NER for
  South and South East Asian Languages,Hyderabad, India. Asian Federation of
  Natural Language Processing}}\ (\BPGS\ 25--32).
\PrintBackRefs{\CurrentBib}

\bibitem [\protect \citeauthoryear {%
Gella%
, Cook%
\BCBL {}\ \BBA {} Baldwin%
}{%
Gella%
\ \protect \BOthers {.}}{%
{\protect \APACyear {2014}}%
}]{%
Gella+14}
\APACinsertmetastar {%
Gella+14}%
\begin{APACrefauthors}%
Gella, S.%
, Cook, P.%
\BCBL {}\ \BBA {} Baldwin, T.%
\end{APACrefauthors}%
\unskip\
\newblock
\APACrefYearMonthDay{2014}{}{}.
\newblock
{\BBOQ}\APACrefatitle {{One Sense per Tweeter... and Other Lexical Semantic
  Tales of Twitter}} {{One Sense per Tweeter... and Other Lexical Semantic
  Tales of Twitter}}.{\BBCQ}
\newblock
\APACjournalVolNumPages{Proceedings of the 14th Conference of the European
  Chapter of the Association for Computational Linguistics}{}{}{215--220}.
\newblock
\begin{APACrefURL} \url{http://www.aclweb.org/anthology/E14-4042}
  \end{APACrefURL}
\PrintBackRefs{\CurrentBib}

\bibitem [\protect \citeauthoryear {%
Gella%
, Cook%
\BCBL {}\ \BBA {} Han%
}{%
Gella%
\ \protect \BOthers {.}}{%
{\protect \APACyear {2013}}%
}]{%
Gella+13}
\APACinsertmetastar {%
Gella+13}%
\begin{APACrefauthors}%
Gella, S.%
, Cook, P.%
\BCBL {}\ \BBA {} Han, B.%
\end{APACrefauthors}%
\unskip\
\newblock
\APACrefYearMonthDay{2013}{}{}.
\newblock
{\BBOQ}\APACrefatitle {{Unsupervised Word Usage Similarity in Social Media
  Texts}} {{Unsupervised Word Usage Similarity in Social Media Texts}}.{\BBCQ}
\newblock
\APACjournalVolNumPages{Second Joint Conference on Lexical and Computational
  Semantics (*SEM), Volume 1: Proceedings of the Main Conference and the Shared
  Task: Semantic Textual Similarity}{1}{}{248--253}.
\newblock
\begin{APACrefURL} \url{http://www.aclweb.org/anthology/S13-1036}
  \end{APACrefURL}
\PrintBackRefs{\CurrentBib}

\bibitem [\protect \citeauthoryear {%
Ghosh%
\ \BBA {} Ghosh%
}{%
Ghosh%
\ \BBA {} Ghosh%
}{%
{\protect \APACyear {2016}}%
}]{%
Ghosh+16}
\APACinsertmetastar {%
Ghosh+16}%
\begin{APACrefauthors}%
Ghosh, S.%
\BCBT {}\ \BBA {} Ghosh, K.%
\end{APACrefauthors}%
\unskip\
\newblock
\APACrefYearMonthDay{2016}{}{}.
\newblock
{\BBOQ}\APACrefatitle {{Overview of the FIRE 2016 Microblog track: Information
  Extraction from Microblogs posted during Disasters}} {{Overview of the FIRE
  2016 Microblog track: Information Extraction from Microblogs posted during
  Disasters}}.{\BBCQ}
\newblock
\APACjournalVolNumPages{Working notes of FIRE}{}{}{7--10}.
\newblock
\begin{APACrefURL} \url{http://ceur-ws.org/Vol-1737/T2-1.pdf} \end{APACrefURL}
\PrintBackRefs{\CurrentBib}

\bibitem [\protect \citeauthoryear {%
Ghosh%
\ \protect \BOthers {.}}{%
Ghosh%
\ \protect \BOthers {.}}{%
{\protect \APACyear {2017}}%
}]{%
Ghosh+17}
\APACinsertmetastar {%
Ghosh+17}%
\begin{APACrefauthors}%
Ghosh, S.%
, Ghosh, K.%
, Ganguly, D.%
, Chakraborty, T.%
, Jones, G\BPBI J.%
\BCBL {}\ \BBA {} Moens, M\BHBI F.%
\end{APACrefauthors}%
\unskip\
\newblock
\APACrefYearMonthDay{2017}{}{}.
\newblock
{\BBOQ}\APACrefatitle {{ECIR 2017 Workshop on Exploitation of Social Media for
  Emergency Relief and Preparedness (SMERP 2017)}} {{ECIR 2017 Workshop on
  Exploitation of Social Media for Emergency Relief and Preparedness (SMERP
  2017)}}.{\BBCQ}
\newblock
\BIn{} \APACrefbtitle {{ACM SIGIR Forum}} {{ACM SIGIR Forum}}\ (\BVOL~51,
  \BPGS\ 36--41).
\newblock
\begin{APACrefURL} \url{http://ceur-ws.org/Vol-1832/} \end{APACrefURL}
\PrintBackRefs{\CurrentBib}

\bibitem [\protect \citeauthoryear {%
Gupta%
, Lamba%
, Kumaraguru%
\BCBL {}\ \BBA {} Joshi%
}{%
Gupta%
\ \protect \BOthers {.}}{%
{\protect \APACyear {2013}}%
}]{%
Gupta+13}
\APACinsertmetastar {%
Gupta+13}%
\begin{APACrefauthors}%
Gupta, A.%
, Lamba, H.%
, Kumaraguru, P.%
\BCBL {}\ \BBA {} Joshi, A.%
\end{APACrefauthors}%
\unskip\
\newblock
\APACrefYearMonthDay{2013}{}{}.
\newblock
{\BBOQ}\APACrefatitle {{Faking Sandy: Characterizing and Identifying Fake
  Images on Twitter During Hurricane Sandy}} {{Faking Sandy: Characterizing and
  Identifying Fake Images on Twitter During Hurricane Sandy}}.{\BBCQ}
\newblock
\BIn{} \APACrefbtitle {Proceedings of the 22Nd International Conference on
  World Wide Web} {Proceedings of the 22nd international conference on world
  wide web}\ (\BPGS\ 729--736).
\newblock
\APACaddressPublisher{{New York, NY, USA}}{ACM}.
\newblock
\begin{APACrefURL} \url{http://doi.acm.org/10.1145/2487788.2488033}
  \end{APACrefURL}
\newblock
\begin{APACrefDOI} \doi{10.1145/2487788.2488033} \end{APACrefDOI}
\PrintBackRefs{\CurrentBib}

\bibitem [\protect \citeauthoryear {%
Henelius%
, Puolam{\"a}ki%
, Bostr{\"o}m%
\BCBL {}\ \BBA {} Papapetrou%
}{%
Henelius%
\ \protect \BOthers {.}}{%
{\protect \APACyear {2016}}%
}]{%
Henelius+16}
\APACinsertmetastar {%
Henelius+16}%
\begin{APACrefauthors}%
Henelius, A.%
, Puolam{\"a}ki, K.%
, Bostr{\"o}m, H.%
\BCBL {}\ \BBA {} Papapetrou, P.%
\end{APACrefauthors}%
\unskip\
\newblock
\APACrefYearMonthDay{2016}{}{}.
\newblock
{\BBOQ}\APACrefatitle {Clustering with Confidence: Finding Clusters with
  Statistical Guarantees} {Clustering with confidence: Finding clusters with
  statistical guarantees}.{\BBCQ}
\newblock
\APACjournalVolNumPages{arXiv preprint arXiv:1612.08714}{}{}{}.
\newblock
\begin{APACrefURL} \url{https://arxiv.org/abs/1612.08714} \end{APACrefURL}
\PrintBackRefs{\CurrentBib}

\bibitem [\protect \citeauthoryear {%
Hmamouche%
, Przymus%
, Alouaoui%
, Casali%
\BCBL {}\ \BBA {} Lakhal%
}{%
Hmamouche%
\ \protect \BOthers {.}}{%
{\protect \APACyear {2019}}%
}]{%
Hmamouche+19}
\APACinsertmetastar {%
Hmamouche+19}%
\begin{APACrefauthors}%
Hmamouche, Y.%
, Przymus, P\BPBI M.%
, Alouaoui, H.%
, Casali, A.%
\BCBL {}\ \BBA {} Lakhal, L.%
\end{APACrefauthors}%
\unskip\
\newblock
\APACrefYearMonthDay{2019}{}{}.
\newblock
{\BBOQ}\APACrefatitle {{Large Multivariate Time Series Forecasting: Survey on
  Methods and Scalability}} {{Large Multivariate Time Series Forecasting:
  Survey on Methods and Scalability}}.{\BBCQ}
\newblock
\BIn{} \APACrefbtitle {Utilizing Big Data Paradigms for Business Intelligence}
  {Utilizing big data paradigms for business intelligence}\ (\BPGS\ 170--197).
\newblock
\APACaddressPublisher{}{IGI Global}.
\PrintBackRefs{\CurrentBib}

\bibitem [\protect \citeauthoryear {%
H{\"u}rriyeto{\u{g}}lu%
, Gudehus%
, Oostdijk%
\BCBL {}\ \BBA {} van~den Bosch%
}{%
H{\"u}rriyeto{\u{g}}lu%
\ \protect \BOthers {.}}{%
{\protect \APACyear {2016}}%
}]{%
Hurriyetoglu+16b}
\APACinsertmetastar {%
Hurriyetoglu+16b}%
\begin{APACrefauthors}%
H{\"u}rriyeto{\u{g}}lu, A.%
, Gudehus, C.%
, Oostdijk, N.%
\BCBL {}\ \BBA {} van~den Bosch, A.%
\end{APACrefauthors}%
\unskip\
\newblock
\APACrefYearMonthDay{2016}{}{}.
\newblock
{\BBOQ}\APACrefatitle {{Relevancer: Finding and Labeling Relevant Information
  in Tweet Collections}} {{Relevancer: Finding and Labeling Relevant
  Information in Tweet Collections}}.{\BBCQ}
\newblock
\BIn{} E.~Spiro\ \BBA {} Y\BHBI Y.~Ahn\ (\BEDS), \APACrefbtitle {{Social
  Informatics: 8th International Conference, SocInfo 2016, Bellevue, WA, USA,
  November 11-14, 2016, Proceedings, Part II}} {{Social Informatics: 8th
  International Conference, SocInfo 2016, Bellevue, WA, USA, November 11-14,
  2016, Proceedings, Part II}}\ (\BPGS\ 210--224).
\newblock
\APACaddressPublisher{Cham}{Springer International Publishing}.
\newblock
\begin{APACrefURL} \url{http://dx.doi.org/10.1007/978-3-319-47874-6_15}
  \end{APACrefURL}
\newblock
\begin{APACrefDOI} \doi{10.1007/978-3-319-47874-6_15} \end{APACrefDOI}
\PrintBackRefs{\CurrentBib}

\bibitem [\protect \citeauthoryear {%
H{\"u}rriyeto{\v{g}}lu%
, Oostdijk%
, Erkan~Ba{\c{s}}ar%
\BCBL {}\ \BBA {} van~den Bosch%
}{%
H{\"u}rriyeto{\v{g}}lu%
\ \protect \BOthers {.}}{%
{\protect \APACyear {2017}}%
}]{%
Hurriyetoglu+17b}
\APACinsertmetastar {%
Hurriyetoglu+17b}%
\begin{APACrefauthors}%
H{\"u}rriyeto{\v{g}}lu, A.%
, Oostdijk, N.%
, Erkan~Ba{\c{s}}ar, M.%
\BCBL {}\ \BBA {} van~den Bosch, A.%
\end{APACrefauthors}%
\unskip\
\newblock
\APACrefYearMonthDay{2017}{}{}.
\newblock
{\BBOQ}\APACrefatitle {{Supporting Experts to Handle Tweet Collections About
  Significant Events}} {{Supporting Experts to Handle Tweet Collections About
  Significant Events}}.{\BBCQ}
\newblock
\BIn{} F.~Frasincar, A.~Ittoo, L\BPBI M.~Nguyen\BCBL {}\ \BBA {} E.~M{\'e}tais\
  (\BEDS), \APACrefbtitle {{Natural Language Processing and Information
  Systems: 22nd International Conference on Applications of Natural Language to
  Information Systems, NLDB 2017, Li{\`e}ge, Belgium, June 21-23, 2017,
  Proceedings}} {{Natural Language Processing and Information Systems: 22nd
  International Conference on Applications of Natural Language to Information
  Systems, NLDB 2017, Li{\`e}ge, Belgium, June 21-23, 2017, Proceedings}}\
  (\BPGS\ 138--141).
\newblock
\APACaddressPublisher{Cham}{Springer International Publishing}.
\newblock
\begin{APACrefURL} \url{https://doi.org/10.1007/978-3-319-59569-6_14}
  \end{APACrefURL}
\newblock
\begin{APACrefDOI} \doi{10.1007/978-3-319-59569-6_14} \end{APACrefDOI}
\PrintBackRefs{\CurrentBib}

\bibitem [\protect \citeauthoryear {%
H{\"u}rriyeto{\u{g}}lu%
, Oostdijk%
\BCBL {}\ \BBA {} van~den Bosch%
}{%
H{\"u}rriyeto{\u{g}}lu%
\ \protect \BOthers {.}}{%
{\protect \APACyear {2018}}%
}]{%
Hurriyetoglu+18}
\APACinsertmetastar {%
Hurriyetoglu+18}%
\begin{APACrefauthors}%
H{\"u}rriyeto{\u{g}}lu, A.%
, Oostdijk, N.%
\BCBL {}\ \BBA {} van~den Bosch, A.%
\end{APACrefauthors}%
\unskip\
\newblock
\APACrefYearMonthDay{2018}{}{}.
\newblock
{\BBOQ}\APACrefatitle {{Estimating Time to Event based on Linguistic Cues on
  Twitter.}} {{Estimating Time to Event based on Linguistic Cues on
  Twitter.}}{\BBCQ}
\newblock
\BIn{} K.~Shaalan, A\BPBI E.~Hassanien\BCBL {}\ \BBA {} F.~Tolba\ (\BEDS),
  \APACrefbtitle {{Intelligent Natural Language Processing: Trends and
  Applications}} {{Intelligent Natural Language Processing: Trends and
  Applications}}\ (\BVOL~740).
\newblock
\APACaddressPublisher{}{Springer International Publishing}.
\newblock
\begin{APACrefURL} \url{http://www.springer.com/cn/book/9783319670553}
  \end{APACrefURL}
\PrintBackRefs{\CurrentBib}

\bibitem [\protect \citeauthoryear {%
H{\"u}rriyeto\u{g}lu%
, Kunneman%
\BCBL {}\ \BBA {} van~den Bosch%
}{%
H{\"u}rriyeto\u{g}lu%
\ \protect \BOthers {.}}{%
{\protect \APACyear {2013}}%
}]{%
Hurriyetoglu+13}
\APACinsertmetastar {%
Hurriyetoglu+13}%
\begin{APACrefauthors}%
H{\"u}rriyeto\u{g}lu, A.%
, Kunneman, F.%
\BCBL {}\ \BBA {} van~den Bosch, A.%
\end{APACrefauthors}%
\unskip\
\newblock
\APACrefYearMonthDay{2013}{}{}.
\newblock
{\BBOQ}\APACrefatitle {{Estimating the Time between Twitter Messages and Future
  Events}} {{Estimating the Time between Twitter Messages and Future
  Events}}.{\BBCQ}
\newblock
\BIn{} \APACrefbtitle {{Proceedings of the 13th Dutch-Belgian Workshop on
  Information Retrieval}} {{Proceedings of the 13th Dutch-Belgian Workshop on
  Information Retrieval}}\ (\BPGS\ 20--23).
\newblock
\begin{APACrefURL} \url{http://ceur-ws.org/Vol-986/paper_23.pdf}
  \end{APACrefURL}
\PrintBackRefs{\CurrentBib}

\bibitem [\protect \citeauthoryear {%
H\"{u}rriyeto\u{g}lu%
\ \BBA {} Oostdijk%
}{%
H\"{u}rriyeto\u{g}lu%
\ \BBA {} Oostdijk%
}{%
{\protect \APACyear {2017}}%
}]{%
Hurriyetoglu+17}
\APACinsertmetastar {%
Hurriyetoglu+17}%
\begin{APACrefauthors}%
H\"{u}rriyeto\u{g}lu, A.%
\BCBT {}\ \BBA {} Oostdijk, N.%
\end{APACrefauthors}%
\unskip\
\newblock
\APACrefYearMonthDay{2017}{April}{}.
\newblock
{\BBOQ}\APACrefatitle {{Extracting Humanitarian Information from Tweets}}
  {{Extracting Humanitarian Information from Tweets}}.{\BBCQ}
\newblock
\BIn{} \APACrefbtitle {{Proceedings of the First International Workshop on
  Exploitation of Social Media for Emergency Relief and Preparedness}.}
  {{Proceedings of the First International Workshop on Exploitation of Social
  Media for Emergency Relief and Preparedness}.}
\newblock
\APACaddressPublisher{Aberdeen, United Kingdom}{}.
\newblock
\begin{APACrefURL}
  \url{http://ceur-ws.org/Vol-1832/SMERP-2017-DC-RU-Retrieval.pdf}
  \end{APACrefURL}
\PrintBackRefs{\CurrentBib}

\bibitem [\protect \citeauthoryear {%
H\"{u}rriyeto\u{g}lu%
, Oostdijk%
\BCBL {}\ \BBA {} van~den Bosch%
}{%
H\"{u}rriyeto\u{g}lu%
\ \protect \BOthers {.}}{%
{\protect \APACyear {2014}}%
}]{%
Hurriyetoglu+14}
\APACinsertmetastar {%
Hurriyetoglu+14}%
\begin{APACrefauthors}%
H\"{u}rriyeto\u{g}lu, A.%
, Oostdijk, N.%
\BCBL {}\ \BBA {} van~den Bosch, A.%
\end{APACrefauthors}%
\unskip\
\newblock
\APACrefYearMonthDay{2014}{April}{}.
\newblock
{\BBOQ}\APACrefatitle {Estimating Time to Event from Tweets Using Temporal
  Expressions} {Estimating time to event from tweets using temporal
  expressions}.{\BBCQ}
\newblock
\BIn{} \APACrefbtitle {{Proceedings of the 5th Workshop on Language Analysis
  for Social Media (LASM)}} {{Proceedings of the 5th Workshop on Language
  Analysis for Social Media (LASM)}}\ (\BPGS\ 8--16).
\newblock
\APACaddressPublisher{Gothenburg, Sweden}{Association for Computational
  Linguistics}.
\newblock
\begin{APACrefURL} \url{http://www.aclweb.org/anthology/W14-1302}
  \end{APACrefURL}
\PrintBackRefs{\CurrentBib}

\bibitem [\protect \citeauthoryear {%
H\"{u}rriyeto\u{g}lu%
, van~den Bosch%
\BCBL {}\ \BBA {} Oostdijk%
}{%
H\"{u}rriyeto\u{g}lu%
\ \protect \BOthers {.}}{%
{\protect \APACyear {2016}}%
{\protect \APACexlab {{\protect \BCnt {2}}}}}]{%
Hurriyetoglu+16c}
\APACinsertmetastar {%
Hurriyetoglu+16c}%
\begin{APACrefauthors}%
H\"{u}rriyeto\u{g}lu, A.%
, van~den Bosch, A.%
\BCBL {}\ \BBA {} Oostdijk, N.%
\end{APACrefauthors}%
\unskip\
\newblock
\APACrefYearMonthDay{2016{\protect \BCnt {2}}}{December}{}.
\newblock
{\BBOQ}\APACrefatitle {{Using Relevancer to Detect Relevant Tweets: The Nepal
  Earthquake Case}} {{Using Relevancer to Detect Relevant Tweets: The Nepal
  Earthquake Case}}.{\BBCQ}
\newblock
\BIn{} \APACrefbtitle {{Working notes of FIRE 2016 - Forum for Information
  Retrieval Evaluation}.} {{Working notes of FIRE 2016 - Forum for Information
  Retrieval Evaluation}.}
\newblock
\APACaddressPublisher{Kolkata, India}{}.
\newblock
\begin{APACrefURL} \url{http://ceur-ws.org/Vol-1737/T2-6.pdf} \end{APACrefURL}
\PrintBackRefs{\CurrentBib}

\bibitem [\protect \citeauthoryear {%
H\"{u}rriyeto\u{g}lu%
, van~den Bosch%
\BCBL {}\ \BBA {} Oostdijk%
}{%
H\"{u}rriyeto\u{g}lu%
\ \protect \BOthers {.}}{%
{\protect \APACyear {2016}}%
{\protect \APACexlab {{\protect \BCnt {1}}}}}]{%
Hurriyetoglu+16a}
\APACinsertmetastar {%
Hurriyetoglu+16a}%
\begin{APACrefauthors}%
H\"{u}rriyeto\u{g}lu, A.%
, van~den Bosch, J\BPBI W\BPBI A.%
\BCBL {}\ \BBA {} Oostdijk, N.%
\end{APACrefauthors}%
\unskip\
\newblock
\APACrefYearMonthDay{2016{\protect \BCnt {1}}}{September}{}.
\newblock
{\BBOQ}\APACrefatitle {{Analysing the Role of Key Term Inflections in Knowledge
  Discovery on Twitter}} {{Analysing the Role of Key Term Inflections in
  Knowledge Discovery on Twitter}}.{\BBCQ}
\newblock
\BIn{} \APACrefbtitle {{Proceedings of the 1st International Workshop on
  Knowledge Discovery on the WEB}.} {{Proceedings of the 1st International
  Workshop on Knowledge Discovery on the WEB}.}
\newblock
\APACaddressPublisher{Cagliari, Italy}{}.
\newblock
\begin{APACrefURL}
  \url{http://www.iascgroup.it/kdweb2016-program/accepted-papers.html}
  \end{APACrefURL}
\PrintBackRefs{\CurrentBib}

\bibitem [\protect \citeauthoryear {%
Imran%
, Castillo%
, Diaz%
\BCBL {}\ \BBA {} Vieweg%
}{%
Imran%
\ \protect \BOthers {.}}{%
{\protect \APACyear {2015}}%
}]{%
Imran+15}
\APACinsertmetastar {%
Imran+15}%
\begin{APACrefauthors}%
Imran, M.%
, Castillo, C.%
, Diaz, F.%
\BCBL {}\ \BBA {} Vieweg, S.%
\end{APACrefauthors}%
\unskip\
\newblock
\APACrefYearMonthDay{2015}{{\APACmonth{06}}}{}.
\newblock
{\BBOQ}\APACrefatitle {Processing Social Media Messages in Mass Emergency: A
  Survey} {Processing social media messages in mass emergency: A
  survey}.{\BBCQ}
\newblock
\APACjournalVolNumPages{ACM Computing Surveys (CSUR)}{47}{4}{67:1--67:38}.
\newblock
\begin{APACrefURL} \url{http://doi.acm.org/10.1145/2771588} \end{APACrefURL}
\newblock
\begin{APACrefDOI} \doi{10.1145/2771588} \end{APACrefDOI}
\PrintBackRefs{\CurrentBib}

\bibitem [\protect \citeauthoryear {%
Jatowt%
\ \BBA {} Au~Yeung%
}{%
Jatowt%
\ \BBA {} Au~Yeung%
}{%
{\protect \APACyear {2011}}%
}]{%
Jatowt+11}
\APACinsertmetastar {%
Jatowt+11}%
\begin{APACrefauthors}%
Jatowt, A.%
\BCBT {}\ \BBA {} Au~Yeung, C\BHBI m.%
\end{APACrefauthors}%
\unskip\
\newblock
\APACrefYearMonthDay{2011}{}{}.
\newblock
{\BBOQ}\APACrefatitle {{Extracting Collective Expectations About the Future
  from Large Text Collections}} {{Extracting Collective Expectations About the
  Future from Large Text Collections}}.{\BBCQ}
\newblock
\BIn{} \APACrefbtitle {{Proceedings of the 20th ACM International Conference on
  Information and Knowledge Management}} {{Proceedings of the 20th ACM
  International Conference on Information and Knowledge Management}}\ (\BPGS\
  1259--1264).
\newblock
\APACaddressPublisher{New York, NY, USA}{ACM}.
\newblock
\begin{APACrefURL} \url{http://doi.acm.org/10.1145/2063576.2063759}
  \end{APACrefURL}
\newblock
\begin{APACrefDOI} \doi{10.1145/2063576.2063759} \end{APACrefDOI}
\PrintBackRefs{\CurrentBib}

\bibitem [\protect \citeauthoryear {%
Jatowt%
, Au~Yeung%
\BCBL {}\ \BBA {} Tanaka%
}{%
Jatowt%
\ \protect \BOthers {.}}{%
{\protect \APACyear {2013}}%
}]{%
Jatowt+13}
\APACinsertmetastar {%
Jatowt+13}%
\begin{APACrefauthors}%
Jatowt, A.%
, Au~Yeung, C\BHBI M.%
\BCBL {}\ \BBA {} Tanaka, K.%
\end{APACrefauthors}%
\unskip\
\newblock
\APACrefYearMonthDay{2013}{}{}.
\newblock
{\BBOQ}\APACrefatitle {Estimating Document Focus Time} {Estimating document
  focus time}.{\BBCQ}
\newblock
\BIn{} \APACrefbtitle {{Proceedings of the 22Nd ACM International Conference on
  Conference on Information \& Knowledge Management}} {{Proceedings of the 22Nd
  ACM International Conference on Conference on Information \& Knowledge
  Management}}\ (\BPGS\ 2273--2278).
\newblock
\APACaddressPublisher{New York, NY, USA}{ACM}.
\newblock
\begin{APACrefURL} \url{http://doi.acm.org/10.1145/2505515.2505655}
  \end{APACrefURL}
\newblock
\begin{APACrefDOI} \doi{10.1145/2505515.2505655} \end{APACrefDOI}
\PrintBackRefs{\CurrentBib}

\bibitem [\protect \citeauthoryear {%
Java%
, Song%
, Finin%
\BCBL {}\ \BBA {} Tseng%
}{%
Java%
\ \protect \BOthers {.}}{%
{\protect \APACyear {2007}}%
}]{%
Java+07}
\APACinsertmetastar {%
Java+07}%
\begin{APACrefauthors}%
Java, A.%
, Song, X.%
, Finin, T.%
\BCBL {}\ \BBA {} Tseng, B.%
\end{APACrefauthors}%
\unskip\
\newblock
\APACrefYearMonthDay{2007}{}{}.
\newblock
{\BBOQ}\APACrefatitle {Why We Twitter: Understanding Microblogging Usage and
  Communities} {Why we twitter: Understanding microblogging usage and
  communities}.{\BBCQ}
\newblock
\BIn{} \APACrefbtitle {{Proceedings of the 9th WebKDD and 1st SNA-KDD 2007
  Workshop on Web Mining and Social Network Analysis}} {{Proceedings of the 9th
  WebKDD and 1st SNA-KDD 2007 Workshop on Web Mining and Social Network
  Analysis}}\ (\BPGS\ 56--65).
\newblock
\APACaddressPublisher{New York, NY, USA}{ACM}.
\newblock
\begin{APACrefURL} \url{http://doi.acm.org/10.1145/1348549.1348556}
  \end{APACrefURL}
\newblock
\begin{APACrefDOI} \doi{10.1145/1348549.1348556} \end{APACrefDOI}
\PrintBackRefs{\CurrentBib}

\bibitem [\protect \citeauthoryear {%
Kallus%
}{%
Kallus%
}{%
{\protect \APACyear {2014}}%
}]{%
Kallus14}
\APACinsertmetastar {%
Kallus14}%
\begin{APACrefauthors}%
Kallus, N.%
\end{APACrefauthors}%
\unskip\
\newblock
\APACrefYearMonthDay{2014}{}{}.
\newblock
{\BBOQ}\APACrefatitle {Predicting Crowd Behavior with Big Public Data}
  {Predicting crowd behavior with big public data}.{\BBCQ}
\newblock
\BIn{} \APACrefbtitle {{Proceedings of the Companion Publication of the 23rd
  International Conference on World Wide Web Companion}} {{Proceedings of the
  Companion Publication of the 23rd International Conference on World Wide Web
  Companion}}\ (\BPGS\ 625--630).
\newblock
\APACaddressPublisher{Republic and Canton of Geneva, Switzerland}{International
  World Wide Web Conferences Steering Committee}.
\newblock
\begin{APACrefURL} \url{http://dx.doi.org/10.1145/2567948.2579233}
  \end{APACrefURL}
\newblock
\begin{APACrefDOI} \doi{10.1145/2567948.2579233} \end{APACrefDOI}
\PrintBackRefs{\CurrentBib}

\bibitem [\protect \citeauthoryear {%
Kanhabua%
, Romano%
\BCBL {}\ \BBA {} Stewart%
}{%
Kanhabua%
\ \protect \BOthers {.}}{%
{\protect \APACyear {2012}}%
}]{%
Kanhabua+12}
\APACinsertmetastar {%
Kanhabua+12}%
\begin{APACrefauthors}%
Kanhabua, N.%
, Romano, S.%
\BCBL {}\ \BBA {} Stewart, A.%
\end{APACrefauthors}%
\unskip\
\newblock
\APACrefYearMonthDay{2012}{}{}.
\newblock
{\BBOQ}\APACrefatitle {Identifying Relevant Temporal Expressions for Real-World
  Events} {Identifying relevant temporal expressions for real-world
  events}.{\BBCQ}
\newblock
\BIn{} \APACrefbtitle {{Proceedings of The SIGIR 2012 Workshop on Time-aware
  Information Access, Portland, OR}.} {{Proceedings of The SIGIR 2012 Workshop
  on Time-aware Information Access, Portland, OR}.}
\PrintBackRefs{\CurrentBib}

\bibitem [\protect \citeauthoryear {%
Kavanaugh%
, Tedesco%
\BCBL {}\ \BBA {} Madondo%
}{%
Kavanaugh%
\ \protect \BOthers {.}}{%
{\protect \APACyear {2014}}%
}]{%
Kavanaugh+14}
\APACinsertmetastar {%
Kavanaugh+14}%
\begin{APACrefauthors}%
Kavanaugh, A.%
, Tedesco, J\BPBI C.%
\BCBL {}\ \BBA {} Madondo, K.%
\end{APACrefauthors}%
\unskip\
\newblock
\APACrefYearMonthDay{2014}{}{}.
\newblock
{\BBOQ}\APACrefatitle {Social Media vs. Traditional Internet Use for Community
  Involvement: Toward Broadening Participation} {Social media vs. traditional
  internet use for community involvement: Toward broadening
  participation}.{\BBCQ}
\newblock
\BIn{} E.~Tambouris, A.~Macintosh\BCBL {}\ \BBA {} F.~Bannister\ (\BEDS),
  \APACrefbtitle {{Electronic Participation: 6th IFIP WG 8.5 International
  Conference, ePart 2014, Dublin, Ireland, September 2-3, 2014. Proceedings}}
  {{Electronic Participation: 6th IFIP WG 8.5 International Conference, ePart
  2014, Dublin, Ireland, September 2-3, 2014. Proceedings}}\ (\BPGS\ 1--12).
\newblock
\APACaddressPublisher{Berlin, Heidelberg}{Springer Berlin Heidelberg}.
\newblock
\begin{APACrefURL} \url{http://dx.doi.org/10.1007/978-3-662-44914-1_1}
  \end{APACrefURL}
\newblock
\begin{APACrefDOI} \doi{10.1007/978-3-662-44914-1_1} \end{APACrefDOI}
\PrintBackRefs{\CurrentBib}

\bibitem [\protect \citeauthoryear {%
Kawai%
, Jatowt%
, Tanaka%
, Kunieda%
\BCBL {}\ \BBA {} Yamada%
}{%
Kawai%
\ \protect \BOthers {.}}{%
{\protect \APACyear {2010}}%
}]{%
Kawai+10}
\APACinsertmetastar {%
Kawai+10}%
\begin{APACrefauthors}%
Kawai, H.%
, Jatowt, A.%
, Tanaka, K.%
, Kunieda, K.%
\BCBL {}\ \BBA {} Yamada, K.%
\end{APACrefauthors}%
\unskip\
\newblock
\APACrefYearMonthDay{2010}{}{}.
\newblock
{\BBOQ}\APACrefatitle {ChronoSeeker: Search Engine for Future and Past Events}
  {Chronoseeker: Search engine for future and past events}.{\BBCQ}
\newblock
\BIn{} \APACrefbtitle {{Proceedings of the 4th International Conference on
  Uniquitous Information Management and Communication}} {{Proceedings of the
  4th International Conference on Uniquitous Information Management and
  Communication}}\ (\BPGS\ 25:1--25:10).
\newblock
\APACaddressPublisher{New York, NY, USA}{ACM}.
\newblock
\begin{APACrefURL} \url{http://doi.acm.org/10.1145/2108616.2108647}
  \end{APACrefURL}
\newblock
\begin{APACrefDOI} \doi{10.1145/2108616.2108647} \end{APACrefDOI}
\PrintBackRefs{\CurrentBib}

\bibitem [\protect \citeauthoryear {%
Khoury%
, Khoury%
\BCBL {}\ \BBA {} Hamou-Lhadj%
}{%
Khoury%
\ \protect \BOthers {.}}{%
{\protect \APACyear {2014}}%
}]{%
Khoury+14}
\APACinsertmetastar {%
Khoury+14}%
\begin{APACrefauthors}%
Khoury, R.%
, Khoury, R.%
\BCBL {}\ \BBA {} Hamou-Lhadj, A.%
\end{APACrefauthors}%
\unskip\
\newblock
\APACrefYearMonthDay{2014}{}{}.
\newblock
{\BBOQ}\APACrefatitle {Microtext Processing} {Microtext processing}.{\BBCQ}
\newblock
\BIn{} R.~Alhajj\ \BBA {} J.~Rokne\ (\BEDS), \APACrefbtitle {{Encyclopedia of
  Social Network Analysis and Mining}} {{Encyclopedia of Social Network
  Analysis and Mining}}\ (\BPGS\ 894--904).
\newblock
\APACaddressPublisher{New York, NY}{Springer New York}.
\newblock
\begin{APACrefURL} \url{http://dx.doi.org/10.1007/978-1-4614-6170-8_353}
  \end{APACrefURL}
\newblock
\begin{APACrefDOI} \doi{10.1007/978-1-4614-6170-8_353} \end{APACrefDOI}
\PrintBackRefs{\CurrentBib}

\bibitem [\protect \citeauthoryear {%
Kleinberg%
}{%
Kleinberg%
}{%
{\protect \APACyear {2006}}%
}]{%
Kleinberg06}
\APACinsertmetastar {%
Kleinberg06}%
\begin{APACrefauthors}%
Kleinberg, J.%
\end{APACrefauthors}%
\unskip\
\newblock
\APACrefYearMonthDay{2006}{}{}.
\newblock
{\BBOQ}\APACrefatitle {Temporal Dynamics of {On-Line} Information Streams}
  {Temporal dynamics of {On-Line} information streams}.{\BBCQ}
\newblock
\BIn{} \APACrefbtitle {{Data Stream Management: Processing High-speed Data}.}
  {{Data Stream Management: Processing High-speed Data}.}
\newblock
\begin{APACrefURL}
  \url{http://citeseerx.ist.psu.edu/viewdoc/summary?doi=10.1.1.62.600}
  \end{APACrefURL}
\PrintBackRefs{\CurrentBib}

\bibitem [\protect \citeauthoryear {%
Krumm%
, Davies%
\BCBL {}\ \BBA {} Narayanaswami%
}{%
Krumm%
\ \protect \BOthers {.}}{%
{\protect \APACyear {2008}}%
}]{%
Krumm+08}
\APACinsertmetastar {%
Krumm+08}%
\begin{APACrefauthors}%
Krumm, J.%
, Davies, N.%
\BCBL {}\ \BBA {} Narayanaswami, C.%
\end{APACrefauthors}%
\unskip\
\newblock
\APACrefYearMonthDay{2008}{Oct}{}.
\newblock
{\BBOQ}\APACrefatitle {{User-Generated Content}} {{User-Generated
  Content}}.{\BBCQ}
\newblock
\APACjournalVolNumPages{IEEE Pervasive Computing}{7}{4}{10-11}.
\newblock
\begin{APACrefDOI} \doi{10.1109/MPRV.2008.85} \end{APACrefDOI}
\PrintBackRefs{\CurrentBib}

\bibitem [\protect \citeauthoryear {%
Kunneman%
\ \BBA {} {Van den Bosch}%
}{%
Kunneman%
\ \BBA {} {Van den Bosch}%
}{%
{\protect \APACyear {2012}}%
}]{%
Kunneman+12}
\APACinsertmetastar {%
Kunneman+12}%
\begin{APACrefauthors}%
Kunneman, F.%
\BCBT {}\ \BBA {} {Van den Bosch}, A.%
\end{APACrefauthors}%
\unskip\
\newblock
\APACrefYearMonthDay{2012}{}{}.
\newblock
{\BBOQ}\APACrefatitle {Leveraging unscheduled event prediction through mining
  scheduled event tweets} {Leveraging unscheduled event prediction through
  mining scheduled event tweets}.{\BBCQ}
\newblock
\BIn{} N.~Roos, M.~Winands\BCBL {}\ \BBA {} J.~Uiterwijk\ (\BEDS),
  \APACrefbtitle {{Proceedings of the 24th Benelux Conference on Artficial
  Intelligence}} {{Proceedings of the 24th Benelux Conference on Artficial
  Intelligence}}\ (\BPGS\ 147--154).
\newblock
\APACaddressPublisher{Maastricht, The Netherlands}{}.
\PrintBackRefs{\CurrentBib}

\bibitem [\protect \citeauthoryear {%
Kunneman%
\ \BBA {} {Van den Bosch}%
}{%
Kunneman%
\ \BBA {} {Van den Bosch}%
}{%
{\protect \APACyear {2014}}%
}]{%
Kunneman+14b}
\APACinsertmetastar {%
Kunneman+14b}%
\begin{APACrefauthors}%
Kunneman, F.%
\BCBT {}\ \BBA {} {Van den Bosch}, A.%
\end{APACrefauthors}%
\unskip\
\newblock
\APACrefYearMonthDay{2014}{}{}.
\newblock
{\BBOQ}\APACrefatitle {{Event detection in Twitter: A machine-learning approach
  based on term pivoting}} {{Event detection in Twitter: A machine-learning
  approach based on term pivoting}}.{\BBCQ}
\newblock
\BIn{} F.~Grootjen, M.~Otworowska\BCBL {}\ \BBA {} J.~Kwisthout\ (\BEDS),
  \APACrefbtitle {{Proceedings of the 26th Benelux Conference on Artificial
  Intelligence}} {{Proceedings of the 26th Benelux Conference on Artificial
  Intelligence}}\ (\BPGS\ 65--72).
\PrintBackRefs{\CurrentBib}

\bibitem [\protect \citeauthoryear {%
Kwak%
, Lee%
, Park%
\BCBL {}\ \BBA {} Moon%
}{%
Kwak%
\ \protect \BOthers {.}}{%
{\protect \APACyear {2010}}%
}]{%
Kwak+10}
\APACinsertmetastar {%
Kwak+10}%
\begin{APACrefauthors}%
Kwak, H.%
, Lee, C.%
, Park, H.%
\BCBL {}\ \BBA {} Moon, S.%
\end{APACrefauthors}%
\unskip\
\newblock
\APACrefYearMonthDay{2010}{}{}.
\newblock
{\BBOQ}\APACrefatitle {What is Twitter, a Social Network or a News Media?}
  {What is twitter, a social network or a news media?}{\BBCQ}
\newblock
\BIn{} \APACrefbtitle {{Proceedings of the 19th International Conference on
  World Wide Web}} {{Proceedings of the 19th International Conference on World
  Wide Web}}\ (\BPGS\ 591--600).
\newblock
\APACaddressPublisher{New York, NY, USA}{ACM}.
\newblock
\begin{APACrefURL} \url{http://doi.acm.org/10.1145/1772690.1772751}
  \end{APACrefURL}
\newblock
\begin{APACrefDOI} \doi{10.1145/1772690.1772751} \end{APACrefDOI}
\PrintBackRefs{\CurrentBib}

\bibitem [\protect \citeauthoryear {%
Lau%
\ \protect \BOthers {.}}{%
Lau%
\ \protect \BOthers {.}}{%
{\protect \APACyear {2012}}%
}]{%
Lau+12}
\APACinsertmetastar {%
Lau+12}%
\begin{APACrefauthors}%
Lau, J\BPBI H.%
, Cook, P.%
, McCarthy, D.%
, Newman, D.%
, Baldwin, T.%
\BCBL {}\ \BBA {} Computing, L.%
\end{APACrefauthors}%
\unskip\
\newblock
\APACrefYearMonthDay{2012}{}{}.
\newblock
{\BBOQ}\APACrefatitle {{Word sense induction for novel sense detection}} {{Word
  sense induction for novel sense detection}}.{\BBCQ}
\newblock
\APACjournalVolNumPages{Proceedings of the 13th Conference of the European
  Chapter of the Association for computational Linguistics (EACL
  2012)}{}{}{591--601}.
\PrintBackRefs{\CurrentBib}

\bibitem [\protect \citeauthoryear {%
Lee%
, Surdeanu%
, Maccartney%
\BCBL {}\ \BBA {} Jurafsky%
}{%
Lee%
\ \protect \BOthers {.}}{%
{\protect \APACyear {2014}}%
}]{%
Lee+14}
\APACinsertmetastar {%
Lee+14}%
\begin{APACrefauthors}%
Lee, H.%
, Surdeanu, M.%
, Maccartney, B.%
\BCBL {}\ \BBA {} Jurafsky, D.%
\end{APACrefauthors}%
\unskip\
\newblock
\APACrefYearMonthDay{2014}{may}{}.
\newblock
{\BBOQ}\APACrefatitle {{On the Importance of Text Analysis for Stock Price
  Prediction}} {{On the Importance of Text Analysis for Stock Price
  Prediction}}.{\BBCQ}
\newblock
\BIn{} N\BPBI C\BPBI C.~Chair)\ \BOthers {.}\ (\BEDS), \APACrefbtitle
  {{Proceedings of the Ninth International Conference on Language Resources and
  Evaluation (LREC'14)}.} {{Proceedings of the Ninth International Conference
  on Language Resources and Evaluation (LREC'14)}.}
\newblock
\APACaddressPublisher{Reykjavik, Iceland}{European Language Resources
  Association (ELRA)}.
\newblock
\begin{APACrefURL}
  \url{http://www.lrec-conf.org/proceedings/lrec2014/summaries/1065.html}
  \end{APACrefURL}
\PrintBackRefs{\CurrentBib}

\bibitem [\protect \citeauthoryear {%
{Lee}%
, {Lee}%
\BCBL {}\ \BBA {} {Lee}%
}{%
{Lee}%
\ \protect \BOthers {.}}{%
{\protect \APACyear {2012}}%
}]{%
Lee+12}
\APACinsertmetastar {%
Lee+12}%
\begin{APACrefauthors}%
{Lee}, K\BPBI M.%
, {Lee}, K\BPBI M.%
\BCBL {}\ \BBA {} {Lee}, C\BPBI H.%
\end{APACrefauthors}%
\unskip\
\newblock
\APACrefYearMonthDay{2012}{Nov}{}.
\newblock
{\BBOQ}\APACrefatitle {Statistical cluster validity indexes to consider
  cohesion and separation} {Statistical cluster validity indexes to consider
  cohesion and separation}.{\BBCQ}
\newblock
\BIn{} \APACrefbtitle {2012 International conference on Fuzzy Theory and Its
  Applications (iFUZZY2012)} {2012 international conference on fuzzy theory and
  its applications (ifuzzy2012)}\ (\BPG~228-232).
\newblock
\begin{APACrefDOI} \doi{10.1109/iFUZZY.2012.6409706} \end{APACrefDOI}
\PrintBackRefs{\CurrentBib}

\bibitem [\protect \citeauthoryear {%
Lkhagvasuren%
, Gon{\c{c}}alves%
\BCBL {}\ \BBA {} Saias%
}{%
Lkhagvasuren%
\ \protect \BOthers {.}}{%
{\protect \APACyear {2016}}%
}]{%
lkhagvasuren+16}
\APACinsertmetastar {%
lkhagvasuren+16}%
\begin{APACrefauthors}%
Lkhagvasuren, G.%
, Gon{\c{c}}alves, T.%
\BCBL {}\ \BBA {} Saias, J.%
\end{APACrefauthors}%
\unskip\
\newblock
\APACrefYearMonthDay{2016}{}{}.
\newblock
{\BBOQ}\APACrefatitle {{Semi-automatic keyword based approach for FIRE 2016
  Microblog Track}} {{Semi-automatic keyword based approach for FIRE 2016
  Microblog Track}}.{\BBCQ}
\newblock
\BIn{} \APACrefbtitle {{FIRE (Working Notes)}} {{FIRE (Working Notes)}}\
  (\BPGS\ 91--93).
\newblock
\begin{APACrefURL} \url{http://ceur-ws.org/Vol-1737/T2-11.pdf} \end{APACrefURL}
\PrintBackRefs{\CurrentBib}

\bibitem [\protect \citeauthoryear {%
Mani%
\ \BBA {} Wilson%
}{%
Mani%
\ \BBA {} Wilson%
}{%
{\protect \APACyear {2000}}%
}]{%
Mani+00}
\APACinsertmetastar {%
Mani+00}%
\begin{APACrefauthors}%
Mani, I.%
\BCBT {}\ \BBA {} Wilson, G.%
\end{APACrefauthors}%
\unskip\
\newblock
\APACrefYearMonthDay{2000}{}{}.
\newblock
{\BBOQ}\APACrefatitle {{Robust Temporal Processing of News}} {{Robust Temporal
  Processing of News}}.{\BBCQ}
\newblock
\BIn{} \APACrefbtitle {{Proceedings of the 38th Annual Meeting on Association
  for Computational Linguistics}} {{Proceedings of the 38th Annual Meeting on
  Association for Computational Linguistics}}\ (\BPGS\ 69--76).
\newblock
\APACaddressPublisher{Stroudsburg, PA, USA}{Association for Computational
  Linguistics}.
\newblock
\begin{APACrefURL} \url{http://dx.doi.org/10.3115/1075218.1075228}
  \end{APACrefURL}
\newblock
\begin{APACrefDOI} \doi{10.3115/1075218.1075228} \end{APACrefDOI}
\PrintBackRefs{\CurrentBib}

\bibitem [\protect \citeauthoryear {%
Mccarthy%
, Apidianaki%
\BCBL {}\ \BBA {} Erk%
}{%
Mccarthy%
\ \protect \BOthers {.}}{%
{\protect \APACyear {2016}}%
}]{%
Mccarthy+16}
\APACinsertmetastar {%
Mccarthy+16}%
\begin{APACrefauthors}%
Mccarthy, D.%
, Apidianaki, M.%
\BCBL {}\ \BBA {} Erk, K.%
\end{APACrefauthors}%
\unskip\
\newblock
\APACrefYearMonthDay{2016}{}{}.
\newblock
{\BBOQ}\APACrefatitle {{Word Sense Clustering and Clusterability}} {{Word Sense
  Clustering and Clusterability}}.{\BBCQ}
\newblock
\APACjournalVolNumPages{Computational Linguistics}{42}{2}{4943}.
\newblock
\begin{APACrefDOI} \doi{10.1162/COLI} \end{APACrefDOI}
\PrintBackRefs{\CurrentBib}

\bibitem [\protect \citeauthoryear {%
Melville%
, Gryc%
\BCBL {}\ \BBA {} Lawrence%
}{%
Melville%
\ \protect \BOthers {.}}{%
{\protect \APACyear {2009}}%
}]{%
Melville+09}
\APACinsertmetastar {%
Melville+09}%
\begin{APACrefauthors}%
Melville, P.%
, Gryc, W.%
\BCBL {}\ \BBA {} Lawrence, R\BPBI D.%
\end{APACrefauthors}%
\unskip\
\newblock
\APACrefYearMonthDay{2009}{}{}.
\newblock
{\BBOQ}\APACrefatitle {{Sentiment Analysis of Blogs by Combining Lexical
  Knowledge with Text Classification}} {{Sentiment Analysis of Blogs by
  Combining Lexical Knowledge with Text Classification}}.{\BBCQ}
\newblock
\BIn{} \APACrefbtitle {{Proceedings of the 15th ACM SIGKDD International
  Conference on Knowledge Discovery and Data Mining}} {{Proceedings of the 15th
  ACM SIGKDD International Conference on Knowledge Discovery and Data Mining}}\
  (\BPGS\ 1275--1284).
\newblock
\APACaddressPublisher{New York, NY, USA}{ACM}.
\newblock
\begin{APACrefURL} \url{http://doi.acm.org/10.1145/1557019.1557156}
  \end{APACrefURL}
\newblock
\begin{APACrefDOI} \doi{10.1145/1557019.1557156} \end{APACrefDOI}
\PrintBackRefs{\CurrentBib}

\bibitem [\protect \citeauthoryear {%
Mikolov%
, Sutskever%
, Chen%
, Corrado%
\BCBL {}\ \BBA {} Dean%
}{%
Mikolov%
\ \protect \BOthers {.}}{%
{\protect \APACyear {2013}}%
}]{%
Mikolov+13}
\APACinsertmetastar {%
Mikolov+13}%
\begin{APACrefauthors}%
Mikolov, T.%
, Sutskever, I.%
, Chen, K.%
, Corrado, G.%
\BCBL {}\ \BBA {} Dean, J.%
\end{APACrefauthors}%
\unskip\
\newblock
\APACrefYearMonthDay{2013}{}{}.
\newblock
{\BBOQ}\APACrefatitle {{Distributed Representations of Words and Phrases and
  Their Compositionality}} {{Distributed Representations of Words and Phrases
  and Their Compositionality}}.{\BBCQ}
\newblock
\BIn{} \APACrefbtitle {Proceedings of the 26th International Conference on
  Neural Information Processing Systems - Volume 2} {Proceedings of the 26th
  international conference on neural information processing systems - volume
  2}\ (\BPGS\ 3111--3119).
\newblock
\APACaddressPublisher{USA}{Curran Associates Inc.}
\newblock
\begin{APACrefURL} \url{http://dl.acm.org/citation.cfm?id=2999792.2999959}
  \end{APACrefURL}
\PrintBackRefs{\CurrentBib}

\bibitem [\protect \citeauthoryear {%
Morency%
}{%
Morency%
}{%
{\protect \APACyear {2006}}%
}]{%
Morency06}
\APACinsertmetastar {%
Morency06}%
\begin{APACrefauthors}%
Morency, P.%
\end{APACrefauthors}%
\unskip\
\newblock
\APACrefYearMonthDay{2006}{}{}.
\newblock
\APACrefbtitle {When temporal expressions don't tell time: A pragmatic approach
  to temporality, argumentation and subjectivity.} {When temporal expressions
  don't tell time: A pragmatic approach to temporality, argumentation and
  subjectivity.}
\newblock
\begin{APACrefURL}
  \url{https://www2.unine.ch/files/content/sites/cognition/files/shared/documents/patrickmorency-thesisproject.pdf}
  \end{APACrefURL}
\PrintBackRefs{\CurrentBib}

\bibitem [\protect \citeauthoryear {%
Muthiah%
}{%
Muthiah%
}{%
{\protect \APACyear {2014}}%
}]{%
Muthiah14}
\APACinsertmetastar {%
Muthiah14}%
\begin{APACrefauthors}%
Muthiah, S.%
\end{APACrefauthors}%
\unskip\
\newblock
\APACrefYear{2014}.
\unskip\
\newblock
\APACrefbtitle {Forecasting Protests by Detecting Future Time Mentions in News
  and Social Media} {Forecasting protests by detecting future time mentions in
  news and social media}\ \APACtypeAddressSchool {\BMTh}{}{Virginia Polytechnic
  Institute and State University}.
\unskip\
\newblock
\begin{APACrefURL} \url{http://vtechworks.lib.vt.edu/handle/10919/25430}
  \end{APACrefURL}
\PrintBackRefs{\CurrentBib}

\bibitem [\protect \citeauthoryear {%
Nakajima%
, Ptaszynski%
, Honma%
\BCBL {}\ \BBA {} Masui%
}{%
Nakajima%
\ \protect \BOthers {.}}{%
{\protect \APACyear {2014}}%
}]{%
Nakajima+14}
\APACinsertmetastar {%
Nakajima+14}%
\begin{APACrefauthors}%
Nakajima, Y.%
, Ptaszynski, M.%
, Honma, H.%
\BCBL {}\ \BBA {} Masui, F.%
\end{APACrefauthors}%
\unskip\
\newblock
\APACrefYearMonthDay{2014}{}{}.
\newblock
{\BBOQ}\APACrefatitle {Investigation of Future Reference Expressions in Trend
  Information} {Investigation of future reference expressions in trend
  information}.{\BBCQ}
\newblock
\BIn{} \APACrefbtitle {{2014 AAAI Spring Symposium Series}} {{2014 AAAI Spring
  Symposium Series}}\ (\BPGS\ 32--38).
\newblock
\begin{APACrefURL}
  \url{http://www.aaai.org/ocs/index.php/SSS/SSS14/paper/view/7691}
  \end{APACrefURL}
\PrintBackRefs{\CurrentBib}

\bibitem [\protect \citeauthoryear {%
Navigli%
}{%
Navigli%
}{%
{\protect \APACyear {2009}}%
}]{%
Navigli09}
\APACinsertmetastar {%
Navigli09}%
\begin{APACrefauthors}%
Navigli, R.%
\end{APACrefauthors}%
\unskip\
\newblock
\APACrefYearMonthDay{2009}{}{}.
\newblock
{\BBOQ}\APACrefatitle {Word sense disambiguation: A survey} {Word sense
  disambiguation: A survey}.{\BBCQ}
\newblock
\APACjournalVolNumPages{ACM Computing Surveys (CSUR)}{41}{2}{10}.
\PrintBackRefs{\CurrentBib}

\bibitem [\protect \citeauthoryear {%
Nguyen-Son%
\ \protect \BOthers {.}}{%
Nguyen-Son%
\ \protect \BOthers {.}}{%
{\protect \APACyear {2014}}%
}]{%
Nguyen-Son+14}
\APACinsertmetastar {%
Nguyen-Son+14}%
\begin{APACrefauthors}%
Nguyen-Son, H\BHBI Q.%
, Hoang, A\BHBI T.%
, Tran, M\BHBI T.%
, Yoshiura, H.%
, Sonehara, N.%
\BCBL {}\ \BBA {} Echizen, I.%
\end{APACrefauthors}%
\unskip\
\newblock
\APACrefYearMonthDay{2014}{}{}.
\newblock
{\BBOQ}\APACrefatitle {Anonymizing Temporal Phrases in Natural Language Text to
  be Posted on Social Networking Services} {Anonymizing temporal phrases in
  natural language text to be posted on social networking services}.{\BBCQ}
\newblock
\BIn{} Y\BPBI Q.~Shi, H\BHBI J.~Kim\BCBL {}\ \BBA {} F.~Pérez-González\
  (\BEDS), \APACrefbtitle {{Digital-Forensics and Watermarking}}
  {{Digital-Forensics and Watermarking}}\ (\BPG~437-451).
\newblock
\APACaddressPublisher{}{Springer Berlin Heidelberg}.
\newblock
\begin{APACrefURL} \url{http://dx.doi.org/10.1007/978-3-662-43886-2\_31}
  \end{APACrefURL}
\newblock
\begin{APACrefDOI} \doi{10.1007/978-3-662-43886-2\_31} \end{APACrefDOI}
\PrintBackRefs{\CurrentBib}

\bibitem [\protect \citeauthoryear {%
Noce%
, Zamberletti%
, Gallo%
, Piccoli%
\BCBL {}\ \BBA {} Rodriguez%
}{%
Noce%
\ \protect \BOthers {.}}{%
{\protect \APACyear {2014}}%
}]{%
Noce+14}
\APACinsertmetastar {%
Noce+14}%
\begin{APACrefauthors}%
Noce, L.%
, Zamberletti, A.%
, Gallo, I.%
, Piccoli, G.%
\BCBL {}\ \BBA {} Rodriguez, J.%
\end{APACrefauthors}%
\unskip\
\newblock
\APACrefYearMonthDay{2014}{}{}.
\newblock
{\BBOQ}\APACrefatitle {Automatic Prediction of Future Business Conditions}
  {Automatic prediction of future business conditions}.{\BBCQ}
\newblock
\BIn{} A.~Przepiórkowski\ \BBA {} M.~Ogrodniczuk\ (\BEDS), \APACrefbtitle
  {{Advances in Natural Language Processing}} {{Advances in Natural Language
  Processing}}\ (\BVOL\ 8686, \BPG~371-383).
\newblock
\APACaddressPublisher{}{Springer International Publishing}.
\newblock
\begin{APACrefURL} \url{http://dx.doi.org/10.1007/978-3-319-10888-9\_37}
  \end{APACrefURL}
\newblock
\begin{APACrefDOI} \doi{10.1007/978-3-319-10888-9\_37} \end{APACrefDOI}
\PrintBackRefs{\CurrentBib}

\bibitem [\protect \citeauthoryear {%
Noro%
, Inui%
, Takamura%
\BCBL {}\ \BBA {} Okumura%
}{%
Noro%
\ \protect \BOthers {.}}{%
{\protect \APACyear {2006}}%
}]{%
Noro+06}
\APACinsertmetastar {%
Noro+06}%
\begin{APACrefauthors}%
Noro, T.%
, Inui, T.%
, Takamura, H.%
\BCBL {}\ \BBA {} Okumura, M.%
\end{APACrefauthors}%
\unskip\
\newblock
\APACrefYearMonthDay{2006}{}{}.
\newblock
{\BBOQ}\APACrefatitle {Time period identification of events in text} {Time
  period identification of events in text}.{\BBCQ}
\newblock
\BIn{} \APACrefbtitle {{Proceedings of the 21st International Conference on
  Computational Linguistics and the 44th annual meeting of the Association for
  Computational Linguistics}} {{Proceedings of the 21st International
  Conference on Computational Linguistics and the 44th annual meeting of the
  Association for Computational Linguistics}}\ (\BPGS\ 1153--1160).
\newblock
\APACaddressPublisher{Stroudsburg, PA, USA}{Association for Computational
  Linguistics}.
\newblock
\begin{APACrefURL} \url{http://dx.doi.org/10.3115/1220175.1220320}
  \end{APACrefURL}
\newblock
\begin{APACrefDOI} \doi{10.3115/1220175.1220320} \end{APACrefDOI}
\PrintBackRefs{\CurrentBib}

\bibitem [\protect \citeauthoryear {%
Olteanu%
, Castillo%
, Diaz%
\BCBL {}\ \BBA {} Kiciman%
}{%
Olteanu%
\ \protect \BOthers {.}}{%
{\protect \APACyear {2016}}%
}]{%
Olteanu+16}
\APACinsertmetastar {%
Olteanu+16}%
\begin{APACrefauthors}%
Olteanu, A.%
, Castillo, C.%
, Diaz, F.%
\BCBL {}\ \BBA {} Kiciman, E.%
\end{APACrefauthors}%
\unskip\
\newblock
\APACrefYearMonthDay{2016}{}{}.
\newblock
{\BBOQ}\APACrefatitle {Social Data: Biases, Methodological Pitfalls, and
  Ethical Boundaries} {Social data: Biases, methodological pitfalls, and
  ethical boundaries}.{\BBCQ}
\newblock
\APACjournalVolNumPages{SSRN}{}{}{}.
\newblock
\begin{APACrefURL} \url{https://ssrn.com/abstract=2886526} \end{APACrefURL}
\PrintBackRefs{\CurrentBib}

\bibitem [\protect \citeauthoryear {%
Olteanu%
, Castillo%
, Diaz%
\BCBL {}\ \BBA {} Vieweg%
}{%
Olteanu%
\ \protect \BOthers {.}}{%
{\protect \APACyear {2014}}%
}]{%
Olteanu+14}
\APACinsertmetastar {%
Olteanu+14}%
\begin{APACrefauthors}%
Olteanu, A.%
, Castillo, C.%
, Diaz, F.%
\BCBL {}\ \BBA {} Vieweg, S.%
\end{APACrefauthors}%
\unskip\
\newblock
\APACrefYearMonthDay{2014}{1}{}.
\newblock
{\BBOQ}\APACrefatitle {CrisisLex: A lexicon for collecting and filtering
  Microblogged communications in crises} {Crisislex: A lexicon for collecting
  and filtering microblogged communications in crises}.{\BBCQ}
\newblock
\BIn{} \APACrefbtitle {{Proceedings of the 8th International Conference on
  Weblogs and Social Media, ICWSM 2014}} {{Proceedings of the 8th International
  Conference on Weblogs and Social Media, ICWSM 2014}}\ (\BPGS\ 376--385).
\newblock
\APACaddressPublisher{}{The AAAI Press}.
\PrintBackRefs{\CurrentBib}

\bibitem [\protect \citeauthoryear {%
Oostdijk%
, H{\"u}rriyeto\u{g}lu%
, Puts%
, Daas%
\BCBL {}\ \BBA {} van~den Bosch%
}{%
Oostdijk%
\ \protect \BOthers {.}}{%
{\protect \APACyear {2016}}%
}]{%
Oostdijk+16}
\APACinsertmetastar {%
Oostdijk+16}%
\begin{APACrefauthors}%
Oostdijk, N.%
, H{\"u}rriyeto\u{g}lu, A.%
, Puts, M.%
, Daas, P.%
\BCBL {}\ \BBA {} van~den Bosch, A.%
\end{APACrefauthors}%
\unskip\
\newblock
\APACrefYearMonthDay{2016}{}{}.
\newblock
{\BBOQ}\APACrefatitle {{Information extraction from social media: A
  linguistically motivated approach}} {{Information extraction from social
  media: A linguistically motivated approach}}.{\BBCQ}
\newblock
\APACjournalVolNumPages{PARIS Inalco du 4 au 8 juillet 2016}{10}{}{21--33}.
\PrintBackRefs{\CurrentBib}

\bibitem [\protect \citeauthoryear {%
Oostdijk%
\ \BBA {} van Halteren%
}{%
Oostdijk%
\ \BBA {} van Halteren%
}{%
{\protect \APACyear {2013}}%
{\protect \APACexlab {{\protect \BCnt {1}}}}}]{%
Oostdijk+13a}
\APACinsertmetastar {%
Oostdijk+13a}%
\begin{APACrefauthors}%
Oostdijk, N.%
\BCBT {}\ \BBA {} van Halteren, H.%
\end{APACrefauthors}%
\unskip\
\newblock
\APACrefYearMonthDay{2013{\protect \BCnt {1}}}{}{}.
\newblock
{\BBOQ}\APACrefatitle {N-gram-based recognition of threatening tweets}
  {N-gram-based recognition of threatening tweets}.{\BBCQ}
\newblock
\BIn{} A.~Gelbukh\ (\BED), \APACrefbtitle {{CICLing 2013, Part II, LNCS7817}}
  {{CICLing 2013, Part II, LNCS7817}}\ (\BPGS\ 183--196).
\newblock
\APACaddressPublisher{}{Springer Verlag, Berlin -- Heidelberg}.
\PrintBackRefs{\CurrentBib}

\bibitem [\protect \citeauthoryear {%
Oostdijk%
\ \BBA {} van Halteren%
}{%
Oostdijk%
\ \BBA {} van Halteren%
}{%
{\protect \APACyear {2013}}%
{\protect \APACexlab {{\protect \BCnt {2}}}}}]{%
Oostdijk+13b}
\APACinsertmetastar {%
Oostdijk+13b}%
\begin{APACrefauthors}%
Oostdijk, N.%
\BCBT {}\ \BBA {} van Halteren, H.%
\end{APACrefauthors}%
\unskip\
\newblock
\APACrefYearMonthDay{2013{\protect \BCnt {2}}}{}{}.
\newblock
{\BBOQ}\APACrefatitle {Shallow parsing for recognizing threats in Dutch tweets}
  {Shallow parsing for recognizing threats in dutch tweets}.{\BBCQ}
\newblock
\BIn{} \APACrefbtitle {{Proceedings of the 2013 IEEE/ACM International
  Conference on Advances in Social Networks Analysis and Mining (ASONAM 2013,
  Niagara Falls, Canada, August 25-28, 2013)}} {{Proceedings of the 2013
  IEEE/ACM International Conference on Advances in Social Networks Analysis and
  Mining (ASONAM 2013, Niagara Falls, Canada, August 25-28, 2013)}}\ (\BPGS\
  1034--1041).
\PrintBackRefs{\CurrentBib}

\bibitem [\protect \citeauthoryear {%
Ozdikis%
, Senkul%
\BCBL {}\ \BBA {} Oguztuzun%
}{%
Ozdikis%
\ \protect \BOthers {.}}{%
{\protect \APACyear {2012}}%
}]{%
Ozdikis+12}
\APACinsertmetastar {%
Ozdikis+12}%
\begin{APACrefauthors}%
Ozdikis, O.%
, Senkul, P.%
\BCBL {}\ \BBA {} Oguztuzun, H.%
\end{APACrefauthors}%
\unskip\
\newblock
\APACrefYearMonthDay{2012}{}{}.
\newblock
{\BBOQ}\APACrefatitle {Semantic expansion of hashtags for enhanced event
  detection in twitter} {Semantic expansion of hashtags for enhanced event
  detection in twitter}.{\BBCQ}
\newblock
\BIn{} \APACrefbtitle {{Proceedings of the 1st International Workshop on Online
  Social Systems}.} {{Proceedings of the 1st International Workshop on Online
  Social Systems}.}
\PrintBackRefs{\CurrentBib}

\bibitem [\protect \citeauthoryear {%
Pang%
, Lee%
\BCBL {}\ \BBA {} Vaithyanathan%
}{%
Pang%
\ \protect \BOthers {.}}{%
{\protect \APACyear {2002}}%
}]{%
Pang+2002}
\APACinsertmetastar {%
Pang+2002}%
\begin{APACrefauthors}%
Pang, B.%
, Lee, L.%
\BCBL {}\ \BBA {} Vaithyanathan, S.%
\end{APACrefauthors}%
\unskip\
\newblock
\APACrefYearMonthDay{2002}{}{}.
\newblock
{\BBOQ}\APACrefatitle {Thumbs Up?: Sentiment Classification Using Machine
  Learning Techniques} {Thumbs up?: Sentiment classification using machine
  learning techniques}.{\BBCQ}
\newblock
\BIn{} \APACrefbtitle {{Proceedings of the ACL-02 Conference on Empirical
  Methods in Natural Language Processing - Volume 10}} {{Proceedings of the
  ACL-02 Conference on Empirical Methods in Natural Language Processing -
  Volume 10}}\ (\BPGS\ 79--86).
\newblock
\APACaddressPublisher{Stroudsburg, PA, USA}{Association for Computational
  Linguistics}.
\newblock
\begin{APACrefURL} \url{https://doi.org/10.3115/1118693.1118704}
  \end{APACrefURL}
\newblock
\begin{APACrefDOI} \doi{10.3115/1118693.1118704} \end{APACrefDOI}
\PrintBackRefs{\CurrentBib}

\bibitem [\protect \citeauthoryear {%
Pedersen%
}{%
Pedersen%
}{%
{\protect \APACyear {2006}}%
}]{%
Pedersen+06}
\APACinsertmetastar {%
Pedersen+06}%
\begin{APACrefauthors}%
Pedersen, T.%
\end{APACrefauthors}%
\unskip\
\newblock
\APACrefYearMonthDay{2006}{}{}.
\newblock
{\BBOQ}\APACrefatitle {Unsupervised corpus-based methods for WSD} {Unsupervised
  corpus-based methods for wsd}.{\BBCQ}
\newblock
\APACjournalVolNumPages{Word sense disambiguation}{}{}{133--166}.
\PrintBackRefs{\CurrentBib}

\bibitem [\protect \citeauthoryear {%
Pedregosa%
\ \protect \BOthers {.}}{%
Pedregosa%
\ \protect \BOthers {.}}{%
{\protect \APACyear {2011}}%
}]{%
Pedregosa+11}
\APACinsertmetastar {%
Pedregosa+11}%
\begin{APACrefauthors}%
Pedregosa, F.%
, Varoquaux, G.%
, Gramfort, A.%
, Michel, V.%
, Thirion, B.%
, Grisel, O.%
\BDBL {}Duchesnay, E.%
\end{APACrefauthors}%
\unskip\
\newblock
\APACrefYearMonthDay{2011}{}{}.
\newblock
{\BBOQ}\APACrefatitle {{Scikit-learn: Machine Learning in {P}ython}}
  {{Scikit-learn: Machine Learning in {P}ython}}.{\BBCQ}
\newblock
\APACjournalVolNumPages{Journal of Machine Learning
  Research}{12}{}{2825--2830}.
\PrintBackRefs{\CurrentBib}

\bibitem [\protect \citeauthoryear {%
Potts%
, Seitzinger%
, Jones%
\BCBL {}\ \BBA {} Harrison%
}{%
Potts%
\ \protect \BOthers {.}}{%
{\protect \APACyear {2011}}%
}]{%
Potts+2011}
\APACinsertmetastar {%
Potts+2011}%
\begin{APACrefauthors}%
Potts, L.%
, Seitzinger, J.%
, Jones, D.%
\BCBL {}\ \BBA {} Harrison, A.%
\end{APACrefauthors}%
\unskip\
\newblock
\APACrefYearMonthDay{2011}{}{}.
\newblock
{\BBOQ}\APACrefatitle {{Tweeting Disaster: Hashtag Constructions and
  Collisions}} {{Tweeting Disaster: Hashtag Constructions and
  Collisions}}.{\BBCQ}
\newblock
\BIn{} \APACrefbtitle {{Proceedings of the 29th ACM International Conference on
  Design of Communication}} {{Proceedings of the 29th ACM International
  Conference on Design of Communication}}\ (\BPGS\ 235--240).
\newblock
\APACaddressPublisher{New York, NY, USA}{ACM}.
\newblock
\begin{APACrefURL} \url{http://doi.acm.org/10.1145/2038476.2038522}
  \end{APACrefURL}
\newblock
\begin{APACrefDOI} \doi{10.1145/2038476.2038522} \end{APACrefDOI}
\PrintBackRefs{\CurrentBib}

\bibitem [\protect \citeauthoryear {%
Praveen%
\ \BBA {} Ravi~Kiran%
}{%
Praveen%
\ \BBA {} Ravi~Kiran%
}{%
{\protect \APACyear {2008}}%
}]{%
Praveen+08}
\APACinsertmetastar {%
Praveen+08}%
\begin{APACrefauthors}%
Praveen, K\BPBI P.%
\BCBT {}\ \BBA {} Ravi~Kiran, V.%
\end{APACrefauthors}%
\unskip\
\newblock
\APACrefYearMonthDay{2008}{}{}.
\newblock
{\BBOQ}\APACrefatitle {{A Hybrid Named Entity Recognition System for South
  Asian Languages}} {{A Hybrid Named Entity Recognition System for South Asian
  Languages}}.{\BBCQ}
\newblock
\BIn{} \APACrefbtitle {{Proceedings of the IJCNLP-08 Workshop on NER for South
  and South East Asian Languages,Hyderabad, India. Asian Federation of Natural
  Language Processing}} {{Proceedings of the IJCNLP-08 Workshop on NER for
  South and South East Asian Languages,Hyderabad, India. Asian Federation of
  Natural Language Processing}}\ (\BPGS\ 83--88).
\newblock
\begin{APACrefURL} \url{https://www.aclweb.org/anthology/I08-5012}
  \end{APACrefURL}
\PrintBackRefs{\CurrentBib}

\bibitem [\protect \citeauthoryear {%
Radinsky%
, Davidovich%
\BCBL {}\ \BBA {} Markovitch%
}{%
Radinsky%
\ \protect \BOthers {.}}{%
{\protect \APACyear {2012}}%
}]{%
Radinsky+12}
\APACinsertmetastar {%
Radinsky+12}%
\begin{APACrefauthors}%
Radinsky, K.%
, Davidovich, S.%
\BCBL {}\ \BBA {} Markovitch, S.%
\end{APACrefauthors}%
\unskip\
\newblock
\APACrefYearMonthDay{2012}{}{}.
\newblock
{\BBOQ}\APACrefatitle {Learning causality for news events prediction} {Learning
  causality for news events prediction}.{\BBCQ}
\newblock
\BIn{} \APACrefbtitle {{Proceedings of the 21st international conference on
  World Wide Web}} {{Proceedings of the 21st international conference on World
  Wide Web}}\ (\BPGS\ 909--918).
\newblock
\APACaddressPublisher{New York, NY, USA}{ACM}.
\newblock
\begin{APACrefURL} \url{http://dx.doi.org/10.1145/2187836.2187958}
  \end{APACrefURL}
\newblock
\begin{APACrefDOI} \doi{10.1145/2187836.2187958} \end{APACrefDOI}
\PrintBackRefs{\CurrentBib}

\bibitem [\protect \citeauthoryear {%
Ramakrishnan%
\ \protect \BOthers {.}}{%
Ramakrishnan%
\ \protect \BOthers {.}}{%
{\protect \APACyear {2014}}%
}]{%
Ramakrishnan+14}
\APACinsertmetastar {%
Ramakrishnan+14}%
\begin{APACrefauthors}%
Ramakrishnan, N.%
, Butler, P.%
, Muthiah, S.%
, Self, N.%
, Khandpur, R.%
, Saraf, P.%
\BDBL {}Mares, D.%
\end{APACrefauthors}%
\unskip\
\newblock
\APACrefYearMonthDay{2014}{}{}.
\newblock
{\BBOQ}\APACrefatitle {{'Beating the news' with EMBERS: Forecasting Civil
  Unrest using Open Source Indicators}} {{'Beating the news' with EMBERS:
  Forecasting Civil Unrest using Open Source Indicators}}.{\BBCQ}
\newblock
\APACjournalVolNumPages{CoRR}{abs/1402.7035}{}{}.
\PrintBackRefs{\CurrentBib}

\bibitem [\protect \citeauthoryear {%
Ratner%
\ \protect \BOthers {.}}{%
Ratner%
\ \protect \BOthers {.}}{%
{\protect \APACyear {2018}}%
}]{%
Ratner+18}
\APACinsertmetastar {%
Ratner+18}%
\begin{APACrefauthors}%
Ratner, A.%
, Bach, S\BPBI H.%
, Ehrenberg, H.%
, Fries, J.%
, Wu, S.%
\BCBL {}\ \BBA {} R{\'e}, C.%
\end{APACrefauthors}%
\unskip\
\newblock
\APACrefYearMonthDay{2018}{}{}.
\newblock
{\BBOQ}\APACrefatitle {Snorkel: Rapid training data creation with weak
  supervision} {Snorkel: Rapid training data creation with weak
  supervision}.{\BBCQ}
\newblock
\APACjournalVolNumPages{Proceedings of the VLDB Endowment}{}{}{}.
\newblock
\begin{APACrefURL} \url{http://www.vldb.org/pvldb/vol11/p269-ratner.pdf}
  \end{APACrefURL}
\PrintBackRefs{\CurrentBib}

\bibitem [\protect \citeauthoryear {%
Redd%
\ \protect \BOthers {.}}{%
Redd%
\ \protect \BOthers {.}}{%
{\protect \APACyear {2013}}%
}]{%
Redd+13}
\APACinsertmetastar {%
Redd+13}%
\begin{APACrefauthors}%
Redd, A.%
, Carter, M.%
, Divita, G.%
, Shen, S.%
, Palmer, M.%
, Samore, M.%
\BCBL {}\ \BBA {} Gundlapalli, A\BPBI V.%
\end{APACrefauthors}%
\unskip\
\newblock
\APACrefYearMonthDay{2013}{}{}.
\newblock
{\BBOQ}\APACrefatitle {Detecting earlier indicators of homelessness in the free
  text of medical records} {Detecting earlier indicators of homelessness in the
  free text of medical records}.{\BBCQ}
\newblock
\APACjournalVolNumPages{Studies in health technology and
  informatics}{202}{}{153--156}.
\PrintBackRefs{\CurrentBib}

\bibitem [\protect \citeauthoryear {%
Ritter%
, Mausam%
, Etzioni%
\BCBL {}\ \BBA {} Clark%
}{%
Ritter%
\ \protect \BOthers {.}}{%
{\protect \APACyear {2012}}%
}]{%
Ritter+12}
\APACinsertmetastar {%
Ritter+12}%
\begin{APACrefauthors}%
Ritter, A.%
, Mausam%
, Etzioni, O.%
\BCBL {}\ \BBA {} Clark, S.%
\end{APACrefauthors}%
\unskip\
\newblock
\APACrefYearMonthDay{2012}{}{}.
\newblock
{\BBOQ}\APACrefatitle {Open domain event extraction from Twitter} {Open domain
  event extraction from twitter}.{\BBCQ}
\newblock
\BIn{} \APACrefbtitle {{Proceedings of the 18th ACM SIGKDD international
  conference on Knowledge discovery and data mining}} {{Proceedings of the 18th
  ACM SIGKDD international conference on Knowledge discovery and data mining}}\
  (\BPGS\ 1104--1112).
\newblock
\APACaddressPublisher{New York, NY, USA}{ACM}.
\newblock
\begin{APACrefURL} \url{http://dx.doi.org/10.1145/2339530.2339704}
  \end{APACrefURL}
\newblock
\begin{APACrefDOI} \doi{10.1145/2339530.2339704} \end{APACrefDOI}
\PrintBackRefs{\CurrentBib}

\bibitem [\protect \citeauthoryear {%
Saha%
, Chatterji%
, Dandapat%
, Sarkar%
\BCBL {}\ \BBA {} Mitra%
}{%
Saha%
\ \protect \BOthers {.}}{%
{\protect \APACyear {2008}}%
}]{%
Saha+08}
\APACinsertmetastar {%
Saha+08}%
\begin{APACrefauthors}%
Saha, S\BPBI K.%
, Chatterji, S.%
, Dandapat, S.%
, Sarkar, S.%
\BCBL {}\ \BBA {} Mitra, P.%
\end{APACrefauthors}%
\unskip\
\newblock
\APACrefYearMonthDay{2008}{}{}.
\newblock
{\BBOQ}\APACrefatitle {{A hybrid approach for named entity recognition in
  Indian languages}} {{A hybrid approach for named entity recognition in Indian
  languages}}.{\BBCQ}
\newblock
\BIn{} \APACrefbtitle {{Proceedings of the IJCNLP-08 Workshop on NER for South
  and South East Asian Languages,Hyderabad, India. Asian Federation of Natural
  Language Processing}} {{Proceedings of the IJCNLP-08 Workshop on NER for
  South and South East Asian Languages,Hyderabad, India. Asian Federation of
  Natural Language Processing}}\ (\BPGS\ 17--24).
\PrintBackRefs{\CurrentBib}

\bibitem [\protect \citeauthoryear {%
Srihari%
, Niu%
\BCBL {}\ \BBA {} Li%
}{%
Srihari%
\ \protect \BOthers {.}}{%
{\protect \APACyear {2000}}%
}]{%
Srihari+00}
\APACinsertmetastar {%
Srihari+00}%
\begin{APACrefauthors}%
Srihari, R.%
, Niu, C.%
\BCBL {}\ \BBA {} Li, W.%
\end{APACrefauthors}%
\unskip\
\newblock
\APACrefYearMonthDay{2000}{}{}.
\newblock
{\BBOQ}\APACrefatitle {{A Hybrid Approach for Named Entity and Sub-type
  Tagging}} {{A Hybrid Approach for Named Entity and Sub-type Tagging}}.{\BBCQ}
\newblock
\BIn{} \APACrefbtitle {{Proceedings of the Sixth Conference on Applied Natural
  Language Processing}} {{Proceedings of the Sixth Conference on Applied
  Natural Language Processing}}\ (\BPGS\ 247--254).
\newblock
\APACaddressPublisher{Stroudsburg, PA, USA}{Association for Computational
  Linguistics}.
\newblock
\begin{APACrefURL} \url{https://doi.org/10.3115/974147.974181} \end{APACrefURL}
\newblock
\begin{APACrefDOI} \doi{10.3115/974147.974181} \end{APACrefDOI}
\PrintBackRefs{\CurrentBib}

\bibitem [\protect \citeauthoryear {%
Str{\"o}tgen%
\ \BBA {} Gertz%
}{%
Str{\"o}tgen%
\ \BBA {} Gertz%
}{%
{\protect \APACyear {2013}}%
}]{%
Strotgen+13}
\APACinsertmetastar {%
Strotgen+13}%
\begin{APACrefauthors}%
Str{\"o}tgen, J.%
\BCBT {}\ \BBA {} Gertz, M.%
\end{APACrefauthors}%
\unskip\
\newblock
\APACrefYearMonthDay{2013}{Jun}{}.
\newblock
{\BBOQ}\APACrefatitle {{Multilingual and cross-domain temporal tagging}}
  {{Multilingual and cross-domain temporal tagging}}.{\BBCQ}
\newblock
\APACjournalVolNumPages{Language Resources and Evaluation}{47}{2}{269-298}.
\newblock
\begin{APACrefURL} \url{http://dx.doi.org/10.1007/s10579-012-9179-y}
  \end{APACrefURL}
\newblock
\begin{APACrefDOI} \doi{10.1007/s10579-012-9179-y} \end{APACrefDOI}
\PrintBackRefs{\CurrentBib}

\bibitem [\protect \citeauthoryear {%
Sutton%
, Palen%
\BCBL {}\ \BBA {} Shklovski%
}{%
Sutton%
\ \protect \BOthers {.}}{%
{\protect \APACyear {2008}}%
}]{%
Sutton+08}
\APACinsertmetastar {%
Sutton+08}%
\begin{APACrefauthors}%
Sutton, J.%
, Palen, L.%
\BCBL {}\ \BBA {} Shklovski, I.%
\end{APACrefauthors}%
\unskip\
\newblock
\APACrefYearMonthDay{2008}{}{}.
\newblock
{\BBOQ}\APACrefatitle {Backchannels on the front lines: Emergent uses of social
  media in the 2007 southern California wildfires} {Backchannels on the front
  lines: Emergent uses of social media in the 2007 southern california
  wildfires}.{\BBCQ}
\newblock
\BIn{} \APACrefbtitle {{Proceedings of the 5th International ISCRAM
  Conference}} {{Proceedings of the 5th International ISCRAM Conference}}\
  (\BPGS\ 624--632).
\PrintBackRefs{\CurrentBib}

\bibitem [\protect \citeauthoryear {%
Tanguy%
, Tulechki%
, Urieli%
, Hermann%
\BCBL {}\ \BBA {} Raynal%
}{%
Tanguy%
\ \protect \BOthers {.}}{%
{\protect \APACyear {2016}}%
}]{%
Tanguy+16}
\APACinsertmetastar {%
Tanguy+16}%
\begin{APACrefauthors}%
Tanguy, L.%
, Tulechki, N.%
, Urieli, A.%
, Hermann, E.%
\BCBL {}\ \BBA {} Raynal, C.%
\end{APACrefauthors}%
\unskip\
\newblock
\APACrefYearMonthDay{2016}{}{}.
\newblock
{\BBOQ}\APACrefatitle {Natural language processing for aviation safety reports:
  From classification to interactive analysis} {Natural language processing for
  aviation safety reports: From classification to interactive analysis}.{\BBCQ}
\newblock
\APACjournalVolNumPages{Computers in Industry}{78}{}{80 - 95}.
\newblock
\begin{APACrefURL}
  \url{http://www.sciencedirect.com/science/article/pii/S0166361515300464}
  \end{APACrefURL}
\newblock
\APACrefnote{Natural Language Processing and Text Analytics in Industry}
\newblock
\begin{APACrefDOI} \doi{http://dx.doi.org/10.1016/j.compind.2015.09.005}
  \end{APACrefDOI}
\PrintBackRefs{\CurrentBib}

\bibitem [\protect \citeauthoryear {%
Tjong Kim~Sang%
\ \BBA {} van~den Bosch%
}{%
Tjong Kim~Sang%
\ \BBA {} van~den Bosch%
}{%
{\protect \APACyear {2013}}%
}]{%
TjongKimSang+13}
\APACinsertmetastar {%
TjongKimSang+13}%
\begin{APACrefauthors}%
Tjong Kim~Sang, E.%
\BCBT {}\ \BBA {} van~den Bosch, A.%
\end{APACrefauthors}%
\unskip\
\newblock
\APACrefYearMonthDay{2013}{12/2013}{}.
\newblock
{\BBOQ}\APACrefatitle {Dealing with big data: The case of Twitter} {Dealing
  with big data: The case of twitter}.{\BBCQ}
\newblock
\APACjournalVolNumPages{{Computational Linguistics in the Netherlands
  Journal}}{3}{}{121-134}.
\newblock
\begin{APACrefURL}
  \url{http://www.clinjournal.org/sites/clinjournal.org/files/08-TjongKimSang-vandenBosch-CLIN2013.pdf}
  \end{APACrefURL}
\PrintBackRefs{\CurrentBib}

\bibitem [\protect \citeauthoryear {%
Tops%
, van~den Bosch%
\BCBL {}\ \BBA {} Kunneman%
}{%
Tops%
\ \protect \BOthers {.}}{%
{\protect \APACyear {2013}}%
}]{%
Tops+13}
\APACinsertmetastar {%
Tops+13}%
\begin{APACrefauthors}%
Tops, H.%
, van~den Bosch, A.%
\BCBL {}\ \BBA {} Kunneman, F.%
\end{APACrefauthors}%
\unskip\
\newblock
\APACrefYearMonthDay{2013}{}{}.
\newblock
{\BBOQ}\APACrefatitle {Predicting time-to-event from Twitter messages}
  {Predicting time-to-event from twitter messages}.{\BBCQ}
\newblock
\APACjournalVolNumPages{BNAIC 2013 The 24th Benelux Conference on Artificial
  Intelligence}{}{}{207--2014}.
\PrintBackRefs{\CurrentBib}

\bibitem [\protect \citeauthoryear {%
Tufekci%
}{%
Tufekci%
}{%
{\protect \APACyear {2014}}%
}]{%
Tufekci14}
\APACinsertmetastar {%
Tufekci14}%
\begin{APACrefauthors}%
Tufekci, Z.%
\end{APACrefauthors}%
\unskip\
\newblock
\APACrefYearMonthDay{2014}{}{}.
\newblock
{\BBOQ}\APACrefatitle {Big Questions for Social Media Big Data:
  Representativeness, Validity and Other Methodological Pitfalls} {Big
  questions for social media big data: Representativeness, validity and other
  methodological pitfalls}.{\BBCQ}
\newblock
\BIn{} E.~Adar, P.~Resnick, M\BPBI D.~Choudhury, B.~Hogan\BCBL {}\ \BBA {}
  A.~Oh\ (\BEDS), \APACrefbtitle {{Proceedings of the Eighth International
  Conference on Weblogs and Social Media, ICWSM 2014, Ann Arbor, Michigan, USA,
  June 1-4, 2014}.} {{Proceedings of the Eighth International Conference on
  Weblogs and Social Media, ICWSM 2014, Ann Arbor, Michigan, USA, June 1-4,
  2014}.}
\newblock
\APACaddressPublisher{}{The AAAI Press}.
\newblock
\begin{APACrefURL}
  \url{http://www.aaai.org/ocs/index.php/ICWSM/ICWSM14/paper/view/8062}
  \end{APACrefURL}
\PrintBackRefs{\CurrentBib}

\bibitem [\protect \citeauthoryear {%
Vallor%
}{%
Vallor%
}{%
{\protect \APACyear {2016}}%
}]{%
Vallor16}
\APACinsertmetastar {%
Vallor16}%
\begin{APACrefauthors}%
Vallor, S.%
\end{APACrefauthors}%
\unskip\
\newblock
\APACrefYearMonthDay{2016}{}{}.
\newblock
{\BBOQ}\APACrefatitle {{Social Networking and Ethics}} {{Social Networking and
  Ethics}}.{\BBCQ}
\newblock
\BIn{} E\BPBI N.~Zalta\ (\BED), \APACrefbtitle {{The Stanford Encyclopedia of
  Philosophy}} {{The Stanford Encyclopedia of Philosophy}}\
  (\PrintOrdinal{Winter 2016}\ \BEd).
\newblock
\APACaddressPublisher{}{Metaphysics Research Lab, Stanford University}.
\newblock
\APAChowpublished
  {\url{https://plato.stanford.edu/archives/win2016/entries/ethics-social-networking/}}.
\PrintBackRefs{\CurrentBib}

\bibitem [\protect \citeauthoryear {%
van~den Hoven%
, Blaauw%
, Pieters%
\BCBL {}\ \BBA {} Warnier%
}{%
van~den Hoven%
\ \protect \BOthers {.}}{%
{\protect \APACyear {2016}}%
}]{%
vanDenHoven+16}
\APACinsertmetastar {%
vanDenHoven+16}%
\begin{APACrefauthors}%
van~den Hoven, J.%
, Blaauw, M.%
, Pieters, W.%
\BCBL {}\ \BBA {} Warnier, M.%
\end{APACrefauthors}%
\unskip\
\newblock
\APACrefYearMonthDay{2016}{}{}.
\newblock
{\BBOQ}\APACrefatitle {Privacy and Information Technology} {Privacy and
  information technology}.{\BBCQ}
\newblock
\BIn{} E\BPBI N.~Zalta\ (\BED), \APACrefbtitle {{The Stanford Encyclopedia of
  Philosophy}} {{The Stanford Encyclopedia of Philosophy}}\
  (\PrintOrdinal{Spring 2016}\ \BEd).
\newblock
\APACaddressPublisher{}{Metaphysics Research Lab, Stanford University}.
\newblock
\APAChowpublished
  {\url{https://plato.stanford.edu/archives/spr2016/entries/it-privacy/}}.
\PrintBackRefs{\CurrentBib}

\bibitem [\protect \citeauthoryear {%
van Noord%
, Kunneman%
\BCBL {}\ \BBA {} van~den Bosch%
}{%
van Noord%
\ \protect \BOthers {.}}{%
{\protect \APACyear {2017}}%
}]{%
Noord+17}
\APACinsertmetastar {%
Noord+17}%
\begin{APACrefauthors}%
van Noord, R.%
, Kunneman, F\BPBI A.%
\BCBL {}\ \BBA {} van~den Bosch, A.%
\end{APACrefauthors}%
\unskip\
\newblock
\APACrefYearMonthDay{2017}{}{}.
\newblock
{\BBOQ}\APACrefatitle {{Predicting Civil Unrest by Categorizing Dutch Twitter
  Events}} {{Predicting Civil Unrest by Categorizing Dutch Twitter
  Events}}.{\BBCQ}
\newblock
\BIn{} T.~Bosse\ \BBA {} B.~Bredeweg\ (\BEDS), \APACrefbtitle {{BNAIC 2016:
  Artificial Intelligence}} {{BNAIC 2016: Artificial Intelligence}}\ (\BPGS\
  3--16).
\newblock
\APACaddressPublisher{Cham}{Springer International Publishing}.
\PrintBackRefs{\CurrentBib}

\bibitem [\protect \citeauthoryear {%
Vieweg%
, Hughes%
, Starbird%
\BCBL {}\ \BBA {} Palen%
}{%
Vieweg%
\ \protect \BOthers {.}}{%
{\protect \APACyear {2010}}%
}]{%
Vieweg+10}
\APACinsertmetastar {%
Vieweg+10}%
\begin{APACrefauthors}%
Vieweg, S.%
, Hughes, A\BPBI L.%
, Starbird, K.%
\BCBL {}\ \BBA {} Palen, L.%
\end{APACrefauthors}%
\unskip\
\newblock
\APACrefYearMonthDay{2010}{}{}.
\newblock
{\BBOQ}\APACrefatitle {Microblogging During Two Natural Hazards Events: What
  Twitter May Contribute to Situational Awareness} {Microblogging during two
  natural hazards events: What twitter may contribute to situational
  awareness}.{\BBCQ}
\newblock
\BIn{} \APACrefbtitle {{Proceedings of the SIGCHI Conference on Human Factors
  in Computing Systems}} {{Proceedings of the SIGCHI Conference on Human
  Factors in Computing Systems}}\ (\BPGS\ 1079--1088).
\newblock
\APACaddressPublisher{New York, NY, USA}{ACM}.
\newblock
\begin{APACrefURL} \url{http://doi.acm.org/10.1145/1753326.1753486}
  \end{APACrefURL}
\newblock
\begin{APACrefDOI} \doi{10.1145/1753326.1753486} \end{APACrefDOI}
\PrintBackRefs{\CurrentBib}

\bibitem [\protect \citeauthoryear {%
S.~Wang%
\ \BBA {} Manning%
}{%
S.~Wang%
\ \BBA {} Manning%
}{%
{\protect \APACyear {2012}}%
}]{%
WangS+12}
\APACinsertmetastar {%
WangS+12}%
\begin{APACrefauthors}%
Wang, S.%
\BCBT {}\ \BBA {} Manning, C\BPBI D.%
\end{APACrefauthors}%
\unskip\
\newblock
\APACrefYearMonthDay{2012}{}{}.
\newblock
{\BBOQ}\APACrefatitle {{Baselines and Bigrams: Simple, Good Sentiment and Topic
  Classification}} {{Baselines and Bigrams: Simple, Good Sentiment and Topic
  Classification}}.{\BBCQ}
\newblock
\BIn{} \APACrefbtitle {{Proceedings of the 50th Annual Meeting of the
  Association for Computational Linguistics: Short Papers - Volume 2}}
  {{Proceedings of the 50th Annual Meeting of the Association for Computational
  Linguistics: Short Papers - Volume 2}}\ (\BPGS\ 90--94).
\newblock
\APACaddressPublisher{Stroudsburg, PA, USA}{Association for Computational
  Linguistics}.
\newblock
\begin{APACrefURL} \url{http://dl.acm.org/citation.cfm?id=2390665.2390688}
  \end{APACrefURL}
\PrintBackRefs{\CurrentBib}

\bibitem [\protect \citeauthoryear {%
X.~Wang%
, Tokarchuk%
, Cuadrado%
\BCBL {}\ \BBA {} Poslad%
}{%
X.~Wang%
\ \protect \BOthers {.}}{%
{\protect \APACyear {2013}}%
}]{%
Wang+13}
\APACinsertmetastar {%
Wang+13}%
\begin{APACrefauthors}%
Wang, X.%
, Tokarchuk, L.%
, Cuadrado, F.%
\BCBL {}\ \BBA {} Poslad, S.%
\end{APACrefauthors}%
\unskip\
\newblock
\APACrefYearMonthDay{2013}{}{}.
\newblock
{\BBOQ}\APACrefatitle {Exploiting Hashtags for Adaptive Microblog Crawling}
  {Exploiting hashtags for adaptive microblog crawling}.{\BBCQ}
\newblock
\BIn{} \APACrefbtitle {{Proceedings of the 2013 IEEE/ACM International
  Conference on Advances in Social Networks Analysis and Mining}} {{Proceedings
  of the 2013 IEEE/ACM International Conference on Advances in Social Networks
  Analysis and Mining}}\ (\BPGS\ 311--315).
\newblock
\APACaddressPublisher{New York, NY, USA}{ACM}.
\newblock
\begin{APACrefURL} \url{http://doi.acm.org/10.1145/2492517.2492624}
  \end{APACrefURL}
\newblock
\begin{APACrefDOI} \doi{10.1145/2492517.2492624} \end{APACrefDOI}
\PrintBackRefs{\CurrentBib}

\bibitem [\protect \citeauthoryear {%
Weerkamp%
\ \BBA {} De~Rijke%
}{%
Weerkamp%
\ \BBA {} De~Rijke%
}{%
{\protect \APACyear {2012}}%
}]{%
Weerkamp+12}
\APACinsertmetastar {%
Weerkamp+12}%
\begin{APACrefauthors}%
Weerkamp, W.%
\BCBT {}\ \BBA {} De~Rijke, M.%
\end{APACrefauthors}%
\unskip\
\newblock
\APACrefYearMonthDay{2012}{{\APACmonth{08}}}{}.
\newblock
{\BBOQ}\APACrefatitle {Activity Prediction: A Twitter-based Exploration}
  {Activity prediction: A twitter-based exploration}.{\BBCQ}
\newblock
\BIn{} \APACrefbtitle {{Proceedings of the SIGIR 2012 Workshop on Time-aware
  Information Access, TAIA-2012}.} {{Proceedings of the SIGIR 2012 Workshop on
  Time-aware Information Access, TAIA-2012}.}
\PrintBackRefs{\CurrentBib}

\bibitem [\protect \citeauthoryear {%
Wilson%
, Wiebe%
\BCBL {}\ \BBA {} Hoffmann%
}{%
Wilson%
\ \protect \BOthers {.}}{%
{\protect \APACyear {2005}}%
}]{%
Wilson+05}
\APACinsertmetastar {%
Wilson+05}%
\begin{APACrefauthors}%
Wilson, T.%
, Wiebe, J.%
\BCBL {}\ \BBA {} Hoffmann, P.%
\end{APACrefauthors}%
\unskip\
\newblock
\APACrefYearMonthDay{2005}{}{}.
\newblock
{\BBOQ}\APACrefatitle {Recognizing Contextual Polarity in Phrase-level
  Sentiment Analysis} {Recognizing contextual polarity in phrase-level
  sentiment analysis}.{\BBCQ}
\newblock
\BIn{} \APACrefbtitle {{Proceedings of the Conference on Human Language
  Technology and Empirical Methods in Natural Language Processing}}
  {{Proceedings of the Conference on Human Language Technology and Empirical
  Methods in Natural Language Processing}}\ (\BPGS\ 347--354).
\newblock
\APACaddressPublisher{Stroudsburg, PA, USA}{Association for Computational
  Linguistics}.
\newblock
\begin{APACrefURL} \url{https://doi.org/10.3115/1220575.1220619}
  \end{APACrefURL}
\newblock
\begin{APACrefDOI} \doi{10.3115/1220575.1220619} \end{APACrefDOI}
\PrintBackRefs{\CurrentBib}

\bibitem [\protect \citeauthoryear {%
Xu%
, Hong%
, Tsujii%
\BCBL {}\ \BBA {} Chang%
}{%
Xu%
\ \protect \BOthers {.}}{%
{\protect \APACyear {2012}}%
}]{%
Xu+12}
\APACinsertmetastar {%
Xu+12}%
\begin{APACrefauthors}%
Xu, Y.%
, Hong, K.%
, Tsujii, J.%
\BCBL {}\ \BBA {} Chang, E\BPBI I\BHBI C.%
\end{APACrefauthors}%
\unskip\
\newblock
\APACrefYearMonthDay{2012}{}{}.
\newblock
{\BBOQ}\APACrefatitle {Feature engineering combined with machine learning and
  rule-based methods for structured information extraction from narrative
  clinical discharge summaries} {Feature engineering combined with machine
  learning and rule-based methods for structured information extraction from
  narrative clinical discharge summaries}.{\BBCQ}
\newblock
\APACjournalVolNumPages{Journal of the American Medical Informatics
  Association}{19}{5}{824-832}.
\newblock
\begin{APACrefURL} \url{+ http://dx.doi.org/10.1136/amiajnl-2011-000776}
  \end{APACrefURL}
\newblock
\begin{APACrefDOI} \doi{10.1136/amiajnl-2011-000776} \end{APACrefDOI}
\PrintBackRefs{\CurrentBib}

\bibitem [\protect \citeauthoryear {%
Yang%
\ \BBA {} Eisenstein%
}{%
Yang%
\ \BBA {} Eisenstein%
}{%
{\protect \APACyear {2015}}%
}]{%
Yang+15}
\APACinsertmetastar {%
Yang+15}%
\begin{APACrefauthors}%
Yang, Y.%
\BCBT {}\ \BBA {} Eisenstein, J.%
\end{APACrefauthors}%
\unskip\
\newblock
\APACrefYearMonthDay{2015}{}{}.
\newblock
{\BBOQ}\APACrefatitle {{Putting Things in Context: Community-specific Embedding
  Projections for Sentiment Analysis}} {{Putting Things in Context:
  Community-specific Embedding Projections for Sentiment Analysis}}.{\BBCQ}
\newblock
\APACjournalVolNumPages{CoRR}{abs/1511.0}{}{}.
\newblock
\begin{APACrefURL} \url{http://arxiv.org/abs/1511.06052} \end{APACrefURL}
\PrintBackRefs{\CurrentBib}

\bibitem [\protect \citeauthoryear {%
Yin%
, Lampert%
, Cameron%
, Robinson%
\BCBL {}\ \BBA {} Power%
}{%
Yin%
\ \protect \BOthers {.}}{%
{\protect \APACyear {2012}}%
}]{%
Yin+12}
\APACinsertmetastar {%
Yin+12}%
\begin{APACrefauthors}%
Yin, J.%
, Lampert, A.%
, Cameron, M.%
, Robinson, B.%
\BCBL {}\ \BBA {} Power, R.%
\end{APACrefauthors}%
\unskip\
\newblock
\APACrefYearMonthDay{2012}{{\APACmonth{11}}}{}.
\newblock
{\BBOQ}\APACrefatitle {Using Social Media to Enhance Emergency Situation
  Awareness} {Using social media to enhance emergency situation
  awareness}.{\BBCQ}
\newblock
\APACjournalVolNumPages{IEEE Intelligent Systems}{27}{6}{52--59}.
\newblock
\begin{APACrefURL} \url{http://dx.doi.org/10.1109/MIS.2012.6} \end{APACrefURL}
\newblock
\begin{APACrefDOI} \doi{10.1109/MIS.2012.6} \end{APACrefDOI}
\PrintBackRefs{\CurrentBib}

\bibitem [\protect \citeauthoryear {%
Yu%
\ \BBA {} Kak%
}{%
Yu%
\ \BBA {} Kak%
}{%
{\protect \APACyear {2012}}%
}]{%
Yu+12}
\APACinsertmetastar {%
Yu+12}%
\begin{APACrefauthors}%
Yu, S.%
\BCBT {}\ \BBA {} Kak, S.%
\end{APACrefauthors}%
\unskip\
\newblock
\APACrefYearMonthDay{2012}{{\APACmonth{03}}}{7}.
\newblock
{\BBOQ}\APACrefatitle {A Survey of Prediction Using Social Media} {A survey of
  prediction using social media}.{\BBCQ}
\newblock
\APACjournalVolNumPages{CoRR}{abs/1203.1647}{}{}.
\newblock
\begin{APACrefURL} \url{http://arxiv.org/abs/1203.1647} \end{APACrefURL}
\PrintBackRefs{\CurrentBib}

\bibitem [\protect \citeauthoryear {%
Zanzotto%
, Pennacchiotti%
\BCBL {}\ \BBA {} Tsioutsiouliklis%
}{%
Zanzotto%
\ \protect \BOthers {.}}{%
{\protect \APACyear {2011}}%
}]{%
Zanzotto+11}
\APACinsertmetastar {%
Zanzotto+11}%
\begin{APACrefauthors}%
Zanzotto, F\BPBI M.%
, Pennacchiotti, M.%
\BCBL {}\ \BBA {} Tsioutsiouliklis, K.%
\end{APACrefauthors}%
\unskip\
\newblock
\APACrefYearMonthDay{2011}{}{}.
\newblock
{\BBOQ}\APACrefatitle {Linguistic Redundancy in Twitter} {Linguistic redundancy
  in twitter}.{\BBCQ}
\newblock
\BIn{} \APACrefbtitle {{Proceedings of the Conference on Empirical Methods in
  Natural Language Processing}} {{Proceedings of the Conference on Empirical
  Methods in Natural Language Processing}}\ (\BPGS\ 659--669).
\newblock
\APACaddressPublisher{Stroudsburg, PA, USA}{Association for Computational
  Linguistics}.
\newblock
\begin{APACrefURL} \url{http://dl.acm.org/citation.cfm?id=2145432.2145509}
  \end{APACrefURL}
\PrintBackRefs{\CurrentBib}

\bibitem [\protect \citeauthoryear {%
D.~Zhao%
\ \BBA {} Rosson%
}{%
D.~Zhao%
\ \BBA {} Rosson%
}{%
{\protect \APACyear {2009}}%
}]{%
Zhao+09}
\APACinsertmetastar {%
Zhao+09}%
\begin{APACrefauthors}%
Zhao, D.%
\BCBT {}\ \BBA {} Rosson, M\BPBI B.%
\end{APACrefauthors}%
\unskip\
\newblock
\APACrefYearMonthDay{2009}{}{}.
\newblock
{\BBOQ}\APACrefatitle {{How and Why People Twitter: The Role That
  Micro-blogging Plays in Informal Communication at Work}} {{How and Why People
  Twitter: The Role That Micro-blogging Plays in Informal Communication at
  Work}}.{\BBCQ}
\newblock
\BIn{} \APACrefbtitle {{Proceedings of the ACM 2009 International Conference on
  Supporting Group Work}} {{Proceedings of the ACM 2009 International
  Conference on Supporting Group Work}}\ (\BPGS\ 243--252).
\newblock
\APACaddressPublisher{New York, NY, USA}{ACM}.
\newblock
\begin{APACrefURL} \url{http://doi.acm.org/10.1145/1531674.1531710}
  \end{APACrefURL}
\newblock
\begin{APACrefDOI} \doi{10.1145/1531674.1531710} \end{APACrefDOI}
\PrintBackRefs{\CurrentBib}

\bibitem [\protect \citeauthoryear {%
W\BPBI X.~Zhao%
\ \protect \BOthers {.}}{%
W\BPBI X.~Zhao%
\ \protect \BOthers {.}}{%
{\protect \APACyear {2011}}%
}]{%
Zhao+11b}
\APACinsertmetastar {%
Zhao+11b}%
\begin{APACrefauthors}%
Zhao, W\BPBI X.%
, Jiang, J.%
, Weng, J.%
, He, J.%
, Lim, E\BHBI P.%
, Yan, H.%
\BCBL {}\ \BBA {} Li, X.%
\end{APACrefauthors}%
\unskip\
\newblock
\APACrefYearMonthDay{2011}{}{}.
\newblock
{\BBOQ}\APACrefatitle {Comparing Twitter and Traditional Media Using Topic
  Models} {Comparing twitter and traditional media using topic models}.{\BBCQ}
\newblock
\BIn{} P.~Clough\ \BOthers {.}\ (\BEDS), \APACrefbtitle {{Advances in
  Information Retrieval: 33rd European Conference on IR Research, ECIR 2011,
  Dublin, Ireland, April 18-21, 2011. Proceedings}} {{Advances in Information
  Retrieval: 33rd European Conference on IR Research, ECIR 2011, Dublin,
  Ireland, April 18-21, 2011. Proceedings}}\ (\BPGS\ 338--349).
\newblock
\APACaddressPublisher{Berlin, Heidelberg}{Springer Berlin Heidelberg}.
\newblock
\begin{APACrefURL} \url{http://dx.doi.org/10.1007/978-3-642-20161-5_34}
  \end{APACrefURL}
\newblock
\begin{APACrefDOI} \doi{10.1007/978-3-642-20161-5_34} \end{APACrefDOI}
\PrintBackRefs{\CurrentBib}

\end{thebibliography}

\lhead{\emph{samenvatting}}
\chapter{Samenvatting}

Microblogs zoals Twitter vormen een krachtige bron van informatie. Een deel van deze informatie kan worden geaggregeerd buiten het niveau van individuele berichten. Een deel van deze geaggregeerde informatie verwijst naar gebeurtenissen die kunnen of moeten worden aangepakt in het belang van bijvoorbeeld e-governance, openbare veiligheid of andere niveaus van openbaar belang. Bovendien kan een aanzienlijk deel van deze informatie, indien samengevoegd, bestaande informatienetwerken op niet-triviale wijze aanvullen. In dit proefschrift wordt een semi-automatische methode voorgesteld voor het extraheren van bruikbare informatie die dit doel dient.

We rapporteren drie belangrijke bijdragen en een eindconclusie die in een apart hoofdstuk van dit proefschrift worden gepresenteerd. Ten eerste laten we zien dat het voorspellen van de tijd totdat een gebeurtenis plaatsvindt mogelijk is voor zowel binnen-domein als cross-domein scenario's. Ten tweede stellen we een methode voor die de definitie van relevantie in de context van een analist vergemakkelijkt en de analist in staat stelt om de definitie te gebruiken om nieuwe gegevens te analyseren. Ten slotte stellen we een methode voor relevante informatie op basis van machinaal leren te integreren met een regelgebaseerde informatieclassificatietechniek om microteksten (tweets) te classificeren.

In Hoofdstuk 2 doen we verslag van ons onderzoek dat erop gericht is om de informatie over de begintijd van een gebeurtenis op sociale media te karakteriseren en deze automatisch te gebruiken om een methode te ontwikkelen die de tijd tot de gebeurtenis betrouwbaar kan voorspellen. Verschillende algoritmes voor feature-extractie en machinaal leren werden vergeleken om de beste implementatie voor deze taak te vinden. Op die manier hebben we een methode ontwikkeld die nauwkeurige schattingen maakt met behulp van skipgrammen en, indien beschikbaar, kan profiteren van de tijdsinformatie in de tekst van de berichten. Er is gebruik gemaakt van time series analysetechnieken om deze kenmerken te combineren en om een schatting te doen welke vervolgens geïntegreerd werd met de eerdere schattingen voor die gebeurtenis.

In Hoofdstuk 3 beschrijven we een studie dat tot doel heeft relevante inhoud op sociale media te identificeren op een door een expert bepaald granulariteitsniveau. De evolutie van de tool, die bij elke volgende use case werd geïllustreerd, geeft weer hoe elke stap van de analyse werd ontwikkeld als antwoord op de behoeften die zich voordoen in het proces van het analyseren van een tweetcollectie. De nieuwigheid van onze methodologie is het behandelen van alle stappen van de analyse, van het verzamelen van gegevens tot het hebben van een machine learning classifier die gebruikt kan worden om relevante informatie uit microtekstverzamelingen te detecteren. Het stelt de experts in staat om te ontdekken wat er in een verzameling zit en beslissingen te nemen over de granulariteit van hun analyse. Het resultaat is een schaalbare tool die alle analyses behandelt, van het verzamelen van gegevens tot het hebben van een classificator die kan worden gebruikt om relevante informatie te detecteren tegen een redelijk aantal correcte inschattingen.

Ons werk in het kader van Hoofdstuk 4 heeft aangetoond dat microtekstclassificatie van waarde is voor het detecteren van bruikbare inzichten uit tweetcollecties in rampscenario’s, maar zowel machine learning als regelgebaseerde benaderingen vertonen zwakke punten. We hebben uitgezocht hoe de prestaties van deze benaderingen samenhangen met beschikbare middelen. Onze conclusie is dat de beschikbaarheid van geannoteerde data, het domein en linguïstische expertise samen van invloed zijn om de beste aanpak en de prestaties die kunnen worden behaald in een gegeven casus.

Het volledig automatiseren van microtekstanalyse is ons doel sinds de eerste dag van dit onderzoeksproject. Onze inspanningen in deze richting hebben ons inzicht gegeven in de mate waarin deze automatisering kan worden gerealiseerd. We ontwikkelden meestal eerst een geautomatiseerde aanpak, waarna we deze uitbreidden en verbeterden door menselijke interventie te integreren in verschillende stappen van de geautomatiseerde aanpak. Onze ervaring bevestigt eerder werk dat stelt dat een goed ontworpen menselijke interventie of bijdrage in het ontwerp, de realisatie of evaluatie van een informatiesysteem de prestaties ervan verbetert of de realisatie ervan mogelijk maakt. Nadat onze studies en resultaten ons hebben gericht op de noodzaak en de waarde ervan, werden we geïnspireerd door eerdere studies in het ontwerpen van menselijke betrokkenheid en pasten we onze aanpak aan om te profiteren van de menselijke inbreng. Bijgevolg is onze bijdrage aan bestaand onderzoek in deze lijn de bevestiging geworden van de waarde van menselijk ingrijpen in het extraheren van bruikbare informatie uit microteksten.
\lhead{\emph{Summary}}
\chapter{Summary}

Microblogs such as Twitter represent a powerful source of information. Part of this information can be aggregated beyond the level of individual posts. Some of this aggregated information is referring to events that could or should be acted upon in the interest of e-governance, public safety, or other levels of public interest. Moreover, a significant amount of this information, if aggregated, could complement existing information networks in a non-trivial way. This dissertation proposes a semi-automatic method for extracting actionable information that serves this purpose.

We report three main contributions and a final conclusion that are presented in a separate chapter of this dissertation. First, we show that predicting time to event is possible for both in-domain and cross-domain scenarios. Second, we suggest a method which facilitates the definition of relevance for an analyst’s context and the use of this definition to analyze new data. Finally, we propose a method to integrate the machine learning based relevant information classification method with a rule-based information classification technique to classify microtexts.

In Chapter 2 we reported on our research that aims at characterizing the event information about start time of an event on social media and automatically using it to develop a method that can reliably predict time to event in this chapter. Various feature extraction and machine learning algorithms were explored in order to find the best combination for this task. As a result, we developed a method that produces accurate estimates using skipgrams and, in case available, is able to benefit from temporal information available in the text of the posts. Time series analysis techniques were used to combine these features for generating an estimate and integrate that estimate with the previous estimates for that event.

In Chapter 3 we presented our research that aims at identifying relevant content on social media at a granularity level an expert determines. The evolution of the tool, which was illustrated at each subsequent use case, reflects how each step of the analysis was developed as a response to the needs that arise in the process of analyzing a tweet collection. The novelty of our methodology is in dealing with all steps of the analysis from data collection to having a classifier that can be used to detect relevant information from microtext collections. It enables the experts to discover what is in a collection and make decisions about the granularity of their analysis. The result is a scalable tool that deals with all analysis, from data collection to having a classifier that can be used to detect relevant information at a reasonable performance.

Our work in the scope of Chapter 4 showed that microtext classification can facilitate detecting actionable insights from tweet collections in disaster scenarios, but both machine-learning and rule-based approaches show weaknesses. We determined how the performance of these approaches depends on the starting conditions in terms of available resources. As a result, our conclusion is that availability of annotated data, domain and linguistic expertise together determine the path that should be followed and the performance that can be obtained in a use case.

Fully automatizing microtext analysis has been our goal since the first day of this research project. Our efforts in this direction informed us about the extent this automation can be realized. We mostly first developed an automated approach, then we extended and improved it by integrating human intervention at various steps of the automated approach. Our experience confirms previous work that states that a well-designed human intervention or contribution in design, realization, or evaluation of an information system either improves its performance or enables its realization. As our studies and results directed us toward its necessity and value, we were inspired from previous studies in designing human involvement and customized our approaches to benefit from human input. Consequently, our contribution to existing body of research in this line has become the confirmation of the value of human intervention in extracting actionable information from microtexts.

\lhead{\emph{Curriculum Vitae}}
\chapter{Curriculum Vitae}

Ali H\"{u}rriyeto\v{g}lu has a computer engineering bachelor’s degree from Ege University, İzmir, Turkey, and a Master of Science degree in Cognitive Science from Middle East Technical University, Ankara, Turkey. Moreover, he spent one year as an exchange student at TH Mittelhessen University of Applied Sciences, Giessen, Germany during his computer engineering education. During his master education, he worked at METU as a research assistant for 22 months, which enabled him to understand the context of working at a university. He has started working in research projects in computational linguistics in various settings since the beginning of his MSc education at the European Union Joint Research Center between October 2010 and October 2011. His responsibility was to support a team of computational linguists to develop and integrate Turkish NLP tools in Europe Media Monitor, a multilingual (70 languages) news monitoring and analysis system. Then, he performed his Ph.D. research work at Radboud University, Nijmegen, the Netherlands. His Ph.D. project was embedded in a national ICT project in the Netherlands, so he had the opportunity to collaborate with other researchers and industrial partners in the Netherlands, establish research partnerships with domain experts, and acquire demo development grants for his work with the project partners, and be engaged in a World Bank project as the academic partner for that project. Ali spent 5 months working on sentiment analysis at Netbase Solutions, Inc. in Mountain View, CA, USA during his Ph.D. studies. Finally, after completing his full-time Ph.D. research, he continued working on his dissertation as a guest researcher at Radboud University and worked at the Center for Big Data Statistics at Statistics Netherlands, which is the national official statistics institute in Netherlands. At Statistics Netherlands, he gained experience in working in a governmental organization to apply text mining techniques to official reports and web data in a big data setting for creating national statistics and provide trainings for his colleagues from other national statistics offices around Europe in the line of his work. Ali has been leading a work package on collecting protest information from local news published in India, China, Brazil, Mexico, and South Africa in the scope of a European Research Council (ERC) project since December 1, 2017.

\chapter[SIKS Dissertation Series]{\Huge SIKS Dissertation Series}

\scriptsize
\setlist[enumerate]{topsep=0pt,itemsep=-1ex,partopsep=1ex,parsep=1ex}
\begin{multicols}{2}

\subsubsection*{1998}

\begin{enumerate}

    \item Johan van den Akker (CWI) {\em DEGAS - An Active, Temporal Database of Autonomous Objects}

    \item Floris Wiesman (UM) {\em Information Retrieval by Graphically Browsing Meta-Information}

    \item Ans Steuten (TUD) {\em A Contribution to the Linguistic Analysis of Busine} {\em within the Language/Action Perspective}

    \item Dennis Breuker (UM) {\em Memory versus Search in Games}
    \item E.W. Oskamp (RUL) {\em Computerondersteuning bij Straftoemeting}

\end{enumerate}

\subsubsection*{1999}

\begin{enumerate}

    \item Mark Sloof (VU) {\em Physiology of Quality Change Modelling;} {\em Automated modelling of Quality Change of Agricultural Products}

    \item Rob Potharst (EUR) {\em Classification using decision trees and neural nets}

    \item Don Beal (UM) {\em The Nature of Minimax Search}

    \item Jacques Penders (UM) {\em The practical Art of Moving Physical Objects}

    \item Aldo de Moor (KUB) {\em Empowering Communities: A Method for the Legitimate User-Driven} {\em Specification of Network Information Systems}

    \item Niek J.E. Wijngaards (VU) {\em Re-design of compositional systems}

    \item David Spelt (UT) {\em Verification support for object database design}

    \item Jacques H.J. Lenting (UM) {\em Informed Gambling: Conception and Analysis of a Multi-Agent} {\em Mechanism for Discrete Reallocation}

\end{enumerate}

\subsubsection*{2000}

\begin{enumerate}

    \item Frank Niessink (VU) {\em Perspectives on Improving Software Maintenance}

    \item Koen Holtman (TUE) {\em Prototyping of CMS Storage Management}

    \item Carolien M.T. Metselaar (UVA) {\em Sociaal-organisatorische gevolgen van kennistechnologie;} {\em een procesbenadering en actorperspectief}

    \item Geert de Haan (VU) {\em ETAG, A Formal Model of Competence Knowledge for User Interface Design}

    \item Ruud van der Pol (UM) {\em Knowledge-based Query Formulation in Information Retrieval}

    \item Rogier van Eijk (UU) {\em Programming Languages for Agent Communication}

    \item Niels Peek (UU) {\em Decision-theoretic Planning of Clinical Patient Management}

    \item Veerle Coup (EUR) {\em Sensitivity Analyis of Decision-Theoretic Networks}

    \item Florian Waas (CWI) {\em Principles of Probabilistic Query Optimization}

    \item Niels Nes (CWI) {\em Image Database Management System Design Considerations, Algorithms and Architecture}

    \item Jonas Karlsson (CWI) {\em Scalable Distributed Data Structures for Database Management}

\end{enumerate}

\subsubsection*{2001}

\begin{enumerate}

    \item Silja Renooij (UU) {\em Qualitative Approaches to Quantifying Probabilistic Networks}

    \item Koen Hindriks (UU) {\em Agent Programming Languages: Programming with Mental Models}

    \item Maarten van Someren (UvA) {\em Learning as problem solving}

    \item Evgueni Smirnov (UM) {\em Conjunctive and Disjunctive Version Spaces with } {\em Instance-Based Boundary Sets}

    \item Jacco van Ossenbruggen (VU) {\em Processing Structured Hypermedia: A Matter of Style }

    \item Martijn van Welie (VU) {\em Task-based User Interface Design}

    \item Bastiaan Schonhage (VU) {\em Diva: Architectural Perspectives on Information Visualization}

    \item Pascal van Eck (VU) {\em A Compositional Semantic Structure for Multi-Agent Systems Dynamics}

    \item Pieter Jan 't Hoen (RUL) {\em Towards Distributed Development of Large Object-Oriented Models,} {\em Views of Packages as Classes  }

    \item Maarten Sierhuis (UvA) {\em Modeling and Simulating Work Practice} {\em BRAHMS: a multiagent modeling and simulation language } {\em for work practice analysis and design}

    \item Tom M. van Engers (VUA) {\em Knowledge Management: } {\em The Role of Mental Models in Business Systems Design}

\end{enumerate}

\subsubsection*{2002}

\begin{enumerate}

    \item Nico Lassing (VU) {\em Architecture-Level Modifiability Analysis}

    \item Roelof van Zwol (UT) {\em Modelling and searching web-based document collections}

    \item Henk Ernst Blok (UT) {\em Database Optimization Aspects for Information Retrieval}

    \item Juan Roberto Castelo Valdueza (UU) {\em The Discrete Acyclic Digraph Markov Model in Data Mining}

    \item Radu Serban (VU) {\em The Private Cyberspace Modeling Electronic Environments } {\em inhabited by Privacy-concerned Agents}

    \item Laurens Mommers (UL) {\em Applied legal epistemology; } {\em Building a knowledge-based ontology of the legal domain }

    \item Peter Boncz (CWI) {\em Monet: A Next-Generation DBMS Kernel For Query-Intensive Applications}

    \item Jaap Gordijn (VU) {\em Value Based Requirements Engineering: Exploring Innovative} {\em E-Commerce Ideas }

    \item Willem-Jan van den Heuvel(KUB) {\em Integrating Modern Business Applications with Objectified Legacy Systems }

    \item Brian Sheppard (UM) {\em Towards Perfect Play of Scrabble }

    \item Wouter C.A. Wijngaards (VU) {\em Agent Based Modelling of Dynamics: Biological and Organisational Applications}

    \item Albrecht Schmidt (Uva) {\em Processing XML in Database Systems}

    \item Hongjing Wu (TUE) {\em A Reference Architecture for Adaptive Hypermedia Applications}

    \item Wieke de Vries (UU) {\em Agent Interaction: Abstract Approaches to Modelling, Programming and Verifying Multi-Agent Systems }

    \item Rik Eshuis (UT) {\em Semantics and Verification of UML Activity Diagrams for Workflow Modelling}

    \item Pieter van Langen (VU) {\em The Anatomy of Design: Foundations, Models and Applications  }

    \item Stefan Manegold (UVA) {\em Understanding, Modeling, and Improving Main-Memory Database Performance }

\end{enumerate}

\subsubsection*{2003}

\begin{enumerate}

    \item Heiner Stuckenschmidt (VU) {\em Ontology-Based Information Sharing in Weakly Structured Environments}

    \item Jan Broersen (VU) {\em Modal Action Logics for Reasoning About Reactive Systems }

    \item Martijn Schuemie (TUD) {\em Human-Computer Interaction and Presence in Virtual Reality Exposure Therapy}

    \item Milan Petkovic (UT) {\em Content-Based Video Retrieval Supported by Database Technology }

    \item Jos Lehmann (UVA) {\em Causation in Artificial Intelligence and Law - A modelling approach}

    \item Boris van Schooten (UT) {\em Development and specification of virtual environments}

    \item Machiel Jansen (UvA) {\em Formal Explorations of Knowledge Intensive Tasks}

    \item Yongping Ran (UM) {\em Repair Based Scheduling }

    \item Rens Kortmann (UM) {\em The resolution of visually guided behaviour }

    \item Andreas Lincke (UvT) {\em Electronic Business Negotiation: Some experimental studies on t} {\em between medium, innovation context and culture }

    \item Simon Keizer (UT) {\em Reasoning under Uncertainty in Natural Language Dialogue using Bayesian Networks}

    \item Roeland Ordelman (UT) {\em Dutch speech recognition in multimedia information retrieval}

    \item Jeroen Donkers (UM) {\em Nosce Hostem - Searching with Opponent Models}

    \item Stijn Hoppenbrouwers (KUN) {\em Freezing Language: Conceptualisation Processes across ICT-Supported Organisations}

    \item Mathijs de Weerdt (TUD) {\em Plan Merging in Multi-Agent Systems}

    \item Menzo Windhouwer (CWI) {\em Feature Grammar Systems - Incremental Maintenance of Indexes to } {\em Digital Media Warehouses}

    \item David Jansen (UT) {\em Extensions of Statecharts with Probability, Time, and Stochastic Timing}

    \item Levente Kocsis (UM) {\em Learning Search Decisions }

\end{enumerate}

\subsubsection*{2004}

\begin{enumerate}

    \item Virginia Dignum (UU) {\em A Model for Organizational Interaction: Based on Agents, Founded in Logic}

    \item Lai Xu (UvT) {\em Monitoring Multi-party Contracts for E-business }

    \item Perry Groot (VU) {\em A Theoretical and Empirical Analysis of Approximation in Symbolic Problem Solving }

    \item Chris van Aart (UVA) {\em Organizational Principles for Multi-Agent Architectures}

    \item Viara Popova (EUR) {\em Knowledge discovery and monotonicity}

    \item Bart-Jan Hommes (TUD) {\em The Evaluation of Business Process Modeling Techniques }

    \item Elise Boltjes (UM) {\em Voorbeeldig onderwijs; voorbeeldgestuurd onderwijs, een opstap naar abstract denken, vooral voor meisjes}

    \item Joop Verbeek(UM) {\em Politie en de Nieuwe Internationale Informatiemarkt, Grensregionale} {\em politi\"{e}le gegevensuitwisseling en digitale expertise }

    \item Martin Caminada (VU) {\em For the Sake of the Argument; explorations into argument-based reasoning}

    \item Suzanne Kabel (UVA) {\em Knowledge-rich indexing of learning-objects }

    \item Michel Klein (VU) {\em Change Management for Distributed Ontologies}

    \item The Duy Bui (UT) {\em Creating emotions and facial expressions for embodied agents }

    \item Wojciech Jamroga (UT) {\em Using Multiple Models of Reality: On Agents who Know how to Play }

    \item Paul Harrenstein (UU) {\em Logic in Conflict. Logical Explorations in Strategic Equilibrium }

    \item Arno Knobbe (UU) {\em Multi-Relational Data Mining }

    \item Federico Divina (VU) {\em Hybrid Genetic Relational Search for Inductive Learning }

    \item Mark Winands (UM) {\em Informed Search in Complex Games }

    \item Vania Bessa Machado (UvA) {\em Supporting the Construction of Qualitative Knowledge Models }

    \item Thijs Westerveld (UT) {\em Using generative probabilistic models for multimedia retrieval }

    \item Madelon Evers (Nyenrode) {\em Learning from Design: facilitating multidisciplinary design teams }

\end{enumerate}

\subsubsection*{2005}

\begin{enumerate}

    \item Floor Verdenius (UVA) {\em Methodological Aspects of Designing Induction-Based Applications }

    \item Erik van der Werf (UM)) {\em AI techniques for the game of Go }

    \item Franc Grootjen (RUN) {\em A Pragmatic Approach to the Conceptualisation of Language }

    \item Nirvana Meratnia (UT) {\em Towards Database Support for Moving Object data }

    \item Gabriel Infante-Lopez (UVA) {\em Two-Level Probabilistic Grammars for Natural Language Parsing }

    \item Pieter Spronck (UM) {\em Adaptive Game AI}

    \item Flavius Frasincar (TUE) {\em Hypermedia Presentation Generation for Semantic Web Information Systems}

    \item Richard Vdovjak (TUE) {\em A Model-driven Approach for Building Distributed Ontology-based Web Applications}

    \item Jeen Broekstra (VU) {\em Storage, Querying and Inferencing for Semantic Web Languages}

    \item Anders Bouwer (UVA) {\em Explaining Behaviour: Using Qualitative Simulation in Interactive Learning Environments }

    \item Elth Ogston (VU) {\em Agent Based Matchmaking and Clustering - A Decentralized Approach to Search }

    \item Csaba Boer (EUR) {\em Distributed Simulation in Industry }

    \item Fred Hamburg (UL) {\em Een Computermodel voor het Ondersteunen van Euthanasiebeslissingen }

    \item Borys Omelayenko (VU) {\em Web-Service configuration on the Semantic Web; Exploring how semantics meets pragmatics }

    \item Tibor Bosse (VU) {\em Analysis of the Dynamics of Cognitive Processes }

    \item Joris Graaumans (UU) {\em Usability of XML Query Languages}

    \item Boris Shishkov (TUD) {\em Software Specification Based on Re-usable Business Components }

    \item Danielle Sent (UU) {\em Test-selection strategies for probabilistic networks}

    \item Michel van Dartel (UM) {\em Situated Representation }

    \item Cristina Coteanu (UL) {\em Cyber Consumer Law, State of the Art and Perspectives }

    \item Wijnand Derks (UT) {\em Improving Concurrency and Recovery in Database Systems by Exploiting Application Semantics }

\end{enumerate}

\subsubsection*{2006}

\begin{enumerate}

    \item Samuil Angelov (TUE) {\em Foundations of B2B Electronic Contracting }

    \item Cristina Chisalita (VU) {\em Contextual issues in the design and use of information technology in organizations }

    \item Noor Christoph (UVA) {\em The role of metacognitive skills in learning to solve problems }

    \item Marta Sabou (VU) {\em Building Web Service Ontologies }

    \item Cees Pierik (UU) {\em Validation Techniques for Object-Oriented Proof Outlines }

    \item Ziv Baida (VU) {\em Software-aided Service Bundling - Intelligent Methods \& Tools } {\em for Graphical Service Modeling }

    \item Marko Smiljanic (UT) {\em XML schema matching -- balancing efficiency and effectiveness by means of clustering }

    \item Eelco Herder (UT) {\em Forward, Back and Home Again - Analyzing User Behavior on the Web}

    \item Mohamed Wahdan (UM) {\em Automatic Formulation of the Auditor's Opinion }

    \item Ronny Siebes (VU) {\em Semantic Routing in Peer-to-Peer Systems}

    \item Joeri van Ruth (UT) {\em Flattening Queries over Nested Data Types }

    \item Bert Bongers (VU) {\em Interactivation - Towards an e-cology of people, our technological environment, and the arts }

    \item Henk-Jan Lebbink (UU) {\em Dialogue and Decision Games for Information Exchanging Agents }

    \item Johan Hoorn (VU) {\em Software Requirements: Update, Upgrade, Redesign - towards a Theory of Requirements Change }

    \item Rainer Malik (UU) {\em CONAN: Text Mining in the Biomedical Domain }

    \item Carsten Riggelsen (UU) {\em Approximation Methods for Efficient Learning of Bayesian Networks }

    \item Stacey Nagata (UU) {\em User Assistance for Multitasking with Interruptions on a Mobile Device }

    \item Valentin Zhizhkun (UVA) {\em Graph transformation for Natural Language Processing }

    \item Birna van Riemsdijk (UU) {\em Cognitive Agent Programming: A Semantic Approach }

    \item Marina Velikova (UvT) {\em Monotone models for prediction in data mining }

    \item Bas van Gils (RUN) {\em Aptness on the Web }

    \item Paul de Vrieze (RUN) {\em Fundaments of Adaptive Personalisation }

    \item Ion Juvina (UU) {\em Development of Cognitive Model for Navigating on the Web} 

    \item Laura Hollink (VU) {\em Semantic Annotation for Retrieval of Visual Resources }

    \item Madalina Drugan (UU) {\em Conditional log-likelihood MDL and Evolutionary MCMC }

    \item Vojkan Mihajlovi\'{c} (UT) {\em Score Region Algebra: A Flexible Framework for Structured Information Retrieval }

    \item Stefano Bocconi (CWI) {\em Vox Populi: generating video documentaries from semantically annotated media repositories}

    \item Borkur Sigurbjornsson (UVA) {\em Focused Information Access using XML Element Retrieval}

\end{enumerate}

\subsubsection*{2007}

\begin{enumerate}

    \item Kees Leune (UvT) {\em Access Control and Service-Oriented Architectures}

    \item Wouter Teepe (RUG) {\em Reconciling Information Exchange and Confidentiality: A Formal Approach}

    \item Peter Mika (VU) {\em Social Networks and the Semantic Web }

    \item Jurriaan van Diggelen (UU) {\em Achieving Semantic Interoperability in Multi-agent Systems: a dialogue-based approach }

    \item Bart Schermer (UL) {\em Software Agents, Surveillance, and the Right to Privacy: a Legislative Framework for Agent-enabled Surveillance }

    \item Gilad Mishne (UVA) {\em Applied Text Analytics for Blogs}

    \item Natasa Jovanovic' (UT) {\em To Whom It May Concern - Addressee Identification in Face-to-Face Meetings }

    \item Mark Hoogendoorn (VU) {\em Modeling of Change in Multi-Agent Organizations}

    \item David Mobach (VU) {\em Agent-Based Mediated Service Negotiation }

    \item Huib Aldewereld (UU) {\em Autonomy vs. Conformity: an Institutional Perspective on Norms and Protocols }

    \item Natalia Stash (TUE) {\em Incorporating Cognitive/Learning Styles in a General-Purpose Adaptive Hypermedia System}

    \item Marcel van Gerven (RUN) {\em Bayesian Networks for Clinical Decision Support: A Rational Approach to Dynamic Decision-Making under Uncertainty }

    \item Rutger Rienks (UT) {\em Meetings in Smart Environments; Implications of Progressing Technology}

    \item Niek Bergboer (UM) {\em Context-Based Image Analysis }

    \item Joyca Lacroix (UM) {\em NIM: a Situated Computational Memory Model}

    \item Davide Grossi (UU) {\em Designing Invisible Handcuffs. Formal investigations in Institutions and Organizations for Multi-agent Systems }

    \item Theodore Charitos (UU) {\em Reasoning with Dynamic Networks in Practice }

    \item Bart Orriens (UvT) {\em On the development an management of adaptive business collaborations}

    \item David Levy (UM) {\em Intimate relationships with artificial partners }

    \item Slinger Jansen (UU) {\em Customer Configuration Updating in a Software Supply Network}

    \item Karianne Vermaas (UU) {\em Fast diffusion and broadening use: A research on residential adoption and usage of broadband internet in the Netherlands between 2001 and 2005 }

    \item Zlatko Zlatev (UT) {\em Goal-oriented design of value and process models from patterns}

    \item Peter Barna (TUE) {\em Specification of Application Logic in Web Information Systems}

    \item Georgina Ramírez Camps (CWI) {\em Structural Features in XML Retrieval}

    \item Joost Schalken (VU) {\em Empirical Investigations in Software Process Improvement}

\end{enumerate}

\subsubsection*{2008}

\begin{enumerate}

    \item Katalin Boer-Sorbán (EUR) {\em Agent-Based Simulation of Financial Markets: A modular, continuous-time approach}

    \item Alexei Sharpanskykh (VU) {\em On Computer-Aided Methods for Modeling and Analysis of Organizations}

    \item Vera Hollink (UVA) {\em Optimizing hierarchical menus: a usage-based approach}

    \item Ander de Keijzer (UT) {\em Management of Uncertain Data - towards unattended integration}

    \item Bela Mutschler (UT) {\em Modeling and simulating causal dependencies on process-aware information systems from a cost perspective}

    \item Arjen Hommersom (RUN) {\em On the Application of Formal Methods to Clinical Guidelines, an Artificial Intelligence Perspective}

    \item Peter van Rosmalen (OU) {\em Supporting the tutor in the design and support of adaptive e-learning}

    \item Janneke Bolt (UU) {\em Bayesian Networks: Aspects of Approximate Inference}

    \item Christof van Nimwegen (UU) {\em The paradox of the guided user: assistance can be counter-effective}

    \item Wauter Bosma (UT) {\em Discourse oriented summarization}

    \item Vera Kartseva (VU) {\em Designing Controls for Network Organizations: A Value-Based Approach}

    \item Jozsef Farkas (RUN) {\em A Semiotically Oriented Cognitive Model of Knowledge Representation}

    \item Caterina Carraciolo (UVA) {\em Topic Driven Access to Scientific Handbooks}

    \item Arthur van Bunningen (UT) {\em Context-Aware Querying; Better Answers with Less Effort}

    \item Martijn van Otterlo (UT) {\em The Logic of Adaptive Behavior: Knowledge Representation and Algorithms for the Markov Decision Process Framework in First-Order Domains}

    \item Henriette van Vugt (VU) {\em Embodied agents from a user's perspective}

    \item Martin Op 't Land (TUD) {\em Applying Architecture and Ontology to the Splitting and Allying of Enterprises}

    \item Guido de Croon (UM) {\em Adaptive Active Vision}

    \item Henning Rode (UT) {\em From Document to Entity Retrieval: Improving Precision and Performance of Focused Text Search}

    \item Rex Arendsen (UVA) {\em Geen bericht, goed bericht. Een onderzoek naar de effecten van de introductie van elektronisch berichtenverkeer met de overheid op de administratieve lasten van bedrijven}

    \item Krisztian Balog (UVA) {\em People Search in the Enterprise}

    \item Henk Koning (UU) {\em Communication of IT-Architecture}

    \item Stefan Visscher (UU) {\em Bayesian network models for the management of ventilator-associated pneumonia}

    \item Zharko Aleksovski (VU) {\em Using background knowledge in ontology matching}

    \item Geert Jonker (UU) {\em Efficient and Equitable Exchange in Air Traffic Management Plan Repair using Spender-signed Currency}

    \item Marijn Huijbregts (UT) {\em Segmentation, Diarization and Speech Transcription: Surprise Data Unraveled}

    \item Hubert Vogten (OU) {\em Design and Implementation Strategies for IMS Learning Design}

    \item Ildiko Flesch (RUN) {\em On the Use of Independence Relations in Bayesian Networks}

    \item Dennis Reidsma (UT) {\em Annotations and Subjective Machines - Of Annotators, Embodied Agents, Users, and Other Humans}

    \item Wouter van Atteveldt (VU) {\em Semantic Network Analysis: Techniques for Extracting, Representing and Querying Media Content}

    \item Loes Braun (UM) {\em Pro-Active Medical Information Retrieval}

    \item Trung H. Bui (UT) {\em Toward Affective Dialogue Management using Partially Observable Markov Decision Processes}

    \item Frank Terpstra (UVA) {\em Scientific Workflow Design; theoretical and practical issues}

    \item Jeroen de Knijf (UU) {\em Studies in Frequent Tree Mining}

    \item Ben Torben Nielsen (UvT) {\em Dendritic morphologies: function shapes structure}

\end{enumerate}

\subsubsection*{2009}

\begin{enumerate}

    \item Rasa Jurgelenaite (RUN) {\em Symmetric Causal Independence Models}

    \item Willem Robert van Hage (VU) {\em Evaluating Ontology-Alignment Techniques}

    \item Hans Stol (UvT) {\em A Framework for Evidence-based Policy Making Using IT}

    \item Josephine Nabukenya (RUN) {\em Improving the Quality of Organisational Policy Making using Collaboration Engineering}

    \item Sietse Overbeek (RUN) {\em Bridging Supply and Demand for Knowledge Intensive Tasks - Based on Knowledge, Cognition, and Quality}

    \item Muhammad Subianto (UU) {\em Understanding Classification}

    \item Ronald Poppe (UT) {\em Discriminative Vision-Based Recovery and Recognition of Human Motion}

    \item Volker Nannen (VU) {\em Evolutionary Agent-Based Policy Analysis in Dynamic Environments}

    \item Benjamin Kanagwa (RUN) {\em Design, Discovery and Construction of Service-oriented Systems}

    \item Jan Wielemaker (UVA) {\em Logic programming for knowledge-intensive interactive applications}

    \item Alexander Boer (UVA) {\em Legal Theory, Sources of Law \& the Semantic Web }

    \item Peter Massuthe (TUE, Humboldt-Universitaet zu Berlin) {\em perating Guidelines for Services}

    \item Steven de Jong (UM) {\em Fairness in Multi-Agent Systems }

    \item Maksym Korotkiy (VU) {\em From ontology-enabled services to service-enabled ontologies (making ontologies work in e-science with ONTO-SOA) }

    \item Rinke Hoekstra (UVA) {\em Ontology Representation - Design Patterns and Ontologies that Make Sense}

    \item Fritz Reul (UvT) {\em New Architectures in Computer Chess}

    \item Laurens van der Maaten (UvT) {\em Feature Extraction from Visual Data}

    \item Fabian Groffen (CWI) {\em Armada, An Evolving Database System}

    \item Valentin Robu (CWI) {\em Modeling Preferences, Strategic Reasoning and Collaboration in Agent-Mediated Electronic Markets}

    \item Bob van der Vecht (UU) {\em Adjustable Autonomy: Controling Influences on Decision Making}

    \item Stijn Vanderlooy (UM) {\em Ranking and Reliable Classification}

    \item Pavel Serdyukov (UT) {\em Search For Expertise: Going beyond direct evidence}

    \item Peter Hofgesang (VU) {\em Modelling Web Usage in a Changing Environment}

    \item Annerieke Heuvelink (VUA) {\em Cognitive Models for Training Simulations}

    \item Alex van Ballegooij (CWI) {\em "RAM: Array Database Management through Relational Mapping"}

    \item Fernando Koch (UU) {\em An Agent-Based Model for the Development of Intelligent Mobile Services}

    \item Christian Glahn (OU) {\em Contextual Support of social Engagement and Reflection on the Web}

    \item Sander Evers (UT) {\em Sensor Data Management with Probabilistic Models}

    \item Stanislav Pokraev (UT) {\em Model-Driven Semantic Integration of Service-Oriented Applications}

    \item Marcin Zukowski (CWI) {\em Balancing vectorized query execution with bandwidth-optimized storage}

    \item Sofiya Katrenko (UVA) {\em A Closer Look at Learning Relations from Text}

    \item Rik Farenhorst (VU) and Remco de Boer (VU) {\em Architectural Knowledge Management: Supporting Architects and Auditors}

    \item Khiet Truong (UT) {\em How Does Real Affect Affect Affect Recognition In Speech?}

    \item Inge van de Weerd (UU) {\em Advancing in Software Product Management: An Incremental Method Engineering Approach}

    \item Wouter Koelewijn (UL) {\em Privacy en Politiegegevens; Over geautomatiseerde normatieve informatie-uitwisseling}

    \item Marco Kalz (OUN) {\em Placement Support for Learners in Learning Networks}

    \item Hendrik Drachsler (OUN) {\em Navigation Support for Learners in Informal Learning Networks}

    \item Riina Vuorikari (OU) {\em Tags and self-organisation: a metadata ecology for learning resources in a multilingual context}

    \item Christian Stahl (TUE, Humboldt-Universitaet zu Berlin) {\em Service Substitution -- A Behavioral Approach Based on Petri Nets }

    \item Stephan Raaijmakers (UvT) {\em Multinomial Language Learning: Investigations into the Geometry of Language}

    \item Igor Berezhnyy (UvT) {\em Digital Analysis of Paintings}

    \item Toine Bogers {\em Recommender Systems for Social Bookmarking}

    \item Virginia Nunes Leal Franqueira (UT) {\em Finding Multi-step Attacks in Computer Networks using Heuristic Search and Mobile Ambients}

    \item Roberto Santana Tapia (UT) {\em Assessing Business-IT Alignment in Networked Organizations}

    \item Jilles Vreeken (UU) {\em Making Pattern Mining Useful}

    \item Loredana Afanasiev (UvA) {\em Querying XML: Benchmarks and Recursion}

\end{enumerate}

\subsubsection*{2010}

\begin{enumerate}

    \item Matthijs van Leeuwen (UU) {\em Patterns that Matter}

    \item Ingo Wassink (UT) {\em Work flows in Life Science}

    \item Joost Geurts (CWI) {\em A Document Engineering Model and Processing Framework for Multimedia documents}
    \item Olga Kulyk (UT) {\em Do You Know What I Know? Situational Awareness of Co-located Teams in Multidisplay Environments}

    \item Claudia Hauff (UT) {\em Predicting the Effectiveness of Queries and Retrieval Systems}

    \item Sander Bakkes (UvT) {\em Rapid Adaptation of Video Game AI}

    \item Wim Fikkert (UT) {\em Gesture interaction at a Distance}

    \item Krzysztof Siewicz (UL) {\em Towards an Improved Regulatory Framework of Free Software. Protecting user freedoms in a world of software communities and eGovernments}

    \item Hugo Kielman (UL) {\em A Politiele gegevensverwerking en Privacy, Naar een effectieve waarborging}

    \item Rebecca Ong (UL) {\em Mobile Communication and Protection of Childr}
    \item Adriaan Ter Mors (TUD) {\em The world according to MARP: Multi-Agent Route Planning }

    \item Susan van den Braak (UU) {\em Sensemaking software for crime analysis}

    \item Gianluigi Folino (RUN) {\em High Performance Data Mining using Bio-inspired techniques }

    \item Sander van Splunter (VU) {\em Automated Web Service Reconfiguration}

    \item Lianne Bodenstaff (UT) {\em Managing Dependency Relations in Inter-Organizational Models }

    \item Sicco Verwer (TUD) {\em Efficient Identification of Timed Automata, theory and practice }

    \item Spyros Kotoulas (VU) {\em Scalable Discovery of Networked Resources: Algorithms, Infrastructure, Applications}

    \item Charlotte Gerritsen (VU) {\em Caught in the Act: Investigating Crime by Agent-Based Simulation}

    \item Henriette Cramer (UvA) {\em People's Responses to Autonomous and Adaptive Systems}

    \item Ivo Swartjes (UT) {\em Whose Story Is It Anyway? How Improv Informs Agency and Authorship of Emergent Narrative }

    \item Harold van Heerde (UT) {\em Privacy-aware data management by means of data degradation}

    \item Michiel Hildebrand (CWI) {\em End-user Support for Access to Heterogeneous Linked Data }

    \item Bas Steunebrink (UU) {\em The Logical Structure of Emotions}
    \item Dmytro Tykhonov (TUD) {\em Designing Generic and Efficient Negotiation Strategies}

    \item Zulfiqar Ali Memon (VU) {\em Modelling Human-Awareness for Ambient Agents: A Human Mindreading Perspective}

    \item Ying Zhang (CWI) {\em XRPC: Efficient Distributed Query Processing on Heterogeneous XQuery Engines}
    \item Marten Voulon (UL) {\em Automatisch contracteren}

    \item Arne Koopman (UU) {\em Characteristic Relational Patterns}

    \item Stratos Idreos(CWI) {\em Database Cracking: Towards Auto-tuning Database Kernels }

    \item Marieke van Erp (UvT) {\em Accessing Natural History - Discoveries in data cleaning, structuring, and retrieval }

    \item Victor de Boer (UVA) {\em Ontology Enrichment from Heterogeneous Sources on the Web}

    \item Marcel Hiel (UvT) {\em An Adaptive Service Oriented Architecture: Automatically solving Interoperability Problems }

    \item Robin Aly (UT) {\em Modeling Representation Uncertainty in Concept-Based Multimedia Retrieval}

    \item Teduh Dirgahayu (UT) {\em Interaction Design in Service Compositions}

    \item Dolf Trieschnigg (UT) {\em Proof of Concept: Concept-based Biomedical Information Retrieval  }

    \item Jose Janssen (OU) {\em Paving the Way for Lifelong Learning; Facilitating competence development through a learning path specification}

    \item Niels Lohmann (TUE) {\em Correctness of services and their composition}

    \item Dirk Fahland (TUE) {\em From Scenarios to components}

    \item Ghazanfar Farooq Siddiqui (VU) {\em Integrative modeling of emotions in virtual agents}

    \item Mark van Assem (VU) {\em Converting and Integrating Vocabularies for the Semantic Web}

    \item Guillaume Chaslot (UM) {\em Monte-Carlo Tree Search}

    \item Sybren de Kinderen (VU) {\em Needs-driven service bundling in a multi-supplier setting - the computational e3-service approach}

    \item Peter van Kranenburg (UU) {\em A Computational Approach to Content-Based Retrieval of Folk Song Melodies}

    \item Pieter Bellekens (TUE) {\em An Approach towards Context-sensitive and User-adapted Access to Heterogeneous Data Sources, Illustrated in the Television Domain}

    \item Vasilios Andrikopoulos (UvT) {\em A theory and model for the evolution of software services}

    \item Vincent Pijpers (VU) {\em  e3alignment: Exploring Inter-Organizational Business-ICT Alignment}

    \item Chen Li (UT) {\em Mining Process Model Variants: Challenges, Techniques, Examples}

    \item Milan Lovric (EUR) {\em Behavioral Finance and Agent-Based Artificial Markets}

    \item Jahn-Takeshi Saito (UM) {\em Solving difficult game positions }

    \item Bouke Huurnink (UVA) {\em Search in Audiovisual Broadcast Archives}
    \item Alia Khairia Amin (CWI) {\em Understanding and supporting information seeking tasks in multiple sources }

    \item Peter-Paul van Maanen (VU) {\em Adaptive Support for Human-Computer Teams: Exploring the Use of Cognitive Models of Trust and Attention }

    \item Edgar Meij (UVA) {\em Combining Concepts and Language Models for Information Access}

\end{enumerate}

\subsubsection*{2011}

\begin{enumerate}

    \item Botond Cseke (RUN) {\em Variational Algorithms for Bayesian Inference in Latent Gaussian Models}

    \item Nick Tinnemeier(UU) {\em Organizing Agent Organizations. Syntax and Operational Semantics of an Organization-Oriented Programming Language}

    \item Jan Martijn van der Werf (TUE) {\em Compositional Design and Verification of Component-Based Information Systems}

    \item Hado van Hasselt (UU) {\em Insights in Reinforcement Learning; Formal analysis and empirical evaluation of temporal-difference learning algorithms}

    \item Base van der Raadt (VU) {\em Enterprise Architecture Coming of Age - Increasing the Performance of an Emerging Discipline}

    \item Yiwen Wang (TUE) {\em Semantically-Enhanced Recommendations in Cultural Heritage}

    \item Yujia Cao (UT) {\em Multimodal Information Presentation for High Load Human Computer Interaction}

    \item Nieske Vergunst (UU) {\em BDI-based Generation of Robust Task-Oriented Dialogues}
    \item Tim de Jong (OU) {\em Contextualised Mobile Media for Learning}

    \item Bart Bogaert (UvT) {\em Cloud Content Contention}

    \item Dhaval Vyas (UT) {\em Designing for Awareness: An Experience-focused HCI Perspective}

    \item Carmen Bratosin (TUE) {\em Grid Architecture for Distributed Process Mining}

    \item Xiaoyu Mao (UvT) {\em Airport under Control. Multiagent Scheduling for Airport Ground Handling}

    \item Milan Lovric (EUR) {\em Behavioral Finance and Agent-Based Artificial Markets}

    \item Marijn Koolen (UvA) {\em The Meaning of Structure: the Value of Link Evidence for Information Retrieval}

    \item Maarten Schadd (UM) {\em Selective Search in Games of Different Complexity}

    \item Jiyin He (UVA) {\em Exploring Topic Structure: Coherence, Diversity and Relatedness}

    \item Mark Ponsen (UM) {\em Strategic Decision-Making in complex games }

    \item Ellen Rusman (OU) {\em The Mind ' s Eye on Personal Profiles}

    \item Qing Gu (VU) {\em Guiding service-oriented software engineering - A view-based approach}

    \item Linda Terlouw (TUD) {\em Modularization and Specification of Service-Oriented Systems}

    \item Junte Zhang (UVA) {\em System Evaluation of Archival Description and Access}

    \item Wouter Weerkamp (UVA) {\em Finding People and their Utterances in Social Media}

    \item Herwin van Welbergen (UT) {\em Behavior Generation for Interpersonal Coordination with Virtual Humans On Specifying, Scheduling and Realizing Multimodal Virtual Human Behavior}

    \item Syed Waqar ul Qounain Jaffry (VU) {\em Analysis and Validation of Models for Trust Dynamics}

    \item Matthijs Aart Pontier (VU) {\em Virtual Agents for Human Communication - Emotion Regulation and Involvement-Distance Trade-Offs in Embodied Conversational Agents and Robots }

    \item Aniel Bhulai (VU) {\em Dynamic website optimization through autonomous management of design patterns}

    \item Rianne Kaptein(UVA) {\em Effective Focused Retrieval by Exploiting Query Context and Document Structure }

    \item Faisal Kamiran (TUE) {\em Discrimination-aware Classification}

    \item Egon van den Broek (UT) {\em Affective Signal Processing (ASP): Unraveling the mystery of emotions}

    \item Ludo Waltman (EUR) {\em Computational and Game-Theoretic Approaches for Modeling Bounded Rationality}
    \item Nees-Jan van Eck (EUR) {\em Methodological Advances in Bibliometric Mapping of Science }

    \item Tom van der Weide (UU) {\em Arguing to Motivate Decisions}

    \item Paolo Turrini (UU) {\em Strategic Reasoning in Interdependence: Logical and Game-theoretical Investigations}

    \item Maaike Harbers (UU) {\em Explaining Agent Behavior in Virtual Training }

    \item Erik van der Spek (UU) {\em Experiments in serious game design: a cognitive approach }
    \item Adriana Burlutiu (RUN) {\em Machine Learning for Pairwise Data, Applications for Preference Learning and Supervised Network Inference }

    \item Nyree Lemmens (UM) {\em Bee-inspired Distributed Optimization}

    \item Joost Westra (UU) {\em Organizing Adaptation using Agents in Serious Games }

    \item Viktor Clerc (VU) {\em Architectural Knowledge Management in Global Software Development }

    \item Luan Ibraimi (UT) {\em Cryptographically Enforced Distributed Data Access Control }

    \item Michal Sindlar (UU) {\em Explaining Behavior through Mental State Attribution}

    \item Henk van der Schuur (UU) {\em Process Improvement through Software Operation Knowledge}

    \item Boris Reuderink (UT) {\em Robust Brain-Computer Interfaces }
    \item Herman Stehouwer (UvT) {\em Statistical Language Models for Alternative Sequence Selection }

    \item Beibei Hu (TUD) {\em Towards Contextualized Information Delivery: A Rule-based Architecture for the Domain of Mobile Police Work }

    \item Azizi Bin Ab Aziz(VU) {\em Exploring Computational Models for Intelligent Support of Persons with Depression}

    \item Mark Ter Maat (UT) {\em Response Selection and Turn-taking for a Sensitive Artificial Listening Agent }
    \item Andreea Niculescu (UT) {\em Conversational interfaces for task-oriented spoken dialogues: design aspects influencing interaction quality}

\end{enumerate}

\subsubsection*{2012}

\begin{enumerate}
    \item Terry Kakeeto (UvT) {\em Relationship Marketing for SMEs in Uganda}

    \item Muhammad Umair(VU) {\em Adaptivity, emotion, and Rationality in Human and Ambient Agent Models}

    \item Adam Vanya (VU) {\em Supporting Architecture Evolution by Mining Software Repositories}

    \item Jurriaan Souer (UU) {\em Development of Content Management System-based Web Applications}

    \item Marijn Plomp (UU) {\em Maturing Interorganisational Information Systems}

    \item Wolfgang Reinhardt (OU) {\em Awareness Support for Knowledge Workers in Research Networks}

    \item Rianne van Lambalgen (VU) {\em When the Going Gets Tough: Exploring Agent-based Models of Human Performance under Demanding Conditions}

    \item Gerben de Vries (UVA) {\em Kernel Methods for Vessel Trajectories}

    \item Ricardo Neisse (UT) {\em Trust and Privacy Management Support for Context-Aware Service Platforms}

    \item David Smits (TUE) {\em Towards a Generic Distributed Adaptive Hypermedia Environment}

    \item J.C.B. Rantham Prabhakara (TUE) {\em Process Mining in the Large: Preprocessing, Discovery, and Diagnostics}

    \item Kees van der Sluijs (TUE) {\em Model Driven Design and Data Integration in Semantic Web Information Systems}

    \item Suleman Shahid (UvT) {\em Fun and Face: Exploring non-verbal expressions of emotion during playful interactions}

    \item Evgeny Knutov(TUE) {\em Generic Adaptation Framework for Unifying Adaptive Web-based Systems}

    \item Natalie van der Wal (VU) {\em Social Agents. Agent-Based Modelling of Integrated Internal and Social Dynamics of Cognitive and Affective Processes}

    \item Fiemke Both (VU) {\em Helping people by understanding them - Ambient Agents supporting task execution and depression treatment}

    \item Amal Elgammal (UvT) {\em Towards a Comprehensive Framework for Business Process Compliance}

    \item Eltjo Poort (VU) {\em Improving Solution Architecting Practices }

    \item Helen Schonenberg (TUE) {\em What's Next? Operational Support for Business Process Execution}

    \item Ali Bahramisharif (RUN) {\em Covert Visual Spatial Attention, a Robust Paradigm for Brain-Computer Interfacing}

    \item Roberto Cornacchia (TUD) {\em Querying Sparse Matrices for Information Retrieval}

    \item Thijs Vis (UvT) {\em Intelligence, politie en veiligheidsdienst: verenigbare grootheden? }

    \item Christian Muehl (UT) {\em Toward Affective Brain-Computer Interfaces: Exploring the Neurophysiology of Affect during Human Media Interaction}

    \item Laurens van der Werff (UT) {\em Evaluation of Noisy Transcripts for Spoken Document Retrieval}
    \item Silja Eckartz (UT) {\em Managing the Business Case Development in Inter-Organizational IT Projects: A Methodology and its Application}
    \item Emile de Maat (UVA) {\em Making Sense of Legal Text}
    \item Hayrettin Gurkok (UT) {\em Mind the Sheep! User Experience Evaluation \& Brain-Computer Interface Games}
    \item Nancy Pascall (UvT) {\em Engendering Technology Empowering Women}
    \item Almer Tigelaar (UT) {\em Peer-to-Peer Information Retrieval}
    \item Alina Pommeranz (TUD) {\em Designing Human-Centered Systems for Reflective Decision Making}
    \item Emily Bagarukayo (RUN) {\em A Learning by Construction Approach for Higher Order Cognitive Skills Improvement, Building Capacity and Infrastructure}
    \item Wietske Visser (TUD) {\em Qualitative multi-criteria preference representation and reasoning}
    \item Rory Sie (OUN) {\em Coalitions in Cooperation Networks (COCOON)}
    \item Pavol Jancura (RUN) {\em Evolutionary analysis in PPI networks and applications}
    \item Evert Haasdijk (VU) {\em Never Too Old To Learn -- On-line Evolution of Controllers in Swarm- and Modular Robotics}
    \item Denis Ssebugwawo (RUN) {\em Analysis and Evaluation of Collaborative Modeling Processes}
    \item Agnes Nakakawa (RUN) {\em A Collaboration Process for Enterprise Architecture Creation}
    \item Selmar Smit (VU) {\em Parameter Tuning and Scientific Testing in Evolutionary Algorithms}
    \item Hassan Fatemi (UT) {\em Risk-aware design of value and coordination networks}
    \item Agus Gunawan (UvT) {\em Information Access for SMEs in Indonesia}
    \item Sebastian Kelle (OU) {\em Game Design Patterns for Learning}
    \item Dominique Verpoorten (OU) {\em Reflection Amplifiers in self-regulated Learning}
    \item Anna Tordai (VU) {\em On Combining Alignment Techniques}
    \item Benedikt Kratz (UvT) {\em A Model and Language for Business-aware Transactions}
    \item Simon Carter (UVA) {\em Exploration and Exploitation of Multilingual Data for Statistical Machine Translation}
    \item Manos Tsagkias (UVA) {\em Mining Social Media: Tracking Content and Predicting Behavior}
    \item Jorn Bakker (TUE) {\em Handling Abrupt Changes in Evolving Time-series Data}
    \item Michael Kaisers (UM) {\em Learning against Learning - Evolutionary dynamics of reinforcement learning algorithms in strategic interactions}
    \item Steven van Kervel (TUD) {\em Ontologogy driven Enterprise Information Systems Engineering}
    \item Jeroen de Jong (TUD) {\em Heuristics in Dynamic Sceduling; a practical framework with a case study in elevator dispatching}
\end{enumerate}

\subsubsection*{2013}

\begin{enumerate}
    \item Viorel Milea (EUR) {\em News Analytics for Financial Decision Support}
    \item Erietta Liarou (CWI) {\em MonetDB/DataCell: Leveraging the Column-store Database Technology for Efficient and Scalable Stream Processing}
    \item Szymon Klarman (VU) {\em Reasoning with Contexts in Description Logics}
    \item Chetan Yadati(TUD) {\em Coordinating autonomous planning and scheduling}
    \item Dulce Pumareja (UT) {\em Groupware Requirements Evolutions Patterns}
    \item Romulo Goncalves(CWI) {\em The Data Cyclotron: Juggling Data and Queries for a Data Warehouse Audience}
    \item Giel van Lankveld (UvT) {\em Quantifying Individual Player Differences}
    \item Robbert-Jan Merk(VU) {\em Making enemies: cognitive modeling for opponent agents in fighter pilot simulators}
    \item Fabio Gori (RUN) {\em Metagenomic Data Analysis: Computational Methods and Applications}
    \item Jeewanie Jayasinghe Arachchige(UvT) {\em A Unified Modeling Framework for Service Design}
    \item Evangelos Pournaras(TUD) {\em Multi-level Reconfigurable Self-organization in Overlay Services}
    \item Marian Razavian(VU) {\em Knowledge-driven Migration to Services}
    \item Mohammad Safiri(UT) {\em Service Tailoring: User-centric creation of integrated IT-based homecare services to support independent living of elderly}
    \item Jafar Tanha (UVA) {\em Ensemble Approaches to Semi-Supervised Learning Learning}
    \item Daniel Hennes (UM) {\em Multiagent Learning - Dynamic Games and Applications}
    \item Eric Kok (UU) {\em Exploring the practical benefits of argumentation in multi-agent deliberation}
    \item Koen Kok (VU) {\em The PowerMatcher: Smart Coordination for the Smart Electricity Grid}
    \item Jeroen Janssens (UvT) {\em Outlier Selection and One-Class Classification}
    \item Renze Steenhuizen (TUD) {\em Coordinated Multi-Agent Planning and Scheduling}
    \item Katja Hofmann (UvA) {\em Fast and Reliable Online Learning to Rank for Information Retrieval}
    \item Sander Wubben (UvT) {\em Text-to-text generation by monolingual machine translation}
    \item Tom Claassen (RUN) {\em Causal Discovery and Logic}
    \item Patricio de Alencar Silva(UvT) {\em Value Activity Monitoring}
    \item Haitham Bou Ammar (UM) {\em Automated Transfer in Reinforcement Learning}
    \item Agnieszka Anna Latoszek-Berendsen (UM) {\em Intention-based Decision Support. A new way of representing and implementing clinical guidelines in a Decision Support System}
    \item Alireza Zarghami (UT) {\em Architectural Support for Dynamic Homecare Service Provisioning}
    \item Mohammad Huq (UT) {\em Inference-based Framework Managing Data Provenance}
    \item Frans van der Sluis (UT) {\em When Complexity becomes Interesting: An Inquiry into the Information eXperience}
    \item Iwan de Kok (UT) {\em Listening Heads}
    \item Joyce Nakatumba (TUE) {\em Resource-Aware Business Process Management: Analysis and Support}
    \item Dinh Khoa Nguyen (UvT) {\em Blueprint Model and Language for Engineering Cloud Applications}
    \item Kamakshi Rajagopal (OUN) {\em Networking For Learning; The role of Networking in a Lifelong Learner's Professional Development}
    \item Qi Gao (TUD) {\em User Modeling and Personalization in the Microblogging Sphere}
    \item Kien Tjin-Kam-Jet (UT) {\em Distributed Deep Web Search}
    \item Abdallah El Ali (UvA) {\em Minimal Mobile Human Computer Interaction}
    \item Than Lam Hoang (TUe) {\em Pattern Mining in Data Streams}
    \item Dirk B\"{u}rner (OUN) {\em Ambient Learning Displays}
    \item Eelco den Heijer (VU) {\em Autonomous Evolutionary Art}
    \item Joop de Jong (TUD) {\em A Method for Enterprise Ontology based Design of Enterprise Information Systems}
    \item Pim Nijssen (UM) {\em Monte-Carlo Tree Search for Multi-Player Games}
    \item Jochem Liem (UVA) {\em Supporting the Conceptual Modelling of Dynamic Systems: A Knowledge Engineering Perspective on Qualitative Reasoning}
    \item L\'{e}on Planken (TUD) {\em Algorithms for Simple Temporal Reasoning}
    \item Marc Bron (UVA) {\em Exploration and Contextualization through Interaction and Concepts}
\end{enumerate}

\subsubsection*{2014}

\begin{enumerate}
    \item Nicola Barile (UU) {\em Studies in Learning Monotone Models from Data}
    \item Fiona Tuliyano (RUN) {\em Combining System Dynamics with a Domain Modeling Method}
    \item Sergio Raul Duarte Torres (UT) {\em Information Retrieval for Children: Search Behavior and Solutions}
    \item Hanna Jochmann-Mannak (UT) {\em Websites for children: search strategies and interface design - Three studies on children's search performance and evaluation}
    \item Jurriaan van Reijsen (UU) {\em Knowledge Perspectives on Advancing Dynamic Capability}
    \item Damian Tamburri (VU) {\em Supporting Networked Software Development}
    \item Arya Adriansyah (TUE) {\em Aligning Observed and Modeled Behavior}
    \item Samur Araujo (TUD) {\em Data Integration over Distributed and Heterogeneous Data Endpoints}
    \item Philip Jackson (UvT) {\em Toward Human-Level Artificial Intelligence: Representation and Computation of Meaning in Natural Language}
    \item Ivan Salvador Razo Zapata (VU) {\em Service Value Networks}
    \item Janneke van der Zwaan (TUD) {\em An Empathic Virtual Buddy for Social Support}
    \item Willem van Willigen (VU) {\em Look Ma, No Hands: Aspects of Autonomous Vehicle Control}
    \item Arlette van Wissen (VU) {\em Agent-Based Support for Behavior Change: Models and Applications in Health and Safety Domains}
    \item Yangyang Shi (TUD) {\em Language Models With Meta-information}
    \item Natalya Mogles (VU) {\em Agent-Based Analysis and Support of Human Functioning in Complex Socio-Technical Systems: Applications in Safety and Healthcare}
    \item Krystyna Milian (VU) {\em Supporting trial recruitment and design by automatically interpreting eligibility criteria}
    \item Kathrin Dentler (VU) {\em Computing healthcare quality indicators automatically: Secondary Use of Patient Data and Semantic Interoperability}
    \item Mattijs Ghijsen (UVA) {\em Methods and Models for the Design and Study of Dynamic Agent Organizations}
    \item Vinicius Ramos (TUE) {\em Adaptive Hypermedia Courses: Qualitative and Quantitative Evaluation and Tool Support}
    \item Mena Habib (UT) {\em Named Entity Extraction and Disambiguation for Informal Text: The Missing Link}
    \item Kassidy Clark (TUD) {\em Negotiation and Monitoring in Open Environments}
    \item Marieke Peeters (UU) {\em Personalized Educational Games - Developing agent-supported scenario-based training}
    \item Eleftherios Sidirourgos (UvA/CWI) {\em Space Efficient Indexes for the Big Data Era}
    \item Davide Ceolin (VU) {\em Trusting Semi-structured Web Data}
    \item Martijn Lappenschaar (RUN) {\em New network models for the analysis of disease interaction}
    \item Tim Baarslag (TUD) {\em What to Bid and When to Stop}
    \item Rui Jorge Almeida (EUR) {\em Conditional Density Models Integrating Fuzzy and Probabilistic Representations of Uncertainty}
    \item Anna Chmielowiec (VU) {\em Decentralized k-Clique Matching}
    \item Jaap Kabbedijk (UU) {\em Variability in Multi-Tenant Enterprise Software}
    \item Peter de Cock (UvT) {\em Anticipating Criminal Behaviour}
    \item Leo van Moergestel (UU) {\em Agent Technology in Agile Multiparallel Manufacturing and Product Support}
    \item Naser Ayat (UvA) {\em On Entity Resolution in Probabilistic Data}
    \item Tesfa Tegegne (RUN) {\em Service Discovery in eHealth}
    \item Christina Manteli(VU) {\em The Effect of Governance in Global Software Development: Analyzing Transactive Memory Systems}
    \item Joost van Ooijen (UU) {\em Cognitive Agents in Virtual Worlds: A Middleware Design Approach}
    \item Joos Buijs (TUE) {\em Flexible Evolutionary Algorithms for Mining Structured Process Models}
    \item Maral Dadvar (UT) {\em Experts and Machines United Against Cyberbullying}
    \item Danny Plass-Oude Bos (UT) {\em Making brain-computer interfaces better: improving usability through post-processing.}
    \item Jasmina Maric (UvT) {\em Web Communities, Immigration, and Social Capital}
    \item Walter Omona (RUN) {\em A Framework for Knowledge Management Using ICT in Higher Education}
    \item Frederic Hogenboom (EUR) {\em Automated Detection of Financial Events in News Text}
    \item Carsten Eijckhof (CWI/TUD) {\em Contextual Multidimensional Relevance Models}
    \item Kevin Vlaanderen (UU) {\em Supporting Process Improvement using Method Increments}
    \item Paulien Meesters (UvT) {\em Intelligent Blauw. Met als ondertitel: Intelligence-gestuurde politiezorg in gebiedsgebonden eenheden}
    \item Birgit Schmitz (OUN) {\em Mobile Games for Learning: A Pattern-Based Approach}
    \item Ke Tao (TUD) {\em Social Web Data Analytics: Relevance, Redundancy, Diversity}
    \item Shangsong Liang (UVA) {\em Fusion and Diversification in Information Retrieval}
\end{enumerate}

\subsubsection*{2015}

\begin{enumerate}
    \item Niels Netten (UvA) {\em Machine Learning for Relevance of Information in Crisis Response}
    \item Faiza Bukhsh (UvT) {\em Smart auditing: Innovative Compliance Checking in Customs Controls}
    \item Twan van Laarhoven (RUN) {\em Machine learning for network data}
    \item Howard Spoelstra (OUN) {\em Collaborations in Open Learning Environments}
    \item Christoph B\"{o}sch(UT) {\em Cryptographically Enforced Search Pattern Hiding}
    \item Farideh Heidari (TUD) {\em Business Process Quality Computation - Computing Non-Functional Requirements to Improve Business Processes}
    \item Maria-Hendrike Peetz(UvA) {\em Time-Aware Online Reputation Analysis}
    \item Jie Jiang (TUD) {\em Organizational Compliance: An agent-based model for designing and evaluating organizational interactions}
    \item Randy Klaassen(UT) {\em HCI Perspectives on Behavior Change Support Systems}
    \item Henry Hermans (OUN) {\em OpenU: design of an integrated system to support lifelong learning}
    \item Yongming Luo(TUE) {\em Designing algorithms for big graph datasets: A study of computing bisimulation and joins}
    \item Julie M. Birkholz (VU) {\em Modi Operandi of Social Network Dynamics: The Effect of Context on Scientific Collaboration Networks}
    \item Giuseppe Procaccianti(VU) {\em Energy-Efficient Software}
    \item Bart van Straalen (UT) {\em A cognitive approach to modeling bad news conversations}
    \item Klaas Andries de Graaf (VU) {\em Ontology-based Software Architecture Documentation}
    \item Changyun Wei (UT) {\em Cognitive Coordination for Cooperative Multi-Robot Teamwork}
    \item Andr\'{e}' van Cleeff (UT) {\em Physical and Digital Security Mechanisms: Properties, Combinations and Trade-offs}
    \item Holger Pirk (CWI) {\em Waste Not, Want Not! - Managing Relational Data in Asymmetric Memories}
    \item Bernardo Tabuenca (OUN) {\em Ubiquitous Technology for Lifelong Learners}
    \item Lo\"{i}s Vanh\'{e}e(UU) {\em Using Culture and Values to Support Flexible Coordination}
    \item Sibren Fetter (OUN) {\em Using Peer-Support to Expand and Stabilize Online Learning}
    \item Zhemin Zhu(UT) {\em Co-occurrence Rate Networks}
    \item Luit Gazendam (VU) {\em Cataloguer Support in Cultural Heritage}
    \item Richard Berendsen (UVA) {\em Finding People, Papers, and Posts: Vertical Search Algorithms and Evaluation}
    \item Steven Woudenberg (UU) {\em Bayesian Tools for Early Disease Detection}
    \item Alexander Hogenboom (EUR) {\em Sentiment Analysis of Text Guided by Semantics and Structure}
    \item S\'{a}ndor H\'{e}man (CWI) {\em Updating compressed colomn stores}
    \item Janet Bagorogoza(TiU) {\em KNOWLEDGE MANAGEMENT AND HIGH PERFORMANCE; The Uganda Financial Institutions Model for HPO}
    \item Hendrik Baier (UM) {\em Monte-Carlo Tree Search Enhancements for One-Player and Two-Player Domains}
    \item Kiavash Bahreini(OU) {\em Real-time Multimodal Emotion Recognition in E-Learning}
    \item Yakup Ko\c{c} (TUD) {\em On the robustness of Power Grids}
    \item Jerome Gard(UL) {\em Corporate Venture Management in SMEs}
    \item Frederik Schadd (TUD) {\em Ontology Mapping with Auxiliary Resources}
    \item Victor de Graaf(UT) {\em Gesocial Recommender Systems}
    \item Jungxao Xu (TUD) {\em Affective Body Language of Humanoid Robots: Perception and Effects in Human Robot Interaction}
\end{enumerate}

\subsubsection*{2016}

\begin{enumerate}
    \item Syed Saiden Abbas (RUN) {\em Recognition of Shapes by Humans and Machines}
    \item Michiel Christiaan Meulendijk (UU) {\em Optimizing medication reviews through decision support: prescribing a better pill to swallow}
    \item Maya Sappelli (RUN) {\em Knowledge Work in Context: User Centered Knowledge Worker Support}
    \item Laurens Rietveld (VU) {\em Publishing and Consuming Linked Data}
    \item Evgeny Sherkhonov (UVA) {\em Expanded Acyclic Queries: Containment and an Application in Explaining Missing Answers}
    \item Michel Wilson (TUD) {\em Robust scheduling in an uncertain environment}
    \item Jeroen de Man (VU) {\em Measuring and modeling negative emotions for virtual training}
    \item Matje van de Camp (TiU) {\em A Link to the Past: Constructing Historical Social Networks from Unstructured Data}
    \item Archana Nottamkandath (VU) {\em Trusting Crowdsourced Information on Cultural Artefacts}
    \item George Karafotias (VUA) {\em Parameter Control for Evolutionary Algorithms}
    \item Anne Schuth (UVA) {\em Search Engines that Learn from Their Users}
    \item Max Knobbout (UU) {\em Logics for Modelling and Verifying Normative Multi-Agent Systems}
    \item Nana Baah Gyan (VU) {\em The Web, Speech Technologies and Rural Development in West Africa - An ICT4D Approach}
    \item Ravi Khadka (UU) {\em Revisiting Legacy Software System Modernization}
    \item Steffen Michels (RUN) {\em Hybrid Probabilistic Logics - Theoretical Aspects, Algorithms and Experiments}
    \item Guangliang Li (UVA) {\em Socially Intelligent Autonomous Agents that Learn from Human Reward}
    \item Berend Weel (VU) {\em Towards Embodied Evolution of Robot Organisms}
    \item Albert Mero\~{n}o Pe\~{n}uela (VU) {\em Refining Statistical Data on the Web}
    \item Julia Efremova (Tu/e) {\em Mining Social Structures from Genealogical Data}
    \item Daan Odijk (UVA) {\em Context \& Semantics in News \& Web Search}
    \item Alejandro Moreno C\'{e}lleri (UT) {\em From Traditional to Interactive Playspaces: Automatic Analysis of Player Behavior in the Interactive Tag Playground}
    \item Grace Lewis (VU) {\em Software Architecture Strategies for Cyber-Foraging Systems}
    \item Fei Cai (UVA) {\em Query Auto Completion in Information Retrieval}
    \item Brend Wanders (UT) {\em Repurposing and Probabilistic Integration of Data; An Iterative and data model independent approach}
    \item Julia Kiseleva (TU/e) {\em Using Contextual Information to Understand Searching and Browsing Behavior}
    \item Dilhan Thilakarathne (VU) {\em In or Out of Control: Exploring Computational Models to Study the Role of Human Awareness and Control in Behavioural Choices, with Applications in Aviation and Energy Management Domains}
    \item Wen Li (TUD) {\em Understanding Geo-spatial Information on Social Media}
    \item Mingxin Zhang (TUD) {\em Large-scale Agent-based Social Simulation - A study on epidemic prediction and control}
    \item Nicolas H\"{o}ning (TUD) {\em Peak reduction in decentralised electricity systems -Markets and prices for flexible planning}
    \item Ruud Mattheij (UvT) {\em The Eyes Have It}
    \item Mohammad Khelghati (UT) {\em Deep web content monitoring}
    \item Eelco Vriezekolk (UT) {\em Assessing Telecommunication Service Availability Risks for Crisis Organisations}
    \item Peter Bloem (UVA) {\em Single Sample Statistics, exercises in learning from just one example}
    \item Dennis Schunselaar (TUE) {\em Configurable Process Trees: Elicitation, Analysis, and Enactment}
    \item Zhaochun Ren (UVA) {\em Monitoring Social Media: Summarization, Classification and Recommendation}
    \item Daphne Karreman (UT) {\em Beyond R2D2: The design of nonverbal interaction behavior optimized for robot-specific morphologies}
    \item Giovanni Sileno (UvA) {\em Aligning Law and Action - a conceptual and computational inquiry}
    \item Andrea Minuto (UT) {\em MATERIALS THAT MATTER  -  Smart Materials meet Art \& Interaction Design}
    \item Merijn Bruijnes (UT) {\em Believable Suspect Agents; Response and Interpersonal Style Selection for an Artificial Suspect}
    \item Christian Detweiler (TUD) {\em Accounting for Values in Design}
    \item Thomas King (TUD) {\em Governing Governance: A Formal Framework for Analysing Institutional Design and Enactment Governance}
    \item Spyros Martzoukos (UVA) {\em Combinatorial and Compositional Aspects of Bilingual Aligned Corpora}
    \item Saskia Koldijk (RUN) {\em Context-Aware Support for Stress Self-Management: From Theory to Practice}
    \item Thibault Sellam (UVA) {\em Automatic Assistants for Database Exploration}
    \item Bram van de Laar (UT) {\em Experiencing Brain-Computer Interface Control}
    \item Jorge Gallego Perez (UT) {\em Robots to Make you Happy}
    \item Christina Weber (UL) {\em Real-time foresight - Preparedness for dynamic innovation networks}
    \item Tanja Buttler (TUD) {\em Collecting Lessons Learned}
    \item Gleb Polevoy (TUD) {\em Participation and Interaction in Projects. A Game-Theoretic Analysis}
    \item Yan Wang (UVT) {\em The Bridge of Dreams: Towards a Method for Operational Performance Alignment in IT-enabled Service Supply Chains}
\end{enumerate}

\subsubsection*{2017}

\begin{enumerate}
    \item Jan-Jaap Oerlemans (UL) {\em Investigating Cybercrime}
    \item Sjoerd Timmer (UU) {\em Designing and Understanding Forensic Bayesian Networks using Argumentation}
    \item Daniël Harold Telgen (UU) {\em Grid Manufacturing; A Cyber-Physical Approach with Autonomous Products and Reconfigurable Manufacturing Machines}
    \item Mrunal Gawade (CWI) {\em MULTI-CORE PARALLELISM IN A COLUMN-STORE}
    \item Mahdieh Shadi (UVA) {\em Collaboration Behavior}
    \item Damir Vandic (EUR) {\em Intelligent Information Systems for Web Product Search}
    \item Roel Bertens (UU) {\em Insight in Information: from Abstract to Anomaly}
    \item Rob Konijn (VU) {\em Detecting Interesting Differences:Data Mining in Health Insurance Data using Outlier Detection and Subgroup Discovery}
    \item Dong Nguyen (UT) {\em Text as Social and Cultural Data: A Computational Perspective on Variation in Text}
    \item Robby van Delden (UT) {\em (Steering) Interactive Play Behavior}
    \item Florian Kunneman (RUN) {\em Modelling patterns of time and emotion in Twitter \#anticipointment}
	\item Sander Leemans (TUE) {\em Robust Process Mining with Guarantees}
 	\item Gijs Huisman (UT) {\em Social Touch Technology - Extending the reach of social touch through haptic technology}
 	\item Shoshannah Tekofsky (UvT) {\em You Are Who You Play You Are: Modelling Player Traits from Video Game Behavior}
 	\item Peter Berck (RUN) {\em Memory-Based Text Correction}
 	\item Aleksandr Chuklin (UVA) {\em Understanding and Modeling Users of Modern Search Engines}
 	\item Daniel Dimov (UL) {\em Crowdsourced Online Dispute Resolution}
 	\item Ridho Reinanda (UVA) {\em Entity Associations for Search}
 	\item Jeroen Vuurens (UT) {\em Proximity of Terms, Texts and Semantic Vectors in Information Retrieval}
 	\item Mohammadbashir Sedighi (TUD) {\em Fostering Engagement in Knowledge Sharing: The Role of Perceived Benefits, Costs and Visibility}
 	\item Jeroen Linssen (UT) {\em Meta Matters in Interactive Storytelling and Serious Gaming (A Play on Worlds)}
 	\item Sara Magliacane (VU) {\em Logics for causal inference under uncertainty}
 	\item David Graus (UVA) {\em Entities of Interest --- Discovery in Digital Traces}
 	\item Chang Wang (TUD) {\em Use of Affordances for Efficient Robot Learning}
 	\item Veruska Zamborlini (VU) {\em Knowledge Representation for Clinical Guidelines, with applications to Multimorbidity Analysis and Literature Search}
 	\item Merel Jung (UT) {\em Socially intelligent robots that understand and respond to human touch}
 	\item Michiel Joosse (UT) {\em Investigating Positioning and Gaze Behaviors of Social Robots: People's Preferences, Perceptions and Behaviors}
 	\item John Klein (VU) {\em Architecture Practices for Complex Contexts}
 	\item Adel Alhuraibi (UvT) {\em From IT-BusinessStrategic Alignment to Performance: A Moderated Mediation Model of Social Innovation, and Enterprise Governance of IT"}
 	\item Wilma Latuny (UvT) {\em The Power of Facial Expressions}
 	\item Ben Ruijl (UL) {\em Advances in computational methods for QFT calculations}
 	\item Thaer Samar (RUN) {\em Access to and Retrievability of Content in Web Archives}
 	\item Brigit van Loggem (OU) {\em Towards a Design Rationale for Software Documentation: A Model of Computer-Mediated Activity}
 	\item Maren Scheffel (OU) {\em The Evaluation Framework for Learning Analytics}
 	\item Martine de Vos (VU) {\em Interpreting natural science spreadsheets}
 	\item Yuanhao Guo (UL) {\em Shape Analysis for Phenotype Characterisation from High-throughput Imaging}
 	\item Alejandro Montes Garcia (TUE) {\em WiBAF: A Within Browser Adaptation Framework that Enables Control over Privacy}
 	\item Alex Kayal (TUD) {\em Normative Social Applications}
 	\item Sara Ahmadi (RUN) {\em Exploiting properties of the human auditory system and compressive sensing methods to increase noise robustness in ASR}
 	\item Altaf Hussain Abro (VUA) {\em Steer your Mind: Computational Exploration of Human Control in Relation to Emotions, Desires and Social Support For applications in human-aware support systems}
 	\item Adnan Manzoor (VUA) {\em Minding a Healthy Lifestyle: An Exploration of Mental Processes and a Smart Environment to Provide Support for a Healthy Lifestyle}
 	\item Elena Sokolova (RUN) {\em Causal discovery from mixed and missing data with applications on ADHD datasets}
 	\item Maaike de Boer (RUN) {\em Semantic Mapping in Video Retrieval}
 	\item Garm Lucassen (UU) {\em Understanding User Stories - Computational Linguistics in Agile Requirements Engineering}
 	\item Bas Testerink	(UU) {\em Decentralized Runtime Norm Enforcement}
 	\item Jan Schneider	(OU) {\em Sensor-based Learning Support}
 	\item Jie Yang (TUD) {\em Crowd Knowledge Creation Acceleration}
 	\item Angel Suarez (OU) {\em Collaborative inquiry-based learning}   
\end{enumerate}

\subsubsection*{2018}

\begin{enumerate}
	\item Han van der Aa (VUA) {\em Comparing and Aligning Process Representations}
 	\item Felix Mannhardt (TUE) {\em Multi-perspective Process Mining}
 	\item Steven Bosems (UT) {\em Causal Models For Well-Being: Knowledge Modeling, Model-Driven Development of Context-Aware Applications, and Behavior Prediction}
 	\item Jordan Janeiro (TUD) {\em Flexible Coordination Support for Diagnosis Teams in Data-Centric Engineering Tasks}
 	\item Hugo Huurdeman (UVA) {\em Supporting the Complex Dynamics of the Information Seeking Process}
 	\item Dan Ionita (UT) {\em Model-Driven Information Security Risk Assessment of Socio-Technical Systems}
 	\item Jieting Luo (UU) {\em A formal account of opportunism in multi-agent systems}
 	\item Rick Smetsers (RUN) {\em Advances in Model Learning for Software Systems}
 	\item Xu Xie	(TUD) {\em Data Assimilation in Discrete Event Simulations}
 	\item Julienka Mollee (VUA) {\em Moving forward: supporting physical activity behavior change through intelligent technology}
 	\item Mahdi Sargolzaei (UVA) {\em Enabling Framework for Service-oriented Collaborative Networks}
 	\item Xixi Lu (TUE) {\em Using behavioral context in process mining}
 	\item Seyed Amin Tabatabaei (VUA) {\em Computing a Sustainable Future}
 	\item Bart Joosten (UVT) {\em Detecting Social Signals with Spatiotemporal Gabor Filters}
 	\item Naser Davarzani (UM) {\em Biomarker discovery in heart failure}
 	\item Jaebok Kim (UT) {\em Automatic recognition of engagement and emotion in a group of children}
 	\item Jianpeng Zhang (TUE) {\em On Graph Sample Clustering}
 	\item Henriette Nakad (UL) {\em De Notaris en Private Rechtspraak}
 	\item Minh Duc Pham (VUA) {\em Emergent relational schemas for RDF}
 	\item Manxia Liu (RUN) {\em Time and Bayesian Networks}
 	\item Aad Slootmaker (OUN) {\em EMERGO: a generic platform for authoring and playing scenario-based serious games}
 	\item Eric Fernandes de Mello Araujo (VUA) {\em Contagious: Modeling the Spread of Behaviours, Perceptions and Emotions in Social Networks}
 	\item Kim Schouten (EUR) {\em Semantics-driven Aspect-Based Sentiment Analysis}
 	\item Jered Vroon (UT) {\em Responsive Social Positioning Behaviour for Semi-Autonomous Telepresence Robots}
 	\item Riste Gligorov (VUA) {\em Serious Games in Audio-Visual Collections}
 	\item Roelof Anne Jelle de Vries (UT) {\em Theory-Based and Tailor-Made: Motivational Messages for Behavior Change Technology}
 	\item Maikel Leemans (TUE) {\em Hierarchical Process Mining for Scalable Software Analysis}
 	\item Christian Willemse (UT) {\em Social Touch Technologies: How they feel and how they make you feel}
 	\item Yu Gu (UVT) {\em Emotion Recognition from Mandarin Speech}
 	\item Wouter Beek {\em The "K" in "semantic web" stands for "knowledge": scaling semantics to the web}
\end{enumerate}

\subsubsection*{2019}

\begin{enumerate}
	\item Rob van Eijk (UL), {\em Comparing and Aligning Process Representations}
 	\item Emmanuelle Beauxis Aussalet (CWI, UU), {\em Statistics and Visualizations for Assessing Class Size Uncertainty}
 	\item Eduardo Gonzalez Lopez de Murillas (TUE) {\em Process Mining on Databases: Extracting Event Data from Real Life Data Sources}
 	\item Ridho Rahmadi (RUN) {\em Finding stable causal structures from clinical data}
 	\item Sebastiaan van Zelst (TUE) {\em Process Mining with Streaming Data}
 	\item Chris Dijkshoorn (VU) {\em Nichesourcing for Improving Access to Linked Cultural Heritage Datasets}
 	\item Soude Fazeli (TUD)
 	\item Frits de Nijs (TUD) {\em Resource-constrained Multi-agent Markov Decision Processes}
 	\item Fahimeh Alizadeh Moghaddam (UVA) {\em Self-adaptation for energy efficiency in software systems}
 	\item Qing Chuan Ye (EUR) {\em Multi-objective Optimization Methods for Allocation and Prediction}
 	\item Yue Zhao (TUD) {\em Learning Analytics Technology to Understand Learner Behavioral Engagement in MOOCs}
 	\item Jacqueline Heinerman (VU) {\em Better Together}
 	\item Guanliang Chen (TUD) {\em MOOC Analytics: Learner Modeling and Content Generation}
 	\item Daniel Davis (TUD) {\em Large-Scale Learning Analytics: Modeling Learner Behavior \& Improving Learning Outcomes in Massive Open Online Courses}
 	\item Erwin Walraven (TUD) {\em Planning under Uncertainty in Constrained and Partially}
 	\item Guangming Li (TUE) {\em Process Mining based on Object-Centric Behavioral Constraint (OCBC) Models}
 	\item Ali H\"{u}rriyeto\v{g}lu (RUN) {\em Extracting Actionable Information from Microtexts}
\end{enumerate}

\end{multicols}

\end{document}